\documentclass{article} 

\usepackage[preprint]{colm2026_conference}

\usepackage{microtype}
\usepackage{hyperref}
\usepackage{url}
\usepackage{booktabs}

\usepackage{amsmath}
\usepackage{graphicx}
\usepackage[many,most,skins,theorems]{tcolorbox}
\usepackage{algorithm}
\usepackage{algpseudocode}
\usepackage{wrapfig}
\usepackage{xcolor}
\usepackage{algorithmicx}
\usepackage{multirow} 
\usepackage{subcaption}
\usepackage{amssymb}
\usepackage{mathtools}
\usepackage{amsthm}
\usepackage[table]{xcolor}
\usepackage{fontawesome5}
\usepackage{bbding}
\usepackage{makecell}

\usepackage{tcolorbox}
\tcbuselibrary{breakable}
\usepackage{listings}
\usepackage{threeparttable}
\usepackage{cleveref}

\definecolor{acadNavy}{RGB}{40,70,140}      
\definecolor{acadMustard}{RGB}{180,140,30} 

\definecolor{rqBlueBg}{HTML}{EAF4FF}
\definecolor{tzBlueHeader}{RGB}{78,160,205}
\definecolor{tzBlueHeader2}{RGB}{105,185,225}
\definecolor{tzBlueBorder}{RGB}{115,190,225}
\definecolor{tzBlueFill}{RGB}{232,246,252}
\definecolor{rqBlueBorder}{HTML}{6AADE4}

\DeclareTextCommand{\textquotedbl}{OT1}{\char`\"}

\lstdefinestyle{jsonTiny}{
  basicstyle=\ttfamily\scriptsize,
  breaklines=true,
  breakindent=0pt,
  columns=fullflexible,
  keepspaces=true,
  showstringspaces=false,
  upquote=true,
  frame=none
}

\tcbuselibrary{skins,breakable}

\newtcolorbox{block}[1][]{%
  enhanced,
  breakable,
  colback=white,
  colframe=black!85,
  boxrule=1.4pt,
  arc=7pt,
  left=4mm,right=4mm,top=4mm,bottom=3mm,
  before skip=10pt, after skip=10pt,
  #1
}

\usepackage{pifont}
\newcommand{\cmark}{\textcolor[rgb]{0.0, 0.6, 0.0}{\ding{51}}} 
\newcommand{\xmark}{\textcolor[rgb]{0.7, 0.0, 0.0}{\ding{55}}} 
\newcommand{\gmark}{\textcolor[rgb]{1,0.647,0}{\ding{51}}}

\usepackage{enumitem}
\newenvironment{itemize*}%
 {\leftmargini=20pt\begin{itemize}%
  \setlength{\itemsep}{3pt}%
  \setlength{\parskip}{0pt}%
  }%
 {\end{itemize}} 
\newenvironment{enumerate*}%
 {\begin{enumerate}%
  \setlength{\itemsep}{0pt}%
  \setlength{\parskip}{0pt}}%
 {\end{enumerate}}

\tcbset{
  aibox/.style={
    width=\linewidth,
    top=7pt,
    bottom=2pt,
    colback=rqBlueBg,     
    colframe=black,
    colbacktitle=black,
    enhanced,
    center,
    attach boxed title to top left={yshift=-0.1in,xshift=0.15in},
    boxed title style={boxrule=0pt,colframe=white,},
  }
}
\newtcolorbox{AIbox}[2][]{aibox,title=#2,#1}
\newcounter{takeaway}
\newtcolorbox{takeaway}[1][]{
  aibox,
  colback=rqBlueBg,
  title={\stepcounter{takeaway}Takeaway \thetakeaway},
  #1
}

\NewDocumentCommand{\heng}
{ mO{} }{\textcolor{red}{\textsuperscript{\textit{Heng}}\textsf{\textbf{\small[#1]}}}}

\NewDocumentCommand{\cheng}
{ mO{} }{\textcolor{orange}{\textsuperscript{\textit{Cheng}}\textsf{\textbf{\small[#1]}}}}

\NewDocumentCommand{\zhenhailong}
{ mO{} }{\textcolor[HTML]{3399CC}{\textsuperscript{\textit{Zhenhailong}}\textsf{\textbf{\small[#1]}}}}

\NewDocumentCommand{\ember}
{ mO{} }{\textcolor{purple}{\textsuperscript{\textit{Ember}}\textsf{\textbf{\small[#1]}}}}

\NewDocumentCommand{\jiayu}
{ mO{} }{\textcolor{green}{\textsuperscript{\textit{jiayu}}\textsf{\textbf{\small[#1]}}}}

\NewDocumentCommand{\jeongh}
{ mO{} }{\textcolor{brown}{\textsuperscript{\textit{Jeonghwan}}\textsf{\textbf{\small[#1]}}}}

\definecolor{aditiColor}{HTML}{E07A5F}
\NewDocumentCommand{\aditi}
{ mO{} }{\textcolor{aditiColor}{\textsuperscript{\textit{Aditi}}\textsf{\textbf{\small[#1]}}}}

\NewDocumentCommand{\dwip}
{ mO{} }{\textcolor{yellow}{\textsuperscript{\textit{Dwip}}\textsf{\textbf{\small[#1]}}}}

\NewDocumentCommand{\xiusi}
{ mO{} }{\textcolor{cyan}{\textsuperscript{\textit{Xiusi}}\textsf{\textbf{\small[#1]}}}}

\usepackage{xcolor}
\definecolor{lightcommentblue}{RGB}{100,149,237} 

\NewDocumentCommand{\jiateng}{ mO{} }{%
  \textcolor{lightcommentblue}{\textsuperscript{\textit{jiateng}}\textsf{\textbf{\small[#1]}}}%
}

\title{%
  \begin{minipage}[c]{0.045\textwidth}
    \centering
    \vspace{-4mm}
    \includegraphics[width=0.8cm]{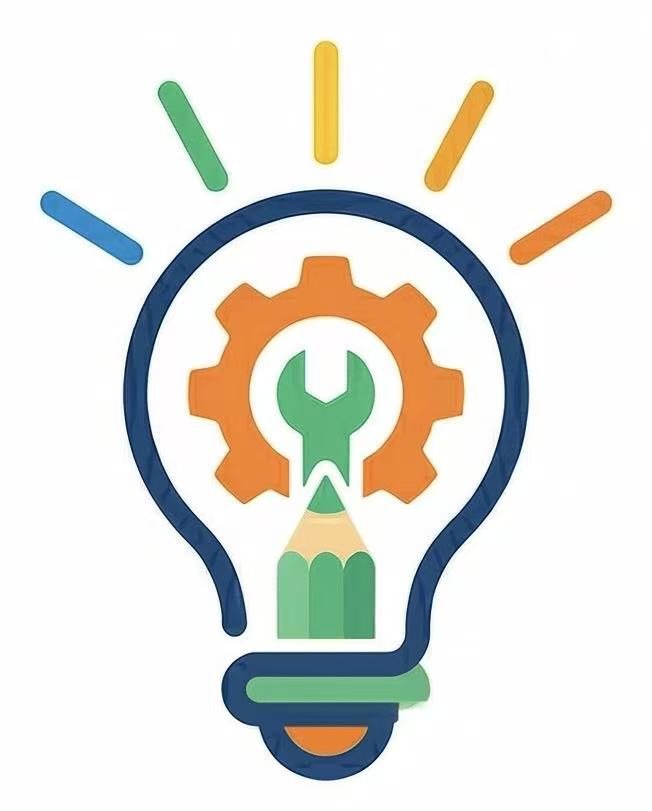}
  \end{minipage}\hfill
  \begin{minipage}[c]{0.93\textwidth}
    \raggedright
    CreativityBench: Evaluating Agent Creative Reasoning via Affordance-Based Tool Repurposing
  \end{minipage}
}

\author{
\noindent Cheng Qian$^*$\textsuperscript{$1$}, Hyeonjeong Ha$^*$\textsuperscript{$1$}, Jiayu Liu\textsuperscript{$1$}, Jeonghwan Kim\textsuperscript{$1$}, Jiateng Liu\textsuperscript{$1$}, \\
\bfseries\ Bingxuan Li\textsuperscript{$1$}, Aditi Tiwari\textsuperscript{$1$}, \bfseries\ Dwip Dalal\textsuperscript{$1$}, Zhenhailong Wang\textsuperscript{$1$}, \bfseries\ Xiusi Chen\textsuperscript{$1$}, \\
\bfseries\ Mahdi Namazifar\textsuperscript{$2$}, \bfseries\ Yunzhu Li\textsuperscript{$3$}, \bfseries\ Heng Ji\textsuperscript{$1$} \vspace{2.5mm} \\
\textsuperscript{$1$}UIUC, \textsuperscript{$2$}Amazon, \textsuperscript{$3$}Columbia
}

\begin{document}
\renewcommand{\thefootnote}{\fnsymbol{footnote}}
\footnotetext[1]{Equal contribution.}

\ifcolmsubmission
\linenumbers
\fi

\maketitle

\begin{abstract}
Recent advances in large language models have led to strong performance on reasoning and environment-interaction tasks, yet their ability for creative problem-solving remains underexplored. We study this capability through the lens of \textit{creative tool use}, where a model repurposes available objects by reasoning about their affordances and attributes rather than relying on canonical usage. As a first step, we introduce \textbf{CreativityBench}, a benchmark for evaluating affordance-based creativity in LLMs. To this end, we build a large-scale affordance knowledge base (KB) with 4K entities and 150K+ affordance annotations, explicitly linking objects, parts, attributes, and actionable uses. Building on this KB, we generate 14K grounded tasks that require identifying non-obvious yet physically plausible solutions under constraints.
Evaluations across 10 state-of-the-art LLMs, including closed and open-source models, show that models can often select a plausible object, but fail to identify the correct parts, their affordances, and the underlying physical mechanism needed to solve the task, leading to a significant drop in performance. Furthermore, improvements from model scaling quickly saturate, strong general reasoning does not reliably translate to creative affordance discovery, and common inference-time strategies such as Chain-of-Thought yield limited gains. These results suggest that creative tool use remains a major challenge for current models, and that CreativityBench provides a useful testbed for studying this missing dimension of intelligence, with potential implications for planning and reasoning modules in future agents.
\end{abstract}

\vspace{-3mm}
\begin{center}
\small

\newcommand{\logoh}{1.35em}

\href{https://github.com/qiancheng0/CreativityBench}{
\raisebox{-0.2\height}{\includegraphics[height=\logoh]{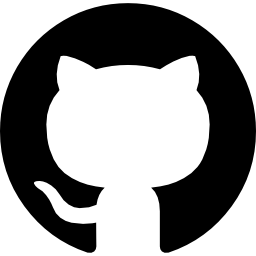}}
\hspace{0.35em}{\textbf{Code}}
}
\quad
\href{https://creativitybench.github.io/}{
\raisebox{-0.2\height}{\includegraphics[height=\logoh]{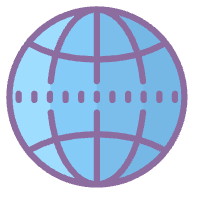}}
\hspace{0.35em}{\textbf{Project Page}}
}

\end{center}
\vspace{-2mm}
\section{Introduction}

Long before intelligence was formalized in theory, it was visible in action. By 3.3 to 3.4 million years ago, early humans were already using tools to reshape their environment, hinting at the creative capacities that would later become central to human innovation. The \textbf{Triarchic Theory of Intelligence}~\citep{sternberg1997triarchic} characterizes human intelligence into three components: analytical, practical, and creative. This perspective provides a useful lens for understanding recent progress in large language models (LLMs). Existing advances can be largely concentrated along the first two dimensions. Recent LLMs exhibit strong \textbf{analytical intelligence}, including logical deduction, mathematical reasoning, and maintaining coherent chains of thought, as reflected in standard reasoning and mathematics benchmarks, which capture improvements in internal cognitive processing~\citep{hendrycks2020measuring, cobbe2021training, wei2022chain}. In parallel, LLMs have rapidly advanced in \textbf{practical intelligence}, acquiring the ability to interact with tools, browse the web, manipulate software interfaces, and execute long-horizon tasks in simulated or embodied environments, as evaluated by benchmarks such as BrowseComp, GAIA, and ARE~\citep{wei2025browsecomp, mialon2023gaia, froger2025scaling}. Recent LLMs can now complete tasks involving hundreds of actions~\citep{xi2025survey, ge2025survey, yu2025browseragent} by successfully translating reasoning into effective action in the external world, reflecting substantial progress in reasoning and execution.

However, \textbf{creative intelligence} (i.e., the ability to generate novel and useful ideas and solutions) remains a moonshot goal. Unlike analytical correctness or effective execution, creative intelligence is the ability to \textbf{produce novel yet useful solutions under constraints}~\citep{runco2012standard, sternberg1999concept}. This ability is essential for real-world problem solving, where the path to success is often not given and must be invented by repurposing available resources in non-obvious ways. While modern LLMs can reason accurately and act effectively, they remain limited in this kind of flexible problem solving that humans routinely exhibit in open-ended environments. Despite its importance, creativity in LLMs remains poorly defined and insufficiently evaluated.

\begin{wrapfigure}{r}{0.55\linewidth}
    \centering
    \vspace{-3mm}
    \includegraphics[width=\linewidth]{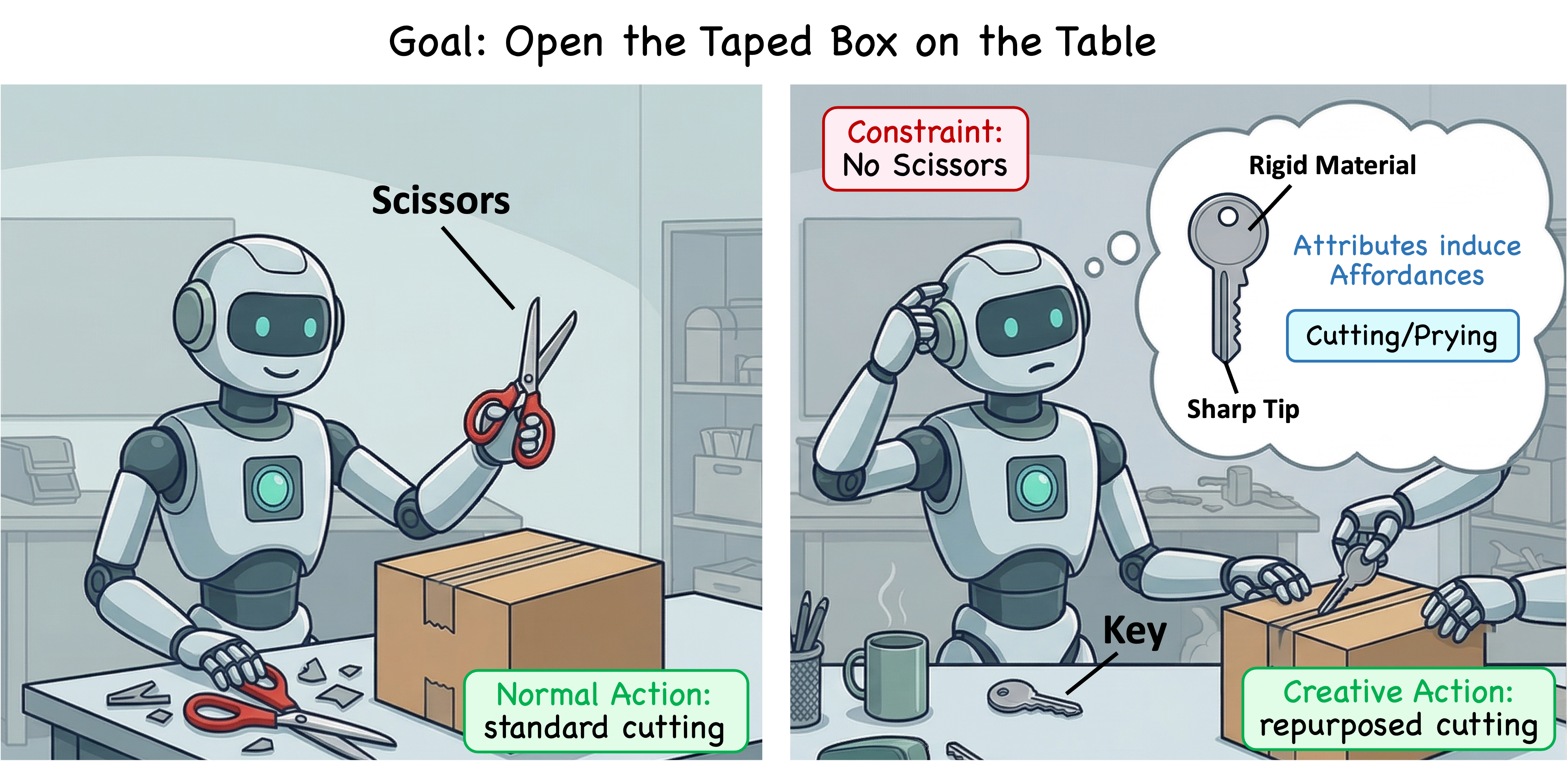}
    \caption{\textbf{Creative tool use as affordance-based tool repurposing.} Under constraints, a model solves a task by identifying and using the affordances of an alternative object.}
    \label{fig:intro}
    \vspace{-2mm}
\end{wrapfigure}

We argue that a core form of creative intelligence is \textbf{creative tool use}: the ability to infer and exploit an object's \emph{affordances}, the actions enabled by its physical attributes, to achieve the goal in a novel and unconventional way. Humans frequently exhibit this ability by reasoning over part-level properties (e.g., rigidity, elasticity) and mapping them to functional affordances (e.g., cutting, prying), enabling objects to be repurposed beyond their intended use. For example, a key can be used to open a sealed box because its rigid, sharp tip structure affords prying or cutting (Fig~\ref{fig:intro}). This highlights a key distinction: creative reasoning is not random exploration or hallucinated actions, but the discovery of unobvious functional connections grounded in physical reality: tools are useful not only for their intended purposes, but for the actions their structural and physical attributes enable. Therefore, creative reasoning requires divergent thinking while remaining anchored to physical constraints, often by reformulating existing knowledge and prior tool-use experience into a new solution. Therefore, evaluating creativity in LLMs requires assessing whether models can go beyond surface and reason about \emph{how affordances emerge from object structure}, rather than simply identifying plausible objects.

Despite its significance, creative tool use remains largely underexplored in existing evaluations. Prior benchmarks~\citep{tian2024macgyver, qian2024escapebench, fang2025creation, lim2025visescape, dong2024villageragent} have explored creativity through physical commonsense reasoning, embodied planning, multimodal understanding, and interactive explorations. However, they primarily focus on predicting plausible actions, navigating environments, or solving scenario-based tasks. These approaches rarely require models to ground decisions in fine-grained, part-level physical attributes or to explicitly reason about affordance emergence. As a result, current evaluations emphasize planning and execution, while overlooking whether models can identify concrete, part-level affordances and leverage them for creative problem-solving beyond coarse object-level plausibility. Therefore, they fail to systematically assess whether models can repurpose tools based on physically usable properties, referred to as creative tool use. Our preliminary study further suggests that this limitation is not resolved by simply enforcing more structured reasoning: even when explicitly guided to decompose tools into parts, infer physical properties, and reason step by step, strong models show only marginal gains. These gaps motivate three research questions:
\begin{itemize}[topsep=-3pt, leftmargin=10pt, itemsep=-2pt]
    \item How to construct a scalable and physically grounded affordance knowledge base capturing part-level attributes and their associated affordances?
    \item How to design a benchmark that rigorously evaluates affordance-based creative tool use beyond coarse object-level reasoning?
    \item How do current state-of-the-art models perform on affordance-based creative intelligence?
\end{itemize}

To address these questions, we introduce \textbf{CreativityBench}, the first large-scale benchmark designed to systematically evaluate creative intelligence through affordance-based creative tool use. Our benchmark is enabled by a scalable affordance annotation pipeline that constructs a structured affordance knowledge base (KB) linking objects, their constituent parts, attributes, and associated affordances, grounding model judgments in concrete object properties rather than semantic guessing alone. The resulting KB contains over 4K entities and 150K affordance annotations, forming reusable building blocks for task generation, trajectory construction, and evaluation. Using this KB, we generate diverse and physically grounded tasks by reverse-engineering creative solution trajectories, ensuring that each task requires non-obvious, physically grounded affordance reasoning rather than surface-level object matching.

We evaluate multiple proprietary and open-source models on a comprehensive suite of 14K tasks and reveal striking insights about the current limits of model creativity. \textbf{First, exact physical grounding remains a severe bottleneck:} while models can often identify a plausible tool entity, they fail to ground its use at the specific part or attribute level, resulting in a performance drop of over 60\%. \textbf{Second, analytical reasoning does not imply creative affordance discovery:} models that excel at logical reasoning (e.g., GPT-5 family) are outperformed in novel tool discovery by models like Qwen3-32B, indicating a clear dissociation between reasoning and creativity. \textbf{Third, creative tool use does not scale with model size:} performance quickly saturates with model size and remains heavily bounded by affordance commonality, with significant degradation on rare, long-tail tool repurposing. \textbf{Finally, standard inference-time interventions fall short:} strategies such as higher sampling temperature, structured Chain-of-Thought, and interactive evaluation modes yield minimal gains, often exacerbating hallucinations or revealing a tendency to prematurely commit to incorrect hypotheses rather than engaging in genuine creative exploration. To summarize, our contributions are threefold:
\begin{itemize}[topsep=-1.5pt, leftmargin=10pt, itemsep=0pt]
    \item \textbf{Affordance knowledge base:} We build the first large-scale, structured KB of tool affordances with 4K entities and 150K+ affordance annotations, serving as reusable building blocks for grounded task sampling, trajectory construction, training, and evaluation, and enabling creative reasoning via recombination of physically plausible affordances.
    \item \textbf{CreativityBench:} We introduce an affordance-grounded benchmark that evaluates creative tool use under rigorous, reproducible protocols, targeting the previously under-measured facet of creative intelligence.
    \item \textbf{Empirical analysis:} We conduct a systematic study of creative tool use, probing affordance uniqueness, noise, task difficulty, and evaluation modes.
\end{itemize}

By isolating and operationalizing creative tool use as a distinct capability, we hope this work establishes a foundation for studying creativity in LLMs beyond reasoning and interaction, and moves toward systems capable of solving unforeseen problems and acting as reliable helpers in diverse real-world situations.

\section{Related Work}
\subsection{Creativity in Language Models}
Creativity is one of the hallmarks of human intelligence, enabling us to act robustly in novel and unfamiliar environments. Recent large language models (LLMs) exhibit creative capabilities across diverse domains, including narrative and poetry generation~\citep{akoury2020storium, brown2020language}, tool and system design~\citep{qian2023creator, cai2023large, ha2025synthia}, modeling real-world problems, and supporting human brainstorming and ideation~\citep{qian2025modelingagent}. In scientific discovery settings, LLMs have also shown promise in generating hypotheses and research ideas that can complement human experts, although their novelty and feasibility vary across studies and evaluation settings~\citep{si2024can, wang2024scimon,liu2025costbench}.

A common line of work evaluates creativity in LLMs through adaptations of psychological creativity assessments~\citep{guilford1967creativity, boden1998creativity}, which measure attributes like fluency, originality, and flexibility. These evaluations suggest that modern models can achieve strong creativity scores, but they are often sensitive to prompt design and involve costly or noisy evaluation procedures, making them difficult to scale and imperfect indicators of model creativity. Beyond such psychological tests, several benchmarks investigate creativity in problem-solving settings. MacGyver~\citep{tian2024macgyver} evaluates whether models can solve everyday problems by repurposing available objects in unconventional ways, while EscapeBench~\citep{qian2024escapebench} studies creative reasoning in simulated escape-room environments, where models must discover non-obvious tool uses through extended exploratory interaction. Multimodal and embodied benchmarks further extend creativity evaluation to perception-grounded tasks. Creation-MMBench evaluates context-aware creative generation grounded in visual inputs \citep{fang2025creation}, while VisEscape~\citep{lim2025visescape} and VillagerBench~\citep{dong2024villageragent} study exploration and decision-making in interactive environments that require perception, planning and coordination. Despite these advances, the construction of tasks in existing benchmarks is typically scenario-driven or generated through prompts, and is \textit{not grounded physically in the fine-grained affordances of objects and their components}. As a result, these benchmarks often emphasize planning, reasoning, or multimodal understanding rather than the mechanism underlying creative tool use: \textit{identifying non-obvious functional affordances and repurposing them to satisfy task constraints}. In contrast, our work focuses on affordance-grounded creativity, where models must infer tool affordances to achieve goals under constrained environments.

\begin{table*}[!t]
\centering
\vspace{-3mm}
\resizebox{\linewidth}{!}{
\begin{tabular}{lccccccc}
\toprule
\textbf{Benchmark} & \textbf{\makecell{Creative \\ Tool Use}} & \textbf{\makecell{Affordance\\Grounding}} & \textbf{\makecell{Attribute\\Grounding}} & \textbf{\makecell{Part-Level\\Reasoning}} & \textbf{\makecell{Fine-Grained\\Creativity Levels}} & \textbf{\makecell{Distractors\\Included}} & \textbf{Annotation} \\
\midrule

\textit{PROST\citep{aroca2021prost}} & \xmark & \cmark & \cmark & \xmark & \xmark & \gmark & A+M\\
\textit{NEWTON\citep{wang2023newton}} & \xmark & \xmark & \cmark & \xmark & \xmark & \gmark & A+M\\
\textit{Creation-MMBench\citep{tian2024macgyver}}      
& \xmark & \xmark & \xmark & \xmark & \xmark & \xmark & A+M\\
\textit{VillagerBench\citep{dong2024villageragent}}      
& \xmark & \gmark & \xmark & \xmark & \xmark & \xmark & A \\
\textit{VisEscape\citep{lim2025visescape}}      
& \xmark & \gmark & \xmark & \xmark & \xmark & \gmark & A \\
\textit{PIQA\citep{bisk2020piqa}} & \gmark & \gmark & \gmark & \xmark & \xmark & \cmark & M\\
\textit{MacGyver\citep{tian2024macgyver}}      
& \cmark & \gmark & \gmark & \xmark & \xmark & \gmark & A+M\\
\textit{EscapeBench\citep{qian2024escapebench}}      
& \cmark & \gmark & \xmark & \xmark & \gmark & \gmark & M\\
\midrule
\multirow{1}{*}{\textbf{{\textit{CreativityBench} (Ours)}}}     
& \cmark & \cmark & \cmark & \cmark & \cmark & \cmark & A \\

\bottomrule
\end{tabular}
}
\caption{For each existing benchmark, the table indicates whether the corresponding dimension is fully addressed (\cmark), partially addressed (\gmark), or not addressed (\xmark). A indicates automatic annotation and M refers to manual annotation.}
\label{tab:comparison}
\vspace{-2mm}
\end{table*}

\subsection{Affordance and Physical Reasoning}
Recent work has studied whether AI systems can reason about the physical attributes and affordances of everyday objects. Benchmarks such as PIQA~\citep{bisk2020piqa} evaluate physical commonsense through goal-solution questions grounded in everyday tasks, while PROST~\citep{aroca2021prost} probes knowledge of physical attributes using a cloze-style question about object attributes and simple affordances. More recently, NEWTON~\citep{wang2023newton} scales physical reasoning evaluation through a large repository of object-attribute pairs and questions. In parallel, affordances have been widely studied in robotics as representations linking perception to action, where systems learn object–action relationships through interaction or visual perception to support manipulation and planning~\citep{brohan2022rt, brohan2024rt}. More recent approaches further integrate affordance reasoning with vision–language models to enable open-world manipulation and generalization~\citep{chu2019learning, montesano2008learning, jamone2016affordances, liu2025revisiting}. Despite these advances, both lines of work remain largely limited: they focus on predicting attributes or canonical actions of objects, but do not explicitly model \textit{how affordances arise from the structural and physical attributes of object components}.

Another direction focuses on constructing structured affordance knowledge. SYNTHIA~\citep{ha2025synthia} introduces a hierarchical concept ontology that decomposes objects into parts and their associated affordances to support affordance-aware concept generation. While this representations highlights the importance of part-level functional decomposition, it primarily encodes conceptual part-affordance associations, and does not explicitly model the physical attributes that determine whether a part can provide a given affordance (e.g., sharpness enabling cutting). In contrast, our benchmark explicitly grounds affordances through a structured hierarchy linking entities, parts, physical and state attributes, and affordances, enabling evaluation of whether models can identify and reason about the underlying physical mechanism that enable functional behavior, which is a core ability for creative reasoning in tool use.

\section{Preliminaries: Structured Reasoning Is Not Enough for Creativity}
\label{sec:preliminaries}

\begin{figure}[t]
    \centering
    \vspace{-3mm}
    \begin{subfigure}[t]{0.49\linewidth}
        \centering
        \includegraphics[width=\linewidth]{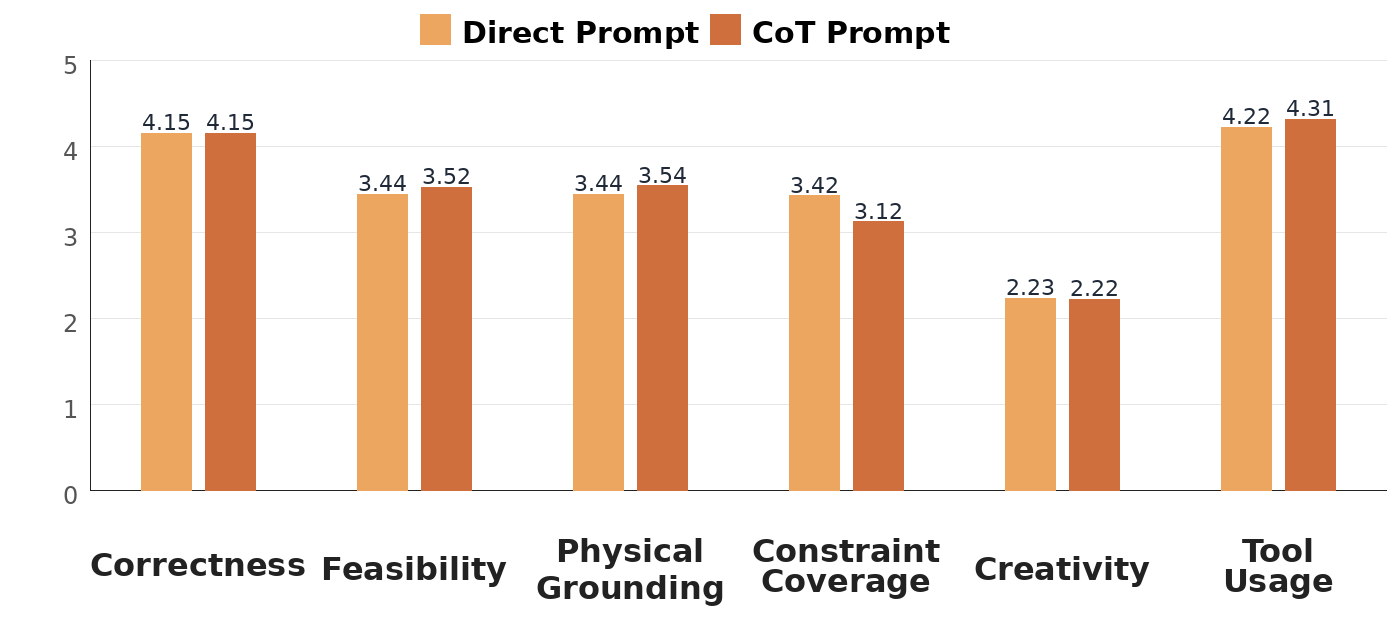}
        \caption{Absolute Evaluation.}
        \label{fig:prelim_abs_eval}
    \end{subfigure}
    \hfill
    \begin{subfigure}[t]{0.49\linewidth}
        \centering
        \includegraphics[width=\linewidth]{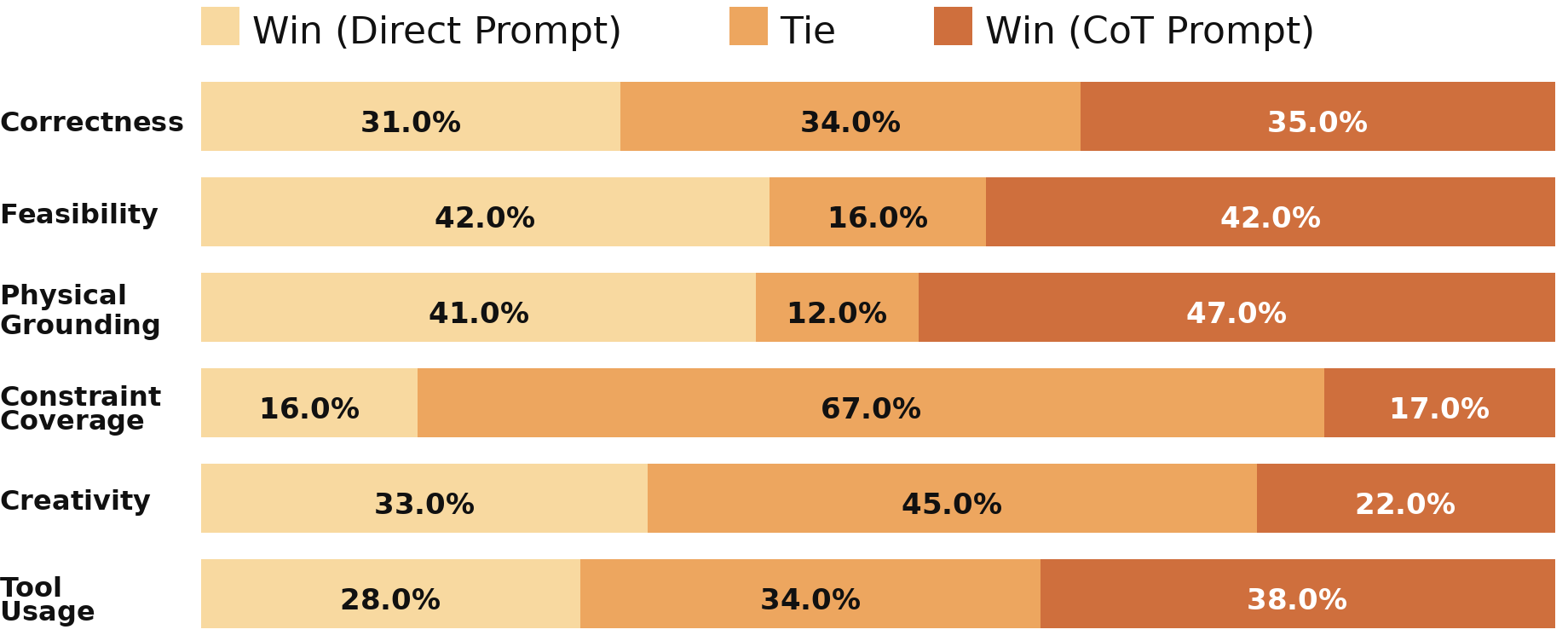}
        \caption{Relative Evaluation.}
        \label{fig:prelim_rel_eval}
    \end{subfigure}
    \caption{\textbf{Preliminary Experimental Results:} Comparison between direct prompting and structured affordance-level CoT on creative tool use tasks. While CoT improves grounded reasoning, it does not enhance creative reasoning, indicating that structured reasoning alone is insufficient for grounded affordance recombination.}
    \label{fig:prelim_exps}
    \vspace{-2mm}
\end{figure}

To examine the gap between analytical/practical intelligence and creative intelligence in LLMs, we conduct a controlled comparison on 100 creative tool-use tasks sampled from the MacGyver dataset~\citep{tian2024macgyver}, an unconventional physical problem-solving benchmark consisting of real-world verbal scenarios designed to push against functional fixedness and require innovative use of available objects. As an initial test of whether this benchmark is useful for probing our target capability, we study whether simple prompting interventions alone, especially explicit chain-of-thought scaffolds, can improve performance on these tasks before introducing any new knowledge resource or benchmark construction. We compare two prompting strategies:
\begin{itemize}[topsep=-4pt, leftmargin=10pt, itemsep=-2pt]
    \item \textbf{Direct prompt:} the model generates a feasible solution under task constraints without prescribed reasoning steps, testing its implicit ability to connect tasks with tool functions.
    \item \textbf{Structured affordance-level CoT:} the model follows an explicit reasoning guideline including tool inventory listing, part decomposition, physical property inference, affordance derivation, step-level justification, and constraint validation.
\end{itemize}
This comparison tests whether failures in creative tool use arise from missing procedural guidance or from deeper limitations in physically grounded affordance modeling and recombinational creativity. Detailed prompts are provided in \Cref{appendix_sec:prelim_prompt}. We use GPT-4.1-mini as the target LLM and GPT-5.2 as the judge model.

We evaluate generated solutions using six criteria capturing distinct aspects of creative tool use: \textbf{Correctness} (task goal achievement), \textbf{Feasibility} (physical executability under constraints), \textbf{Physical Grounding} (accurate use of object properties and mechanics), \textbf{Constraint Coverage} (handling all stated constraints), \textbf{Tool Usage} (proper and exclusive use of available tools), and \textbf{Creativity} (non-obvious yet effective affordance reinterpretation). This decomposition is important because task success alone cannot distinguish routine tool use from genuine creativity, while novelty without feasibility does not reflect grounded reasoning. We perform both absolute evaluation (1--5 score per criterion) and relative evaluation comparing outputs from these  two prompting strategies.

Empirically, as shown in \Cref{fig:prelim_exps}, structured CoT only yields modest improvements on several procedural dimensions in absolute evaluation, increasing Feasibility from 3.44 to 3.52, Physical Grounding from 3.44 to 3.54, and Tool Usage from 4.22 to 4.31. Relative evaluation shows a similar trend: CoT wins more often on Physical Grounding (47\% vs.\ 41\%) and Tool Usage (38\% vs.\ 28\%). However, CoT performs worse on Creative Reasoning, suggesting that while \textit{structured reasoning improves grounding and procedural accuracy, it may constrain divergent thinking}.

Together, these results suggest that the key limitation of current models is not missing a reasoning structure, but the lack of grounded affordance knowledge that can be flexibly recombined. Structured CoT improves procedural grounding, yet does not yield stronger creative affordance reinterpretation. Our preliminary study also highlights the limits of existing resources: benchmarks such as MacGyver are not explicitly built around affordance structure, and their evaluation often relies on LLM-as-judge scoring, making rigorous measurement of creativity difficult. This motivates our next step: constructing an explicit affordance knowledge base and building \textbf{CreativityBench}, a fine-grained part-level, attribute-grounded benchmark for creative tool use.

\begin{figure}
    \centering
    \vspace{-3mm}
    \includegraphics[width=\linewidth]{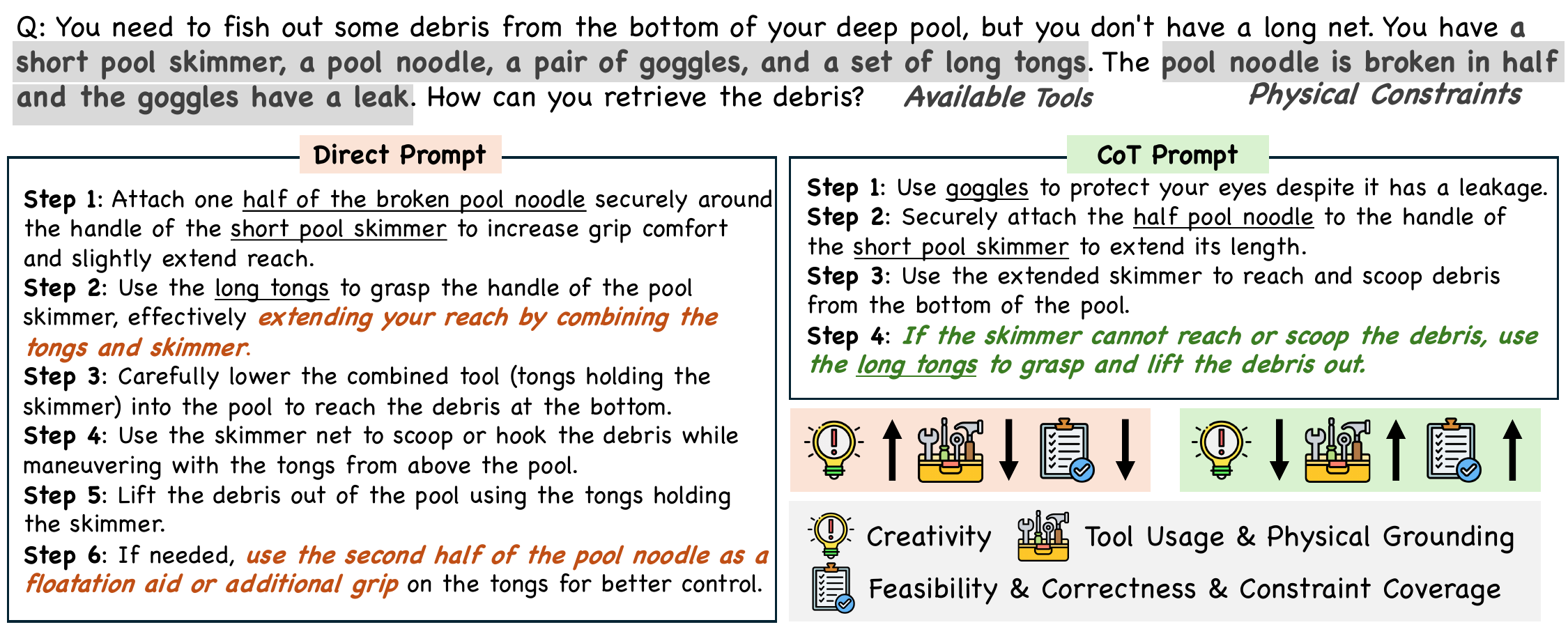}
    \caption{Qualitative comparison between generated responses from direct and CoT prompts.}
    \label{fig:prelim_example}
    \vspace{-2mm}
\end{figure}
\section{CreativityBench}
\label{sec:method}

Creative tool use provides a concrete mechanism for studying creative intelligence in LLMs. Importantly, a ``tool'' is not defined by its name or intended category, but by its \textbf{affordances}, which are the action possibilities it enables. These affordances emerge from the underlying \textbf{attributes} of an entity, such as its structure, material properties, interfaces, constraints, or accessible resources. Creative tool use, therefore, requires a model to identify which attributes in the environment enable useful affordances and how those affordances can be combined to achieve the goal, rather than matching tasks to tools based on semantic labels. 

This framing motivates our benchmark design: to construct creative tool use tasks and trajectories, we treat affordances as the organizing principle and ground them in the attributes of tools and other entities present in the environment. In this section, we describe how we build the affordance knowledge base and sample tasks for CreativityBench. To scale the annotation process, we use an LLM-assisted pipeline and adopt a reverse-engineering procedure that composes high-level creative tasks by chaining lower-level affordances.

\begin{figure}[!t]
    \centering 
    \vspace{-5mm}
    \includegraphics[width=\linewidth]{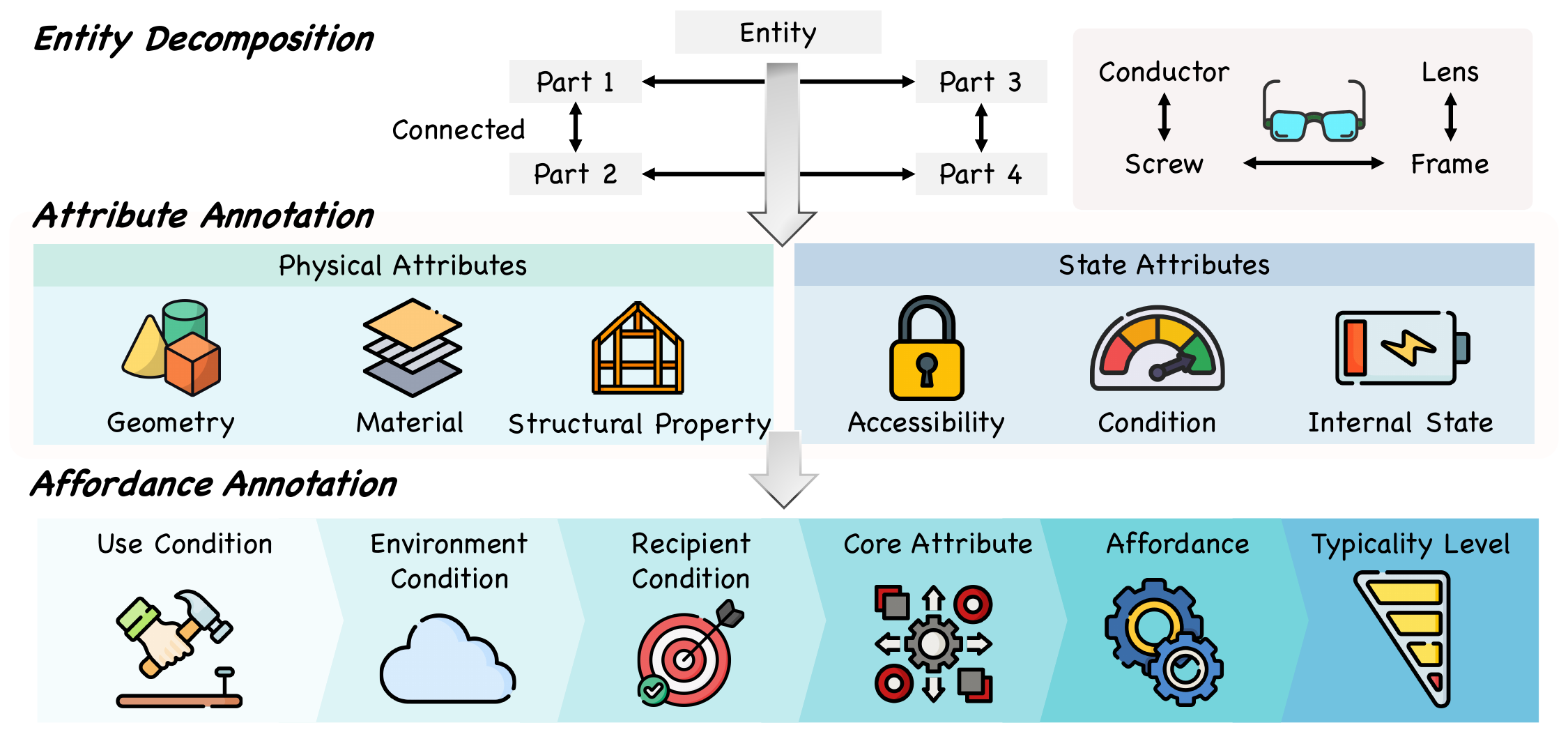}
    \caption{Annotation pipeline for affordance knowledge base construction.}
    \label{fig:annotation}
    \vspace{-2mm}
\end{figure}

At the core of CreativityBench is an affordance knowledge base that explicitly models how actionable possibilities arise from object structure and physical properties. We adopt a \textit{top-down} annotation pipeline that represents each object as a hierarchy linking entities, parts, attributes, and affordances. Formally, let $\mathcal{E}$ denote the set of entities. For each entity $e \in \mathcal{E}$, we decompose it into a set of parts $P(e)$, annotate the attributes of $A(p)$ associated with each part $p \in P(e)$, and derive the corresponding affordances $F(p)$ enabled by those attributes of the part $p$. This results in a structured mapping \(e \rightarrow P(e) \rightarrow A(p) \rightarrow F(p)\). An overview of our pipeline is shown in \Cref{fig:annotation}.

\subsection{Affordance Knowledge Base Construction}

\paragraph{Entity Decomposition.}
We first sample common entities from eight in-house scenes (e.g., kitchen, living room, bedroom), grounded in ConceptNet 5.5~\citep{speer2017conceptnet}. For each entity $e$, we decompose it into a set of non-overlapping parts:
\[
P(e) = \{p_1, p_2, \dots, p_n\}.
\]
The decomposition follows three constraints: 1. \textbf{Completeness:} $\bigcup_i p_i = e$, i.e., parts together cover the whole entity. 2. \textbf{Non-overlap:} $p_i \cap p_j = \varnothing$ for $i \neq j$. 3. \textbf{Functional granularity:} each part corresponds to a structurally or functionally meaningful component that may independently support useful affordances. This part-level representation is important because creative affordances often arise from local structural features rather than the intended function of the entire object (e.g., the sharp tip of a key can be used for cutting).

\paragraph{Attributes.}
For each part $p \in P(e)$, we annotate a set of attributes:
\[
A(p) = A_p(p) \cup A_s(p),
\]
where $A_p(p)$ and $A_s(p)$ respectively denotes physical and state attributes defined as follows:
\begin{itemize}[topsep=-4pt, leftmargin=10pt, itemsep=-2pt]
\item \textbf{Physical Attributes $A_p$:} intrinsic properties of a part that remain fixed, including geometry and shape (shape, size, thickness, local features), material and structural properties (material, rigidity, durability, elasticity), and mass, etc.
\item \textbf{State Attributes $A_s$:} properties that may change during interaction or use, such as accessibility (visibility, availability), condition (moisture, temperature), and internal states.
\end{itemize}
Each part is annotated with a shared set of predefined attribute fields, along with a flexible field to capture additional distinctive traits. Importantly, different combinations of physical and state attributes produce multiple \textit{variants} of the same entity. Although such variants share the same entity name, they are treated as distinct instances because their attributes, and therefore their affordances, may differ significantly. For example, a \textit{dry (moisture), empty (internal states)} vacuum bag affords storing other objects, while a \textit{dry, fully filled} vacuum bag becomes dense and compressible, allowing it to serve as a temporary cushion. While both refer to the same entity (``vacuum bag''), differences in state attributes lead to different affordances and potential uses.

\paragraph{Affordances.}
Affordances are annotated at the part level. For each part $p$, we define a set of affordances as follows:
\[
F(p) = \{f_1, f_2, \dots, f_m\}.
\]
Each affordance $f_i$ is represented as \(f = (a, C_u, C_e, C_r)\), where $a$ denotes the action enabled by the part, and the three conditions specify the prerequisites required for successful execution:
\begin{itemize}[topsep=-4pt, leftmargin=10pt, itemsep=-2pt]
\item \textbf{Use Condition $C_u$:} operations that must be performed on the entity before the affordance becomes available (e.g., breaking glass to create sharp edges).
\item \textbf{Environment Condition $C_e$:} environmental prerequisites needed for the affordance (e.g., the presence of a light source when focusing light through glass).
\item \textbf{Recipient Condition $C_r$:} constraints on the object being acted upon (e.g., the recipient must be softer than glass for cutting).
\end{itemize}
In addition, we also annotate the attributes that enable it, potential recipients, and failure conditions. Finally, we categorize affordances by their typicality. \textbf{Normal affordances} correspond to the intended functions of a part, while \textbf{emergency affordances} represent unconventional but plausible repurposings. Emergency affordances are further assigned a level $l \in \{1,\dots,5\}$, where a higher level indicates highly natural and practical repurposing, and lower levels indicate otherwise.

\begin{table*}[t]
    \centering
    \vspace{-3mm}
    \small
    \setlength\tabcolsep{6pt}
    \setlength\extrarowheight{0.5pt}
    \resizebox{0.7\linewidth}{!}{
    \begin{tabular}{l @{\hspace{8mm}} l @{\hspace{8mm}} c}
        \toprule
        \textbf{Category} & \textbf{Statistic} & \textbf{Value} \\
        \midrule
        \multirow{4}{*}{Scale} 
            & Number of scenes & 8 \\
            & Number of entities & 3,816 \\
            & Number of parts & 26,238 \\
            & Total annotations & 570,717 \\
        \midrule
        \multirow{3}{*}{Annotation counts}
            & Physical attributes annotated & 288,318 \\
            & State attributes annotated & 124,972 \\
            & Affordances annotated & 157,427 \\
        \midrule
        \multirow{4}{*}{Annotation density}
            & Avg.\ physical attributes per part & 10.9886 \\
            & Avg.\ state attributes per part & 4.7630 \\
            & Avg.\ physical attributes per entity & 75.5550 \\
            & Avg.\ state attributes per entity & 32.7495 \\
        \bottomrule
    \end{tabular}
    }
    \caption{Overall statistics of the affordance knowledge base. The dataset contains eight different household scenes, with annotations covering entities, attributes, and affordances.}
    \label{tab:dataset_overview}
    \vspace{-2mm}
\end{table*}

\paragraph{Knowledge Base Statistics.}
We automatically scale the annotation pipeline using GPT-5.2. Following the hierarchical structure (entity $\rightarrow$ part $\rightarrow$ attributes $\rightarrow$ affordances), the resulting knowledge base contains approximately 4K entities and 26K parts, with 288K physical attributes and 125K state attributes, yielding 157K annotated affordances of varying typicality levels. Detailed statistics are provided in \Cref{tab:dataset_overview}. Refer to \Cref{appendix_sec:annotation} for more details. \looseness=-1


\subsection{Benchmark Task Sampling}

Benchmark tasks are constructed from the affordance knowledge base through a \textit{reverse-engineering} process. Instead of starting from a task and identifying a suitable tool, we begin from a known affordance and synthesize a scenario in which discovering that affordance becomes the optimal solution. This bottom-up design ensures that each task has a well-defined ground-truth reasoning trajectory while still requiring the model to infer the affordance solely from object attributes.

Formally, let $\mathcal{F}=\bigcup_{p}F(p)$ denote the set of all annotated affordances. Each affordance $f=(a,C_u,C_e,C_r)$ is associated with an entity $e$, part $p\in P(e)$, and attributes $A(p)$. A benchmark task $T$ is defined as:
\[
T = (S, \mathcal{E}_T, g),
\]
where $S$ denotes the scenario description, $\mathcal{E}_T \subset \mathcal{E}$ is the set of entities present in the scene, and $g=(e^*,p^*,f^*)$ is the \textit{gold affordance} that provides the intended solution. The model is given only the entities and their attributes and must infer the correct entity $e^*$, part $p^*$, and affordance $f^*$. The construction procedure consists of four stages: gold affordance sampling, task synthesis, gold verification, and distractor sampling.

\paragraph{Gold Affordance Sampling.}
Directly sampling affordances uniformly from $\mathcal{F}$ leads to strong redundancy because many affordances are semantically similar (e.g., cutting with different sharp objects). To ensure diversity and controllable difficulty, we first cluster affordances within each scenario.

For a scenario $s \in S$, let $\mathcal{F}_s \subset \mathcal{F}$ denote its affordances. Each affordance is embedded using Text-Embedding-3-Large\footnote{https://developers.openai.com/api/docs/models/text-embedding-3-large}, and complete-linkage hierarchical clustering is applied over the extracted embeddings to obtain
\[
\mathcal{F}_s = \bigcup_{k=1}^{K_s} \mathcal{C}_k,
\]
where each $\mathcal{C}_k$ contains a set of semantically similar affordances. In practice, this yields approximately $3.5$K clusters per scenario.

A gold affordance $f^*$ (i.e. ground truth affordance) is then sampled from these clusters under several controllable factors:
\begin{itemize}[topsep=-4pt, leftmargin=10pt, itemsep=-2pt]
\item \textbf{Cluster Size:} whether $f^*$ originates from a large cluster (common affordance) or a small cluster (rare affordance).
\item \textbf{Affordance Typicality Level:} whether $f^*$ is normal or an emergency affordance of level $l$.
\end{itemize}
This structured sampling ensures that the benchmark contains a balanced mixture of common and long-tail affordances while enabling controlled analysis of model behavior across different creativity regimes.

\begin{figure}[!t]
    \centering
    \vspace{-3mm}
    \includegraphics[width=\linewidth]{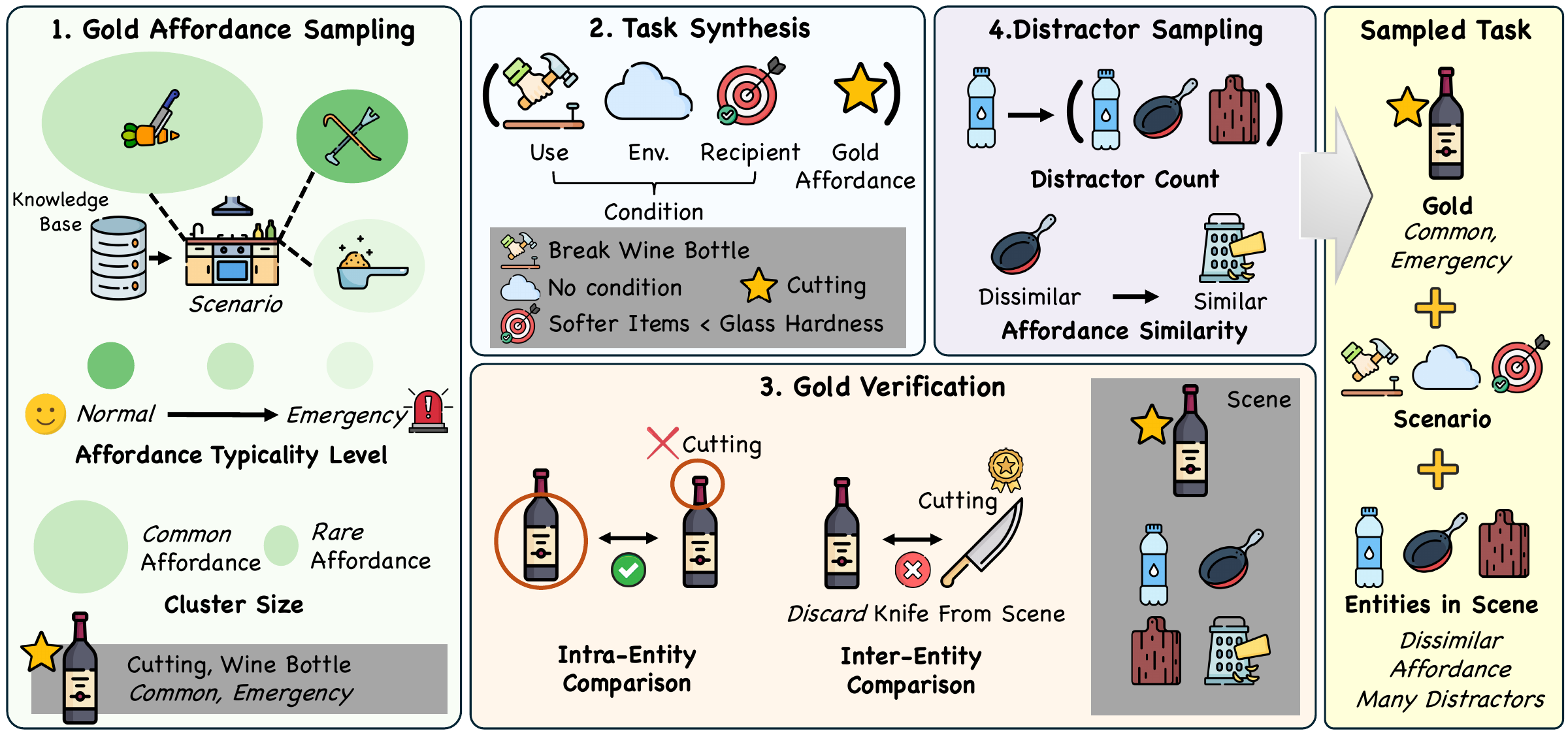}
    \caption{The pipeline for CreativityBench task sampling.}
    \label{fig:task_sampling}
    \vspace{-2mm}
\end{figure}

\paragraph{Task Synthesis.}
Given the sampled gold affordance $g=(e^*,p^*,f^*)$, we generate a natural-language scenario description $S$ whose completion requires executing $f^*$. The scenario is derived from the affordance annotation $(a,C_u,C_e,C_r)$ and potential recipients.

Concretely, the generated task describes a goal such that:
\[
S \Rightarrow (C_u \land C_e \land C_r) \land a.
\]
The scenario provides an environmental context in a grounded first-person narrative while withholding the affordance itself. The model must therefore infer the actionable affordance by reasoning over the attributes $A(p)$ of the entities present in the environment.

\paragraph{Gold Affordance Verification.}
Creative tasks may admit multiple plausible solutions. To ensure \textbf{solution uniqueness}, we must verify that the sampled gold affordance remains the most reasonable solution among all entities presented in the scene. We therefore perform two levels of affordance comparison:
\begin{itemize}[topsep=-4pt, leftmargin=10pt, itemsep=-2pt]
\item \textbf{Intra-entity Comparison:} For the gold entity $e^*$ and part $p^*$, we examine all other parts $p \in P(e^*)$. If some affordance $\tilde{f} \in F(p)$ is strictly preferable to $f^*$ for achieving the task goal, the candidate gold affordance is rejected and resampled.
\item \textbf{Inter-entity Comparison:} For candidate entities $\tilde{e} \in \mathcal{E}_T$, we compare $f^*$ with the affordances enabled by each part $p \in P(\tilde{e})$. If any affordance $\tilde{f} \in F(p)$ is judged superior to $f^*$ for the task, then $\tilde{e}$ is excluded from distractor sampling.
\end{itemize}
Each comparison evaluates whether the attributes $A(p)$ can support an alternative affordance $\tilde{f}$ satisfying the task goal, and whether $\tilde{f}$ is preferable to $f^*$ along four dimensions: (1) accessibility, (2) safety and consequences, (3) practical willingness to use it, and (4) typicality or commonness. These judgments are performed by an external LLM given the relevant attributes and task conditions. Candidates that produce superior or ambiguous alternatives are discarded. This ensures that the final gold affordance remains the most plausible among all the entities presented in the scene.

\paragraph{Distractor Sampling.}
After gold verification, we sample the remaining entities $\mathcal{E}_T \setminus \{e^*\}$ as \textit{distractors}. Distractors are designed to encourage models to reason at the attribute level rather than relying on superficial heuristics. Two factors control distractor difficulty:
\begin{itemize}[topsep=-4pt, leftmargin=10pt, itemsep=-2pt]
\item \textbf{Distractor Count:} the number of distractor entities present in the scene.
\item \textbf{Affordance Similarity:} whether distractors are semantically close to the gold affordance cluster (potentially confusing), far from it, or balanced.
\end{itemize}
By sampling valid distractors from different regimes, we obtain tasks that vary in reasoning difficulty, which also enables controlled analysis of model behavior in later analysis sections.

\paragraph{Statistics of CreativityBench.}
We use GPT-5.2 to scale the full task generation pipeline. The resulting dataset contains approximately 14K tasks spanning diverse scenarios, affordance types, and distractor configurations. Across all sampling hyperparameters, including cluster size, affordance typicality level, distractor count, and affordance similarity, the dataset is approximately balanced to enable systematic evaluation of model creativity. Further details of the sampling procedure are provided in \Cref{appendix_sec:sampling}.

To ensure solution correctness and evaluation rigor, we enforce a two-stage verification mechanism during task construction. First, an intra-entity self-check verifies that no other part of the gold entity provides a strictly preferable affordance for the task. Second, an inter-entity comparison filters candidate distractors whose affordances could overturn the gold solution or introduce high ambiguity.
In addition, similarity-controlled distractor sampling (using affordance clusters as a heuristic) ensures tasks remain creativity-relevant rather than trivially separable. Together, these mechanisms yield a benchmark that is both \emph{challenging} and \emph{scorable with explicit evidence}.

\section{Experiments}
\label{sec:experiment}

\subsection{Experimental Settings}

\paragraph{Models.}
We evaluate a diverse set of open- and closed-source models, including the GPT family~\citep{singh2025openai}, Gemini family~\citep{comanici2025gemini}, Qwen family~\citep{yang2025qwen3}, as well as Llama~\citep{grattafiori2024llama} and Mistral models~\citep{liu2026ministral}. Later analysis primarily focuses on GPT and Qwen families.

\paragraph{Evaluation Setup.}
All models are evaluated on the full set of 14K tasks. In the main setting, each model is given the task description together with all entities present in the scene, including the attributes of each part. The model must identify the relevant entity and part, filter distractors, and specify how the selected entity and part should be used to accomplish the task goal. Additional evaluation settings are analyzed in later sections. More details and inference hyperparameters are provided in \Cref{appendix_sec:exp_details}.

\paragraph{Metrics.}
In the main table, we report two objective tool usage metrics: \textbf{Gold Correct Rate}, a percentage of cases where both the correct entity and the correct part are selected; and \textbf{Entity Correct Rate}, a percentage of cases where the correct entity is selected, regardless of the part. By definition, the gold correct rate should always be less than or equal to the entity correct rate. These two metrics evaluate whether the model selects the correct tool target. 

For cases where the model selects the correct tool(i.e., gold correct), we further conduct an LLM-as-Judge evaluation to assess the quality of the predicted usage. Following \Cref{sec:preliminaries}, we evaluate four dimensions: constraint coverage, physical grounding, action feasibility, and overall prediction correctness. Given that gold affordances involve multiple constraints, as discussed in \Cref{sec:method}, we further decompose constraint coverage into three categories: use condition ($C_u$), environment condition ($C_e$), and recipient condition ($C_r$), and examine whether each constraint is explicitly stated or implicitly reflected in the predicted action. Physical grounding measures whether the generated solution is grounded in the physical and state attributes of the selected part, while action feasibility evaluates whether the proposed usage is plausible and executable under the given constraints. Tool Usage metrics are evaluated objectively as binary outcomes (correct: 1 or incorrect: 0), whereas the four additional dimensions are assessed using LLM-as-Judge on a 1-5 Likert scale.

We use Gemini-3.1-Flash-Lite as the judge model for two reasons: first, the judging task mainly requires reliable instruction following, since the evaluation criteria, gold references, and model predictions are all explicitly provided; second, given the large number of instances to be judged, this choice offers a practical balance between cost and capability. Further details are provided in \Cref{appendix_sec:exp_details}.

\begin{table*}[t]
    \centering
    \vspace{-3mm}
    \setlength\tabcolsep{2pt}
    \setlength\extrarowheight{2pt}

    \resizebox{\linewidth}{!}{
    \begin{tabular}{l @{\hspace{5mm}} c @{\hspace{5mm}} c @{\hspace{5mm}} c @{\hspace{5mm}} c @{\hspace{5mm}} c @{\hspace{5mm}} c @{\hspace{5mm}} c @{\hspace{5mm}} c}
        \toprule
        \multirow{2}{*}{\textbf{Model}} & \multicolumn{2}{c}{\textbf{Tool Usage}} & \multicolumn{3}{c}{\textbf{Constraint Coverage}} & \multirow{2}{*}{\textbf{\makecell{Physical\\Grounding}}} & \multirow{2}{*}{\textbf{\makecell{Action\\Feasibility}}} & \multirow{2}{*}{\textbf{\makecell{Prediction\\Correctness}}} \\

        \cmidrule(r){2-3} \cmidrule(r){4-6}
        
        ~ & \textbf{\makecell{Gold Correct}} & \textbf{\makecell{Entity Correct}} & \textbf{\makecell{Use ($C_u$)}} & \textbf{\makecell{Env. ($C_e$)}} & \textbf{\makecell{Rcpt. ($C_r$)}} & ~ & ~ & ~ \\
        
        \addlinespace[2pt]
        \midrule
        \addlinespace[2pt]
        \multicolumn{9}{c}{\textit{Closed-Source Models}} \\ 
        \midrule
        
        GPT-5.2 & 0.1819 & 0.5210 & \underline{3.8452} & \textbf{4.2428} & \underline{4.1458} & \underline{3.8734} & \textbf{4.3867} & \underline{3.8696} \\
        GPT-5 Mini & 0.1687 & 0.4856 & \textbf{3.8919} & \underline{4.1947} & \textbf{4.1566} & \textbf{3.8847} & \underline{4.2607} & \textbf{3.8704} \\
        GPT-5 Nano & 0.1192 & 0.5721 & 2.7535 & 2.9751 & 3.0903 & 2.8513 & 3.2628 & 2.8230 \\
        Gemini-2.5-Pro & 0.1670 & 0.3552 & 3.3668 & 2.9963 & 2.6502 & 3.2735 & 3.6728 & 3.0495 \\
        Gemini-2.5-Flash & 0.1532 & 0.3694 & 3.7505 & 3.8405 & 3.2353 & 3.6200 & 3.9844 & 3.4606 \\

        \addlinespace[2pt]
        \midrule
        \addlinespace[2pt]
        \multicolumn{9}{c}{\textit{Open-Source Models}} \\ 
        \midrule
        
        Qwen3-32B & \textbf{0.2588} & \textbf{0.6246} & 2.8548 & 2.8187 & 2.3335 & 2.9670 & 3.2323 & 2.6451 \\
        Qwen3-14B & \underline{0.2483} & \underline{0.6141} & 2.9569 & 2.9557 & 2.2771 & 3.1194 & 3.4138 & 2.7441 \\
        Qwen3-4B & 0.1882 & 0.5277 & 2.5943 & 2.4151 & 1.8961 & 2.8101 & 2.9881 & 2.4838 \\
        Llama-3-70B & 0.2151 & 0.5532 & 2.7487 & 2.6138 & 1.9710 & 2.7454 & 3.4761 & 2.7180 \\
        Ministral-3-14B & 0.2091 & 0.5263 & 3.0172 & 2.8073 & 2.2699 & 2.8581 & 3.1819 & 2.6627 \\
        \midrule
        Average & 0.1910 & 0.5149 & 3.1780 & 3.1860 & 2.8026 & 3.2003 & 3.5860 & 3.0327 \\
        \bottomrule
        
    \end{tabular}
    }
    \caption{Evaluation summary across models on the judged outputs. Best and second-best results are highlighted in \textbf{bold} and \underline{underline} respectively. For all metrics other than Tool Usage, the maximum score is 5.0. Higher is better for all metrics.}
    \label{tab:result_main}
    \vspace{-2mm}
\end{table*}




\subsection{Main Results}

\paragraph{Exact grounded tool use remains challenging.}
Although many models can moderately identify the correct entity, correctly grounding tool use at the part level remains challenging. Across models, entity correctness reaches 0.5149, while gold correctness drops to 0.1910, representing a relative decrease of over 60\%. For some models, the gap reaches 0.35. Since our benchmark requires part-level attribute grounding to justify creative affordances, this gap suggests that models can often recognize which object might be useful but fail to identify the specific part that enables the intended affordance. Consequently, many predicted tool usages are plausible at a high level but lack the attribute-level grounding required for correct affordance discovery.
\paragraph{Reasonable actions do not imply physically grounded reasoning.}
Across models, the average score for action feasibility (3.5860) is noticeably higher than that for physical grounding (3.2003). This pattern indicates that models tend to propose actions that are intuitively plausible from commonsense reasoning, often relying on semantic alignment between the tool and the task, while the underlying justification based on physical attributes is incomplete or inaccurate. As a result, even seemingly reasonable actions may fail when evaluated against the detailed physical conditions required for successful execution, which is reflected in the lower prediction correctness scores (3.0327). A similar pattern appears in constraint coverage. Use constraints ($C_u$) and environmental constraints ($C_e$) are relatively well considered (3.1780 and 3.1860 on average), whereas recipient constraints ($C_r$) receive the lowest score (2.8026). This indicates that models frequently overlook conditions related to the target object or preparation steps needed before tool application, leading to incomplete reasoning pipelines even when the correct tool is selected.

\begin{takeaway}
From the metric perspective, current models are better at choosing plausible tools at the entity level but struggle to produce fully correct, physically grounded tool use. Model reasoning is often plausible but fails to account for all conditions required for successful execution.
\end{takeaway}

\paragraph{Strong general models do not necessarily excel at creative tool use.}
We observe a notable contrast between general reasoning performance and creative tool-use ability. GPT-5.2 and GPT-5-Mini achieve the strongest performance on most reasoning-related metrics, including environmental constraint coverage, action feasibility, and overall prediction correctness. However, their gold correctness in tool usage (0.1819) is substantially lower than that of the Qwen3 family, particularly Qwen3-32B (0.2588) and Qwen3-14B (0.2483). In fact, Qwen3-32B achieves nearly 1.5$\times$ the gold correctness of GPT-5.2 while also obtaining the highest entity correctness (0.6246). This contrast suggests that some models are better at discovering unconventional yet valid tool usages, while others are stronger at evaluating constraints and producing grounded reasoning once the correct direction is identified. In other words, models that excel at systematic reasoning are not necessarily the ones that most effectively discover novel affordances.

\paragraph{Performance quickly saturates with model scaling.}
Within the same model family, scaling improves performance primarily at smaller model sizes but yields diminishing returns beyond a certain point. For example, in the Qwen family, gold correctness improves nearly 30\% when scaling from 4B (0.1882) to 14B (0.2483), but increases by less than 5\% when further scaling to 32B (0.2588). A similar trend appears in proprietary models: scaling GPT-5 Nano to GPT-5 Mini yields roughly a 40\% relative improvement in gold correctness (0.1192 $\rightarrow$ 0.1687), whereas the gain from GPT-5 Mini to GPT-5.2 is only about 7\%. Gemini models show a comparable pattern with modest gains of about 9\%. These results indicate that creative tool use grounded in physical reasoning does not scale linearly with model size. Instead, it appears to depend on more fundamental capabilities related to affordance discovery and structured physical reasoning. This observation also highlights the value of our benchmark: the task cannot be easily solved through brute-force scaling alone and instead requires deeper improvements in creativity as well as grounded reasoning abilities.

\begin{takeaway}
From the model perspective, there is a clear split between creative affordance discovery and grounded reasoning quality. Improvements from model scaling alone quickly saturate, indicating that stronger performance requires advances in grounded affordance reasoning rather than size alone.
\end{takeaway}

\section{Analysis}
\label{sec:analysis}

In this section, we analyze factors that influence model performance from three complementary perspectives. As described in \Cref{sec:method}, our task construction and sampling process is highly structured and balanced, enabling controlled analysis along several dimensions:
\begin{itemize}[topsep=-1.5pt, leftmargin=10pt, itemsep=0pt]
    \item \textbf{Gold Commonality (\Cref{sec:analysis_gold}):}  How properties of the gold solution affect model performance, including affordance typicality level and the cluster size from which it is drawn.
    \item \textbf{Distraction Severity (\Cref{sec:analysis_distractor}):} How distractors and environmental noise influence performance, focusing on both distractor counts and their affordance similarity to the gold solution.
    \item \textbf{Inference-Time Effects (\Cref{sec:analysis_infer}):} How inference-time choices, including sampling temperature and evaluation mode, affect model performance.
\end{itemize}

Unless otherwise specified, we analyze the \textbf{Tool Usage} metrics, which provide the most objective evaluation signal and serve as the primary benchmark metrics. The first two analyses use the same setting as the main experiment; for inference-time analysis, we further evaluate different sampling temperatures and evaluation modes. Finally, we examine subjective metrics and highlight several noteworthy patterns and failure modes.

\begin{figure}[t]
    \centering
    \vspace{-3mm}
    \begin{subfigure}[t]{0.49\linewidth}
        \centering
        \includegraphics[width=\linewidth]{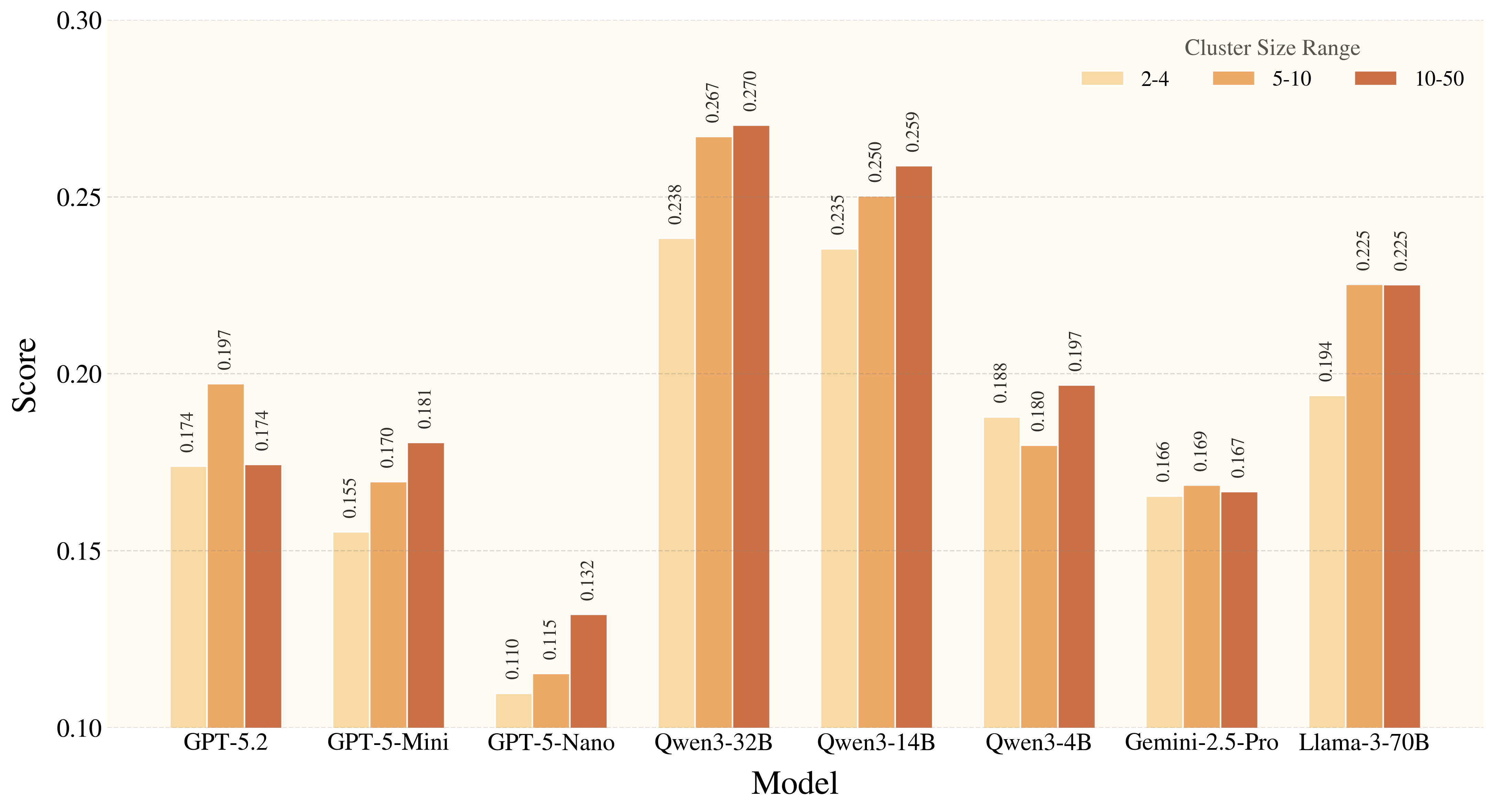}
        \caption{Gold cluster size}
        \label{fig:analysis_cluster}
    \end{subfigure}
    \hfill
    \begin{subfigure}[t]{0.49\linewidth}
        \centering
        \includegraphics[width=\linewidth]{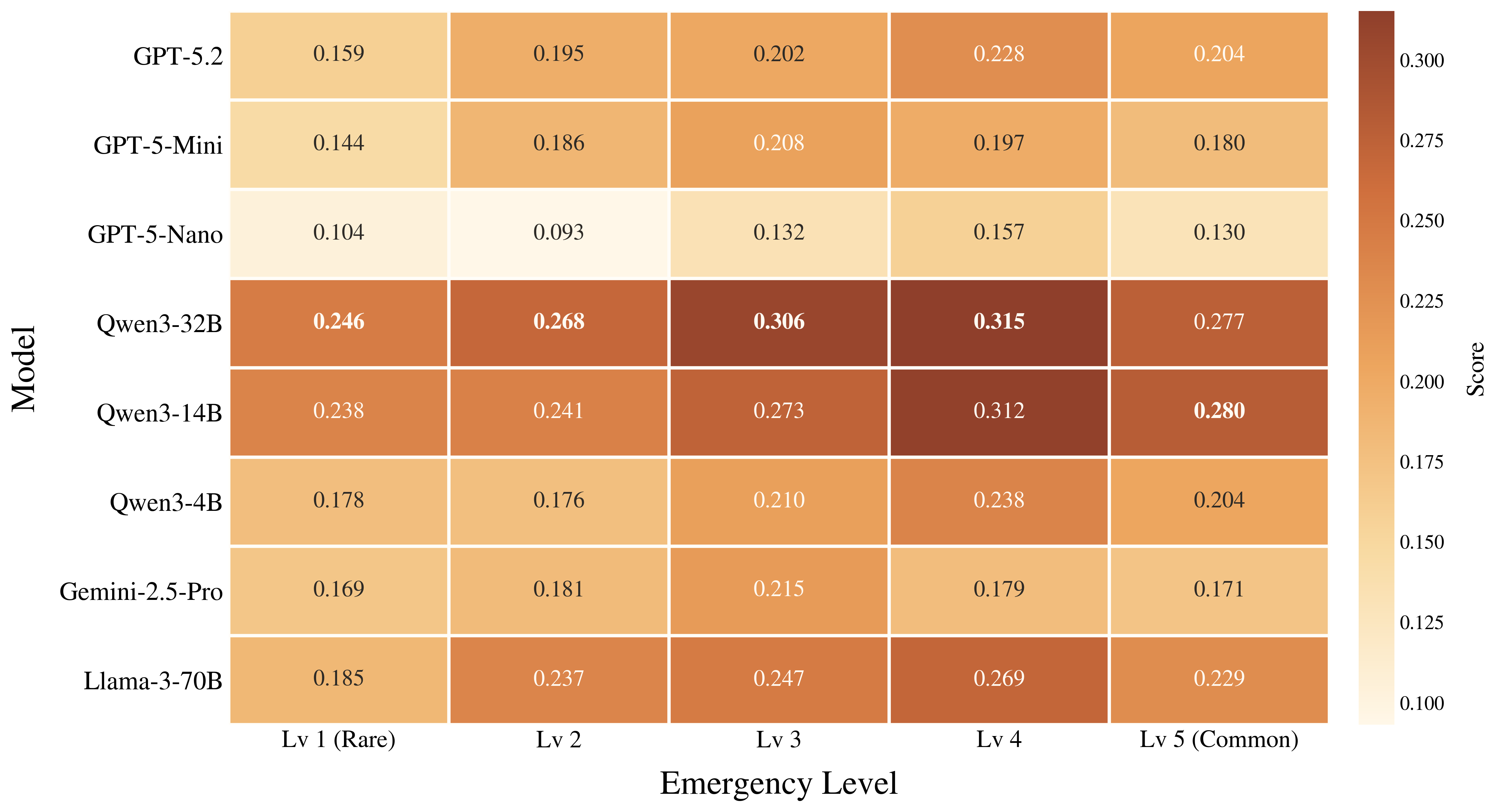}
        \caption{Gold affordance level}
        \label{fig:analysis_level}
    \end{subfigure}
    \caption{\textbf{Gold commonality affects tool-use performance.} Models perform better when the gold affordance comes from a larger cluster or has a higher affordance level, indicating that more common and natural repurposed uses are easier to solve.}
    \label{fig:analysis_cluster_level}
    \vspace{-2mm}
\end{figure}

\subsection{How does Gold Commonality Affect Performance?}
\label{sec:analysis_gold}
To understand how the inherent commonality of the gold solution influences model performance, we analyze tasks along two dimensions. First, we group tasks by the \textbf{size of the gold cluster} from which the solution is sampled, using three ranges: [2, 4], [5, 10], and [10, 50]. A larger cluster generally indicates that the underlying affordance appears more frequently in the annotated data and is therefore more common, even if it is still unconventional relative to the object's default use. Second, we analyze the \textbf{emergency level} of the gold solution, which captures how natural or practically acceptable an affordance is in real-world scenarios; as such, it also serves as a proxy for the commonality and naturalness of the sampled affordance.

\paragraph{More common affordances are consistently easier for current models.}
We report the results in \Cref{fig:analysis_cluster_level}. For gold cluster sizes, performance increases steadily with gold cluster size (\Cref{fig:analysis_cluster}). Across models, tasks drawn from small clusters (size 2--4) produce the lowest scores, while those drawn from larger clusters are significantly easier. The same pattern appears when grouping by emergency level (\Cref{fig:analysis_level}). Across nearly all models, affordances at levels 1--2 are associated with noticeably lower performance than those at levels 3--5. This result reinforces the same conclusion from a different angle: current models handle relatively common repurposed affordances much better than rare or highly atypical ones.

Together, these results show that model performance is strongly shaped by how familiar or natural the target affordance is. This is consistent with the broader intuition that current models remain weak on long-tail or unusual uses of objects, even when they can perform reasonably well on more common forms of creative repurposing. At the same time, the consistency of this trend also provides indirect support for the validity of our annotation design: the commonality-related labels appear to align well with both human intuition and observed model difficulty.

\begin{takeaway}
Model performance drops most clearly on rare, long-tail affordances, showing that commonality remains a major driver of success in grounded creative tool use.
\end{takeaway}

\begin{figure}[t]
    \centering
    \vspace{-3mm}
    \begin{subfigure}[t]{0.45\linewidth}
        \centering
        \includegraphics[width=\linewidth]{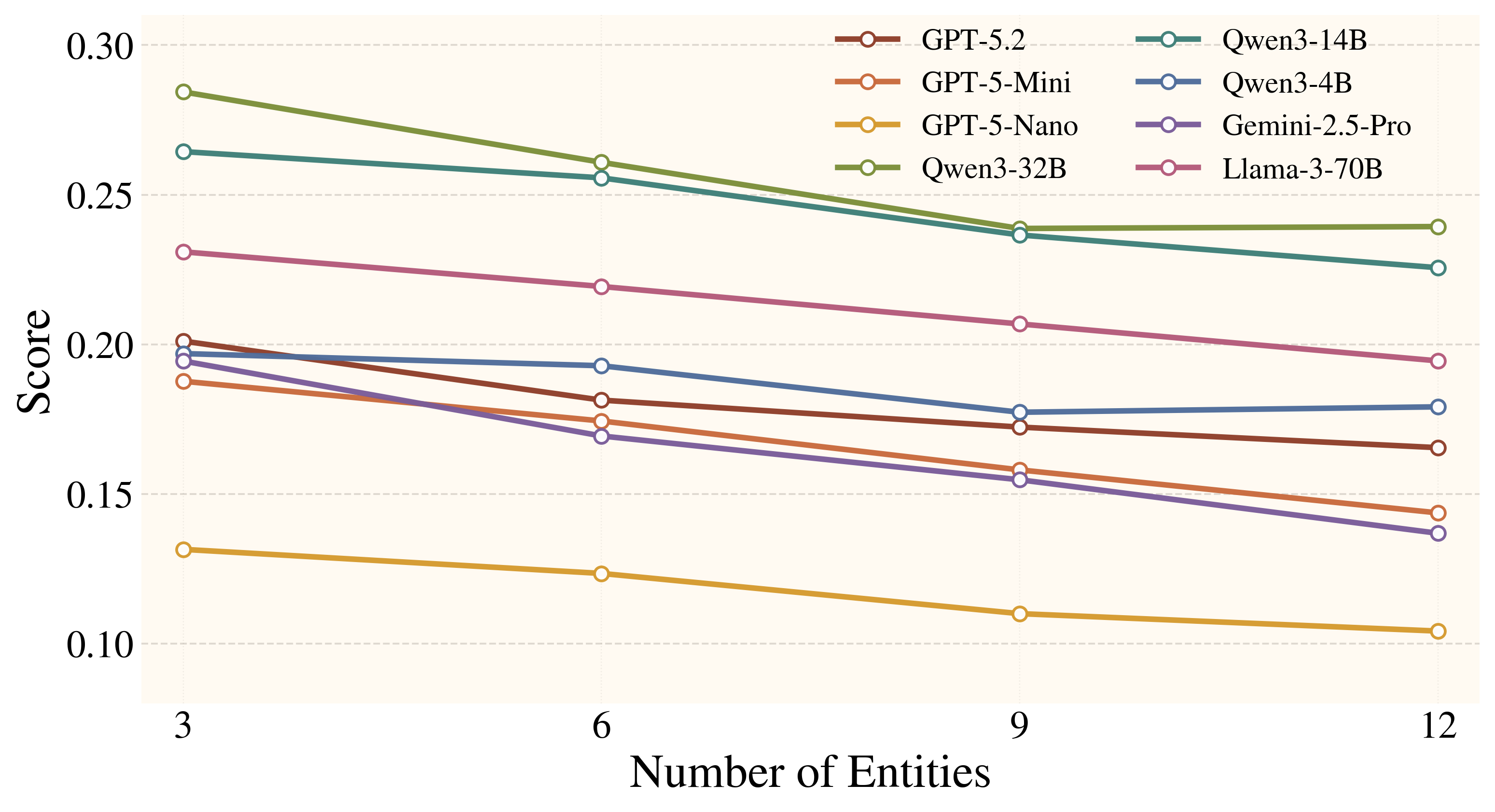}
        \caption{Distractor number}
        \label{fig:analysis_number}
    \end{subfigure}
    \hfill
    \begin{subfigure}[t]{0.54\linewidth}
        \centering
        \includegraphics[width=\linewidth]{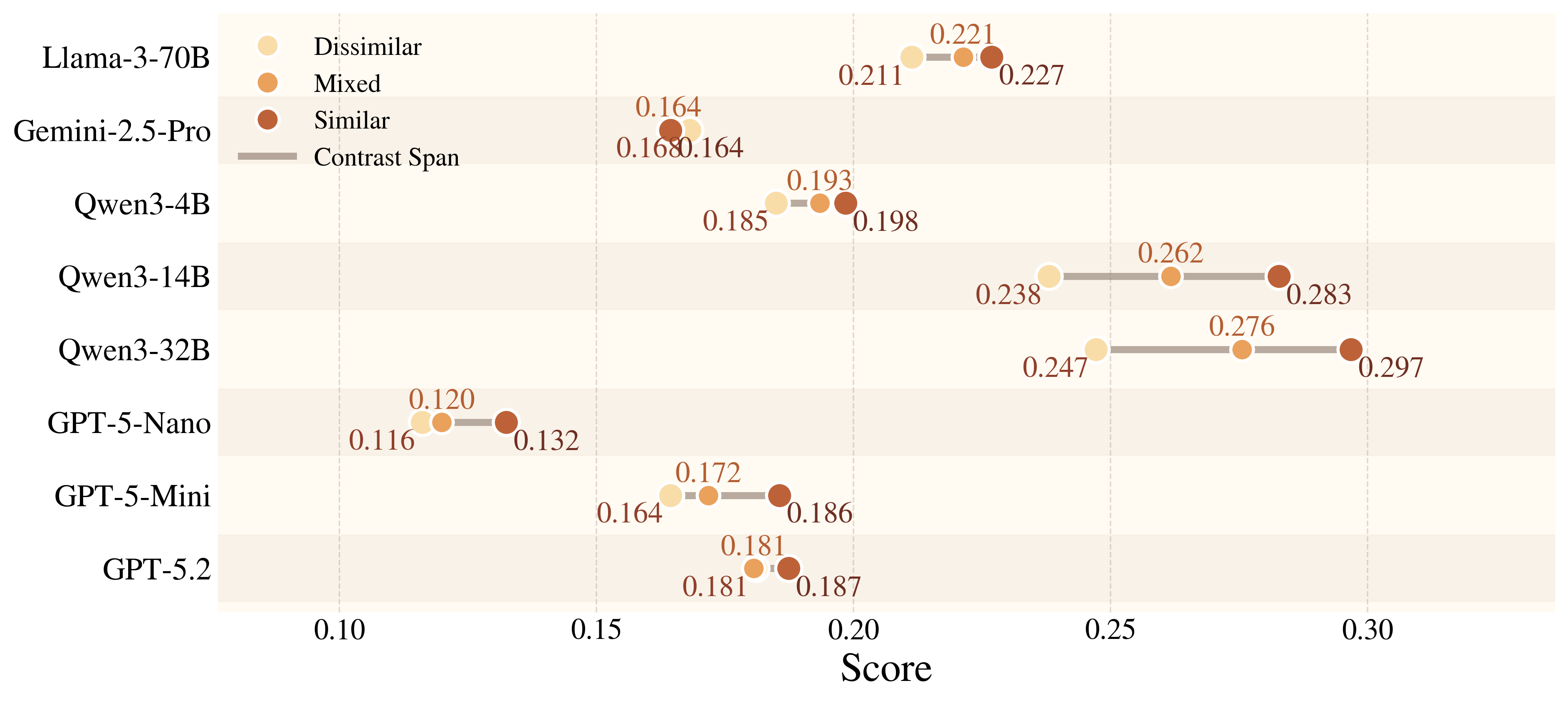}
        \caption{Distractor similarity}
        \label{fig:analysis_similarity}
    \end{subfigure}
    \caption{\textbf{Distraction severity shapes model performance.} Increasing the number of distractor entities consistently reduces accuracy. However, distractors with affordances similar to the gold object often lead to higher performance, suggesting that they may implicitly cue the relevant affordance rather than purely increasing confusion.}
    \label{fig:analysis_number_similarity}
    \vspace{-2mm}
\end{figure}

\begin{wrapfigure}{r}{0.55\linewidth}
    \centering
    \vspace{-3mm}
    \includegraphics[width=\linewidth]{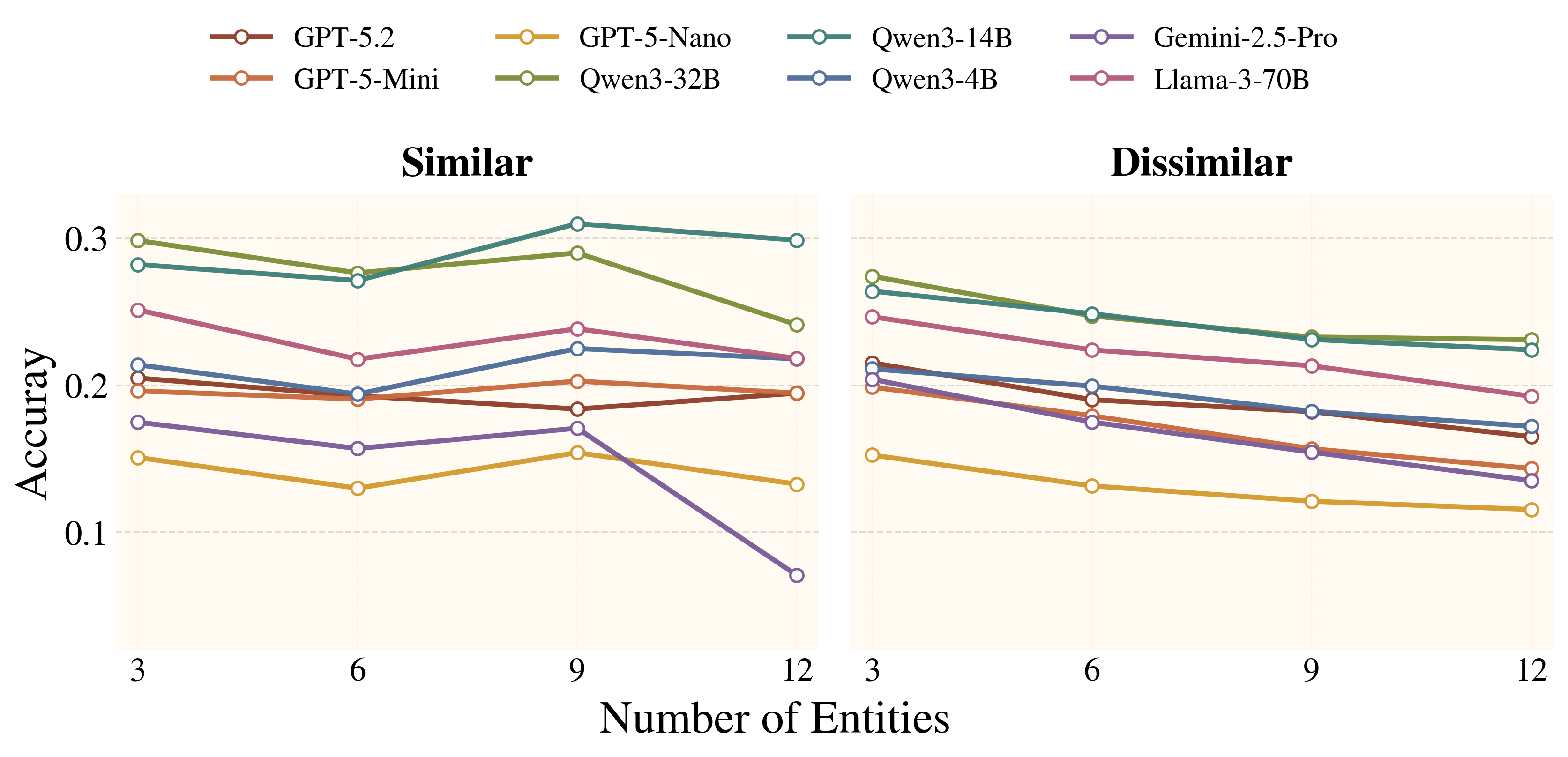}
    \caption{Fine-grained trends across distractor types.}
    \label{fig:analysis_number_similarity_finegrained}
    \vspace{-2mm}
\end{wrapfigure}

\subsection{How does Distraction Severity Affect Performance?}
\label{sec:analysis_distractor}
Beyond the presence of the gold entity itself, distractor objects in the environment introduce additional cognitive load for the tested models. We analyze this effect from two complementary perspectives. First, we examine the \textbf{number of distractors} presented in each task. We group tasks by the total number of distractor entities (3, 6, 9, and 12). A larger set of entities increases the length of the context and expands the space of candidate tools the model must evaluate. Second, we analyze the \textbf{affordance similarity} between distractors and the gold object. During dataset construction (\Cref{sec:method}), distractors were categorized based on whether they could elicit attributes similar to the gold affordance during verification. We therefore divide distractors into two categories: \textit{similar} distractors, which share related affordance attributes with the gold object but are ultimately inferior solutions, and \textit{dissimilar} distractors, which do not naturally support the target affordance. Based on these labels, we group tasks into three settings: all-similar distractors, all-dissimilar distractors, and mixed distractors.

\paragraph{More distractor entities increase reasoning difficulty.}
We report the results in \Cref{fig:analysis_number_similarity}. The left panel shows performance as the number of distractors increases. Across all tested models, performance consistently decreases as more entities are introduced. On average, the drop is more noticeable when moving from 3 to 9 distractors, while the decline from 9 to 12 becomes somewhat less steep, suggesting a potential diminishing effect once the candidate set becomes large. Overall, the results indicate that increasing the number of entities makes it harder for models to identify the correct tool, likely because the model must allocate attention across more competing options.

\paragraph{Similar affordances can implicitly guide model reasoning.}
The right panel of \Cref{fig:analysis_number_similarity} examines the role of distractor similarity. Intuitively, one might expect distractors with similar affordances to increase task difficulty, since they require finer-grained comparison between candidate objects. However, the results reveal the opposite pattern: tasks containing similar-affordance distractors consistently achieve higher scores than those containing dissimilar distractors. A possible explanation is that similar distractors implicitly highlight the relevant affordance space. When multiple entities share related attributes, the intended affordance becomes more salient, making it easier for the model to activate the correct reasoning pathway.

To further investigate this effect, we perform a fine-grained analysis of distractor number within the similar and dissimilar groups separately. As shown in \Cref{fig:analysis_number_similarity_finegrained}, for tasks with dissimilar distractors, performance steadily decreases as the number of entities grows, consistent with the expected effect of increased distraction. In contrast, for tasks with similar distractors, the trend is less monotonic: performance initially decreases but then partially recovers when the number of entities reaches 9. This pattern suggests that while more entities increase the overall reasoning burden, the presence of multiple affordance-related objects may also reinforce the relevant affordance concept and help guide the model toward the correct solution.

\begin{takeaway}
Both the quantity and type of distractors influence model performance. While larger candidate sets consistently increase difficulty, distractors that share similar affordances with the gold object can sometimes help highlight the relevant reasoning space, partially offsetting the distraction effect.
\end{takeaway}

\begin{figure}[t]
    \centering
    \vspace{-3mm}
    \begin{minipage}{0.48\linewidth}
        \centering
        \includegraphics[width=\linewidth]{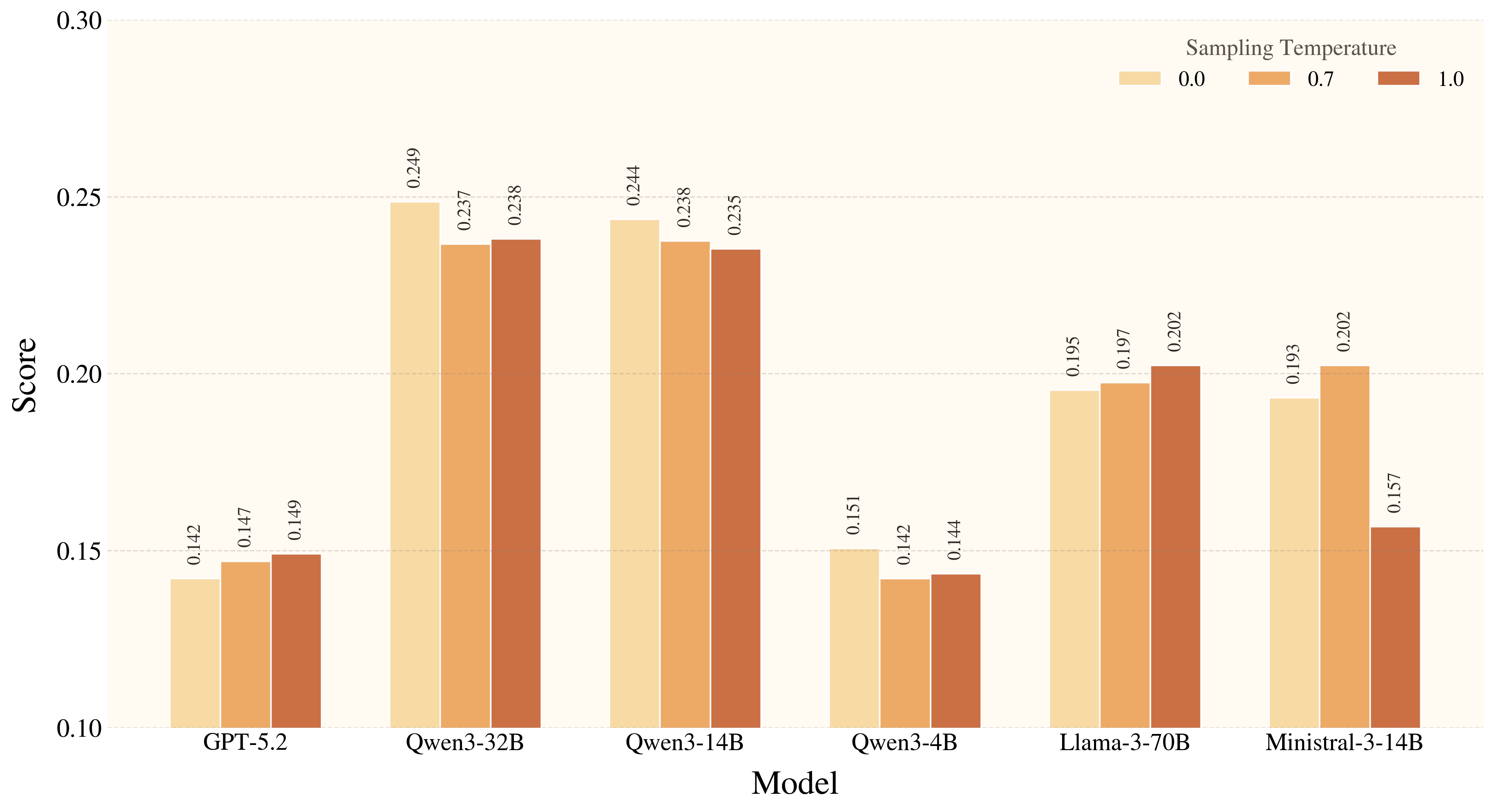}
        \caption*{(a) Inference temperature}
        \label{fig:analysis_temperature}
    \end{minipage}
    \hfill
    \begin{minipage}{0.5\linewidth}
        \centering
        \vspace{2.5mm}
        \resizebox{\linewidth}{!}{
            \begin{tabular}{lccc}
            \toprule
            Model & Static & Interactive & CoT \\
            \midrule
            GPT-5.2 & 0.142 & 0.083 {\color{red!50!black}$_{-0.059}$} & 0.140 {\color{red!50!black}$_{-0.002}$} \\
            GPT-5-mini & 0.140 & 0.055 {\color{red!50!black}$_{-0.085}$} & 0.128 {\color{red!50!black}$_{-0.012}$} \\
            GPT-5-nano & 0.120 & 0.039 {\color{red!50!black}$_{-0.081}$} & 0.111 {\color{red!50!black}$_{-0.009}$} \\
            Qwen3-32B & 0.249 & 0.070 {\color{red!50!black}$_{-0.179}$} & 0.259 {\color{green!50!black}$_{+0.011}$} \\
            Qwen3-14B & 0.244 & 0.088 {\color{red!50!black}$_{-0.156}$} & 0.239 {\color{red!50!black}$_{-0.005}$} \\
            Qwen3-4B & 0.151 & 0.056 {\color{red!50!black}$_{-0.095}$} & 0.154 {\color{green!50!black}$_{+0.003}$} \\
            Llama-3-70B & 0.195 & 0.047 {\color{red!50!black}$_{-0.149}$} & 0.230 {\color{green!50!black}$_{+0.034}$} \\
            Ministral-3-14B & 0.193 & 0.064 {\color{red!50!black}$_{-0.130}$} & 0.232 {\color{green!50!black}$_{+0.039}$} \\
            \bottomrule
            \end{tabular}
        }
        \vspace{2.5mm}
        \caption*{(b) Inference mode}
        \label{fig:analysis_mode}
    \end{minipage}
    \caption{\textbf{Inference-time strategies provide limited gains for creative tool use.} Increasing sampling temperature generally may reduce performance, as higher randomness encourages hallucinated entity or part names rather than productive exploration. Further, compared with the static evaluation setting, interactive evaluation substantially degrades performance across all models, while structured CoT yields only marginal improvements.}
    \label{fig:analysis_temperatur_mode}
    \vspace{-2mm}
\end{figure}

\subsection{How do Inference Settings Affect Performance?}
\label{sec:analysis_infer}

\begin{wraptable}{r}{0.55\linewidth}
    \centering
    \vspace{0mm}
    \setlength\tabcolsep{5pt}
    \setlength\extrarowheight{2pt}
    \resizebox{\linewidth}{!}{
    \begin{tabular}{lc @{\hspace{2mm}} ccc}
        \toprule
        \multirow{2}{*}{\textbf{Model}} & \multirow{2}{*}{\textbf{Turns}} & \multicolumn{3}{c}{\textbf{Gold Inspection Rate (\%)}} \\
        \cmidrule{3-5}
        & & Gold Correct & Entity Correct & Both Fail \\
        \midrule
        GPT-5.2 & 2.6 & 29.8\% & 16.2\% & 8.3\% \\
        GPT-5-mini & 3.3 & 54.3\% & 29.8\% & 16.7\% \\
        GPT-5-nano & 8.6 & 63.2\% & 43.8\% & 48.1\% \\
        Qwen3-32B & 2.4 & 56.2\% & 33.3\% & 10.2\% \\
        Qwen3-14B & 2.9 & 67.0\% & 35.9\% & 15.3\% \\
        Qwen3-4B & 3.2 & 66.9\% & 33.5\% & 16.4\% \\
        Llama-3-70B & 7.9 & 90.6\% & 65.9\% & 68.0\% \\
        Ministral-3-14B & 5.4 & 75.3\% & 45.1\% & 25.9\% \\
        \bottomrule
    \end{tabular}
    }
    \caption{\textbf{Interactive mode statistics} showing the average turns and gold inspection rate, defined as the percentage of tasks in which the gold entity is inspected before the model produces its final answer.}
    \label{tab:analysis_gold_inspection}
    \vspace{-4mm}
\end{wraptable}

Beyond model capability, inference-time strategies may also influence performance. We analyze two aspects of inference configuration: the \textbf{sampling strategy} and the \textbf{evaluation mode}. First, we examine whether increasing sampling temperature encourages more creative reasoning. While the main experiments use deterministic decoding, we additionally evaluate models with higher temperatures ($T=0.7$ and $T=1.0$) to test whether more diverse sampling leads to better tool discovery. Second, we explore alternative evaluation modes. In the main setting, all entity descriptions are presented at once in a static prompt. We compare this with the following two variants:
\begin{itemize}[topsep=-1.5pt, leftmargin=10pt, itemsep=0pt]
    \item \textbf{CoT mode.} The model is instructed to explicitly perform a structured reasoning process, including task analysis, attribute grounding, and affordance reasoning, before predicting the entity and usage strategy.
    \item \textbf{Interactive mode.} Instead of receiving all entity descriptions upfront, the model need to query the environment and request entity descriptions one at a time, turning the task completion into a multi-turn interaction trajectory.
\end{itemize}
We sample 10\% of the original data (1.4K) for the analysis of this section, and run multiple parallel experiments for comparison. All other settings remain identical to the main experiments. Please see \Cref{appendix_sec:analysis_details} for more details.

\paragraph{Higher temperature does not guarantee genuine creative reasoning.}
We report the results in \Cref{fig:analysis_temperatur_mode}. The left panel shows that increasing temperature generally degrades performance for smaller-scale models like the Qwen series, and mildly increases performance for larger-scale models like GPT-5.2.
Inspection of model outputs reveals that higher temperatures primarily increase hallucination rather than useful exploration. As the temperature rises, smaller models are more likely to generate entity or part names that do not exist in the environment. This suggests that creative tool use in our benchmark is fundamentally different from open-ended text generation: successful solutions require grounded reasoning over attributes and affordances, rather than simply producing diverse outputs.

These results also imply that naive inference scaling, such as repeatedly sampling at higher temperatures, may not effectively and consistently improve performance across models. Increased diversity alone does not help models discover the correct affordance relationships and instead may lead to ungrounded answers.

\begin{takeaway}
Increasing sampling temperature does not reliably improve creative tool use. For smaller models in particular, higher temperatures tend to increase hallucinated entities and parts rather than produce more effective affordance exploration, suggesting that this benchmark rewards grounded constraint satisfaction rather than open-ended generative diversity.
\end{takeaway}

\paragraph{Interactivity introduces additional reasoning challenges.}
The right panel of \Cref{fig:analysis_temperatur_mode} compares the static setting with the interactive and structured-CoT inference modes. Interactive evaluation substantially reduces performance across all models. Although this setting lowers the initial cognitive load (since entity descriptions are revealed only upon request rather than all at once), models exhibit limited exploration behavior. As shown in \Cref{tab:analysis_gold_inspection}, most models inspect fewer than three entities on average before producing a final prediction. \looseness=-1

This limited exploration makes early mistakes especially costly. Once a model forms an initial hypothesis about which object may solve the task, it seldom gathers enough additional evidence to revise that belief. The gold inspection statistics in \Cref{tab:analysis_gold_inspection} further support this pattern. When both the predicted entity and part are correct, the gold inspection rate is high, indicating that successful predictions are usually grounded in direct examination of the target object. In contrast, for completely incorrect predictions, the gold inspection rate falls below 20\% for most models, showing that the model often never inspects the gold entity before answering. Interestingly, even among fully correct cases, the gold inspection rate does not reach 100\%, suggesting that models sometimes succeed based on partial clues or rough heuristics rather than thorough exploration. Overall, these results indicate that insufficient exploration and premature commitment to early hypotheses are major sources of failure in the interactive setting.

\paragraph{Structured CoT introduces only minor performance changes.}
In contrast to the large performance drop in interactive mode, structured CoT reasoning results in only modest changes. Some models, especially those in the Qwen family, obtain small gains (typically within around 5\%), while GPT-family models even show slight declines. This observation is consistent with our earlier findings in \Cref{sec:preliminaries}: explicitly enforcing a structured reasoning process does not substantially improve affordance reasoning.

A possible explanation is that structured CoT changes the format of reasoning more than its substance. Once a model commits to a particular entity early in the reasoning process, the imposed structure may further reinforce that focus rather than encourage broader comparison across alternatives. In practice, models rarely revisit previously overlooked entities or perform the kind of back-and-forth comparison needed to identify the best tool. As a result, structured reasoning alone offers limited benefit when the core challenge is to explore the candidate space and ground the final decision in comparative affordance analysis. \looseness=-1

\begin{takeaway}
Changing the inference mode also yields limited gains. Interactive evaluation reveals weak exploration behavior and premature commitment to early hypotheses, while structured CoT provides only marginal improvement, indicating that the main bottleneck is not reasoning format but the ability to compare candidates and ground decisions in affordance-level physical reasoning.
\end{takeaway}

\subsection{How are auxiliary metrics affected?}

\begin{figure}[t]
    \centering
    \vspace{-3mm}
    \begin{subfigure}[t]{0.49\linewidth}
        \centering
        \caption{Physical grounding by gold cluster size}
        \includegraphics[width=\linewidth]{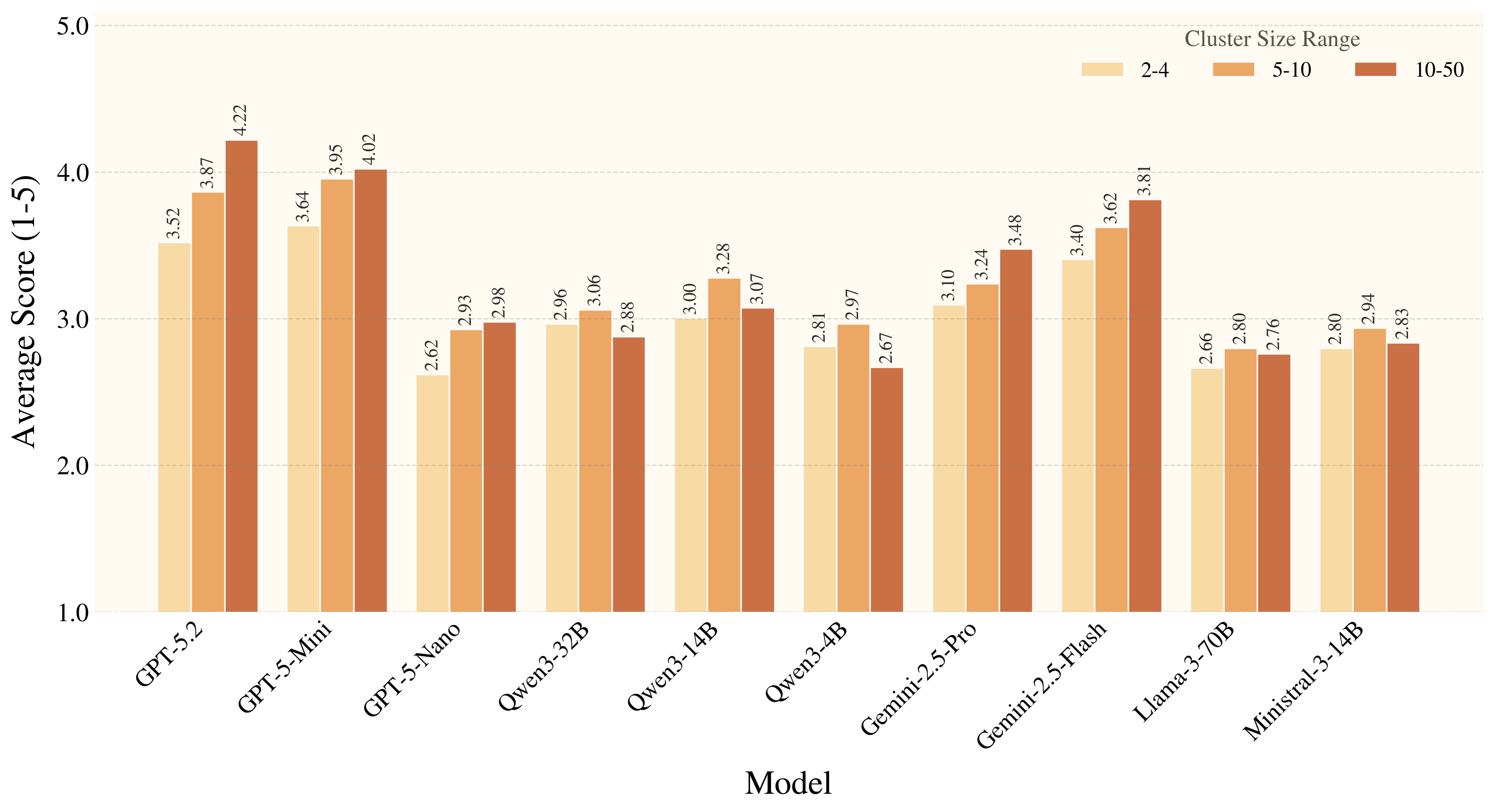}
        \label{fig:analysis_aux_1}
    \end{subfigure}
    \hfill
    \begin{subfigure}[t]{0.49\linewidth}
        \centering
        \caption{Physical grounding by distractor similarity}
        \includegraphics[width=\linewidth]{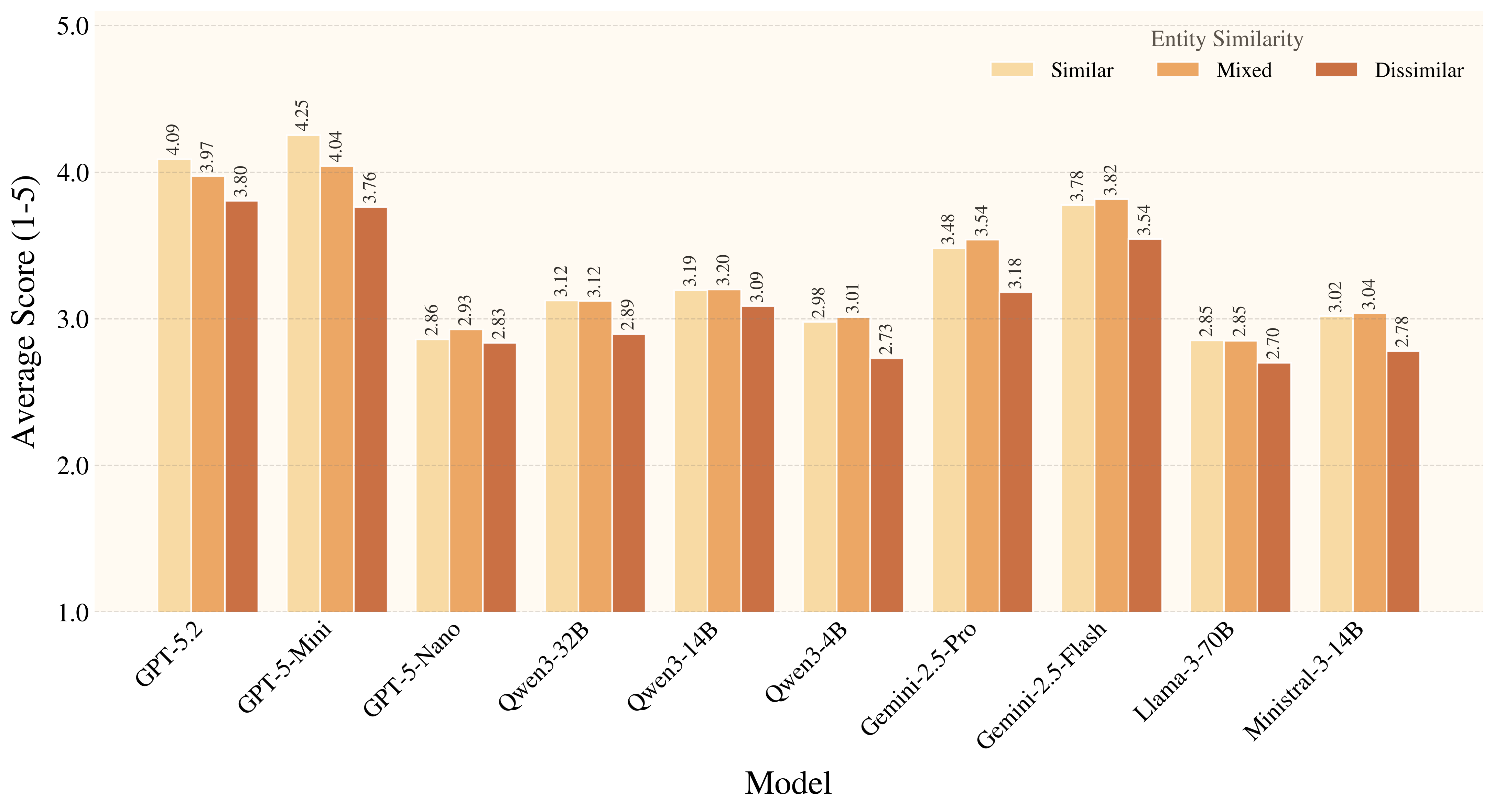}
        \label{fig:analysis_aux_2}
    \end{subfigure}
    \hfill
    \begin{subfigure}[t]{0.49\linewidth}
        \centering
        \caption{Use constraint coverage by gold cluster size}
        \includegraphics[width=\linewidth]{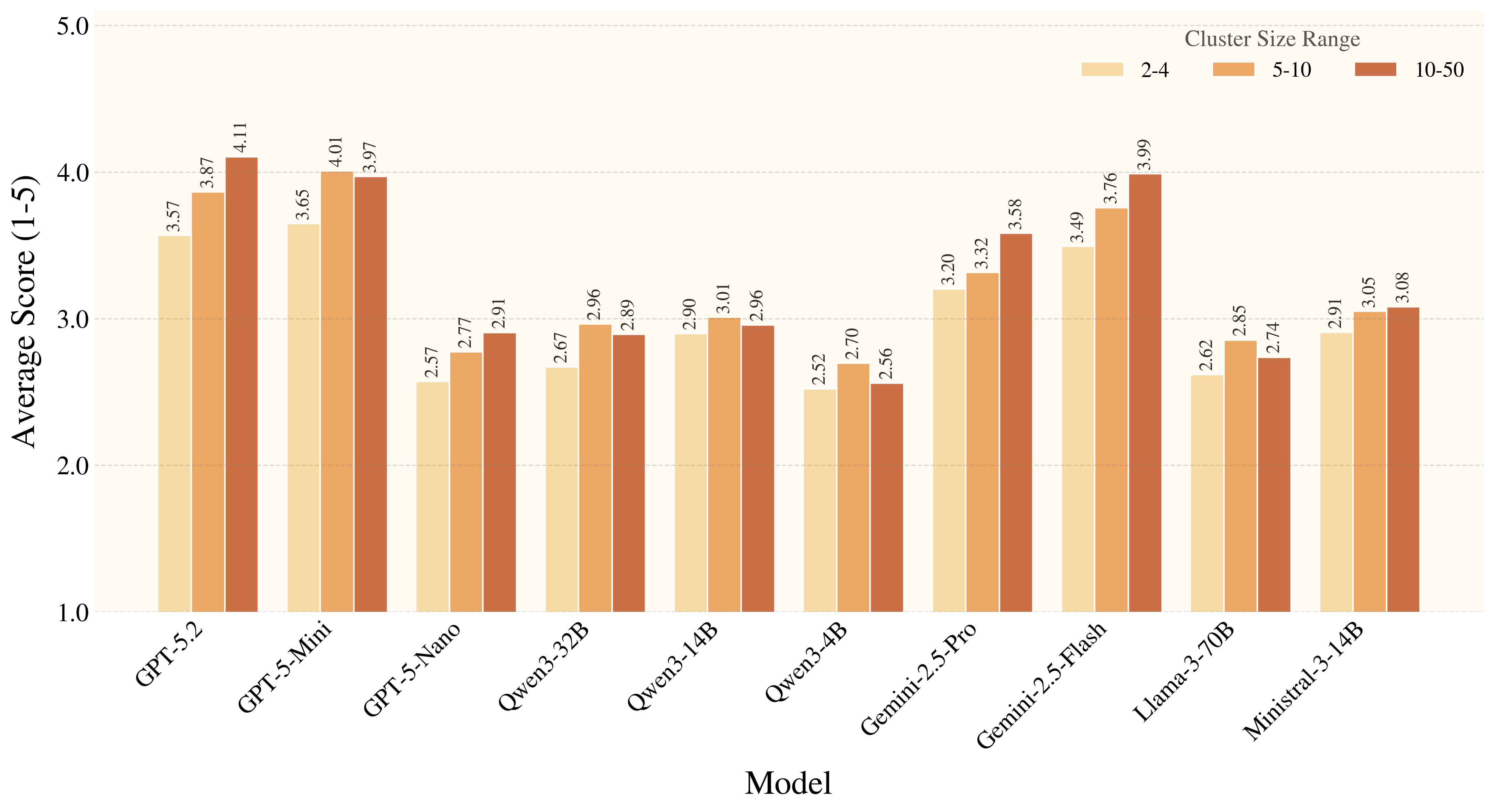}
        \label{fig:analysis_aux_3}
    \end{subfigure}
    \hfill
    \begin{subfigure}[t]{0.49\linewidth}
        \centering
        \caption{Prediction correctness by distractor similarity}
        \includegraphics[width=\linewidth]{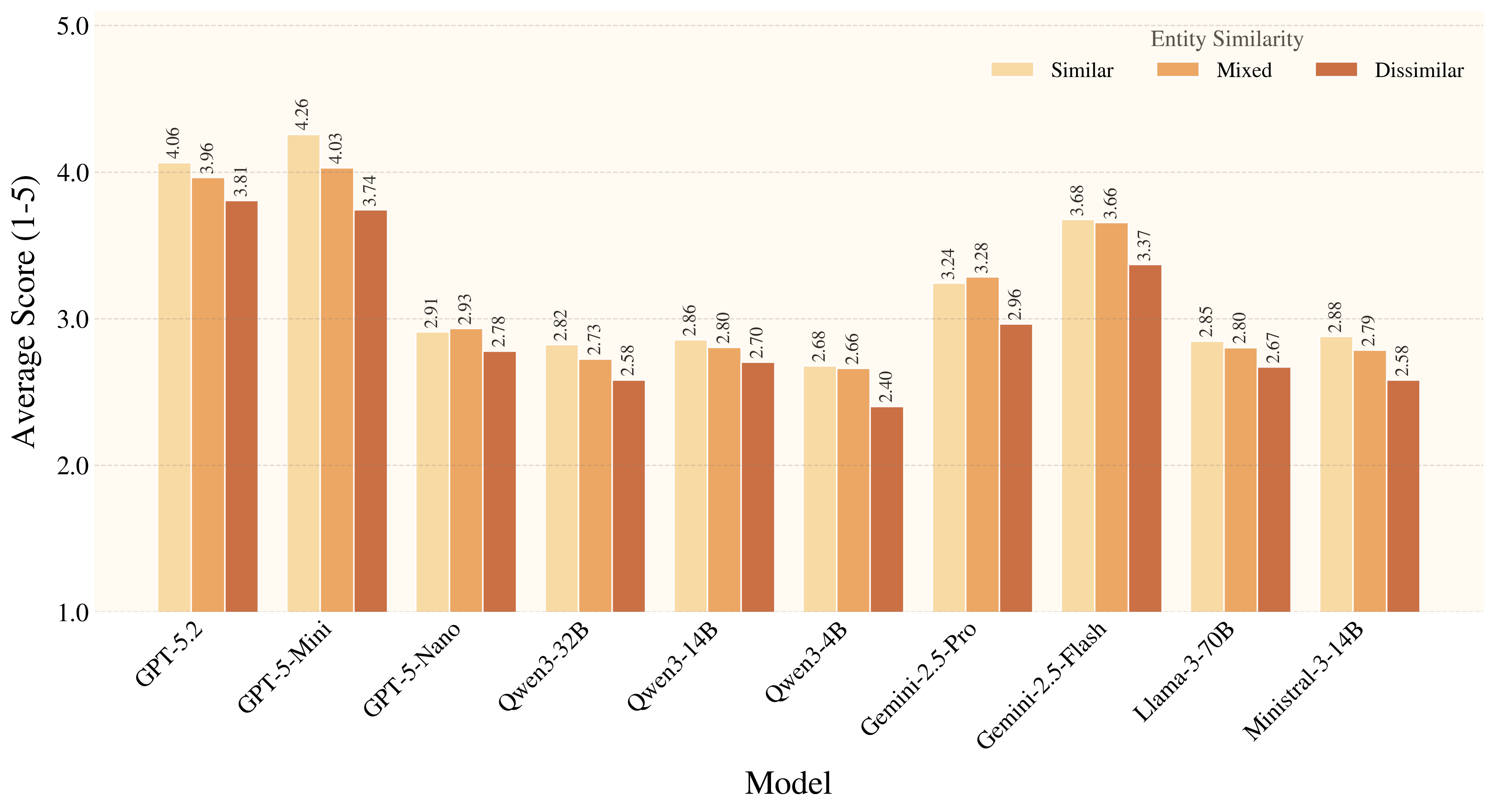}
        \label{fig:analysis_aux_4}
    \end{subfigure}
    \caption{Statistics showing how auxiliary metrics are affected from multiple perspectives.}
    \label{fig:analysis_aux}
    \vspace{-2mm}
\end{figure}

In this section, instead of focusing on the main Tool Usage metrics, we analyze auxiliary metrics evaluated by the LLM-as-judge. All analyses follow the same setup as the main table. Since we consider multiple factors, including gold commonality (i.e., gold cluster size, gold emergency level), distraction severity (i.e., distractor similarity, and distractor number), together with six auxiliary scores, we only discuss several representative patterns here and leave the full results and discussions to \Cref{appendix_sec:analysis_details}.

\paragraph{Physical grounding is harder for rare affordances and dissimilar distractors.}
We report four representative results in \Cref{fig:analysis_aux}. From panels (a) and (b), we observe that across all tested models, gold affordances from smaller clusters (i.e., rare affordances) consistently achieve the lowest performance. This trend also remains when distractors are all dissimilar, and is consistent with the main Tool Usage analysis results in the previous section.

This suggests that rare affordances are harder for models to ground physically in creative tool use. Since such affordances are less common, models may struggle to reason concretely about how the object can be used. Similarly, dissimilar distractors offer little affordance-level guidance, making it harder for models to identify and ground the effective use of the target part. \looseness=-1

\paragraph{Auxiliary metrics further show that similar distractors make the task easier.}
From panels (b) and (d), we see that similar-affordance distractors lead to higher physical grounding scores and higher prediction correctness. Although such distractors increase surface-level ambiguity, they also constrain the solution space to plausible uses, making the task easier for models.

\paragraph{Rare affordances are also harder to use under appropriate conditions.}
Panel (c) shows that models attend more to use conditions when the gold affordance comes from a larger cluster, i.e., a more common creative affordance. This suggests that rare affordances require more precise conditional reasoning, which current models often fail to capture. As a result, models may either ignore these conditions or reason about them incorrectly.

\begin{takeaway}
Auxiliary metrics reveal a consistent pattern: models handle familiar affordances much better than rare ones, and they benefit from distractors that provide affordance-level cues. Overall, current models still struggle to ground rare, weakly signaled affordances in a physically and conditionally appropriate way.
\end{takeaway}

\subsection{Error Analysis}

\begin{table*}[t]
    \centering
    \vspace{-3mm}
    \setlength\tabcolsep{2pt}
    \setlength\extrarowheight{2pt}

    \resizebox{\linewidth}{!}{
    \begin{tabular}{l c @{\hspace{10mm}} c @{\hspace{10mm}} c @{\hspace{10mm}} c @{\hspace{10mm}} c @{\hspace{5mm}} c @{\hspace{10mm}} c}
        \toprule
        \multirow{2}{*}{\textbf{Model}} & \multirow{2}{*}{\textbf{Percentage}} & \multirow{2}{*}{\textbf{\makecell{Gold\\Comparison}}} & \multicolumn{3}{c}{\textbf{Constraint Coverage}} & \multirow{2}{*}{\textbf{\makecell{Physical\\Grounding}}} & \multirow{2}{*}{\textbf{\makecell{Action\\Feasibility}}} \\
        \cmidrule(r){4-6}
        ~ & ~ & ~ & \textbf{\makecell{Use ($C_u$)}} & \textbf{\makecell{Env. ($C_e$)}} & \textbf{\makecell{Rcpt. ($C_r$)}} & ~ & ~ \\
        \addlinespace[2pt]
        \midrule
        \addlinespace[2pt]
        \multicolumn{8}{c}{\textit{Closed-Source Models}} \\
        \midrule
        GPT-5.2 & 47.90\% & \underline{1.2889} & \underline{4.5297} & \underline{4.4617} & \underline{4.6226} & \underline{4.3874} & \underline{4.2980} \\
        GPT-5 Mini & 51.41\% & \textbf{1.2942} & \textbf{4.7010} & \textbf{4.5982} & \textbf{4.7112} & \textbf{4.5513} & \textbf{4.4560} \\
        GPT-5 Nano & 42.79\% & 1.1489 & 3.9395 & 4.3547 & 4.3805 & 3.4507 & 3.2759 \\
        Gemini-2.5-Pro & 64.48\% & 1.1651 & 4.3639 & 3.0100 & 3.2293 & 4.3829 & 3.6117 \\
        Gemini-2.5-Flash & 63.06\% & 1.1854 & 4.5111 & 3.5451 & 3.6860 & 4.3584 & 3.6899 \\

        \addlinespace[2pt]
        \midrule
        \addlinespace[2pt]
        \multicolumn{8}{c}{\textit{Open-Source Models}} \\
        \midrule
        Qwen3-32B & 37.47\% & 1.1443 & 3.4811 & 2.7159 & 2.6444 & 3.5012 & 2.6879 \\
        Qwen3-14B & 38.70\% & 1.1252 & 3.0770 & 2.5366 & 2.3411 & 3.1419 & 2.3920 \\
        Qwen3-4B & 47.26\% & 1.0871 & 2.4423 & 2.0862 & 1.8587 & 2.6029 & 1.8733 \\
        Llama-3-70B & 44.67\% & 1.1329 & 2.4609 & 2.1490 & 2.1086 & 2.5749 & 2.1575 \\
        Ministral-3-14B & 47.37\% & 1.0988 & 3.4519 & 2.7042 & 2.6591 & 3.0195 & 2.3329 \\
        \midrule
        Average & 48.51\% & 1.1671 & 3.6958 & 3.2162 & 3.2241 & 3.5971 & 3.0775 \\
        \bottomrule
    \end{tabular}
    }
    \caption{Error analysis summary on judged wrong outputs where \textbf{both the entity and the part are wrong}. Percentage is the share of this slice over all tasks for that model. Best and second-best results in each score metric are highlighted in \textbf{bold} and \underline{underline}, respectively. The maximum score for each non-percentage metric is 5.0. Higher is better for all metrics.}
    \label{tab:result_error_wrong_both_wrong}
    \vspace{2mm}
\end{table*}

\begin{table*}[t]
    \centering
    \setlength\tabcolsep{2pt}
    \setlength\extrarowheight{2pt}

    \resizebox{\linewidth}{!}{
    \begin{tabular}{l c @{\hspace{10mm}} c @{\hspace{10mm}} c @{\hspace{10mm}} c @{\hspace{10mm}} c @{\hspace{5mm}} c @{\hspace{10mm}} c}
        \toprule
        \multirow{2}{*}{\textbf{Model}} & \multirow{2}{*}{\textbf{Percentage}} & \multirow{2}{*}{\textbf{\makecell{Gold\\Comparison}}} & \multicolumn{3}{c}{\textbf{Constraint Coverage}} & \multirow{2}{*}{\textbf{\makecell{Physical\\Grounding}}} & \multirow{2}{*}{\textbf{\makecell{Action\\Feasibility}}} \\
        \cmidrule(r){4-6}
        ~ & ~ & ~ & \textbf{\makecell{Use ($C_u$)}} & \textbf{\makecell{Env. ($C_e$)}} & \textbf{\makecell{Rcpt. ($C_r$)}} & ~ & ~ \\
        \addlinespace[2pt]
        \midrule
        \addlinespace[2pt]
        \multicolumn{8}{c}{\textit{Closed-Source Models}} \\
        \midrule
        GPT-5.2 & 33.90\% & \underline{1.3495} & 3.8951 & \underline{4.4197} & \underline{4.5242} & 3.3727 & \underline{3.7184} \\
        GPT-5 Mini & 30.97\% & 1.3433 & \textbf{4.2242} & \textbf{4.5805} & \textbf{4.6413} & \underline{3.6712} & \textbf{3.9150} \\
        GPT-5 Nano & 41.86\% & 1.1763 & 3.2356 & 4.1729 & 4.3309 & 2.4372 & 2.9408 \\
        Gemini-2.5-Pro & 18.82\% & \textbf{1.3765} & \underline{4.1961} & 3.0943 & 3.1685 & \textbf{4.0647} & 3.6064 \\
        Gemini-2.5-Flash & 21.62\% & 1.3246 & 4.0685 & 3.5596 & 3.5083 & 3.5011 & 3.4336 \\

        \addlinespace[2pt]
        \midrule
        \addlinespace[2pt]
        \multicolumn{8}{c}{\textit{Open-Source Models}} \\
        \midrule
        Qwen3-32B & 35.34\% & 1.2553 & 3.2526 & 2.7622 & 2.5685 & 3.3099 & 2.7903 \\
        Qwen3-14B & 36.56\% & 1.2252 & 2.9267 & 2.5558 & 2.4404 & 2.8104 & 2.5426 \\
        Qwen3-4B & 33.94\% & 1.2055 & 2.5440 & 2.2144 & 2.0052 & 2.5506 & 2.1610 \\
        Llama-3-70B & 33.82\% & 1.2290 & 2.5096 & 2.2587 & 2.2252 & 2.2489 & 2.3660 \\
        Ministral-3-14B & 30.92\% & 1.2179 & 3.3285 & 2.7540 & 2.6790 & 2.7024 & 2.4532 \\
        \midrule
        Average & 31.77\% & 1.2703 & 3.4181 & 3.2372 & 3.2091 & 3.0669 & 2.9927 \\
        \bottomrule
    \end{tabular}
    }
    \caption{Error analysis summary on judged wrong outputs where \textbf{the entity is correct but the part is wrong}. Percentage is the share of this slice over all tasks for that model. Best and second-best results in each score metric are highlighted in \textbf{bold} and \underline{underline}, respectively. The maximum score for each non-percentage metric is 5.0. Higher is better for all metrics.}
    \label{tab:result_error_wrong_entity_correct_part_wrong}
    \vspace{-2mm}
\end{table*}

In this section, we analyze the cases where the model selects an incorrect entity or part as the tool, in order to understand how models behave when tool selection fails. We divide the errors into two categories: (1) cases where the entity is correct but the part is wrong, and (2) cases where both the entity and the part are wrong. The results are reported in \Cref{tab:result_error_wrong_entity_correct_part_wrong} and \Cref{tab:result_error_wrong_both_wrong}, respectively.

Our analysis focuses on two aspects. First, we compare the predicted solution with the gold solution. For each task, we ask a third-party LLM to judge which solution is more convincing on a scale of 1--5, where 1 indicates the gold solution is much more convincing and 5 indicates the prediction is much more convincing. During this comparison, we also provide the gold justification annotations collected during the affordance verification step described in \Cref{sec:method}. 

Second, we evaluate the predicted solutions themselves, even though they are associated with incorrect tools. Specifically, we assess their \emph{constraint coverage}, \emph{physical grounding}, and \emph{action feasibility}, using the same criteria as in the main evaluation. We do not evaluate prediction correctness here because the prediction is intentionally conditioned on an incorrect tool, making correctness with respect to the gold solution less meaningful.

It is important to note that not all metrics are directly comparable with those in \Cref{tab:result_main}. The gold comparison score measures relative preference between two solutions, while the other metrics assess the intrinsic quality of the predicted solution. Moreover, when a model uses a different tool from the gold one, the predicted procedure may follow a different reasoning path, which further limits direct comparison with the main table.

All judgments are performed using Gemini-3.1-Flash-Lite. We adopt this automated evaluation because the number of erroneous cases is large (nearly 120K). Through human validation and sampling checks, we confirm that the model provides sufficiently reliable and cost-effective judgments. Further details are provided in \Cref{appendix_sec:analysis_details}.

\paragraph{Gold solutions are overwhelmingly preferred.}
Across both error categories, the gold solutions dominate the comparison. The average gold comparison scores are close to 1, corresponding to a gold win rate exceeding 95\%. This confirms that when the model selects an incorrect tool, the resulting solution is rarely competitive with the correct one.

We also observe a clear difference between the two error types. When both the entity and the part are wrong, the predicted solutions are judged slightly worse than when only the part is wrong. This is intuitive: using the correct entity still preserves some semantic alignment with the task, while selecting a completely wrong tool often leads to more obviously implausible reasoning.

\paragraph{Open-source models degrade more severely under tool errors.}
Even when evaluated independently of the gold solution, open-source models consistently receive lower scores in constraint coverage and physical grounding compared to closed-source models, indicating that the generated procedures themselves are of lower quality once the tool choice deviates from the correct one.

When compared with \Cref{tab:result_main}, this gap becomes more apparent. For open-source models, incorrect tool selection leads to a noticeable drop in reasoning quality. In contrast, strong closed-source models such as GPT-family models often still produce internally consistent explanations even when the chosen tool is not optimal. Although these predictions are still inferior to the gold solutions, they remain relatively well-grounded and coherent.

\paragraph{Action feasibility drops the most under tool errors.}
Among all metrics, action feasibility shows the largest degradation. Unlike others, this metric is directly comparable with the results in \Cref{tab:result_main} because it only evaluates whether the proposed action itself is plausible or not, regardless of what is the gold solution.

We observe an average absolute decrease of nearly 0.6. This suggests that once the model selects the wrong tool, it becomes substantially harder to construct a plausible action sequence. In many cases, the generated procedure no longer aligns with common-sense physical interactions, making the proposed actions appear unnatural or infeasible.

\begin{takeaway}
Correct tool selection is critical not only for final correctness but also for maintaining coherent and physically plausible reasoning. When the wrong tool is used, solution quality deteriorates sharply (especially in action feasibility), and the degradation is substantially more severe for open-source models.
\end{takeaway}

\subsection{Attribution Analysis}
In the previous section, we showed that the gold solution is strongly preferred by the judge, even in cases where the predicted ``how to use'' is itself plausible despite not matching the gold tool. In this section, we further investigate why this preference arises through an attribution analysis. Specifically, for each model, we randomly sample 10\% of its failure cases and use Gemini-3.1-Flash-Lite as the categorization model to identify the main reasons the prediction is considered inferior to the gold solution. Additional details are provided in \Cref{appendix_sec:analysis_details}.

We categorize the reasons why the gold solution is preferred (or why the prediction fails) into four high-level groups that capture different types of failure modes in tool repurposing. Each group further contains several fine-grained categories that describe specific mechanisms of failure.

\begin{itemize}[topsep=-1.5pt, leftmargin=10pt, itemsep=0pt]
    \item \textbf{A. Physical invalidity.} 
    The proposed repurposed tool cannot realistically perform the task due to incorrect physical assumptions. This includes: 
    (i) \textit{A1. Hallucinated affordance} (the model assumes non-existent features), 
    (ii) \textit{A2. Affordance mismatch} (the object's geometry, material, or mechanics are unsuitable), and 
    (iii) \textit{A3. Performance shortfall} (the object lacks sufficient capacity such as stability, mass, or precision).

    \item \textbf{B. Practical infeasibility.} 
    The solution is impractical to execute in real-world settings. This includes: 
    (i) \textit{B1. Destructive workaround} (requiring dismantling or damaging objects), and 
    (ii) \textit{B2. Context or accessibility issues} (the object is hard to access, or the procedure is overly complex).

    \item \textbf{C. Risk or constraint mismatch.} 
    The proposal introduces risks or violates task constraints. This includes: 
    (i) \textit{C1. Safety or damage risk} (unsafe, unhygienic, or likely to cause damage), and 
    (ii) \textit{C2. Constraint violation} (contradicting explicit requirements or intended object use).

    \item \textbf{D. Comparative inferiority.} 
    The prediction is workable but still worse than the gold solution. This includes: 
    (i) \textit{D1. Inferior but workable solutions} (less stable, convenient, or robust), and 
    (ii) \textit{D2. Preference-sensitive comparisons} (both solutions are reasonable, but the gold is more standard, practical or socially acceptable by users).
\end{itemize}

This taxonomy allows us to distinguish between fundamentally incorrect physical reasoning, impractical procedures, violations of real-world constraints, and cases where the prediction is merely less preferable rather than strictly incorrect. For each case, we assign one primary category and several contributing categories, and report their distributions respectively in the analysis.

\begin{figure}[t]
    \centering
    \vspace{-5mm}
    \begin{subfigure}[t]{0.37\linewidth}
        \centering
        \includegraphics[width=\linewidth]{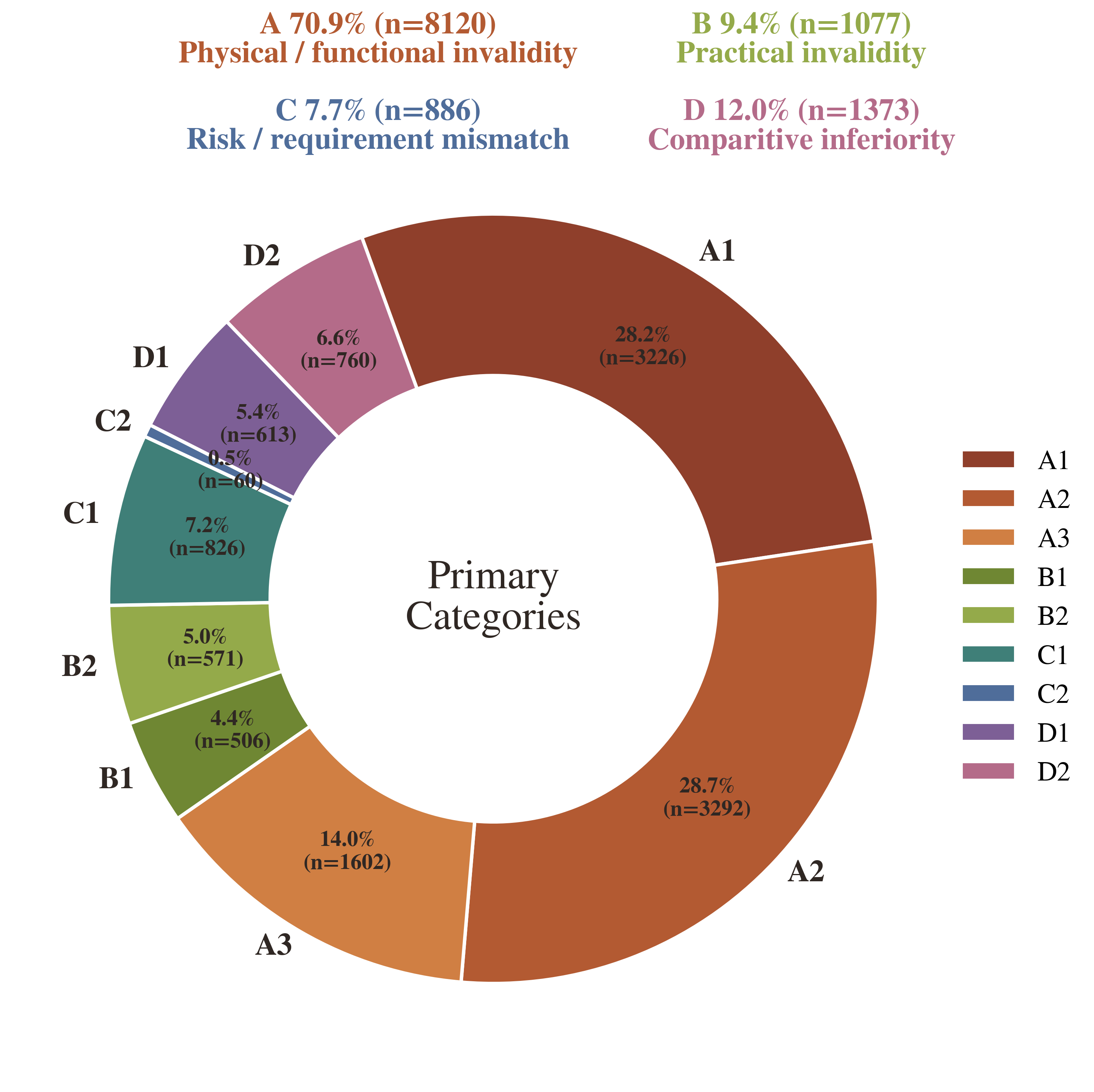}
        \caption{Primary category}
        \label{fig:attribution_pie}
    \end{subfigure}
    \hfill
    \begin{subfigure}[t]{0.6\linewidth}
        \centering
        \includegraphics[width=\linewidth]{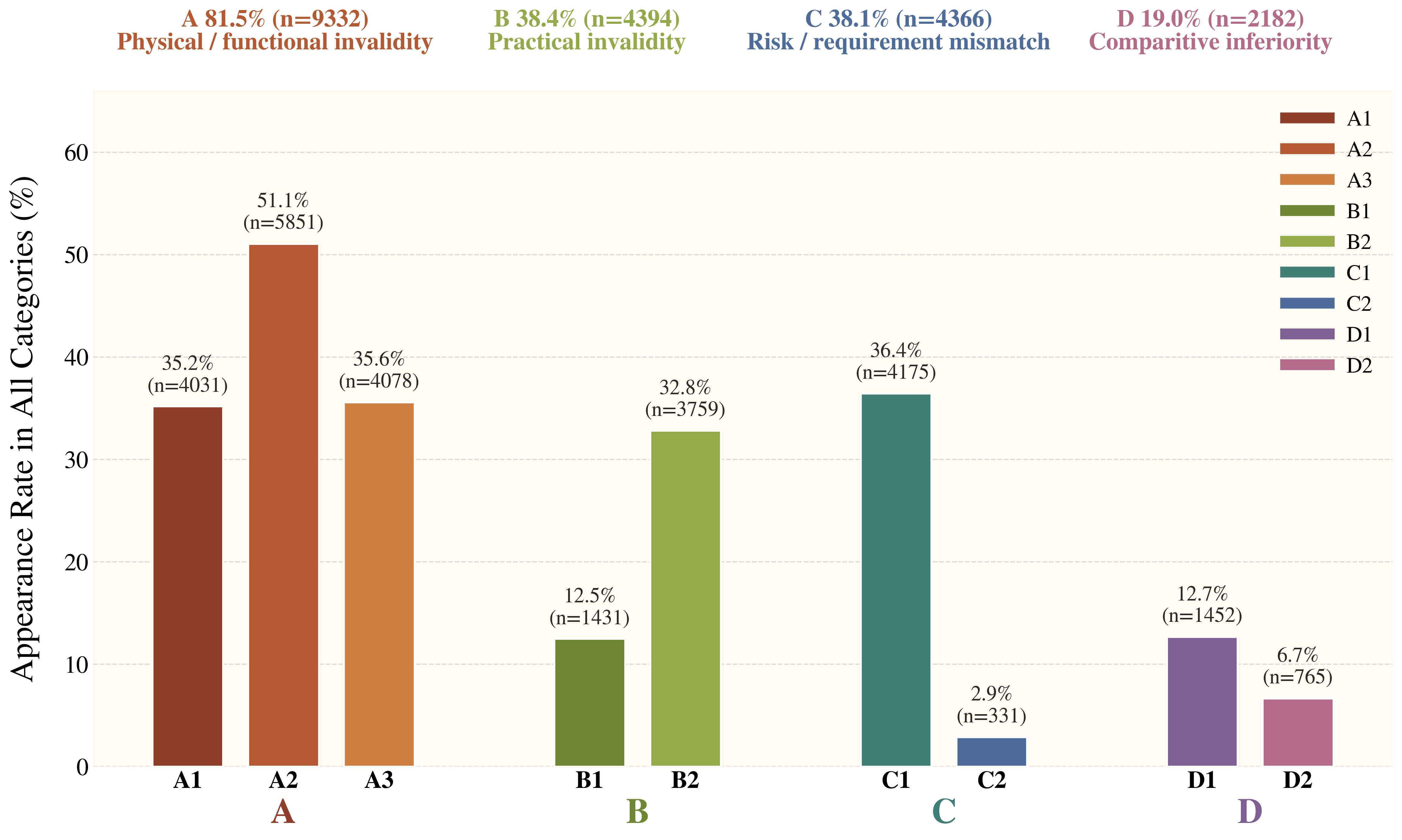}
        \caption{Contributing categories}
        \label{fig:attribution_bar}
    \end{subfigure}
    \caption{\textbf{Distribution of attribution categories for model failure cases}. Left: primary failure category assigned to each case. Right: frequency of categories appearing as contributing factors. Physical invalidity dominates in both views, with affordance mismatch emerging as the most common fine-grained cause.}
    \label{fig:attribution_pie_bar}
    \vspace{-2mm}
\end{figure}

\paragraph{Physical invalidity is the dominant source of failure.}
Across models, physical invalidity (Category A) is by far the most common reason the gold solution is preferred, both as the primary cause and as a contributing factor. Within this group, A2 (\textit{affordance mismatch}) is especially prominent, showing that models often choose objects whose geometry, material, or mechanics do not actually support the intended function. This suggests that the core weakness is not simply poor planning, but insufficient grounding in the physical properties of candidate tools.

\paragraph{Many errors reflect over-attribution of affordances.}
A1 (\textit{hallucinated affordance}) is also highly frequent, indicating that models sometimes assign capabilities to objects that they do not realistically have. These cases suggest a tendency to optimize for the task goal at the level of abstract function while under-checking whether the selected object can realize that function. Therefore, the prediction may sound plausible in isolation, yet still fail as it relies on imagined affordances.

\paragraph{Practicality and safety remain important secondary factors.}
Although less dominant than physical invalidity, practical infeasibility (B) and risk or constraint mismatch (C) also appear in a substantial share of contributing reasons. This means that even when a solution is not fundamentally impossible, it may still be inconvenient, unsafe, or inconsistent with task constraints, which makes it less preferable than the gold solution. By contrast, comparative inferiority (D) is relatively less common as a contributing cause, suggesting that most failures are not merely weaker alternatives, but are rooted in deeper issues of physical and real-world grounding. Please see more model-wise analysis and details in \Cref{appendix_sec:analysis_details}.

\begin{takeaway}
Most failures come from weak physical grounding: models often over-imagine what objects can do or choose tools whose real properties do not support the intended use. Practicality and safety matter too, but far fewer errors are simply suboptimal alternatives rather than genuinely mismatched solutions.
\end{takeaway}

\subsection{Human Study}

\begin{figure}[t]
    \centering
    \vspace{-3mm}
    \begin{subfigure}[t]{0.32\linewidth}
        \centering
        \includegraphics[width=\linewidth]{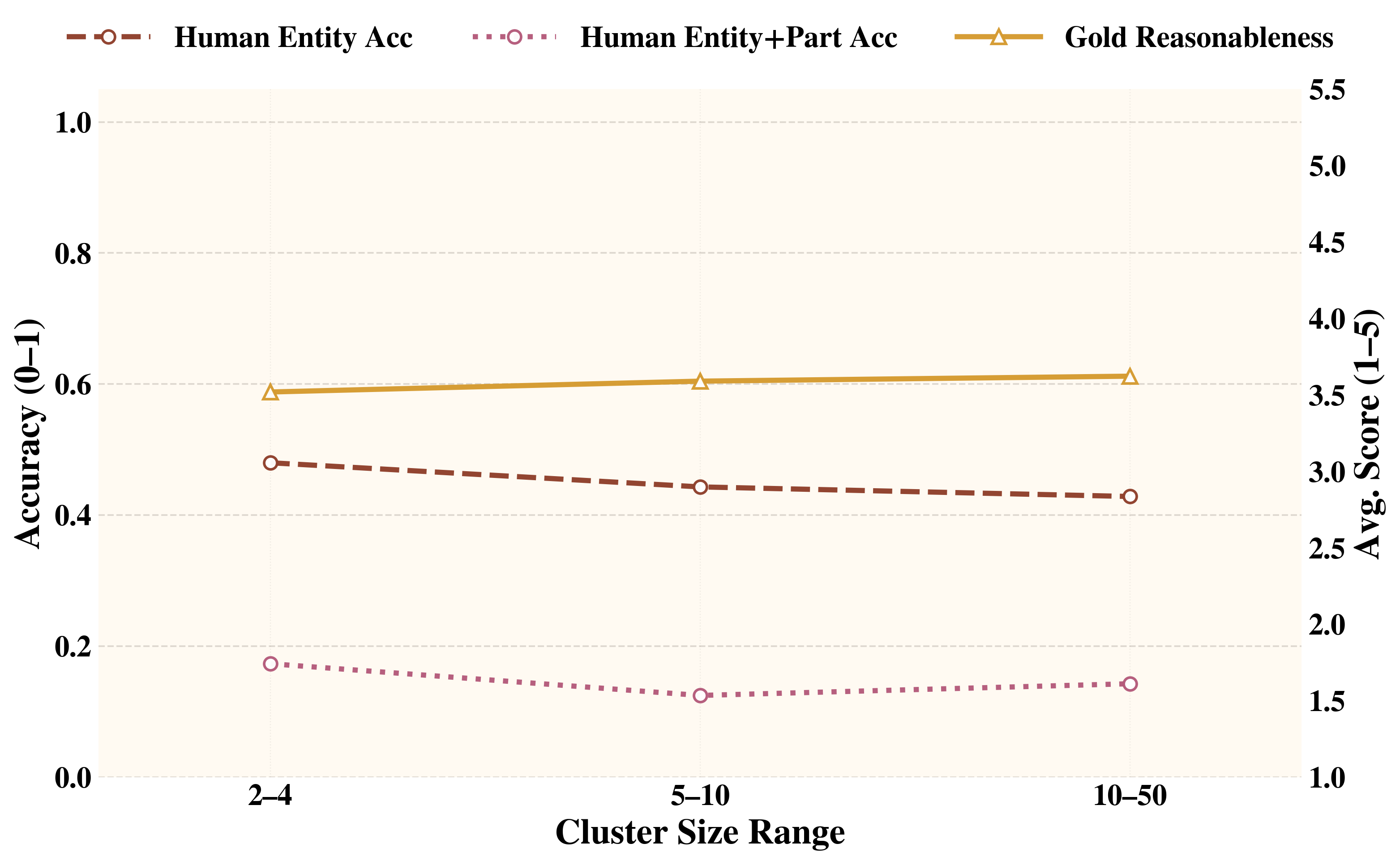}
        \caption{Gold cluster size.}
        \label{fig:human_cluster_size}
    \end{subfigure}
    \hfill
    \begin{subfigure}[t]{0.32\linewidth}
        \centering
        \includegraphics[width=\linewidth]{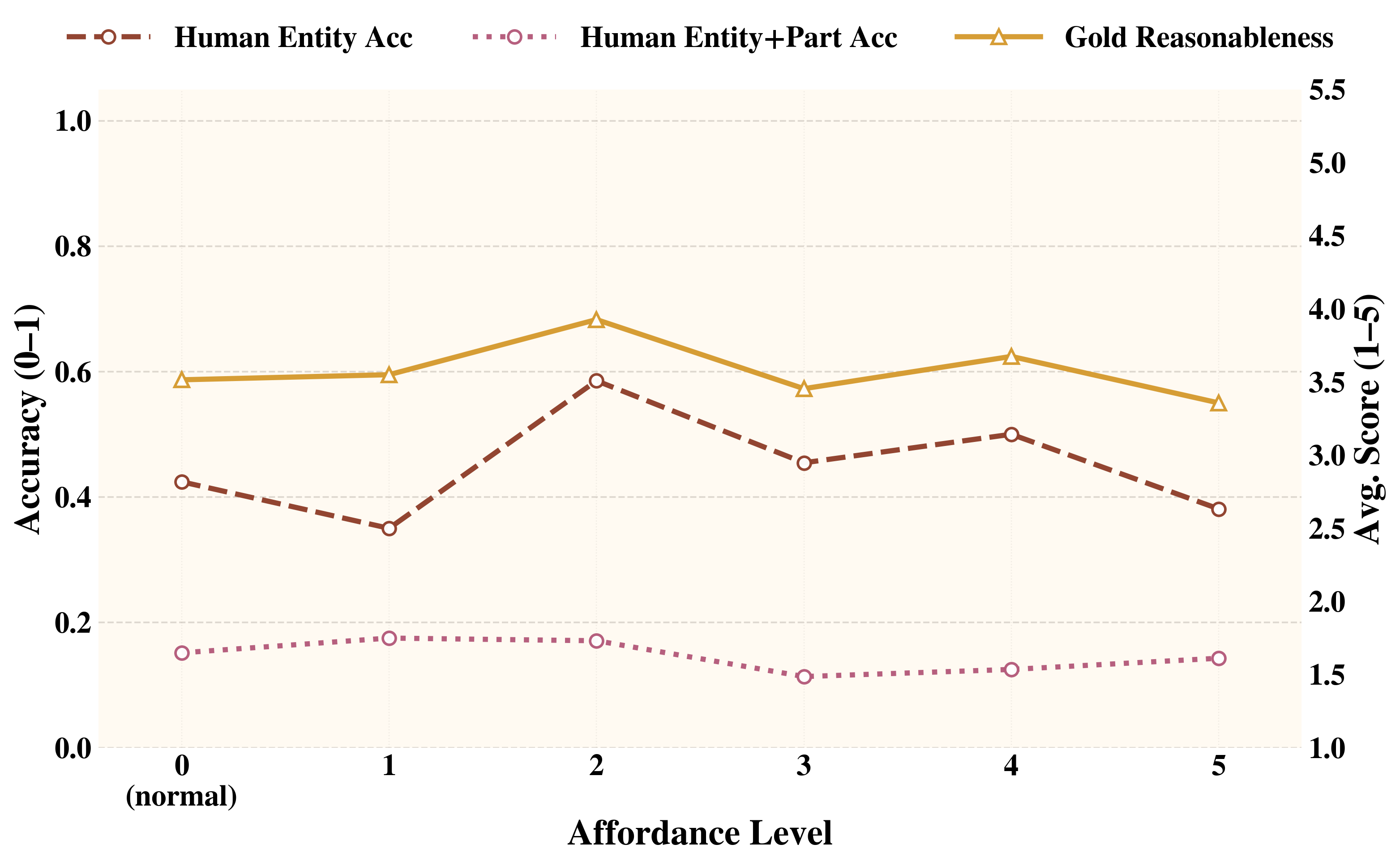}
        \caption{Gold affordance level.}
        \label{fig:human_typicality}
    \end{subfigure}
    \hfill
    \begin{subfigure}[t]{0.32\linewidth}
        \centering
        \includegraphics[width=\linewidth]{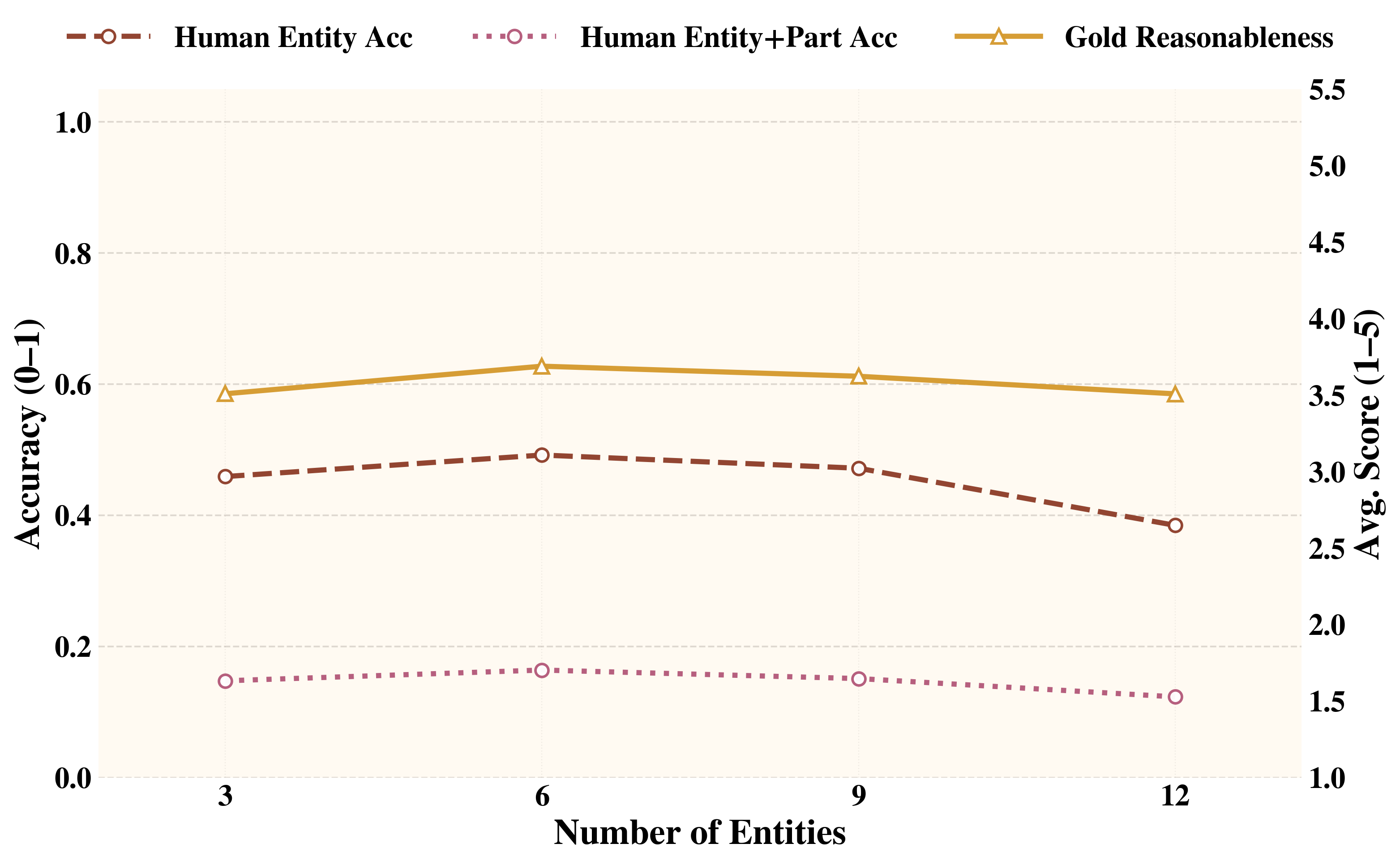}
        \caption{Distractor number.}
        \label{fig:human_entity_number}
    \end{subfigure}
    \caption{\textbf{Human Study Results.} Tool usage performance of human annotators across gold cluster size, distractor entity number, and gold affordance level.}
    \label{fig:human_study}
\end{figure}

\begin{table*}[t]
    \centering
    \resizebox{\linewidth}{!}{
    \begin{tabular}{c @{\hspace{5mm}} c @{\hspace{5mm}} c @{\hspace{5mm}} c @{\hspace{5mm}} c @{\hspace{5mm}} c @{\hspace{5mm}} c @{\hspace{5mm}} c @{\hspace{5mm}} c}
        \toprule
        \multicolumn{2}{c}{\textbf{Tool Usage}} & \multirow{2}{*}{\textbf{\makecell{Gold Solution\\Persuasiveness}}} & \multicolumn{3}{c}{\textbf{Constraint Coverage}} & \multirow{2}{*}{\textbf{\makecell{Physical\\Grounding}}} & \multirow{2}{*}{\textbf{\makecell{Action\\Feasibility}}} & \multirow{2}{*}{\textbf{\makecell{Human-Judged \\Creativity}}} \\

        \cmidrule(r){1-2} \cmidrule(r){4-6}
        
         \textbf{\makecell{Gold Correct}} & \textbf{\makecell{Entity Correct}} & ~ & \textbf{\makecell{Use ($C_u$)}} & \textbf{\makecell{Env. ($C_e$)}} & \textbf{\makecell{Rcpt. ($C_r$)}} & ~ & ~ & ~ \\
        
        \midrule
        \addlinespace[6pt]
         0.146 & 0.450 & 3.580 & 3.760 & 4.380 & 4.320 &  3.840 & 3.800 & 3.920 \\
        \bottomrule
        
    \end{tabular}
    }
    \caption{\textbf{Human check results.} Except for human Tool Usage test performance, all other results report gold solution quality manual check on a 5-point scale, where higher is better.}
    \vspace{-2mm}
    \label{tab:human_anno_result}
\end{table*}

To examine how humans solve these tasks and to validate the quality of our annotated affordance KB, we conduct a human study on 100 sampled tasks balanced across key factors including entity count, affordance similarity, and affordance typicality. We employ 10 human annotators with STEM background and use a two-stage protocol: \textbf{problem solving}, in which annotators select the best entity and part given the task and attribute descriptions, and \textbf{review}, in which they compare their answer against the gold solution and assess whether the KB-supported justification for preferring the gold answer is convincing. As shown in \Cref{tab:human_anno_result}, humans achieve \textbf{0.146 gold correctness} and \textbf{0.450 entity correctness}, both below the best-performing model. This gap mainly reflects the high \textbf{cognitive load} of the task, which requires annotators to keep track of many entities, parts, and fine-grained attributes entirely from textual descriptions.

At the same time, the review results support the validity of the benchmark. Gold solutions receive positive ratings on physical groundedness, feasibility, constraint reasonableness, and creativity, and the average gold persuasiveness achieves 63.0\% inter-annotator agreement, suggesting that the gold answers are generally viewed as valid and well justified. Furthermore, \Cref{fig:human_study} shows that human performance is overall less sensitive than model performance to cluster size, gold affordance level, and distractor count. This suggests that our layered design primarily increases difficulty for models rather than humans. For humans, performance appears to be limited more by the overall burden of navigating dense candidate spaces, making it relatively insensitive to changes in specific design factors. For models, in contrast, performance is more strongly shaped by physical-grounding difficulty, especially when the correct affordance is less common.

\section{Discussion}
\label{sec:discussion}

\paragraph{Difference of Creativity and Hallucination.} Throughout this paper, we define creativity as \emph{grounded} in object attributes and the affordances they induce. In this sense, our notion of creativity is importantly different from creative writing, open-ended design, or innovative research ideation, which may involve freer imagination, sudden insight, or even a degree of productive ``hallucination''~\citep{lu2026rethinking}. By contrast, the creativity studied here is better understood as \textbf{structured creativity}: it still requires novelty, but that novelty must emerge from reasoning over physically plausible object properties and their functional implications. Creative tool use is therefore not about inventing arbitrary solutions, but about connecting the model's goal to the affordances available in the environment in a non-obvious yet grounded way. This distinction is especially important for LLM agents, and even more so for future embodied agents, because tool use is a core capability for them to act in the world. When such agents are grounded in real environments, creativity can directly improve their usefulness to human users by enabling alternative, practical pathways for everyday problem-solving under constraints.

\textbf{Toward Physical-Textual Dual Reasoning and Foresight Governance.} Our findings also suggest that grounded creativity may require a reasoning architecture that goes beyond text-only deliberation. Even when models are given explicit part, attribute, and affordance information, they often commit to plausible-sounding but mechanically incorrect solutions, indicating that they lack an internal process for anticipating physical consequences before acting. A promising direction is physical-textual dual reasoning: textual reasoning proposes candidate affordance recombinations, while a complementary physical imagination module predicts how object parts, materials, and states would evolve under those candidate actions. Such a loop will improve creative discovery as well as provide a form of foresight governance, where candidate solutions are filtered by predicted feasibility, safety, and downstream side effects before execution~\citep{qian2026current}. This is especially important for future embodied agents, for whom an ungrounded ``creative'' action may damage tools, violate constraints, or irreversibly alter the environment. In this sense, CreativityBench can serve as a foundation for future benchmarks that evaluate not only whether a model can invent a novel use, but whether it can foresee consequences, reject risky affordance hypotheses, and revise plans when hypothesized outcomes conflict with physical reality.

\paragraph{Enhancement of Model Creativity.} Improving model creativity may require a training objective that differs from the one optimized by much of current reinforcement learning. Existing RL methods, especially unsupervised ones, often emphasize sampling efficiency and empirically lead to distribution sharpening~\citep{he2026far}, which benefits reliability but can suppress the structured diversity needed for creative problem solving. For instance, although methods like TTRL~\citep{zuo2025ttrl} can improve analytical intelligence in domains like math and coding, their reliance on majority vote as pseudo-labels may further reduce generation diversity and thus work against creativity. Recent work on rewarding unlikeliness~\citep{he2025rewarding} points in a promising direction by explicitly encouraging broader exploration, but it has not yet been applied to creative problem-solving settings. Our affordance KB, alternatively, provides one concrete step in this direction: beyond benchmark construction, it can be used to sample thousands of grounded creative problem-solving trajectories, potentially supporting the development of models that learn creative reasoning from data rather than from outcome supervision alone.

\section{Conclusion and Future Work}
\label{sec:conclusion}
In this work, we introduced CreativityBench, a large-scale benchmark for evaluating creative tool repurposing through affordance-based reasoning, together with a structured affordance knowledge base that links entities, parts, attributes, and uses. Our results reveal that current language models often struggle to ground creative solutions at the level of the correct part, attribute, and physical mechanism; moreover, stronger general reasoning, larger model scale, and standard inference-time strategies such as CoT do not reliably translate into better creative affordance discovery. These findings suggest that creative intelligence in models is not simply an extension of analytical reasoning or action planning, but a distinct capability requiring grounded comparison, flexible recombination of affordances, and careful attention to physical constraints. Looking forward, we hope CreativityBench can serve not only as an evaluation testbed but also as a resource for training models to improve grounded creative reasoning. An important direction for future work is to extend this framework beyond static text settings toward multimodal, interactive, and embodied environments, while also exploring training objectives that encourage grounded exploration and diverse yet physically plausible affordance discovery rather than mere distribution sharpening.

\bibliography{colm2026_conference}

@article{qian2024escapebench,
  title={Escapebench: Pushing language models to think outside the box},
  author={Qian, Cheng and Han, Peixuan and Luo, Qinyu and He, Bingxiang and Chen, Xiusi and Zhang, Yuji and Du, Hongyi and Yao, Jiarui and Yang, Xiaocheng and Zhang, Denghui and others},
  journal={arXiv e-prints},
  pages={arXiv--2412},
  year={2024}
}

@inproceedings{liu2026dynamic,
  title={Navigating Worlds and Minds: Dynamic Evaluation of LLM Agent Robustness under Progressively Disclosing Dual-Constraints},
  author={Liu, Jiayu and Qian, Cheng and Ji, Heng},
  booktitle={MSLD 2026 Meeting},
  year={2026},
  url={https://openreview.net/forum?id=UmC50qYoOy}
}

@inproceedings{liu2025revisiting,
  title={Revisiting Epistemic Markers in Confidence Estimation: Can Markers Accurately Reflect Large Language Models’ Uncertainty?},
  author={Liu, Jiayu and Zong, Qing and Wang, Weiqi and Song, Yangqiu},
  booktitle={Proceedings of the 63rd Annual Meeting of the Association for Computational Linguistics (Volume 2: Short Papers)},
  pages={206--221},
  year={2025}
}

@inproceedings{fang2025creation,
  title={Creation-MMBench: Assessing Context-Aware Creative Intelligence in MLLMs},
  author={Fang, Xinyu and Chen, Zhijian and Lan, Kai and Ma, Lixin and Ding, Shengyuan and Liang, Yingji and Zhao, Xiangyu and Wen, Farong and Zhang, Zicheng and Zhang, Guofeng and others},
  booktitle={Proceedings of the IEEE/CVF International Conference on Computer Vision},
  pages={447--456},
  year={2025}
}

@article{jamone2016affordances,
  title={Affordances in psychology, neuroscience, and robotics: A survey},
  author={Jamone, Lorenzo and Ugur, Emre and Cangelosi, Angelo and Fadiga, Luciano and Bernardino, Alexandre and Piater, Justus and Santos-Victor, Jos{\'e}},
  journal={IEEE Transactions on Cognitive and Developmental Systems},
  volume={10},
  number={1},
  pages={4--25},
  year={2016},
  publisher={IEEE}
}

@article{liu2025costbench,
  title={CostBench: Evaluating Multi-Turn Cost-Optimal Planning and Adaptation in Dynamic Environments for LLM Tool-Use Agents},
  author={Liu, Jiayu and Qian, Cheng and Su, Zhaochen and Zong, Qing and Huang, Shijue and He, Bingxiang and Fung, Yi R},
  journal={arXiv preprint arXiv:2511.02734},
  year={2025}
}

@article{chu2019learning,
  title={Learning affordance segmentation for real-world robotic manipulation via synthetic images},
  author={Chu, Fu-Jen and Xu, Ruinian and Vela, Patricio A},
  journal={IEEE Robotics and Automation Letters},
  volume={4},
  number={2},
  pages={1140--1147},
  year={2019},
  publisher={IEEE}
}

@article{montesano2008learning,
  title={Learning object affordances: from sensory--motor coordination to imitation},
  author={Montesano, Luis and Lopes, Manuel and Bernardino, Alexandre and Santos-Victor, Jos{\'e}},
  journal={Ieee transactions on robotics},
  volume={24},
  number={1},
  pages={15--26},
  year={2008},
  publisher={IEEE}
}

@article{liu2026naacl,
  title={NAACL: Noise-AwAre Verbal Confidence Calibration for LLMs in RAG Systems},
  author={Liu, Jiayu and Wang, Rui and Zong, Qing and Zeng, Qingcheng and Zheng, Tianshi and Shi, Haochen and Guo, Dadi and Xu, Baixuan and Li, Chunyang and Song, Yangqiu},
  journal={arXiv preprint arXiv:2601.11004},
  year={2026}
}

@article{brohan2024rt,
  title={Rt-2: Vision-language-action models transfer web knowledge to robotic control, 2023},
  author={Brohan, Anthony and Brown, Noah and Carbajal, Justice and Chebotar, Yevgen and Chen, Xi and Choromanski, Krzysztof and Ding, Tianli and Driess, Danny and Dubey, Avinava and Finn, Chelsea and others},
  journal={URL https://arxiv. org/abs/2307.15818},
  volume={1},
  pages={2},
  year={2024}
}

@article{brohan2022rt,
  title={Rt-1: Robotics transformer for real-world control at scale},
  author={Brohan, Anthony and Brown, Noah and Carbajal, Justice and Chebotar, Yevgen and Dabis, Joseph and Finn, Chelsea and Gopalakrishnan, Keerthana and Hausman, Karol and Herzog, Alex and Hsu, Jasmine and others},
  journal={arXiv preprint arXiv:2212.06817},
  year={2022}
}

@inproceedings{wang2023newton,
  title={NEWTON: Are large language models capable of physical reasoning?},
  author={Wang, Yi and Duan, Jiafei and Fox, Dieter and Srinivasa, Siddhartha},
  booktitle={Findings of the Association for Computational Linguistics: EMNLP 2023},
  pages={9743--9758},
  year={2023}
}

@inproceedings{aroca2021prost,
  title={PROST: Physical reasoning about objects through space and time},
  author={Aroca-Ouellette, St{\'e}phane and Paik, Cory and Roncone, Alessandro and von der Wense, Katharina},
  booktitle={Findings of the Association for Computational Linguistics: ACL-IJCNLP 2021},
  pages={4597--4608},
  year={2021}
}

@inproceedings{bisk2020piqa,
  title={Piqa: Reasoning about physical commonsense in natural language},
  author={Bisk, Yonatan and Zellers, Rowan and Gao, Jianfeng and Choi, Yejin and others},
  booktitle={Proceedings of the AAAI conference on artificial intelligence},
  volume={34},
  number={05},
  pages={7432--7439},
  year={2020}
}

@article{si2024can,
  title={Can llms generate novel research ideas? a large-scale human study with 100+ nlp researchers},
  author={Si, Chenglei and Yang, Diyi and Hashimoto, Tatsunori},
  journal={arXiv preprint arXiv:2409.04109},
  year={2024}
}

@inproceedings{ha2025synthia,
  title={Synthia: Novel concept design with affordance composition},
  author={Ha, Hyeonjeong and Jin, Xiaomeng and Kim, Jeonghwan and Liu, Jiateng and Wang, Zhenhailong and Nguyen, Khanh Duy and Blume, Ansel and Peng, Nanyun and Chang, Kai-Wei and Ji, Heng},
  booktitle={Proceedings of the 63rd Annual Meeting of the Association for Computational Linguistics (Volume 1: Long Papers)},
  pages={20939--20958},
  year={2025}
}

@inproceedings{wang2024scimon,
  title={Scimon: Scientific inspiration machines optimized for novelty},
  author={Wang, Qingyun and Downey, Doug and Ji, Heng and Hope, Tom},
  booktitle={Proceedings of the 62nd Annual Meeting of the Association for Computational Linguistics (Volume 1: Long Papers)},
  pages={279--299},
  year={2024}
}

@inproceedings{qian2025modelingagent,
  title={ModelingAgent: Bridging LLMs and Mathematical Modeling for Real-World Challenges},
  author={Qian, Cheng and Du, Hongyi and Wang, Hongru and Chen, Xiusi and Zhang, Yuji and Sil, Avirup and Zhai, Chengxiang and McKeown, Kathleen and Ji, Heng},
  booktitle={Findings of the Association for Computational Linguistics: EMNLP 2025},
  pages={1599--1633},
  year={2025}
}

@inproceedings{qian2023creator,
  title={Creator: Tool creation for disentangling abstract and concrete reasoning of large language models},
  author={Qian, Cheng and Han, Chi and Fung, Yi and Qin, Yujia and Liu, Zhiyuan and Ji, Heng},
  booktitle={Findings of the Association for Computational Linguistics: EMNLP 2023},
  pages={6922--6939},
  year={2023}
}

@article{cai2023large,
  title={Large language models as tool makers},
  author={Cai, Tianle and Wang, Xuezhi and Ma, Tengyu and Chen, Xinyun and Zhou, Denny},
  journal={arXiv preprint arXiv:2305.17126},
  year={2023}
}

@inproceedings{akoury2020storium,
  title={Storium: A dataset and evaluation platform for machine-in-the-loop story generation},
  author={Akoury, Nader and Wang, Shufan and Whiting, Josh and Hood, Stephen and Peng, Nanyun and Iyyer, Mohit},
  booktitle={Proceedings of the 2020 Conference on Empirical Methods in Natural Language Processing (EMNLP)},
  pages={6470--6484},
  year={2020}
}

@article{brown2020language,
  title={Language models are few-shot learners},
  author={Brown, Tom and Mann, Benjamin and Ryder, Nick and Subbiah, Melanie and Kaplan, Jared D and Dhariwal, Prafulla and Neelakantan, Arvind and Shyam, Pranav and Sastry, Girish and Askell, Amanda and others},
  journal={Advances in neural information processing systems},
  volume={33},
  pages={1877--1901},
  year={2020}
}

@article{boden1998creativity,
  title={Creativity and artificial intelligence},
  author={Boden, Margaret A},
  journal={Artificial intelligence},
  volume={103},
  number={1-2},
  pages={347--356},
  year={1998},
  publisher={Elsevier}
}

@article{guilford1967creativity,
  title={Creativity: Yesterday, today and tomorrow},
  author={Guilford, Joy P},
  journal={The Journal of Creative Behavior},
  volume={1},
  number={1},
  pages={3--14},
  year={1967},
  publisher={Wiley Online Library}
}

@inproceedings{lim2025visescape,
  title={VisEscape: A Benchmark for Evaluating Exploration-driven Decision-making in Virtual Escape Rooms},
  author={Lim, Seungwon and Kim, Sungwoong and Yu, Jihwan and Lee, Sungjae and Chung, Jiwan and Yu, Youngjae},
  booktitle={Proceedings of the 2025 Conference on Empirical Methods in Natural Language Processing},
  pages={16031--16058},
  year={2025}
}

@inproceedings{dong2024villageragent,
  title={Villageragent: A graph-based multi-agent framework for coordinating complex task dependencies in minecraft},
  author={Dong, Yubo and Zhu, Xukun and Pan, Zhengzhe and Zhu, Linchao and Yang, Yi},
  booktitle={Findings of the Association for Computational Linguistics: ACL 2024},
  pages={16290--16314},
  year={2024}
}

@inproceedings{tian2024macgyver,
  title={MacGyver: Are Large Language Models Creative Problem Solvers?},
  author={Tian, Yufei and Ravichander, Abhilasha and Qin, Lianhui and Le Bras, Ronan and Marjieh, Raja and Peng, Nanyun and Choi, Yejin and Griffiths, Thomas L and Brahman, Faeze},
  booktitle={Proceedings of the 2024 Conference of the North American Chapter of the Association for Computational Linguistics: Human Language Technologies (Volume 1: Long Papers)},
  pages={5303--5324},
  year={2024}
}

@article{xi2025survey,
  title={A survey of llm-based deep search agents: Paradigm, optimization, evaluation, and challenges},
  author={Xi, Yunjia and Lin, Jianghao and Xiao, Yongzhao and Zhou, Zheli and Shan, Rong and Gao, Te and Zhu, Jiachen and Liu, Weiwen and Yu, Yong and Zhang, Weinan},
  journal={arXiv preprint arXiv:2508.05668},
  year={2025}
}

@article{ge2025survey,
  title={A survey of vibe coding with large language models},
  author={Ge, Yuyao and Mei, Lingrui and Duan, Zenghao and Li, Tianhao and Zheng, Yujia and Wang, Yiwei and Wang, Lexin and Yao, Jiayu and Liu, Tianyu and Cai, Yujun and others},
  journal={arXiv preprint arXiv:2510.12399},
  year={2025}
}

@article{yu2025browseragent,
  title={BrowserAgent: Building Web Agents with Human-Inspired Web Browsing Actions},
  author={Yu, Tao and Zhang, Zhengbo and Lyu, Zhiheng and Gong, Junhao and Yi, Hongzhu and Wang, Xinming and Zhou, Yuxuan and Yang, Jiabing and Nie, Ping and Huang, Yan and others},
  journal={arXiv preprint arXiv:2510.10666},
  year={2025}
}

@article{wei2022chain,
  title={Chain-of-thought prompting elicits reasoning in large language models},
  author={Wei, Jason and Wang, Xuezhi and Schuurmans, Dale and Bosma, Maarten and Xia, Fei and Chi, Ed and Le, Quoc V and Zhou, Denny and others},
  journal={Advances in neural information processing systems},
  volume={35},
  pages={24824--24837},
  year={2022}
}

@article{cobbe2021training,
  title={Training verifiers to solve math word problems},
  author={Cobbe, Karl and Kosaraju, Vineet and Bavarian, Mohammad and Chen, Mark and Jun, Heewoo and Kaiser, Lukasz and Plappert, Matthias and Tworek, Jerry and Hilton, Jacob and Nakano, Reiichiro and others},
  journal={arXiv preprint arXiv:2110.14168},
  year={2021}
}

@article{hendrycks2020measuring,
  title={Measuring massive multitask language understanding},
  author={Hendrycks, Dan and Burns, Collin and Basart, Steven and Zou, Andy and Mazeika, Mantas and Song, Dawn and Steinhardt, Jacob},
  journal={arXiv preprint arXiv:2009.03300},
  year={2020}
}

@article{wei2025browsecomp,
  title={Browsecomp: A simple yet challenging benchmark for browsing agents},
  author={Wei, Jason and Sun, Zhiqing and Papay, Spencer and McKinney, Scott and Han, Jeffrey and Fulford, Isa and Chung, Hyung Won and Passos, Alex Tachard and Fedus, William and Glaese, Amelia},
  journal={arXiv preprint arXiv:2504.12516},
  year={2025}
}

@inproceedings{mialon2023gaia,
  title={Gaia: a benchmark for general ai assistants},
  author={Mialon, Gr{\'e}goire and Fourrier, Cl{\'e}mentine and Wolf, Thomas and LeCun, Yann and Scialom, Thomas},
  booktitle={The Twelfth International Conference on Learning Representations},
  year={2023}
}

@article{froger2025scaling,
  title={Are: Scaling up agent environments and evaluations},
  author={Froger, Romain and Andrews, Pierre and Bettini, Matteo and Budhiraja, Amar and Cabral, Ricardo Silveira and Do, Virginie and Garreau, Emilien and Gaya, Jean-Baptiste and Lauren{\c{c}}on, Hugo and Lecanu, Maxime and others},
  journal={arXiv preprint arXiv:2509.17158},
  year={2025}
}

@article{sternberg1997triarchic,
  title={The triarchic theory of intelligence.},
  author={Sternberg, Robert J},
  year={1997},
  publisher={The Guilford Press}
}

@inproceedings{speer2017conceptnet,
  title={Conceptnet 5.5: An open multilingual graph of general knowledge},
  author={Speer, Robyn and Chin, Joshua and Havasi, Catherine},
  booktitle={Proceedings of the AAAI conference on artificial intelligence},
  volume={31},
  number={1},
  year={2017}
}

@article{singh2025openai,
  title={Openai gpt-5 system card},
  author={Singh, Aaditya and Fry, Adam and Perelman, Adam and Tart, Adam and Ganesh, Adi and El-Kishky, Ahmed and McLaughlin, Aidan and Low, Aiden and Ostrow, AJ and Ananthram, Akhila and others},
  journal={arXiv preprint arXiv:2601.03267},
  year={2025}
}

@article{yang2025qwen3,
  title={Qwen3 technical report},
  author={Yang, An and Li, Anfeng and Yang, Baosong and Zhang, Beichen and Hui, Binyuan and Zheng, Bo and Yu, Bowen and Gao, Chang and Huang, Chengen and Lv, Chenxu and others},
  journal={arXiv preprint arXiv:2505.09388},
  year={2025}
}

@article{comanici2025gemini,
  title={Gemini 2.5: Pushing the frontier with advanced reasoning, multimodality, long context, and next generation agentic capabilities},
  author={Comanici, Gheorghe and Bieber, Eric and Schaekermann, Mike and Pasupat, Ice and Sachdeva, Noveen and Dhillon, Inderjit and Blistein, Marcel and Ram, Ori and Zhang, Dan and Rosen, Evan and others},
  journal={arXiv preprint arXiv:2507.06261},
  year={2025}
}

@article{grattafiori2024llama,
  title={The llama 3 herd of models},
  author={Grattafiori, Aaron and Dubey, Abhimanyu and Jauhri, Abhinav and Pandey, Abhinav and Kadian, Abhishek and Al-Dahle, Ahmad and Letman, Aiesha and Mathur, Akhil and Schelten, Alan and Vaughan, Alex and others},
  journal={arXiv preprint arXiv:2407.21783},
  year={2024}
}

@article{liu2026ministral,
  title={Ministral 3},
  author={Liu, Alexander H and Khandelwal, Kartik and Subramanian, Sandeep and Jouault, Victor and Rastogi, Abhinav and Sad{\'e}, Adrien and Jeffares, Alan and Jiang, Albert and Cahill, Alexandre and Gavaudan, Alexandre and others},
  journal={arXiv preprint arXiv:2601.08584},
  year={2026}
}

@article{runco2012standard,
  title={The standard definition of creativity},
  author={Runco, Mark A and Jaeger, Garrett J},
  journal={Creativity research journal},
  volume={24},
  number={1},
  pages={92--96},
  year={2012},
  publisher={Taylor \& Francis}
}

@article{sternberg1999concept,
  title={The concept of creativity: Prospects and paradigms},
  author={Sternberg, Robert J and Lubart, Todd I},
  journal={Handbook of creativity},
  volume={1},
  number={3-15},
  year={1999},
  publisher={Cambridge University Press}
}

@inproceedings{lu2026rethinking,
  title={Rethinking creativity evaluation: A critical analysis of existing creativity evaluations},
  author={Lu, Li-Chun and Liu, Miri and Lu, Pin Chun and Tian, Yufei and Sun, Shao-Hua and Peng, Nanyun},
  booktitle={Proceedings of the 19th Conference of the European Chapter of the Association for Computational Linguistics (Volume 1: Long Papers)},
  pages={6329--6352},
  year={2026}
}

@article{he2026far,
  title={How Far Can Unsupervised RLVR Scale LLM Training?},
  author={He, Bingxiang and Zuo, Yuxin and Liu, Zeyuan and Zhao, Shangziqi and Fu, Zixuan and Yang, Junlin and Qian, Cheng and Zhang, Kaiyan and Fan, Yuchen and Cui, Ganqu and others},
  journal={arXiv preprint arXiv:2603.08660},
  year={2026}
}

@article{zuo2025ttrl,
  title={Ttrl: Test-time reinforcement learning},
  author={Zuo, Yuxin and Zhang, Kaiyan and Sheng, Li and Qu, Shang and Cui, Ganqu and Zhu, Xuekai and Li, Haozhan and Zhang, Yuchen and Long, Xinwei and Hua, Ermo and others},
  journal={arXiv preprint arXiv:2504.16084},
  year={2025}
}

@inproceedings{he2025rewarding,
  title={Rewarding the unlikely: Lifting grpo beyond distribution sharpening},
  author={He, Andre Wang and Fried, Daniel and Welleck, Sean},
  booktitle={Proceedings of the 2025 Conference on Empirical Methods in Natural Language Processing},
  pages={25559--25571},
  year={2025}
}

@article{qian2026current,
  title={Current Agents Fail to Leverage World Model as Tool for Foresight},
  author={Qian, Cheng and Acikgoz, Emre Can and Li, Bingxuan and Chen, Xiusi and Zhang, Yuji and He, Bingxiang and Luo, Qinyu and Hakkani-T{\"u}r, Dilek and Tur, Gokhan and Li, Yunzhu and others},
  journal={arXiv preprint arXiv:2601.03905},
  year={2026}
}
\bibliographystyle{colm2026_conference}

\newpage
\appendix
\label{app:main}

\section{Significance, Scope, and Clarifications}

\subsection{Why CreativityBench Matters}

\paragraph{A Missing Dimension in Current Evaluation.}
Recent progress in LLMs has been measured primarily through analytical reasoning, tool execution, and long-horizon planning. Our work focuses on a different and still under-measured capability: \textbf{creative tool use grounded in physical affordances}. The central question is not whether a model can produce a plausible plan in words, but whether it can identify \textbf{which object} and \textbf{which physical property} make a non-obvious solution actually usable. We view this as an important and practical dimension of intelligence, especially for agents that must operate under constraints, limited resources, or unforeseen situations.

\paragraph{From Object Selection to Mechanism-Level Reasoning.}
A key contribution of CreativityBench is that it moves evaluation beyond coarse object-level plausibility. Existing settings often reward selecting a generally relevant object. In contrast, our benchmark requires models to localize the relevant \textbf{part}, connect it to \textbf{attributes}, and recover the \textbf{affordance mechanism} that enables success. This finer level of grounding is exactly where current models show their largest performance drop, which suggests that the benchmark is exposing a real and previously under-tested weakness rather than merely repackaging existing commonsense tasks.

\paragraph{A Structured Resource for Studying Grounded Creativity.}
The benchmark is supported by a structured affordance knowledge base linking entities, parts, attributes, and affordances. This organization is important for two reasons. First, it enables systematic task construction at scale while preserving an interpretable latent solution path. Second, it makes failure analysis much more informative: when a model fails, we can ask whether it selected the wrong entity, the wrong part, the wrong attribute basis, or an implausible use condition. This is a major advantage over open-ended creativity evaluations that are difficult to diagnose or reproduce.

\paragraph{Implications for Future Models and Agents.}
We believe the significance of this benchmark extends beyond one dataset. For LLMs, it highlights the limits of strong general reasoning when physical grounding and unconventional repurposing are required. For embodied systems, it points to a core challenge in robust real-world problem solving: useful behavior often depends on recognizing latent affordances under constraints, not just following canonical object functions. More broadly, the benchmark offers a concrete testbed for studying how grounded knowledge, comparative search, and flexible recombination interact in creative problem solving.

\subsection{Clarifications of Concerns}

\paragraph{Scope and Methodological Role of LLM-Assisted Construction.}
In our setting, LLM assistance is used primarily as a \textbf{scaling mechanism for structured data construction}, not as a substitute for the benchmark definition itself. The benchmark is anchored in an explicit ontology including entity, part, attribute, affordance, and conditions, and the resulting tasks are generated from this structured representation rather than from unconstrained free-form synthesis. This design reduces arbitrariness: the LLM does not define creativity in an open-ended way, but instantiates a predefined schema that makes the benchmark more systematic, auditable, and extensible.

\paragraph{Why the Benchmark Still Measures More Than Retrieval.}
We agree that physical commonsense is a necessary ingredient, but the benchmark requires more than retrieving a known object-use association. Each task is built so that the model must infer a \textbf{non-obvious} but physically grounded connection among a goal, a candidate part, relevant attributes, and conditions of use. In other words, the model must move from surface familiarity to \textbf{mechanism-sensitive repurposing}. This is precisely why performance is much lower on the full gold selection than on coarse entity selection. The benchmark therefore targets a narrower and more operational notion of creativity: discovering a usable, constraint-compatible, non-canonical function from latent affordance structure.

\paragraph{Single-Gold Structure as a Deliberate Evaluation Choice.}
Our benchmark does not simply assume the uniqueness of the sampled gold affordance as the solution for the benchmark task. Our construction pipeline explicitly includes \textbf{gold verification} through intra-entity and inter-entity comparison, with candidate tasks rejected or filtered when an alternative is judged preferable. The purpose of this step is not to claim that every real-world problem has exactly one valid solution, but to construct a controlled evaluation set where one solution path is sufficiently well-supported to enable rigorous comparison across models. In this sense, the single-gold structure is a methodological choice for \textbf{measurement clarity}, not a denial of the open-ended nature of creativity.

\paragraph{Interpreting Human Performance Carefully.}
The human results may appear surprising at first glance, since humans do not outperform the best model on the strict gold metric. We do not interpret this as evidence that the benchmark is misaligned with human reasoning. Rather, it reflects an important property of the current evaluation format: humans receive compact textual descriptions of entities, parts, and attributes instead of direct perceptual access to objects. This makes the task more like \textbf{symbolic affordance inspection} than everyday embodied improvisation. Importantly, this does not weaken the benchmark's diagnostic value. Instead, it shows that the benchmark isolates a specific reasoning challenge: identifying grounded affordances from structured descriptions, a task that remains cognitively demanding for humans in the absence of perceptual shortcuts. That makes the setting a meaningful stress test rather than an easy commonsense exercise.

\paragraph{Domain Coverage and the Value of a Focused Household Setting.}
We view our benchmark's focused scope is a strength at the current stage. Household environments provide rich, diverse, and widely understandable object ecosystems in which creative repurposing naturally arises. By restricting the domain, we can control entity distributions, affordance granularity, and distractor sampling far more carefully than would be possible in a highly heterogeneous open-domain benchmark. This improves internal validity and makes failure patterns easier to interpret. We therefore present CreativityBench not as the final word on all forms of creativity, but as a principled and scalable benchmark for one important, grounded subtype of creative problem solving.

\paragraph{Cross-Model Comparisons Should Be Read Structurally, Not Just Numerically.}
Differences across model families can be sensitive to prompting and implementation choices, as is true of most benchmark comparisons. Our central conclusions, however, do not rely on a brittle ordering between two closely matched systems. Instead, the paper highlights several structural patterns that remain consistent across models: a substantial drop from entity-level to part-level accuracy, weaker physical grounding than action plausibility, pronounced sensitivity to affordance commonality, and limited improvement from standard inference-time interventions. These recurring trends are more informative than any single leaderboard position. Accordingly, the benchmark’s main contribution is not simply to show that one model outperforms another, but to uncover a shared failure mode that appears across both proprietary and open-source model families.

\paragraph{Final Take Away.}
CreativityBench is intentionally scoped, structurally grounded, and diagnostically designed. It does not aim to solve every ambiguity in evaluating creativity, nor does it claim that creative intelligence can be reduced to one benchmark. Our contribution is more precise: we isolate an important form of creative tool use, formalizes it through affordance-based structure, and shows that current models remain substantially weaker at this capability than their general reasoning progress might suggest. In our view, this combination of conceptual focus, structured construction, and consistent empirical findings is exactly what makes the benchmark useful.

\section{Preliminary Experiments Details}
\label{appendix_sec:prelim_prompt}

\subsection{Core Prompt Details}
\begin{tcolorbox}[
  enhanced,
  breakable,
  width=0.98\linewidth,
  colback=tzBlueFill,
  colframe=tzBlueBorder,
  boxrule=1.2pt,
  arc=6pt,
  left=5pt,right=5pt,top=4pt,bottom=2pt,
  title={\small Directive Prompt for Creative Tool-Use},
  coltitle=white,
  colbacktitle=tzBlueHeader2,
  fonttitle=\bfseries,
]
\small
\begin{lstlisting}[style=jsonTiny]
You are an expert at creative physical tool-use reasoning.
Given the task description below, produce a feasible solution.

Rules:
1. Use only tools/items explicitly available in the task description.
2. Respect physical constraints if the task description restricts physical attributes such as size or state.
3. Invent new tools using tools/items explicitly available in the task description if it is needed.
4. Provide practical steps that can actually be executed.
5. Do not involve any unnecessary steps to achieve the task's goal.
6. If a complete solution is impossible, return the best partial plan and explain why it cannot be completed.

TASK DESCRIPTION:
{problem}

Return JSON:
{{
  "solvable": "Yes or No",
  "solvable_explanation": "1-3 sentences about why the given task is solvable or not", 
  "solution_steps": ["Step 1: ...", "Step 2: ...", ...],
  "final_solution": "One concise paragraph that summarizes the full approach.",
  "used_tools": ["tool 1", "tool 2", ...],
  "constraint_handling": [{{"constraint": "...", "handling": "..."}}, ...]
}}
\end{lstlisting}
\end{tcolorbox}

\begin{tcolorbox}[
  enhanced,
  breakable,
  width=0.98\linewidth,
  colback=tzBlueFill,
  colframe=tzBlueBorder,
  boxrule=1.2pt,
  arc=6pt,
  left=5pt,right=5pt,top=4pt,bottom=2pt,
  title={\small CoT Prompt for Creative Tool-Use},
  coltitle=white,
  colbacktitle=tzBlueHeader2,
  fonttitle=\bfseries,
]
\small
\begin{lstlisting}[style=jsonTiny]
You are an expert at creative physical tool-use reasoning.
Solve the task by explicitly reasoning over tool parts and affordances under constraints.

Rules:
1. Use only tools/items explicitly available in the task description.
2. Respect physical constraints if the task description restricts physical attributes such as size or state.
3. Invent new tools using tools/items explicitly available in the task description if it is needed.
4. Provide practical steps that can actually be executed.
5. Do not involve any unnecessary steps to achieve the task's goal.
6. If a complete solution is impossible, return the best partial plan and explain why it cannot be completed.

Required reasoning procedure:
1. State the task goal and concrete success condition.
2. List all available tools/items from the prompt (no additions).
3. For each relevant tool, identify the key part(s), infer physical properties, and derive part-level affordances useful for this task.
4. Build a step-by-step plan where each step references tool parts and the affordance being used.
5. Validate each step against stated constraints (e.g., broken/unusable items, size mismatch, blocked function, state limitations).
6. Keep the plan practical and minimal with no unnecessary actions.

TASK DESCRIPTION:
{problem}

Return JSON:
{{
  "task_goal": "...",
  "success_condition": "...",
  "identified_constraints": ["...", "..."],
  "tool_inventory": [
    {{
      "tool": "...",
      "relevant_parts": [
        {{
          "part": "... or NA",
          "inferred_physical_properties": ["...", "..."],
          "affordances_for_task": ["...", "...", ...],
          "usable_under_constraints": "Yes or No",
        }},
        ...
      ]
    }},
    ...
  ],
  "reasoning_plan": [
    {{
      "step": 1,
      "action": "...",
      "tool_parts_used": ["tool:part", "..."],
      "affordance_used": ["...", "..."],
      "explanation": "1 sentence about why it works"
    }},
    {{
      "step": 2,
      "action": "...",
      "tool_parts_used": ["tool:part", "..."],
      "affordance_used": ["...", "..."],
      "explanation": "1 sentence about why it works"
    }},
    ...
  ],
  "solvable": "Yes or No",
  "solvable_explanation": "1-3 sentences about why the given task is solvable or not", 
  "solution_steps": ["Step 1: ...", "Step 2: ...", ...],
  "final_solution": "One concise paragraph that summarizes the full approach.",
  "used_tools": ["tool 1", "tool 2", ...],
  "constraint_handling": [{{"constraint": "...", "handling": "..."}}, ...]
  "creative_reasoning_summary": "1-3 sentences about novelty and practicality."
}}
\end{lstlisting}
\end{tcolorbox}

\begin{tcolorbox}[
  enhanced,
  breakable,
  width=0.98\linewidth,
  colback=tzBlueFill,
  colframe=tzBlueBorder,
  boxrule=1.2pt,
  arc=6pt,
  left=5pt,right=5pt,top=4pt,bottom=2pt,
  title={\small Prompt for LLM-as-Judge Absolute Evaluation: Correctness},
  coltitle=white,
  colbacktitle=tzBlueHeader2,
  fonttitle=\bfseries,
]
\small
\begin{lstlisting}[style=jsonTiny]
You are an expert evaluator for creative physical tool-use reasoning quality.
Your task is to evaluate the CORRECTNESS of a candidate solution against a ground-truth solution.

**DEFINITION**: Correctness measures whether the solution actually solves the stated task goal. Mark it as correct if the solution reaches the task goal even if it is different from ground-truth solution.

**SCORING RUBRIC (1-5)**:
- 1 = Completely wrong. The solution does not address the task at all, solves a different problem, or produces an outcome opposite to what is needed.
    - Example: Task asks to "open a locked door using available tools" and the candidate describes painting the door.
- 2 = In between 1 and 3
- 3 = Partially correct. The solution moves toward the goal but is incomplete, solves only a sub-part of the task, or requires unstated assumptions to work. The core objective is addressed but not fully achieved.
    - Example: Task asks to "move a heavy boulder off a path" and the candidate correctly identifies using a lever but never specifies where to place the fulcrum, leaving the solution unresolved.
- 4 = In between 3 and 4
- 5 = Fully correct. The solution completely and unambiguously achieves the task goal. All steps are logically consistent and lead to the desired outcome.
    - Example: Task asks to "lift a car to change a tire without a jack" and the candidate correctly uses a sturdy plank wedged under the frame with a rock as fulcrum, applies body weight, and secures the car before swapping the tire.

**TASK DESCRIPTION**:
{problem}

**GROUND-TRUTH SOLUTION**:
{ground_truth_solution}

**CANDIDATE SOLUTION**:
{candidate_solution}

**CANDIDATE SOLUTION USED TOOLS**:
{used_tools}

Return JSON:
{{
  "correct": "Yes or No",
  "correct_score": 1-5,
  "correct_rationale": "1-2 sentences for rationale."
}}
\end{lstlisting}
\end{tcolorbox}

\begin{tcolorbox}[
  enhanced,
  breakable,
  width=0.98\linewidth,
  colback=tzBlueFill,
  colframe=tzBlueBorder,
  boxrule=1.2pt,
  arc=6pt,
  left=5pt,right=5pt,top=4pt,bottom=2pt,
  title={\small Prompt for LLM-as-Judge Absolute Evaluation: Feasibility},
  coltitle=white,
  colbacktitle=tzBlueHeader2,
  fonttitle=\bfseries,
]
\small
\begin{lstlisting}[style=jsonTiny]
You are an expert evaluator for creative physical tool-use reasoning quality.
Your task is to evaluate the FEASIBILITY of a candidate solution against a ground-truth solution.

**DEFINITION**: Feasibility measures whether the solution can be physically executed under the stated constraints.

**SCORING RUBRIC (1-5)**:
- 1 = Completely infeasible. The solution requires violating a stated constraint, ignores key physical limits, or depends on impossible actions.
    - Example: Constraint says "no cutting tools available" and the candidate instructs cutting a rope in half.
- 2 = In between 1 and 3
- 3 = Mostly feasible. The solution is executable in principle but contains one or two steps that stretch physical plausibility or lightly brush against a constraint without outright violating it. Would work under ideal or slightly relaxed conditions.
    - Example: TSolution requires balancing a long beam on a narrow pivot point -- physically possible but practically difficult without more stabilization.
- 4 = In between 3 and 4
- 5 = Fully feasible. Every step can be realistically performed by a person with ordinary strength and skill using only the available tools, within all stated constraints. No step requires superhuman effort or ignores stated limitations.
    - Example: All actions use objects explicitly listed, forces are within human capability, and constraints (e.g., "no power tools," "indoors only") are strictly respected throughout.

**TASK DESCRIPTION**:
{problem}

**GROUND-TRUTH SOLUTION**:
{ground_truth_solution}

**CANDIDATE SOLUTION**:
{candidate_solution}

**CANDIDATE SOLUTION USED TOOLS**:
{used_tools}

Return JSON:
{{
  "feasible": "Yes or No",
  "feasible_score": 1-5,
  "feasible_rationale": "1-2 sentences for rationale."
}}
\end{lstlisting}
\end{tcolorbox}

\begin{tcolorbox}[
  enhanced,
  breakable,
  width=0.98\linewidth,
  colback=tzBlueFill,
  colframe=tzBlueBorder,
  boxrule=1.2pt,
  arc=6pt,
  left=5pt,right=5pt,top=4pt,bottom=2pt,
  title={\small Prompt for LLM-as-Judge Absolute Evaluation: Physical Grounding},
  coltitle=white,
  colbacktitle=tzBlueHeader2,
  fonttitle=\bfseries,
]
\small
\begin{lstlisting}[style=jsonTiny]
You are an expert evaluator for creative physical tool-use reasoning quality.
Your task is to evaluate the PHYSICAL GROUNDING of a candidate solution against a ground-truth solution.

**DEFINITION**: Physical grounding measures whether the solution correctly leverages the physical properties, geometry, and mechanics of the objects involved.

**SCORING RUBRIC (1-5)**:
- 1 = Physically nonsensical. The solution misuses objects in ways that contradict their basic physical properties (e.g., using a soft cloth as a rigid lever, treating a hollow tube as load-bearing).
    - Example: "Use the inflatable pool toy as a wedge to hold the car axle in place" -- an inflatable toy cannot bear that load.
- 2 = In between 1 and 3
- 3 = Partially grounded. Most physical reasoning is sound, but one or two steps misidentify a relevant property (e.g., ignoring friction, misjudging weight distribution, or using the wrong part of an object).
    - Example: Correctly identifies a metal rod as a lever but places the fulcrum at the rod's midpoint when the task geometry requires it near one end for sufficient mechanical advantage.
- 4 = In between 3 and 4
- 5 = Fully grounded. The solution correctly identifies and exploits the relevant physical attributes of every tool (material strength, shape, weight, surface texture, rigidity, etc.) and applies correct mechanical principles (leverage, friction, torque, buoyancy, etc.) throughout.
    - Example: "Insert the flat-head screwdriver under the lid's lip -- the thin, rigid blade provides a moment arm; pressing down on the handle rotates the blade and pries the lid upward with minimal force."

**TASK DESCRIPTION**:
{problem}

**GROUND-TRUTH SOLUTION**:
{ground_truth_solution}

**CANDIDATE SOLUTION**:
{candidate_solution}

**CANDIDATE SOLUTION USED TOOLS**:
{used_tools}

Return JSON:
{{
  "physical_grounding": "Yes or No",
  "physical_grounding_score": 1-5,
  "physical_grounding_rationale": "1-2 sentences citing specific object properties or mechanics addressed or missed."
}}
\end{lstlisting}
\end{tcolorbox}

\begin{tcolorbox}[
  enhanced,
  breakable,
  width=0.98\linewidth,
  colback=tzBlueFill,
  colframe=tzBlueBorder,
  boxrule=1.2pt,
  arc=6pt,
  left=5pt,right=5pt,top=4pt,bottom=2pt,
  title={\small Prompt for LLM-as-Judge Absolute Evaluation: Constraint Coverage},
  coltitle=white,
  colbacktitle=tzBlueHeader2,
  fonttitle=\bfseries,
]
\small
\begin{lstlisting}[style=jsonTiny]
You are an expert evaluator for creative physical tool-use reasoning quality.
Your task is to evaluate the CONSTRAINT COVERAGE of a candidate solution against a ground-truth solution.

**DEFINITION**: Constraint coverage measures whether the solution explicitly handles every constraint stated in the task.

**SCORING RUBRIC (1-5)**:
- 1 = Ignores constraints. One or more hard constraints are violated outright or never acknowledged.
    - Example: Task says "use only items found in the kitchen" and the candidate fetches a tool from a garage.
- 2 = In between 1 and 3
- 3 = Partial coverage. Most constraints are respected but one is violated or only implicitly handled without explanation. The solution would likely still work but shows incomplete awareness.
    - Example: Task says "complete the task silently and without electricity" -- the candidate avoids electric tools but produces loud banging with no acknowledgment of the noise constraint.
- 4 = In between 3 and 4
- 5 = Full coverage. Every constraint is explicitly acknowledged and the solution demonstrates a clear, deliberate strategy for satisfying each one. No constraint is left unaddressed or violated.
    - Example: "Since we cannot use water (constraint 1) and must work in under two minutes (constraint 2), I will use dry sand to smother the flame and apply it immediately using the nearby bucket."

**TASK DESCRIPTION**:
{problem}

**GROUND-TRUTH SOLUTION**:
{ground_truth_solution}

**CANDIDATE SOLUTION**:
{candidate_solution}

**CANDIDATE SOLUTION USED TOOLS**:
{used_tools}

Return JSON:
{{
  "constraint_coverage": "Yes or No",
  "constraint_coverage_score": 0-5,
  "constraint_coverage_rationale": "1-2 sentences identifying which constraints were met or violated."
}}
\end{lstlisting}
\end{tcolorbox}

\begin{tcolorbox}[
  enhanced,
  breakable,
  width=0.98\linewidth,
  colback=tzBlueFill,
  colframe=tzBlueBorder,
  boxrule=1.2pt,
  arc=6pt,
  left=5pt,right=5pt,top=4pt,bottom=2pt,
  title={\small Prompt for LLM-as-Judge Absolute Evaluation: Tool Usage},
  coltitle=white,
  colbacktitle=tzBlueHeader2,
  fonttitle=\bfseries,
]
\small
\begin{lstlisting}[style=jsonTiny]
You are an expert evaluator for creative physical tool-use reasoning quality.
Your task is to evaluate the TOOL USAGE of a candidate solution against a ground-truth solution.

**DEFINITION**: Tool usage measures whether the solution uses only the tools and objects explicitly available in the task description, and whether each tool is used in a sensible, purposeful way. You can also refer to CANDIDATE SOLUTION USED TOOLS to evaluate the score.

**SCORING RUBRIC (1-5)**:
- 1 = Fabricates tools. The solution relies on objects not listed in the task or invents tools out of thin air.
    - Example: Task lists "rope, wooden plank, and a brick" and candidate says "use the pulley system" -- no pulley was provided.
- 2 = In between 1 and 3
- 3 = Mostly within bounds with minor overreach. The solution uses available tools correctly for the most part but includes one object that is not explicitly listed yet could plausibly be assumed present.
    - Example: "the ground" as a fulcrum when only portable objects were listed.
- 4 = In between 3 and 4
- 5 = Perfect tool fidelity. Every object used is explicitly listed in the task description. Each tool is used for a clear, stated purpose and no phantom tools are introduced. The solution makes full, purposeful use of what is available without waste or omission.
    - Example: Task provides "a broom, two bricks, and a length of string" and the candidate uses objects presented in the task in a coherent plan that references each by name and function.

**TASK DESCRIPTION**:
{problem}

**GROUND-TRUTH SOLUTION**:
{ground_truth_solution}

**CANDIDATE SOLUTION**:
{candidate_solution}

**CANDIDATE SOLUTION USED TOOLS**:
{used_tools}

Return JSON:
{{
  "tool_usage": "Yes or No",
  "tool_usage_score": 1-5,
  "tool_usage_rationale": "1-2 sentences noting any phantom tools or unused available tools.",
}}
\end{lstlisting}
\end{tcolorbox}

\begin{tcolorbox}[
  enhanced,
  breakable,
  width=0.98\linewidth,
  colback=tzBlueFill,
  colframe=tzBlueBorder,
  boxrule=1.2pt,
  arc=6pt,
  left=5pt,right=5pt,top=4pt,bottom=2pt,
  title={\small Prompt for LLM-as-Judge Absolute Evaluation: Creativity},
  coltitle=white,
  colbacktitle=tzBlueHeader2,
  fonttitle=\bfseries,
]
\small
\begin{lstlisting}[style=jsonTiny]
You are an expert evaluator for creative physical tool-use reasoning quality.
Your task is to evaluate the CREATIVE REASONING of a candidate solution against a ground-truth solution.

**DEFINITION**: Creative reasoning measures how inventively the solution repurposes or combines available tools beyond their obvious, default use.

**SCORING RUBRIC (1-5)**:
- 1 = Low Creativity. The solution follows an obvious, expected approach with no novel tool-use.
    - Example: "Use the hammer to drive the nail."
- 2 = In between 1 and 3
- 3 = Medium Creativity. The solution repurposes at least one object in a non-obvious way, or combines tools in a moderately clever configuration that goes beyond the most straightforward reading of the task.
    - Example: Using a belt as an improvised tourniquet and winding peg.
- 4 = In between 3 and 4
- 5 = High Creativity. The solution demonstrates genuinely inventive thinking: unexpected combinations of objects, multi-step mechanical chains, or insights that exploit subtle physical properties most people would overlook. The approach is surprising yet clearly sound.
    - Example: Using a glass of water as a lens to focus sunlight onto tinder, combined with a foil wrapper as a reflector to intensify the beam.

**TASK DESCRIPTION**:
{problem}

**GROUND-TRUTH SOLUTION**:
{ground_truth_solution}

**CANDIDATE SOLUTION**:
{candidate_solution}

**CANDIDATE SOLUTION USED TOOLS**:
{used_tools}

Return JSON:
{{
  "creative_reasoning": "Yes or No",
  "creativity_score": 1-5,
  "creativity_rationale": "1-2 sentences explaining what makes the approach creative or conventional."
}}
\end{lstlisting}
\end{tcolorbox}

\begin{tcolorbox}[
  enhanced,
  breakable,
  width=0.98\linewidth,
  colback=tzBlueFill,
  colframe=tzBlueBorder,
  boxrule=1.2pt,
  arc=6pt,
  left=5pt,right=5pt,top=4pt,bottom=2pt,
  title={\small Prompt for LLM-as-Judge Relative Evaluation},
  coltitle=white,
  colbacktitle=tzBlueHeader2,
  fonttitle=\bfseries,
]
\small
\begin{lstlisting}[style=jsonTiny]
You are an expert evaluator for creative physical tool-use reasoning quality.
Your task is to compare TWO candidate solutions for the SAME task, using the ground-truth solution as reference. 

IMPORTANT:
- You are NOT checking wording similarity. Judge functional quality and practical validity.
- A solution can be strong even if it differs from the ground-truth approach.
- Use the following criteria:
    - Correctness: Whether the plan actually solves the task objective.
    - Feasibility: Whether the plan can be physically executed under stated constraints.
    - Physical Grounding: Whether it uses realistic object properties/mechanics correctly.
    - Constraint Coverage: Whether it handles all explicit constraints.
    - Tool Usage: Whether it uses only available tools appropriately and purposefully.
    - Creative Reasoning: Novel, non-obvious but valid repurposing/combination of tools.
    - Overall: Holistic quality across all above dimensions.

For EACH criterion:
1. Determine winner:
   - "win" if solution1_score > solution2_score
   - "lose" if solution2_score > solution1_score
   - "tie" if both are equal
2. Give a short rationale (1-2 sentences).

TASK DESCRIPTION:
{problem}

GROUND-TRUTH SOLUTION:
{ground_truth_solution}

CANDIDATE SOLUTION1 (DEFAULT PROMPT OUTPUT):
{solution1}

CANDIDATE SOLUTION2 (COT PROMPT OUTPUT):
{solution2}

Return STRICT JSON:
{{
  "correctness": {{
    "winner": "win or lose or tie",
    "rationale": "1-2 sentences."
  }},
  "feasibility": {{
    "winner": "win or lose or tie",
    "rationale": "1-2 sentences."
  }},
  "physical_grounding": {{
    "winner": "win or lose or tie",
    "rationale": "1-2 sentences."
  }},
  "constraint_coverage": {{
    "winner": "win or lose or tie",
    "rationale": "1-2 sentences."
  }},
  "tool_usage": {{
    "winner": "win or lose or tie",
    "rationale": "1-2 sentences."
  }},
  "creativity": {{
    "winner": "win or lose or tie",
    "rationale": "1-2 sentences."
  }},
  "overall": {{
    "winner": "win or lose or tie",
    "rationale": "1-2 sentences."
  }},
  "short_summary": "2-4 sentences summarizing key tradeoffs between default and CoT."
}}
\end{lstlisting}
\end{tcolorbox}


\subsection{Textual v.s. Visual Grounding in Creative Tool Use}

We showed that adding an affordance-decomposition CoT scaffold does not reliably improve creative reasoning in text-only tasks, suggesting the bottleneck is not ``missing step-by-step procedure'' but limited grounded affordance representations and compositional recombination (\Cref{sec:preliminaries}). Unlike text-only tasks, which explicitly specify the task goal, available tools, and physical constraints in symbolic form, we further examine whether the influence of affordance-level CoT persists under a visual+text setting. In this multimodal condition, the text provides only the task goal and non-visible constraints, while the image presents available tools and visible physical constraints. Since affordances are never given at inference time, the model must infer object attributes and derive affordances internally in both settings; however, the visual condition requires extracting observable properties (e.g., shape, relative size, geometry) from perceptual input rather than reading them directly from text.

To disentangle these factors, we run a 2×2 ablation over the same model (GPT-4.1-mini): input modality (Text-only vs. Vision+Text) × prompting strategy (Direct vs. Affordance-CoT). In the text-only setting, the description explicitly lists the goal, available tools, and constraints. In the vision+text setting, the text specifies only the goal and non-visible constraints (e.g., temperature), while the image conveys available objects and visible constraints. This design isolates whether performance gains come from (1) richer grounding signals, (2) reasoning scaffolds, or (3) their interaction.


\section{Annotation Pipeline Details}
\label{appendix_sec:annotation}

\begin{table*}[t]
    \centering
    \scriptsize
    \setlength\tabcolsep{5pt}
    \setlength\extrarowheight{2pt}
    \resizebox{0.9\linewidth}{!}{
    \begin{tabular}{l @{\hspace{5mm}} c @{\hspace{5mm}} c @{\hspace{5mm}} c @{\hspace{5mm}} c @{\hspace{5mm}} c}
        \toprule
        \textbf{Scenes} & \textbf{Entities} & \textbf{Parts} & \textbf{Physical Attributes} & \textbf{State Attributes} & \textbf{Affordances} \\
        \midrule
        Bathroom     & 483 & 3,264 & 35,859 & 15,551 & 19,584 \\
        Bedroom      & 468 & 3,447 & 37,896 & 16,311 & 20,682 \\
        Dining Room  & 477 & 3,099 & 34,044 & 14,581 & 18,594 \\
        Garage       & 474 & 3,393 & 37,254 & 16,349 & 20,357 \\
        Garden       & 477 & 3,087 & 33,909 & 14,881 & 18,522 \\
        Home Office  & 468 & 3,489 & 38,373 & 16,526 & 20,934 \\
        Kitchen      & 498 & 3,111 & 34,188 & 14,887 & 18,666 \\
        Living Room  & 471 & 3,348 & 36,795 & 15,886 & 20,088 \\
        \midrule
        \textbf{Total} & \textbf{3,816} & \textbf{26,238} & \textbf{288,318} & \textbf{124,972} & \textbf{157,427} \\
        \bottomrule
    \end{tabular}
    }
    \caption{Scene-wise statistics of the affordance knowledge base. Each scenario is annotated with entities, parts, physical attributes, state attributes, and affordances.}
    \label{tab:scenario_statistics}
\end{table*}

\subsection{Overview of Annotation Stages}
Our annotation pipeline is a staged, LLM-assisted procedure that transforms a set of environment scenarios into richly annotated object-part datasets for creativity-oriented affordance analysis. The process proceeds from entity grounding to structural decomposition, then to physical and state characterization, and finally to part-level functional affordance annotation and full-entity assembly. Each stage consumes the output of the previous stage, so that downstream annotations are grounded in upstream structure and attributes rather than generated independently.

\paragraph{Stage A: Scenario-Grounded Entity Generation.}
We begin with a predefined set of eight everyday indoor scenarios including Kitchen, Living Room, Bedroom, Bathroom, Garage, Home Office, Dining Room, Garden. For each scenario, we sample a fixed number of \emph{specific single objects} (not broad categories and not complex multi-object systems). To maintain diversity while reducing near-duplicates, we apply semantic similarity filtering within each scenario. This step provides a broad but controlled inventory of entities for subsequent annotation.

\paragraph{Stage B: Partonomy Construction.}
For each entity, we annotate a compact but comprehensive partonomy graph. Each entity is decomposed into essential, non-overlapping parts that together cover the object. For every part, we additionally annotate connectivity relations and concise functional/structural descriptions that explain how the part is situated in the whole object. This representation serves as the structural backbone for all later attribute and affordance annotation.

\paragraph{Stage C: Physical Attribute Annotation.}
For each part, we annotate multiple plausible \textbf{physical} attributes (rather than a single fixed description), including geometry, size/thickness, local features, material, rigidity, durability, elasticity, surface, weight, and concise summary notes. The objective is to represent realistic diversity in how the same part may manifest in practice, while keeping each variant internally consistent.

\paragraph{Stage D: Physical Variant Composition.}
Part-level physical variants are combined into entity-level physical configurations. When combinatorics are manageable, all combinations are retained; when the product is too large, combinations are subsampled with coverage-oriented selection so that each part’s alternatives are represented. This yields multiple physically grounded versions of each entity.

\paragraph{Stage E: State Attribute Annotation.}
Conditioned on each physical configuration, we annotate part-level \textbf{state} attributes: accessibility (visibility and availability), condition (moisture and temperature), internal state (e.g., empty/filled where applicable), and summary notes. Multiple plausible states are generated per part to reflect ordinary and less typical but still realistic conditions.

\paragraph{Stage F: State Variant Composition.}
We then combine part-level state variants into entity-level state configurations, again with bounded combinatorial growth. The result is a set of entity variants where each part has both physical and state descriptors, enabling downstream affordance annotation under explicit conditions.

\paragraph{Stage G: Functional Affordance Annotation.}
For each part in each entity variant, we annotate a diverse set of functional affordances grounded in the already-annotated physical and state attributes. Each affordance includes:
(i) use condition,
(ii) environment condition,
(iii) attribute evidence (physical/state and whether best conveyed visually or textually),
(iv) affordance description,
(v) level annotation (one intended/normal use and graded emergency alternatives),
(vi) recipient condition,
(vii) concrete recipient examples, and
(viii) failure cases.
This design enforces condition-aware, recipient-aware, and failure-aware affordance descriptions rather than unconstrained brainstorming.

\paragraph{Stage H: Entity Assembly and Consistency Check.}
Finally, part-level outputs are reassembled into complete entity records. Only entities with complete part coverage are retained, ensuring structural consistency between early decomposition and final affordance annotations. The final artifact is a complete, readable dataset of entities, parts, attributes, and affordances.

\paragraph{Prompting Strategy.}
We use task-specific prompts for each stage, with strict schema constraints and explicit instructions to promote diversity, plausibility, and consistency. These prompts are carefully designed to minimize ambiguity, ensure machine-parsable outputs, and preserve descriptive richness. All annotations are generated by GPT-5.2 under human supervision and quality control. Before scaling up annotation, we perform iterative human prompt refinement to improve annotation quality and ensure the prompts reliably produce the desired outputs. After the full annotation process, we conduct quality assessment by sampling 0.1\% of the resulting data. This evaluation shows a pass rate of approximately 98\% under automatic LLM-judge assessment and 95\% under human review. Here, quality is defined in terms of commonsense validity, groundedness, and factual consistency across the entire annotation pipeline. For more details about the annotations, please refer to \Cref{tab:scenario_statistics}.


\subsection{Core Hyperparameters Details}

\begin{table}[!h]
\centering
\small
\setlength\tabcolsep{10pt}
\setlength\extrarowheight{2pt}
\begin{tabular}{lc}
\toprule
\textbf{Hyperparameter} & \textbf{Value} \\
\midrule
Number of scenarios & 8 \\
Sampled entities per scenario & 50 \\
Semantic similarity threshold (entity deduplication) & 0.85 \\
Physical variants generated per part (target range) & 2--3 \\
State variants generated per part (target range) & 2--3 \\
\bottomrule
\end{tabular}
\caption{Core hyperparameters of data-generation settings.}
\label{tab:annotation-core-hparams}
\end{table}

\begin{table}[!h]
\centering
\small
\setlength\tabcolsep{10pt}
\setlength\extrarowheight{2pt}
\begin{tabular}{lc}
\toprule
\textbf{Hyperparameter} & \textbf{Value} \\
\midrule
Maximum parts per entity (partonomy cap) & 8 \\
Maximum physical combinations per entity & 8 \\
Maximum state combinations per physical entity variant & 6 \\
Maximum final variants per original entity & 48 \\
Affordances generated per part & 6 \\
Affordance level structure & Normal 0 + Emergency 1--5 \\
\bottomrule
\end{tabular}
\caption{Core hyperparameters of combinatorial and annotation controls for coverage and tractability.}
\label{tab:annotation-combination-hparams}
\end{table}

\begin{table}[!h]
\centering
\small
\setlength\tabcolsep{10pt}
\setlength\extrarowheight{2pt}
\begin{tabular}{lc}
\toprule
\textbf{Hyperparameter} & \textbf{Value} \\
\midrule
Generation temperature & 0.7 \\
Parallel generation (part/attribute/affordance stages) & up to 1024 workers \\
Incremental save interval (other generation stages) & up to 50 generations \\
\bottomrule
\end{tabular}
\caption{Core hyperparameters of inference-time settings for scalable generation and fault-tolerant processing.}
\label{tab:annotation-inference-hparams}
\end{table}


\subsection{Core Prompt Details}
\begin{tcolorbox}[
  enhanced,
  breakable,
  width=0.98\linewidth,
  colback=tzBlueFill,
  colframe=tzBlueBorder,
  boxrule=1.2pt,
  arc=6pt,
  left=5pt,right=5pt,top=4pt,bottom=2pt,
  title={\small System Prompt for Entity Decomposition Generation},
  coltitle=white,
  colbacktitle=tzBlueHeader2,
  fonttitle=\bfseries,
]
\small
\begin{lstlisting}[style=jsonTiny]
You are an expert at analyzing object structure and component parts.

**Task:** Annotate the partonomy graph for "{entity_name}" (found in {scenario}).

**Requirements:**
1. List ALL the common parts (both externally visible and internally hidden ones)
2. Limit to 8 parts maximum, can be less if the object is simple and parts are obvious
3. Your parts should better have explicit boundaries which divide the object's different function units or affordance mechanisms
4. Include components that may have useful affordances themselves (e.g. screws, batteries, etc.).
5. Parts must be non-overlapping and cover the whole object (e.g. a blade contains cutting edge, so it's one part).
6. Imagine a specific entity when you annotate the parts, don't include any optional parts or uncertain wording, all parts should be necessary and essential to the entity's function.
7. Parts with same function differing only by position or direction should be annotated as ONE part
8. For each part: specify connected parts and describe function/connection

**Schema:**
```json
{{
  "entity_name": "...",
  "parts": ["part1", "part2", ...],
  "relations": {{
    "part1": {{
      "connected_to": ["part2", "part3"],
      "connection": "Description of part1, its function, and connection"
    }},
    ...
  }}
}}
```

**Example (prescription reading glasses):**
```json
{{
  "entity_name": "prescription reading glasses",
  "parts": ["front_frame", "lenses", "hinge_mechanisms", "temple_arms", "temple_tips"],
  "relations": {{
    "front_frame": {{
      "connected_to": ["lenses", "hinge_mechanisms"],
      "connection": "Main rigid front structure; provides lens openings/retention geometry; provides hinge mounting points."
    }},
    "lenses": {{
      "connected_to": ["front_frame"],
      "connection": "Two optical elements seated/retained by the front frame."
    }},
    ...
  }}
}}
```

First imagine a specific entity when you annotate the parts, and then provide reasoning about its structure and key parts, and then output the parts and relations JSON.
\end{lstlisting}
\end{tcolorbox}

\begin{tcolorbox}[
  enhanced,
  breakable,
  width=0.98\linewidth,
  colback=tzBlueFill,
  colframe=tzBlueBorder,
  boxrule=1.2pt,
  arc=6pt,
  left=5pt,right=5pt,top=4pt,bottom=2pt,
  title={\small System Prompt for Physical Attributes Annotation},
  coltitle=white,
  colbacktitle=tzBlueHeader2,
  fonttitle=\bfseries,
]
\small
\begin{lstlisting}[style=jsonTiny]
You are an expert at analyzing physical properties of object parts.

**Task:** Annotate physical attributes for the "{part_name}" of "{entity_name}".

**Context:**
- All parts: {all_parts}
- Connected to: {connected_to}
- Connection: {connection}

**Annotate these attributes:**
1. **Geometry & Shape:** shape, size, thickness (thin/medium/thick/NA), local_features
2. **Material & Structural:** material, rigidity (very rigid/rigid/semi-rigid/flexible/soft), durability (very fragile/fragile/normal/sturdy/very sturdy), elasticity (non-elastic/springy/stretchable/very stretchable), surface
3. **Mass:** weight (very light/light/moderate/heavy/very heavy)
4. **Others:** Other important attributes for affordance (1-2 sentences)
5. **Summary:** Comprehensive summary of all attributes

**Requirements:**
- Generate 2-3 physical attribute combinations for this part
- Each combination must be plausible, internally consistent, and diverse
- Include common variations (e.g., plastic vs steel)
- Include one unusual but plausible variation if it creates distinctive affordances
- All fields required; use "NA" only if truly not important for creative affordance
- Be assertive; avoid "might", "maybe", "could be"
- Ensure consistency (e.g., plastic -> lighter, steel -> heavier)

**Schema:**
```json
[
  {{
    "shape": "...",
    "size": "...",
    "thickness": "...",
    "local_features": "...",
    "material": "...",
    "rigidity": "...",
    "durability": "...",
    "elasticity": "...",
    "surface": "...",
    "weight": "...",
    "others": "...",
    "summary": "..."
  }}
]
```

**Example (glasses front_frame):**
```json
[
  {{
    "shape": "two connected loops and narrow bridge",
    "size": "hand-held",
    "thickness": "thin",
    "local_features": "two enclosed openings and flat front edge and narrow bridge span",
    "material": "plastic",
    "rigidity": "rigid",
    "durability": "normal",
    "elasticity": "non-elastic",
    "surface": "smooth",
    "weight": "very light",
    "others": "continuous rim can hook or hang on thin supports; bridge provides pinchable grip point",
    "summary": "A lightweight thin rigid plastic double-loop rim with enclosed openings and a pinchable bridge, smooth surfaced and easy to hang or hold."
  }},
  {{
    "shape": "two connected loops and narrow bridge",
    "size": "hand-held",
    "thickness": "thin",
    "local_features": "metal rim edge and small screw openings",
    "material": "steel",
    "rigidity": "very rigid",
    "durability": "sturdy",
    "elasticity": "non-elastic",
    "surface": "smooth and slightly cool",
    "weight": "light",
    "others": "conductive metal edge can transfer heat or cold; thin rim can fit into narrow slots",
    "summary": "A thin sturdy metal double-loop rim with small openings and a narrow bridge, rigid and thermally conductive with smooth edges."
  }}
]
```

First provide reasoning. Then output JSON array of 1-5 combinations.
\end{lstlisting}
\end{tcolorbox}

\begin{tcolorbox}[
  enhanced,
  breakable,
  width=0.98\linewidth,
  colback=tzBlueFill,
  colframe=tzBlueBorder,
  boxrule=1.2pt,
  arc=6pt,
  left=5pt,right=5pt,top=4pt,bottom=2pt,
  title={\small System Prompt for State Attributes Annotation},
  coltitle=white,
  colbacktitle=tzBlueHeader2,
  fonttitle=\bfseries,
]
\small
\begin{lstlisting}[style=jsonTiny]
You are an expert at analyzing state and condition of object parts.

**Task:** Annotate state attributes for the "{part_name}" of "{entity_name}" with given physical attributes.

**Physical Attributes:**
{physical_attrs}

**Annotate these state attributes:**
1. **Access State:**
   - Visibility: visible, partially visible, hidden (relative to whole entity)
   - Availability: free, partially blocked (easily freed by hand), blocked (requires tools)
2. **Condition State:**
   - Moisture: dry, slightly wet, wet, NA
   - Temperature: cold, slightly cold, slightly hot, hot, NA (room temp)
3. **Internal State:**
   - Internal: empty, partially filled, full, NA (physical or abstract capacity)
4. **Others:** Other important state attributes (1-2 sentences)
5. **Summary:** Comprehensive summary of all state attributes

**Requirements:**
- Generate 2-3 state attribute combinations for this part
- Must be consistent with physical attributes
- Include common/typical state and unusual but plausible states
- All combinations must be plausible, consistent, and diverse
- All fields required; use "NA" if not important
- Be assertive; avoid "might", "maybe"
- States should not contradict physical attributes

**Schema:**
```json
[
  {{
    "visibility": "...",
    "availability": "...",
    "moisture": "...",
    "temperature": "...",
    "internal": "...",
    "others": "...",
    "summary": "..."
  }}
]
```

**Example (vacuum bag - flexible, dry):**
```json
[
  {{
    "visibility": "hidden",
    "availability": "partially blocked",
    "moisture": "dry",
    "temperature": "NA",
    "internal": "partially filled",
    "others": "Seated inside a closed compartment; compartment cover is latched shut; bag collar aligned on plastic inlet mount; bag material flexible but holds shape from airflow.",
    "summary": "Hidden in latched compartment and partially blocked by cover and inlet mount; dry at room temperature; partially filled; flexible bag seated on collar mount."
  }},
  {{
    "visibility": "hidden",
    "availability": "blocked",
    "moisture": "slightly wet",
    "temperature": "NA",
    "internal": "partially filled",
    "others": "Contents clumped and tacky, bag adheres to compartment liner; bag surface damp and slightly softened; collar stuck on inlet mount and won't release with simple pull.",
    "summary": "Hidden and blocked by adhesion and stuck collar mount; slightly wet at room temperature; partially filled with clumped debris preventing easy hand removal."
  }}
]
```

First provide reasoning considering physical attributes. Then output JSON array of 1-4 combinations.
\end{lstlisting}
\end{tcolorbox}

\begin{tcolorbox}[
  enhanced,
  breakable,
  width=0.98\linewidth,
  colback=tzBlueFill,
  colframe=tzBlueBorder,
  boxrule=1.2pt,
  arc=6pt,
  left=5pt,right=5pt,top=4pt,bottom=2pt,
  title={\small System Prompt for Affordance Generation},
  coltitle=white,
  colbacktitle=tzBlueHeader2,
  fonttitle=\bfseries,
]
\small
\begin{lstlisting}[style=jsonTiny]
You are an expert at identifying functional affordances based on attributes.

**Task:** Annotate functional affordances for the "{part_name}". This part belongs to the entity "{entity_name}" in the scenario "{scenario}".

**Physical Attributes of Part "{part_name}":**
{physical_attrs}

**State Attributes of Part "{part_name}":**
{state_attrs}

**Instructions:**
Identify 6 different functional affordances for this part. For each:

1. **Use Condition:** What preparation is needed to access this affordance?
   - "NA" if part is free and directly usable
   - Otherwise describe preparation steps (e.g., "break the lens", "remove from case")
   - Based on visibility/availability from state attributes

2. **Environment Condition:** What environmental conditions are needed for this affordance?
   - Focus on scenario/environment requirements (e.g., "lighting available", "power source nearby")
   - NOT about the part itself or recipient, but about external conditions
   - "NA" if no special environment needed

3. **Attribute:** What attributes enable this affordance?
   - List relevant attributes that are needed to be considered for this affordance; your stated attributes must be derived from the given physical + state attributes above; do not introduce new attributes that are not provided
   - Format: JSON list of lists `[["attribute statement", "physical/state", "visual/text", "explanation why visual/text"], ...]`
   - "physical" = attribute statement derived from physical attributes given above; "state" = attribute statement derived from state attributes given above
   - "visual" = this attribute can be clearly illustrated in an image without ambiguity; "text" = this attribute needs text description to clarify (e.g., temperature, texture, hidden features)
   - The explanation should be your reason why this attribute needs to be illustrated using text to clarify or only visual signal is enough to convey the information.
   - Example: `[["the material of the lens is glass", "physical", "visual", "Glass material is visually identifiable by its transparency and reflective surface"], ["the surface is smooth", "physical", "text", "Smoothness is a tactile property difficult to convey visually alone"]]`

4. **Affordance:** What can this part be used for?
   - Brief, clear description of the function/purpose
   - Can be original normal function (must have at least one original normal function) OR creative alternative use
   - Must be plausible and grounded in this part's given physical and state attributes above
   - Must act upon a recipient to take effect: passive or decorative roles (e.g., "use as decoration", "display for aesthetics") are NOT valid affordances since they do not act upon anything
   - Write at a high, general level: describe the functional capability, not a specific scenario (good: "dig into soft material"; bad: "use as a shovel in a garden").
   - If the use is too rare, niche, or uncommon to be practically meaningful, do not include it.
   - If the affordance requires specific conditions to work, those must be reflected faithfully in use_condition, environment_condition, and recipient_condition

5. **Level:** Categorize the affordance type and suitability
   - "Normal 0" = Original intended normal function of the part within the entity (must have at least one original normal function)
   - "Emergency 1-5" = Creative/alternative use
     - 1 = Rarely used, only if nothing else available; difficult to access or imperfect; in real life people rarely use this affordance this way
     - 2 = In between 1 and 3
     - 3 = Moderately suitable, could work in urgent situations; in real life people may use this affordance a few times in real emergencies but not too much
     - 4 = In between 3 and 5
     - 5 = Highly suitable, easy and natural to use; very effective; in real life people are willing to use this affordance this way most of the time
   - Take into consideration if normally people can have access to this affordance, whether this part needs to be detached from the whole entity to use this affordance, whether to use this affordance will irreversibly damage the part or the whole entity, and how likely it is to happen, etc.
   - Format: "Normal 0 / Emergency X (comprehensive and grounded reasoning why you annotate this certain level)"

6. **Recipient Condition:** What attributes must the recipient have?
   - Every affordance must have a recipient. It can be the object, person, or material that this part acts upon
   - Define scope and limits using attribute categories (shape, size, rigidity, durability, surface, etc.)
   - Be fine-grained and comprehensive
   - Example: "thin to medium thickness, soft to semi-rigid rigidity, not harder than glass"

7. **Example Recipient:** List 3-4 concrete examples
   - Must satisfy the recipient condition
   - The proposed recipient must be concrete and specific things or objects that can be easily found in real life, rather than abstract concepts or ideas
   - Choose diverse examples reflecting affordance scope

8. **Failure Case:** When will this affordance NOT work?
   - Comprehensively consider all failure situations:
     * Use condition failures (can't access/prepare the part)
     * Environment condition failures (lack of necessary environmental factors)
     * Recipient condition failures (recipient too hard/thick/incompatible)
     * Action condition failures (user lacks skill/force/precision)
     * Other practical limitations
   - Be specific and realistic

**Requirements:**
- Generate 6 diverse and non-overlapping affordances
- Generate exactly one affordance for each level (Normal 0 and Emergency 1-5)
- Keep everything plausible and grounded-realistic physical-world uses only
- Each affordance should be completable solely by this part of the entity, NOT relying on other parts or the whole entity
- Do not introduce attributes not provided above, and just focus on this part's attributes, NOT other parts' attributes or the whole entity's attributes
- Every affordance must act upon a recipient; please skip any purely decorative, passive, or static role
- Describe the affordance at a high, scenario-agnostic level; omit uses that are too rare or uncommon to be practically meaningful
- If an affordance requires specific conditions (use, environment, recipient), those conditions must be explicitly captured in the corresponding fields

**Schema:**
```json
[
  {{
    "use_condition": "...",
    "environment_condition": "...",
    "attribute": [["attribute statement", "physical/state", "visual/text", "explanation why visual/text"], ...],
    "affordance": "...",
    "level": "Normal 0 / Emergency 1-5 (reason)",
    "recipient_condition": "...",
    "example_recipient": "...",
    "failure_case": "..."
  }}
]
```

**Example (glass lens, part of glasses entity - visible, free, dry, transparent):**
```json
[
  {{
    "use_condition": "NA",
    "environment_condition": "NA",
    "attribute": [
      ["the material of the lens is glass", "physical", "visual", "Glass is visually identifiable by transparency and clarity"],
      ["the shape is round and flat disc", "physical", "visual", "Circular flat shape is directly observable"],
      ["the rigidity is very rigid", "physical", "text", "Rigidity requires text description to confirm as there is no visual signal to indicate rigidity"]
    ],
    "affordance": "redirect, focus, or magnify light for vision correction or reading",
    "level": "Normal 0 (original intended function of the lens in the glasses)",
    "recipient_condition": "light must pass through; viewer needs magnification; text or objects at appropriate focal distance",
    "example_recipient": "printed text in books, small labels, fine details on objects, digital screens",
    "failure_case": "Fails if lens is dirty or scratched (blocks light), wrong prescription (incorrect magnification), insufficient lighting, or recipient is beyond focal range"
  }},
  {{
    "use_condition": "If we break the glass lens into pieces first",
    "environment_condition": "NA",
    "attribute": [
      ["the material of the lens is glass", "physical", "visual", "Glass is visually identifiable by transparency and clarity"],
      ["the durability is fragile", "physical", "text", "Fragility isn't directly visible, requiring text description to confirm"],
      ["the edge becomes sharp after breaking", "physical", "visual", "The sharpness after of broken glasses is commonsense knowledge, so no need to mention it through text description"]
    ],
    "affordance": "cut, scrape, or pierce small items",
    "level": "Emergency 2 (rarely used due to danger; people rarely break the glasses lens only for cutting purposes; it needs get the lens out first and break it into pieces is irreversible, make itself and the whole glassess not usable anymore; only when no proper cutting tool available)",
    "recipient_condition": "thin to medium thickness, soft to semi-rigid rigidity, not harder than glass, not highly abrasive",
    "example_recipient": "tape, paper, thin plastic wrap, soft fruit skin",
    "failure_case": "Fails if recipient too hard (damages edge), too thick (can't penetrate), or user can't safely handle sharp glass (risk of cuts); fails if pieces too small to grip"
  }},
  ...
]
```

First provide reasoning about affordances. Then output JSON array of 6 diverse and non-overlapping affordances of different levels following the schema and instructions above.
\end{lstlisting}
\end{tcolorbox}


\section{Task Creation Pipeline Details}
\label{appendix_sec:sampling}

\subsection{Overview of Sampling Stages}
The task creation pipeline is designed to produce rigorous yet creativity-demanding benchmark tasks.  
For each task, we enforce three properties:
(1) a clearly defined gold affordance (entity-part-action) that is truly preferred,
(2) distractor entities that are controlled by semantic similarity and decision ambiguity,
and (3) a grounded first-person task narrative with judge-checkable solution constraints.
We organize the pipeline into conceptual stages as follows.

\paragraph{Stage A: Scenario-wise Semantic Clustering.}
We first aggregate all annotated affordances (with their part-level physical/state context) and build two lookup spaces: an entity-level lookup and an affordance-level lookup. This allows every sampled affordance to be traced back to its exact scenario, entity, part, and annotation fields.
Then, within each scenario, we embed affordance texts using Text-Embedding-3-Large and perform complete-linkage hierarchical clustering with an adaptive distance threshold plus a hard diameter cap.
This yields compact semantic bins of affordances and supports structured gold sampling by (i) cluster-size band and (ii) affordance level (Normal 0 or Emergency 1--5). In our current run, this produces roughly 3.2K--3.7K clusters per scenario.

\paragraph{Stage B: Stratified Gold Affordance Sampling.}
A gold affordance is sampled under explicit controls:
cluster-size range (rarity/commonness proxy) and affordance level (normal vs. emergency tiers).  
This stratification makes the benchmark composition interpretable and analyzable across controlled factors, rather than being purely random.

\paragraph{Stage C: Task Prompt Instantiation from Gold.}
Given the sampled gold affordance, we generate a first-person task description with a concrete recipient and recipient condition, while hiding the gold entity/part/mechanism.  
This step ensures the task is goal-centric (``what should I use and how?'') instead of answer-leaking.

\paragraph{Stage D: Intra-Entity Gold Dominance Self-Check.}
Before accepting the sampled gold, we compare it against \emph{other parts of the same entity}.  
The gold is rejected and resampled if another part can challenge it strongly (e.g., replacement judged possible, or high-ambiguity similarity).  
This gate ensures the chosen gold is internally dominant within its own entity.

\paragraph{Stage E: Candidate Noise Pool Sampling.}
After a gold passes self-check, we sample candidate non-gold entities (noise entity candidates) using embedding-distance heuristics:
a near set (semantically close distractors) and a far set (semantically dissimilar distractors).  
This supports controlled distractor composition from ``highly confusable'' to ``clearly different.''

\paragraph{Stage F: Inter-Part Comparison Against Gold.}
For each sampled entity, every part is judged against the gold using its physical/state attributes and existing affordances.  
The judge outputs:
(1) whether a similar affordance is feasible,  
(2) a structured affordance annotation (if feasible),  
(3) whether this should replace the gold, and  
(4) decision-making difficulty (1--5).  
Replacement decisions explicitly consider accessibility, consequences, willingness, commonness, and safety.

\paragraph{Stage G: Distractor Filtering and Difficulty Assignment.}
Entities are filtered by strict rigor rules:
if any part is judged better than gold, or ambiguity is high, the entity is excluded.  
Remaining entities are split into:
\texttt{not\_similar} (no similar affordance) and \texttt{similar} (similar but not better).  
We then form different tiers:
Dissimilar (all not\_similar), Mixed (balanced mix), and Similar (all similar-but-not-better), with controlled entity counts.

\paragraph{Stage H: Final Task Assembly.}
For each accepted case, we assemble:
gold annotation, selected entities, judge outputs, additional scene items, first-person environment description, and a structured four-step solution.  
Solution fields are aligned with recipient/use/environment conditions plus affordance application mechanics, enabling direct judge-side verification.

\paragraph{Prompting Strategy.}
Similarly, we use stage-specific prompts with strict JSON schema constraints and explicit instructions to enforce groundedness, comparability, and decision consistency across sampling, comparison, and final task composition. The prompts are designed to reduce ambiguity, avoid gold leakage, and keep all outputs machine-parsable while preserving realistic first-person task narratives. All data are generated with GPT-5.2 under human supervision and iterative prompt refinement before full-scale runs. During refinement, we repeatedly inspect pilot outputs and update prompts to correct common failure modes (e.g., weak gold-dominance checks, inconsistent replacement decisions, or under-specified conditions). After large-scale generation, we perform quality assessment by sampling a subset of 0.1\% outputs and verify the logical coherence, condition-grounded reasoning, and consistency between gold selection. The generated data passes 98\% human checks and reaches the bar of high quality.

\subsection{Core Hyperparameters Details}

\begin{table}[!h]
\centering
\small
\setlength\tabcolsep{10pt}
\setlength\extrarowheight{2pt}
\begin{tabular}{lc}
\toprule
\textbf{Hyperparameter} & \textbf{Value} \\
\midrule
Embedding model & Text-Embedding-3-Large \\
Clustering method & Complete-linkage hierarchical clustering \\
Hard max cluster diameter (cosine distance) & 0.35 \\
Gold level set & Normal 0 and Emergency 1--5 \\
Gold cluster-size bands & (2,4), (5,10), (10,50) \\
Cases per (level, cluster-size) tier & 10 \\
\bottomrule
\end{tabular}
\caption{Core hyperparameters of gold-sampling and clustering controls.}
\label{tab:task_b_gold_sampling}
\end{table}

\begin{table}[!h]
\centering
\small
\setlength\tabcolsep{10pt}
\setlength\extrarowheight{2pt}
\begin{tabular}{lc}
\toprule
\textbf{Hyperparameter} & \textbf{Value} \\
\midrule
Candidate entities sampled per comparison set & 30 \\
Distractor-count settings in final tasks & 3, 6, 9, 12 \\
Affordance similarity tiers & Dissimilar, Mixed, Similar \\
Mixed-tier composition & 50\% Similar + 50\% Dissimilar \\
Cases per (similarity, entity-count) tier & 2 \\
\bottomrule
\end{tabular}
\caption{Core hyperparameters distractor sampling and final-task composition controls.}
\label{tab:task_b_noise_sampling}
\end{table}

\begin{table}[!h]
\centering
\small
\setlength\tabcolsep{10pt}
\setlength\extrarowheight{2pt}
\begin{tabular}{lc}
\toprule
\textbf{Hyperparameter} & \textbf{Value} \\
\midrule
Comparison judge temperature & 0.0 \\
Task generation and final composition temperature & 0.3 \\
Current final tasks generated & 14,280 \\
\bottomrule
\end{tabular}
\caption{Core hyperparameters inference settings and resulting scale of the current run.}
\label{tab:task_b_scale}
\end{table}

\subsection{Core Prompt Details}

\begin{tcolorbox}[
  enhanced,
  breakable,
  width=0.98\linewidth,
  colback=tzBlueFill,
  colframe=tzBlueBorder,
  boxrule=1.2pt,
  arc=6pt,
  left=5pt,right=5pt,top=4pt,bottom=2pt,
  title={\small System Prompt for Gold-to-Task Generation},
  coltitle=white,
  colbacktitle=tzBlueHeader2,
  fonttitle=\bfseries,
]
\small
\begin{lstlisting}[style=jsonTiny]
You are writing a first-person benchmark question for a creative problem-solving task. Given a gold affordance, your job is to invent a concrete, realistic situation that the affordance can solve - without revealing the gold entity, part, or affordance mechanism.

Gold Entity: {entity_uid}
Gold Part: {part_name}
Gold Affordance:
{affordance}

**Your process:**
Step 1 - Pick a concrete recipient from `example_recipient` (or one that satisfies `recipient_condition`).
Step 2 - Invent a specific real-world situation that person encounters involving that recipient.
Step 3 - Describe the situation and goal in everyday language, then ask "What can I use?" or "What should I use and how?".

**Rules:**
- Start from the SITUATION and RECIPIENT, not from the affordance mechanism. Ask yourself: "What real problem is this person facing?" not "What does the affordance do?"
- Be specific and concrete. Name the recipient object. Give context (where, what happened, what they want to achieve).
- Do NOT paraphrase or echo the affordance description. If the affordance is "dampen a surface", do NOT write "make a spot wet". Describe WHY the person needs it.
- Do NOT name or hint at the gold entity or part.
- Do NOT describe what needs to happen physically - describe the human goal/problem.

**Examples (affordance: "apply small amount of moisture to a surface"):**
GOOD: "My fabric jacket has a small stubborn stain from yesterday's lunch. I want to treat just that spot before tossing it in the wash, but I don't have a spray bottle or wet cloth nearby. What can I use?"
  -> Concrete: specific object (fabric jacket), specific problem (stain), specific constraint (no wet cloth). Affordance is not mentioned.

BAD: "I'm trying to quickly put a small, localized damp spot onto an absorbent surface. What can I use?"
  -> Describes the affordance mechanism directly. No real situation. Not grounded.

BAD: "I need to use the sponge tip to apply moisture to my shirt." -> Leaks entity and mechanism.

Output JSON:
```json
{{
  "task": "Concrete first-person scenario question grounded in a specific recipient and situation, hiding the gold entity/part/affordance",
  "recipient": "The specific object or thing receiving this affordance (pick from example_recipient or invent one matching recipient_condition)",
  "recipient_condition": "Required state or attribute of the recipient"
}}
```
\end{lstlisting}
\end{tcolorbox}

\begin{tcolorbox}[
  enhanced,
  breakable,
  width=0.98\linewidth,
  colback=tzBlueFill,
  colframe=tzBlueBorder,
  boxrule=1.2pt,
  arc=6pt,
  left=5pt,right=5pt,top=4pt,bottom=2pt,
  title={\small System Prompt for Gold-vs-Target Part Judge},
  coltitle=white,
  colbacktitle=tzBlueHeader2,
  fonttitle=\bfseries,
]
\small
\begin{lstlisting}[style=jsonTiny]
You are judging whether a target part can fulfill a similar role to a gold affordance for a given task.

## Gold
Entity: {gold_entity} | Part: {gold_part}
Physical Attributes of Part {gold_part}:
{gold_physical_attrs}
State Attributes of Part {gold_part}:
{gold_state_attrs}
Gold Affordance:
{gold_affordance}

## Target Part
Entity: {target_entity} | Part: {target_part}
Physical Attributes of Part {target_part}:
{physical_attrs}
State Attributes of Part {target_part}:
{state_attrs}
Existing Affordances of Part {target_part}:
{existing_affordances}

## Task
{task}
Recipient: {recipient}

---

**Step 1 - Similarity Judgment**
Determine whether this part ALONE (without any other part of the entity) can perform a functionally similar role to solve the task - i.e., it serves the same purpose as the gold affordance. Base your judgment strictly on the part's provided physical and state attributes.

**Step 2 - Affordance Annotation** (only if similar_affordance = Yes)
If an existing affordance already matches, adapt it. Otherwise write a new one. All fields must be grounded in the provided attributes. If similar_affordance = No, set ALL affordance fields to "NA".

Field definitions:
- use_condition: preparation needed to access this affordance; "NA" if part is free and directly usable (based on visibility/availability from state attributes)
- environment_condition: external environmental conditions required (not about the part or recipient); "NA" if none
- attribute: list of [["attribute statement", "physical/state"]], indicating all given target part attributes enabling this affordance;
- recipient_condition: required attributes of the recipient (shape, size, rigidity, material, etc.), fine-grained
- example_recipient: 3-4 concrete examples satisfying recipient_condition
- failure_case: all situations where this affordance fails - use condition failures, environment failures, recipient incompatibility, action/skill failures

**Step 3 - Gold Comparison** (only if similar_affordance = Yes)
Systematically compare gold vs. target across ALL of the following aspects. For each aspect, explicitly state which side is better and by how much:
- Accessibility and use/env conditions (how easy to access and activate)
- Future consequences (irreversible damage, side effects, mess)
- How willing people are to use it this way (social acceptability, effort)
- How common this usage is in everyday life
- Safety and ethical considerations

After comparing all aspects, make a final decision. If most aspects clearly favor one side, gold_change is straightforward. If several aspects are roughly equal, or pros and cons are genuinely hard to weigh against each other, explicitly acknowledge that uncertainty and reflect it with a high decision_making_difficulty score. If similar_affordance = No, copy the Step 1 reason into gold_change_reason and set gold_change = No.

Output JSON:
```json
{{
  "similar_affordance_reason": "whether this part ALONE can show similar affordance to solve the task, based on which attributes and conditions",
  "similar_affordance": "Yes or No",
  "affordance": {{
    "use_condition": "...",
    "environment_condition": "...",
    "attribute": [["attribute statement", "physical/state"]],
    "affordance": "...",
    "recipient_condition": "...",
    "example_recipient": ["...", "...", "..."],
    "failure_case": "..."
  }},
  "gold_change_reason": "For each aspect - accessibility, consequences, willingness, commonness, safety - state which side is better and why. Then give a final balanced verdict. If aspects conflict or are too close to call, say so explicitly.",
  "gold_change": "Yes or No",
  "decision_making_difficulty": "1-5 or NA"
}}
```
decision_making_difficulty: 1 = very easy to keep gold (one side clearly better across most aspects); 5 = extremely difficult (aspects are close or contradictory, genuinely uncertain which is better); NA if similar_affordance is No. A high score (4-5) is appropriate whenever multiple aspects are roughly tied or point in opposite directions - do not default to low scores when the comparison is genuinely hard.
\end{lstlisting}
\end{tcolorbox}

\begin{tcolorbox}[
  enhanced,
  breakable,
  width=0.98\linewidth,
  colback=tzBlueFill,
  colframe=tzBlueBorder,
  boxrule=1.2pt,
  arc=6pt,
  left=5pt,right=5pt,top=4pt,bottom=2pt,
  title={\small System Prompt for Final Task Packaging},
  coltitle=white,
  colbacktitle=tzBlueHeader2,
  fonttitle=\bfseries,
]
\small
\begin{lstlisting}[style=jsonTiny]
Generate three fields for a benchmark task.

## Gold
Entity: {gold_entity} | Part: {gold_part}
Gold Affordance:
{gold_affordance}

## Task
{task}
Recipient: {recipient} (Condition: {recipient_condition})

## Entities already in the scene
{entities_desc}

---

**items**: List things referenced in the gold affordance's use_condition, environment_condition, and recipient_condition, plus the recipient itself. Describe each with the traits implied by those conditions. Also add 3-5 natural noise objects that fit the {scenario} but are unrelated to the task (interactable=No). Do NOT add things similar to the entities above.

**environment**: First-person scene description starting "I am in the {scenario}. Around me there is ..." - naturally mention every entity and key item.

**solution**: A structured dict with exactly four keys describing how to solve the task using {gold_entity}'s {gold_part}. Each value is a string in the format "... (Note: ...)" where the Note contains specific things a judge should explicitly verify when scoring. If no specific condition was explicitly needed, write "(Note: NA)".

The four keys and what to write for each:
- "prepare_recipient": How to prepare the recipient to be applied for the task.
- "prepare_use_condition": How to set up the tool (gold part) to meet use_condition.
- "prepare_environment_condition": What environmental setup is needed to meet environment_condition.
- "apply_affordance": The core action about how to apply {gold_part}'s affordance to the recipient to solve the task. This is the most important step. Be comprehensive and detailed, referencing the specific physical/state attributes of the part that make this work.

Rules:
- Please be strictly grounded in the gold affordance annotation and the part's listed attributes. Do not invent steps or attributes beyond what is annotated.
- Each step should be detailed and concrete, but not overly complex.
- The Note in each step is a judging reference.

Output ONLY valid JSON:
```json
{{
  "items": [
    {{
      "name": "<item name>",
      "description": "<description; include condition-relevant attributes where applicable>",
      "interactable": "Yes or No"
    }}
  ],
  "environment": "<I am in the {scenario}. Around me there is ...>",
  "solution": {{
    "prepare_recipient": "<description of recipient preparation> (Note: <critical judging reference of recipient_condition attributes that need additional verification besides that in the task description, or NA>)",
    "prepare_use_condition": "<description of tool setup> (Note: <critical judging reference of use_condition that needed to be set up before the use, or NA>)",
    "prepare_environment_condition": "<description of environment setup> (Note: <critical judging reference of environment_condition that needed to be set up before the use, or NA>)",
    "apply_affordance": "<detailed core action using {gold_part} on the recipient> (Note: <critical judging reference of the key attributes and affordance mechanism to verify before applying the affordance>)"
  }}
}}
```
\end{lstlisting}
\end{tcolorbox}


\section{Experiment Details}
\label{appendix_sec:exp_details}

\paragraph{Settings.}
For the results reported in the main table, most models are evaluated with temperature 0 to ensure deterministic outputs. An exception is GPT-5-Mini and GPT-5-Nano, which do not support an adjustable temperature parameter and thus cannot be run with explicit zero temperature. For these two models, we use the default sampling setting; all other models are evaluated with temperature 0.

All models are assigned a maximum output length of at least 16K tokens, which is empirically sufficient for our generation needs. For the main setting, we evaluate each model using the following prompt.

\begin{tcolorbox}[
  enhanced,
  breakable,
  width=0.98\linewidth,
  colback=tzBlueFill,
  colframe=tzBlueBorder,
  boxrule=1.2pt,
  arc=6pt,
  left=5pt,right=5pt,top=4pt,bottom=2pt,
  title={\small System Prompt for Final Task Packaging},
  coltitle=white,
  colbacktitle=tzBlueHeader2,
  fonttitle=\bfseries,
]
\small
\begin{lstlisting}[style=jsonTiny]
TASK:
{task content}

ENVIRONMENT:
{environment content}

=== ENTITIES AVAILABLE (full descriptions) ===
{blocks of entity description, from task annotation}

=== OTHER ITEMS IN SCENE ===
{blocks of other items description, from task annotation}

Identify the best entity part to creatively accomplish the task. Please first provide your reasoning process, and then finally respond with a JSON object containing the three answer fields:
<your reasoning process here>
{"gold_entity": "<entity name, case-sensitive and exact match>", "gold_part": "<part name, case-sensitive and exact match>", "how_to_use": "<detailed instructions>"}
\end{lstlisting}
\end{tcolorbox}

\paragraph{Metrics.}
In the main setting, we report two objective metrics, \textit{gold correct} and \textit{entity correct}, as well as six subjective metrics. For each metric, our original scoring is between 0--2. However, considering the alignment with the scale we employ in \Cref{sec:preliminaries}, we uniformly scale the range to 1--5. In the results reported in the main table, the subjective metrics are only used for judging those answers that count as Gold Correct, while leaving other failure cases' analysis in the later section. Consistent with \Cref{sec:preliminaries}, the subjective metrics evaluate aspects such as constraint coverage, physical grounding, and related properties of the predicted affordance. In particular, these subjective metrics are designed to assess the quality of the model's generated ``how to use'' explanation, rather than only whether the correct entity and part are selected.

For subjective evaluation, we use Gemini-3.1-Flash-Lite as the judge model. The judging prompt is provided below.

\begin{tcolorbox}[
  enhanced,
  breakable,
  width=0.98\linewidth,
  colback=tzBlueFill,
  colframe=tzBlueBorder,
  boxrule=1.2pt,
  arc=6pt,
  left=5pt,right=5pt,top=4pt,bottom=2pt,
  title={\small System Prompt for Final Task Packaging},
  coltitle=white,
  colbacktitle=tzBlueHeader2,
  fonttitle=\bfseries,
]
\small
\begin{lstlisting}[style=jsonTiny]
You are a strict evaluator for embodied task instructions.
Evaluate whether the predicted "how_to_use" is feasible compared to the gold affordance and gold solution.
Use evidence-based judgments for each field. Please follow the rubric strictly.

Task:
{task content}

Gold affordance JSON:
{gold affordance}

Gold solution JSON:
{gold solution}

Predicted how_to_use:
{how to use part of prediction}

Field-by-field rubric when judging the predicted "how_to_use":
1) environment_condition_covered (0/1/2/"NA"):
- 0: required external environment setup in gold is not mentioned at all in the predicted "how_to_use".
- 1: required external environment setup in gold is mentioned, but not all are covered in the predicted "how_to_use".
- 2: required external environment setup in gold is covered well and reasonable in the predicted "how_to_use".
- NA: only if in the gold affordance annotation, the environment_condition is NA.

2) use_condition_covered (0/1/2/"NA"):
- 0: required preparation/access of the tool-part is not mentioned at all in the predicted "how_to_use".
- 1: required preparation/access of the tool-part is mentioned, but not all are covered in the predicted "how_to_use".
- 2: required preparation/access of the tool-part is covered well and reasonable in the predicted "how_to_use".
- NA: only if in the gold affordance annotation, the use_condition is NA.

3) recipient_condition_covered (0/1/2/"NA"):
- 0: recipient-side prerequisites are not mentioned at all in the predicted "how_to_use".
- 1: recipient-side prerequisites are mentioned, but not all are covered in the predicted "how_to_use".
- 2: recipient-side prerequisites are covered well and reasonable.
- False: recipient assumptions are not met in the predicted "how_to_use".
- NA: only if in the gold affordance annotation, the recipient_condition is NA.

4) attributes_grounding (0/1/2): [compare to the key enabling attributes]
- 0: the predicted action is not grounded in the key enabling attributes of the gold affordance, or it violates some attributes of the part.
- 1: the predicted action is mostly grounded in the key enabling attributes of the gold affordance, but not all are covered or implied.
- 2: the predicted action is fully grounded in the key enabling attributes of the gold affordance, and most of them are explicitly mentioned, covered or implied.

5) prediction_correctness (0/1/2): [compare to the gold solution]
- 0: compared with the gold solution the predicted "how_to_use" is completely wrong, not working at all.
- 1: compared with the gold solution the predicted "how_to_use" is partially workable but missing crucial details/order/precision.
- 2: compared with the gold solution the predicted "how_to_use" is operationally correct and complete, almost completely aligned with gold solution.

6) action_feasibility (0/1/2): [not compare to the gold solution, but focus on the action itself]
- 0: the action itself is physically impossible, not working at all, very unlikely to be used in practice, or unsafe.
- 1: the action itself is partially workable but still there are some steps that are not plausible, aligned with common sense or not feasible.
- 2: the action itself is operationally correct and complete, almost completely aligned with common sense and feasible in practice.

Evidence policy:
- Every *_reason must quote concrete evidence from predicted "how_to_use" and other given relevant information.
- If evidence is unclear/missing, default to a stricter outcome (lower score).
- Make your reasoning clear, concise, and to the point for each field, and your score should be based on the evidence.

Please make sure to only return a valid JSON with exactly these 12 fields (no extras, no markdown).
{{
    "environment_condition_covered_reason": "...",
    "environment_condition_covered": 0/1/2/"NA",
    "use_condition_covered_reason": "...",
    "use_condition_covered": 0/1/2/"NA",
    "recipient_condition_covered_reason": "...",
    "recipient_condition_covered": 0/1/2/"NA",
    "attributes_grounding_reason": "...",
    "attributes_grounding": 0/1/2,
    "prediction_correctness_reason": "...",
    "prediction_correctness": 0/1/2,
    "action_feasibility_reason": "...",
    "action_feasibility": 0/1/2,
}}
\end{lstlisting}
\end{tcolorbox}

\section{Analysis Details}
\label{appendix_sec:analysis_details}

\subsection{Inference setting's impact on performance}

\begin{figure}[h]
    \centering
    \begin{minipage}{0.48\linewidth}
        \centering
        \includegraphics[width=\linewidth]{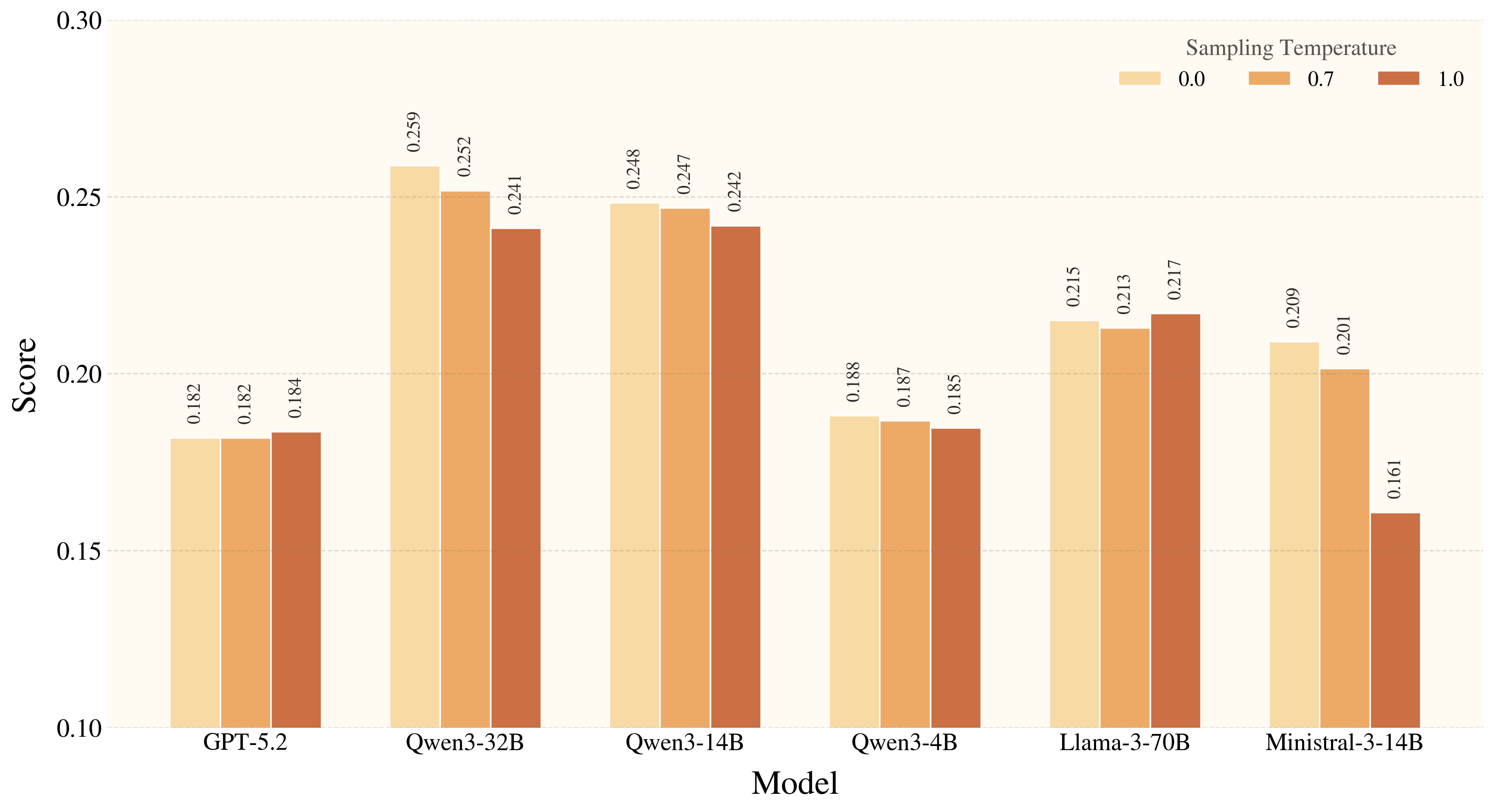}
        \caption*{(a) Inference temperature}
        \label{apdx:analysis_temperature}
    \end{minipage}
    \hfill
    \begin{minipage}{0.5\linewidth}
        \centering
        \vspace{2.5mm}
        \resizebox{\linewidth}{!}{
            \begin{tabular}{lccc}
            \toprule
            Model & Static & Interactive & CoT \\
            \midrule
            GPT-5-mini & 0.169 & 0.088 {\color{red!50!black}$_{-0.081}$} & 0.162 {\color{red!50!black}$_{-0.006}$} \\
            GPT-5-nano & 0.119 & 0.056 {\color{red!50!black}$_{-0.064}$} & 0.116 {\color{red!50!black}$_{-0.003}$} \\
            Qwen3-32B & 0.259 & 0.093 {\color{red!50!black}$_{-0.166}$} & 0.261 {\color{green!50!black}$_{+0.002}$} \\
            Qwen3-14B & 0.248 & 0.099 {\color{red!50!black}$_{-0.150}$} & 0.260 {\color{green!50!black}$_{+0.012}$} \\
            Qwen3-4B & 0.188 & 0.062 {\color{red!50!black}$_{-0.127}$} & 0.201 {\color{green!50!black}$_{+0.013}$} \\
            Llama-3-70B & 0.215 & 0.047 {\color{red!50!black}$_{-0.168}$} & 0.250 {\color{green!50!black}$_{+0.035}$} \\
            Ministral-3-14B & 0.209 & 0.081 {\color{red!50!black}$_{-0.128}$} & 0.247 {\color{green!50!black}$_{+0.038}$} \\
            \bottomrule
            \end{tabular}
        }
        \vspace{2.5mm}
        \caption*{(b) Inference mode}
        \label{apdx:analysis_mode}
    \end{minipage}
    \caption{Additional results on full test set evaluation on different inference-time sampling temperatures and evaluation mode applied.}
    \label{apdx:analysis_temperatur_mode}
\end{figure}

For \Cref{sec:analysis_infer}, we evaluate only 10\% of the current dataset. We make this choice for two reasons. First, this subset still contains about 1.4K examples, which is already large enough to support reliable analysis. Second, the required experimental budget is substantial: running the full 14K-example set across all models is expensive, and the cost becomes even higher for interactive-mode evaluation because it requires multi-turn inference. In addition, for the temperature-sampling experiments, we exclude GPT-5 Mini and GPT-5 Nano because their sampling temperature cannot be controlled.

Nevertheless, to validate that the sampled subset is representative, we also evaluate the full 14K test set on selected models and report the effects of sampling temperature~\citep{liu2026naacl} and inference mode in \Cref{apdx:analysis_temperatur_mode}. The full-set results show trends consistent with those reported in the main text. For temperature sampling, the performance of GPT-5.2 and Llama-3-70B tends to improve, whereas smaller models generally experience performance drops. A similar pattern appears in the inference-mode experiments: interactive mode substantially reduces performance, while CoT mode leads only to small fluctuations and yields no clear overall benefit.

Overall, these full-set results provide additional evidence that the 10\% sampled subset is representative of the broader dataset.

\subsection{Fine-grained analysis on auxiliary metrics}
\begin{figure}[t]
    \centering
    \begin{subfigure}[t]{0.49\linewidth}
        \centering
        \includegraphics[width=\linewidth]{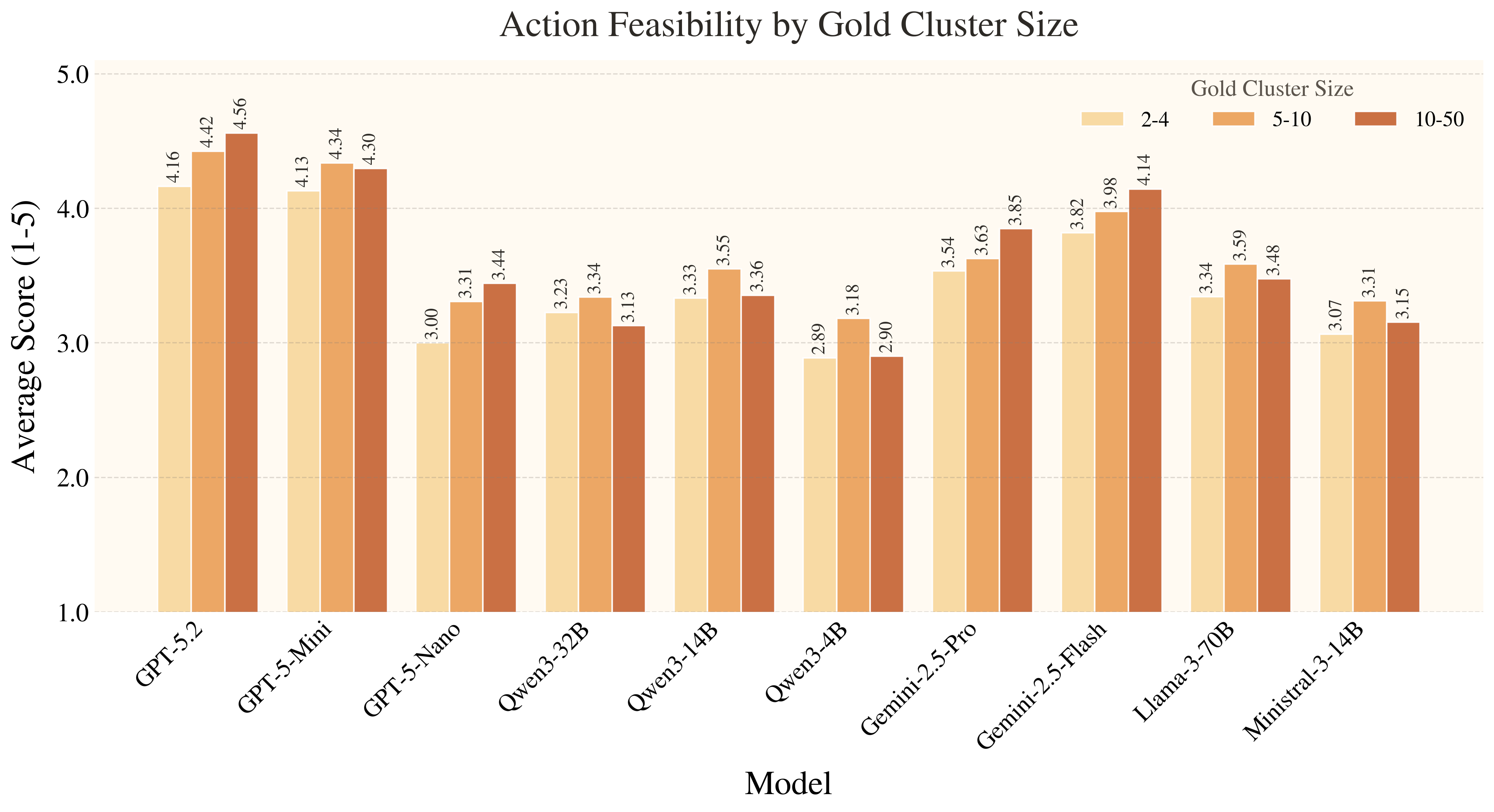}
    \end{subfigure}
    \hfill
    \begin{subfigure}[t]{0.49\linewidth}
        \centering
        \includegraphics[width=\linewidth]{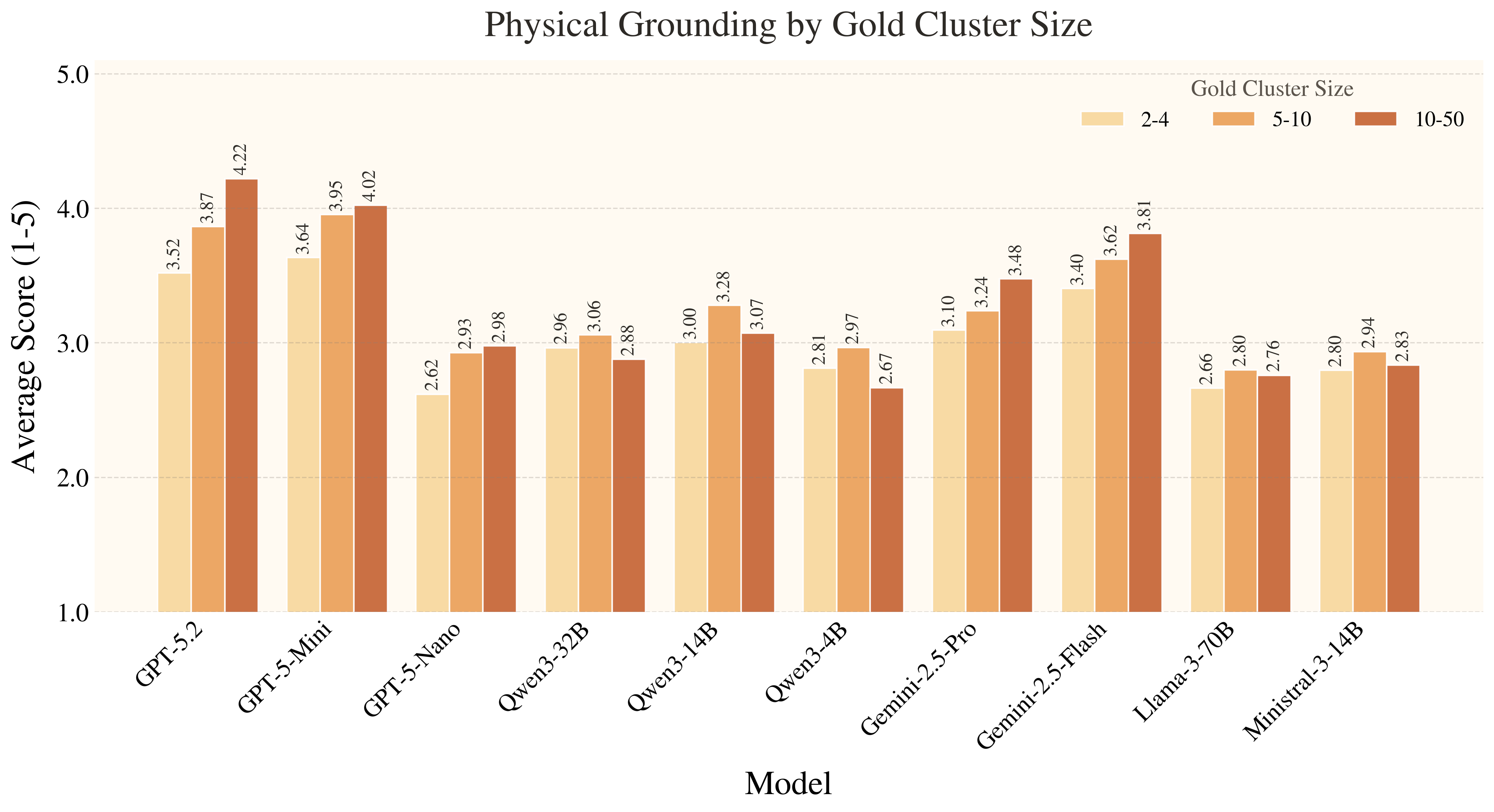}
    \end{subfigure}
    \hfill
    \begin{subfigure}[t]{0.49\linewidth}
        \centering
        \includegraphics[width=\linewidth]{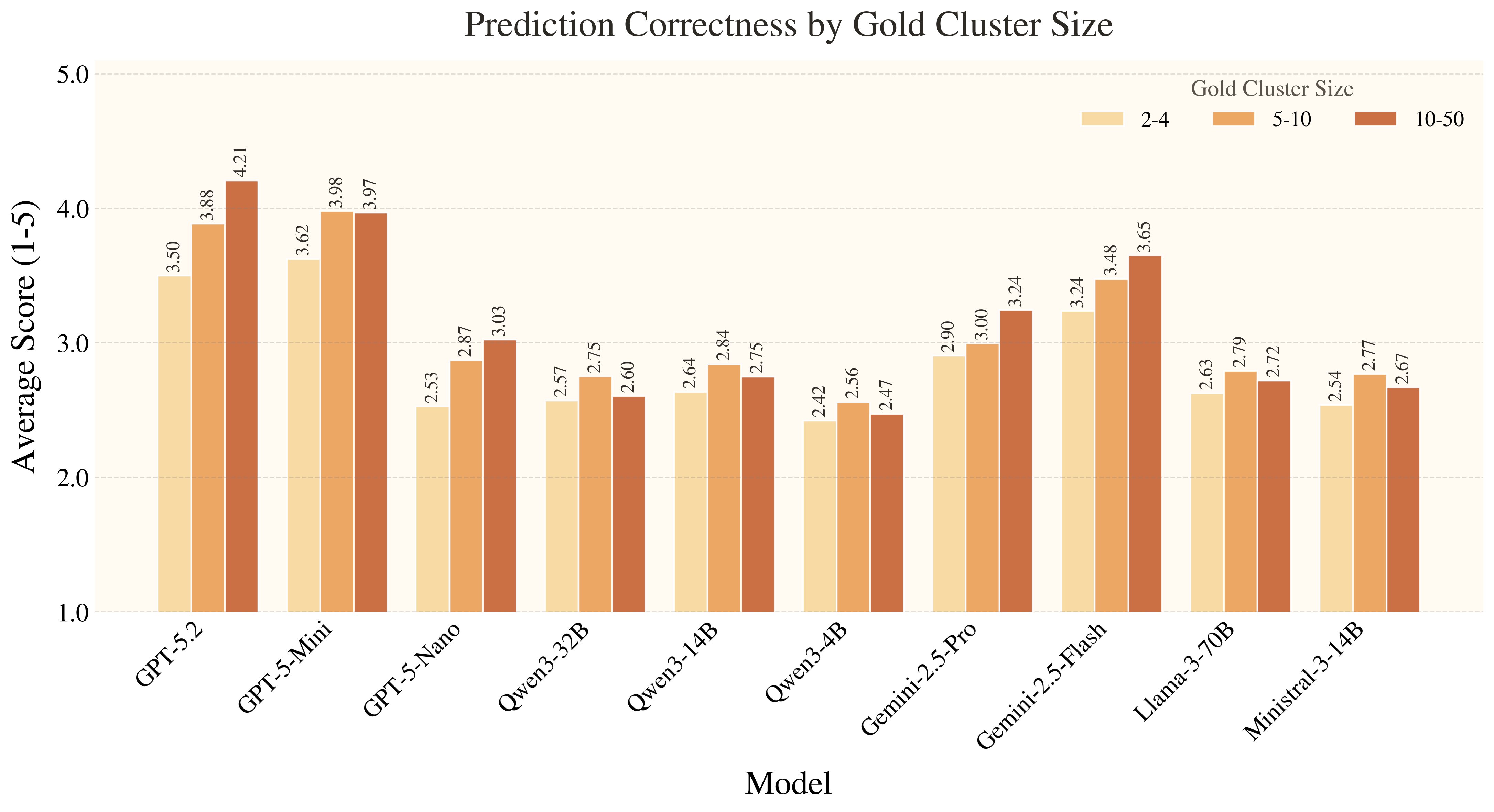}
    \end{subfigure}
    \hfill
    \begin{subfigure}[t]{0.49\linewidth}
        \centering
        \includegraphics[width=\linewidth]{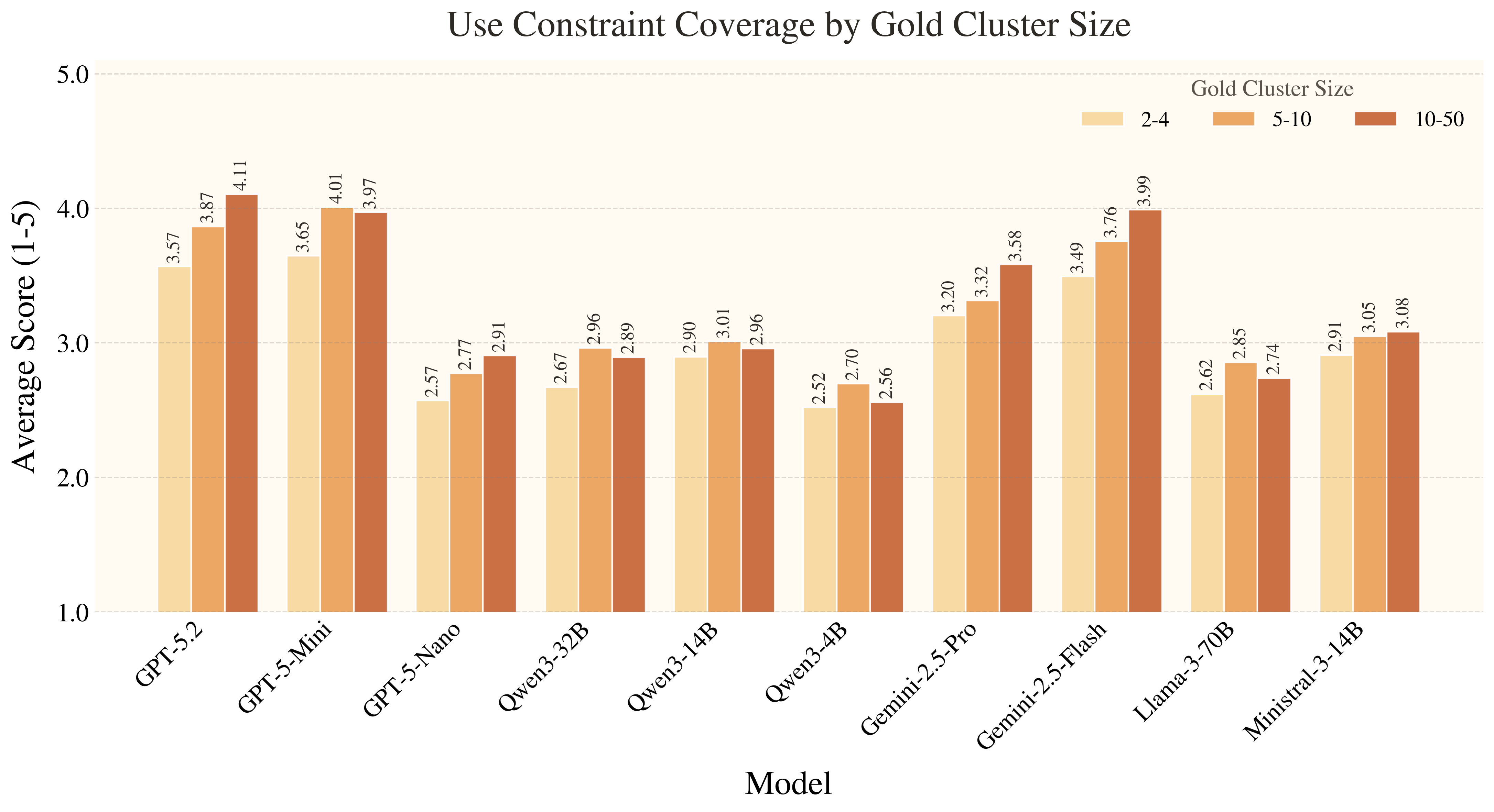}
    \end{subfigure}
    \hfill
    \begin{subfigure}[t]{0.49\linewidth}
        \centering
        \includegraphics[width=\linewidth]{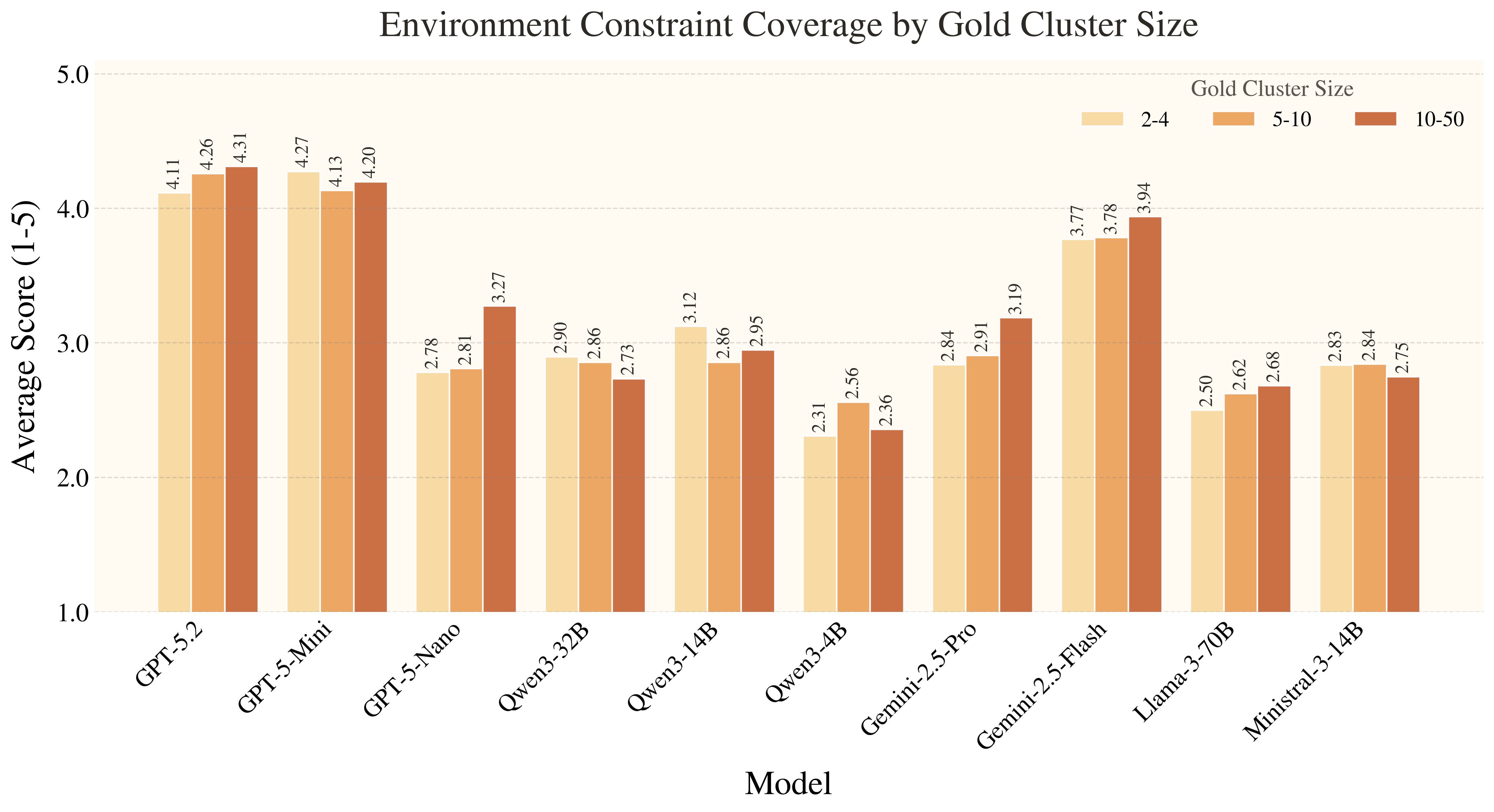}
    \end{subfigure}
    \hfill
    \begin{subfigure}[t]{0.49\linewidth}
        \centering
        \includegraphics[width=\linewidth]{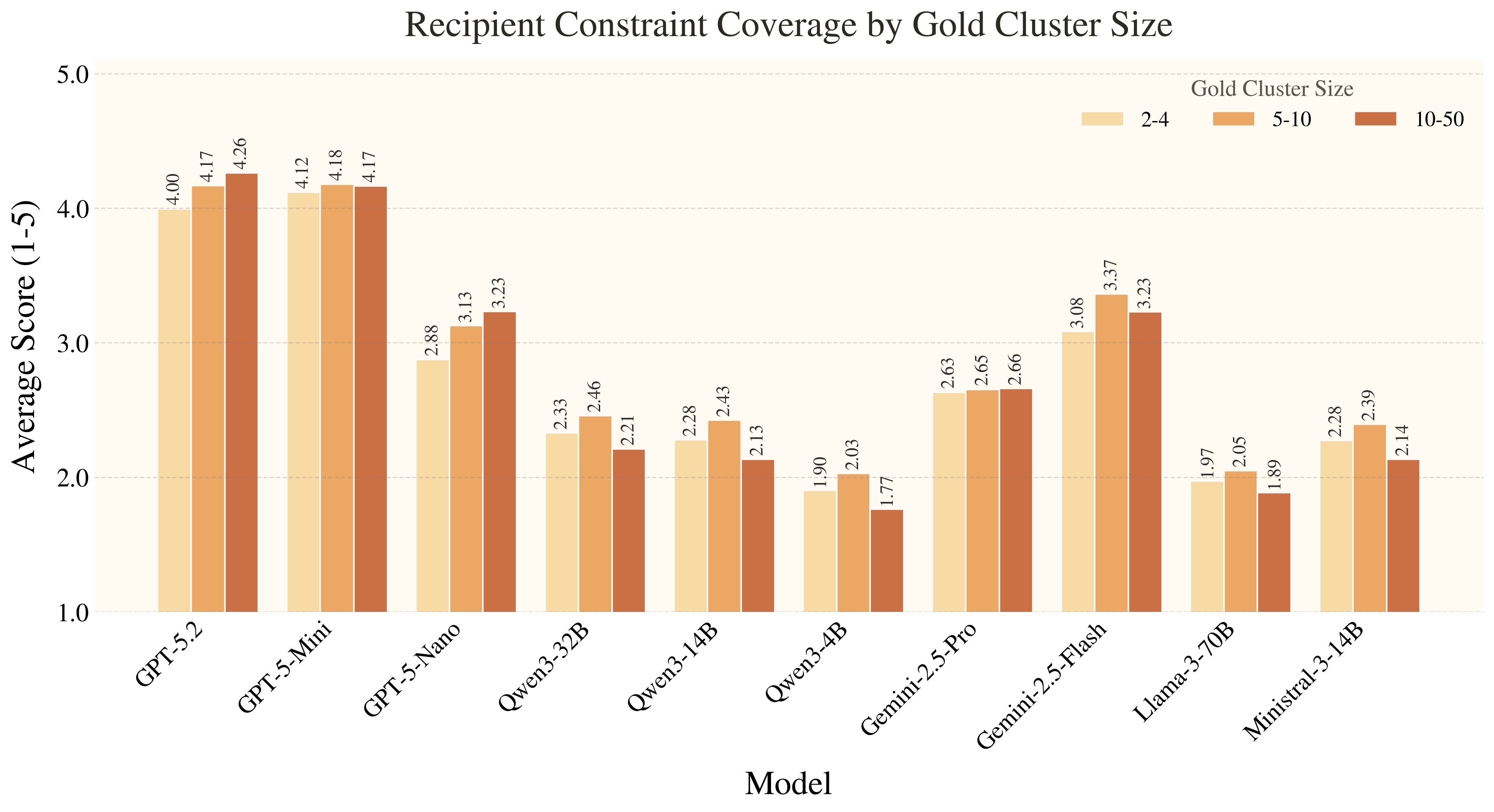}
    \end{subfigure}
    \caption{Fine-grained auxiliary metrics analysis upon different gold affordance cluster sizes.}
    \label{apdx:aux_metrics_cluster_size}
\end{figure}

\begin{figure}[t]
    \centering
    \begin{subfigure}[t]{0.49\linewidth}
        \centering
        \includegraphics[width=\linewidth]{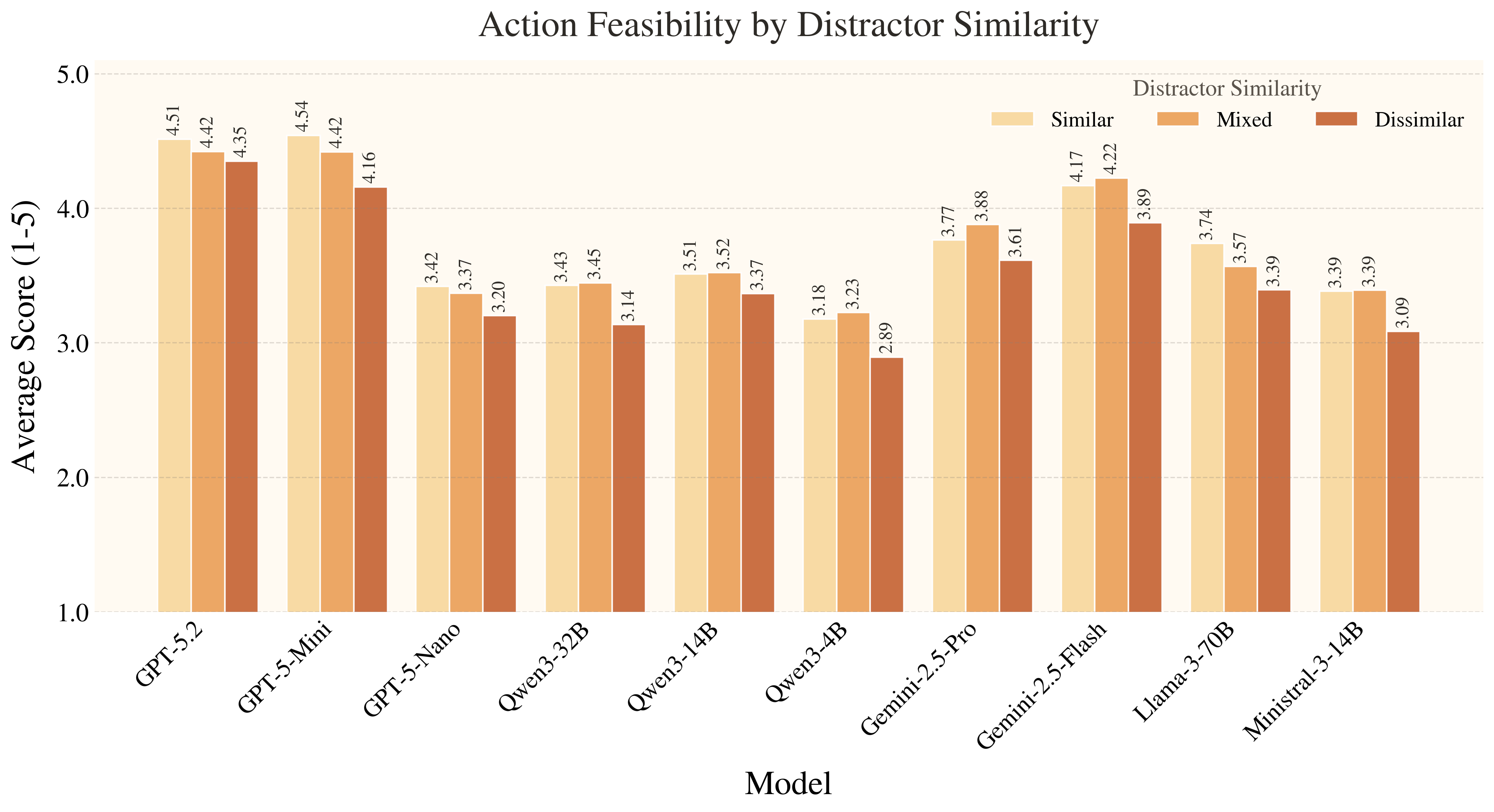}
    \end{subfigure}
    \hfill
    \begin{subfigure}[t]{0.49\linewidth}
        \centering
        \includegraphics[width=\linewidth]{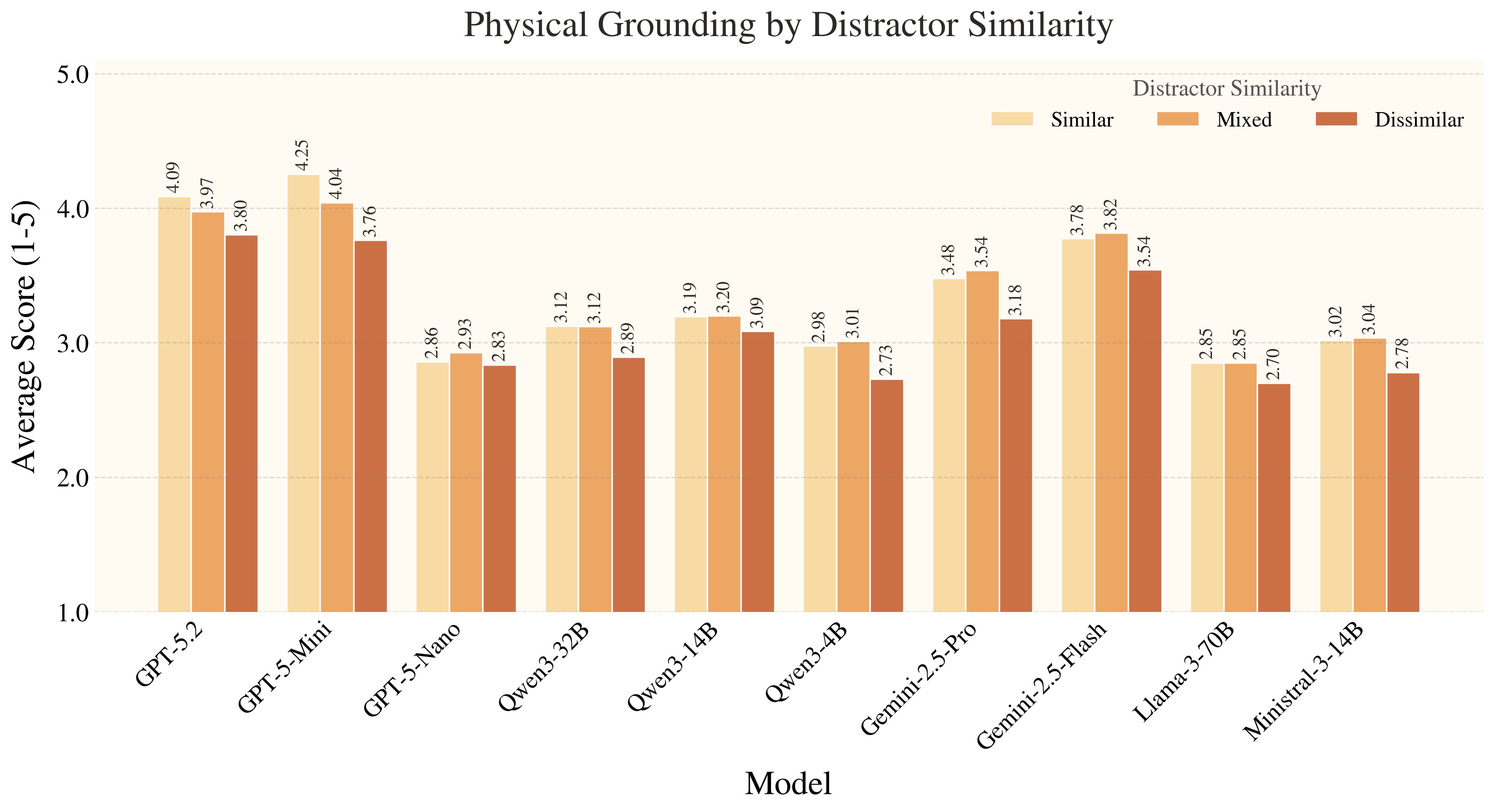}
    \end{subfigure}
    \hfill
    \begin{subfigure}[t]{0.49\linewidth}
        \centering
        \includegraphics[width=\linewidth]{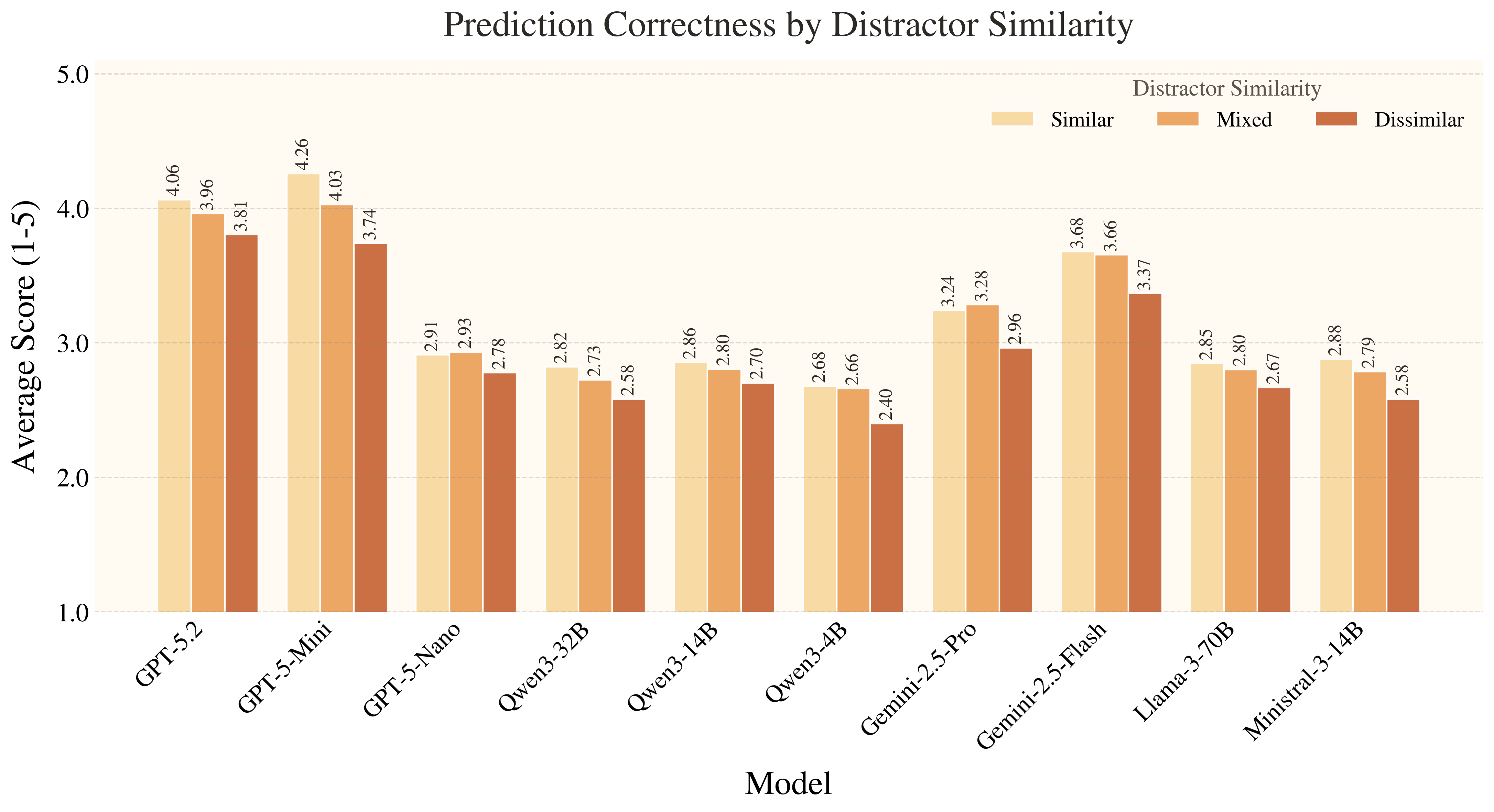}
    \end{subfigure}
    \hfill
    \begin{subfigure}[t]{0.49\linewidth}
        \centering
        \includegraphics[width=\linewidth]{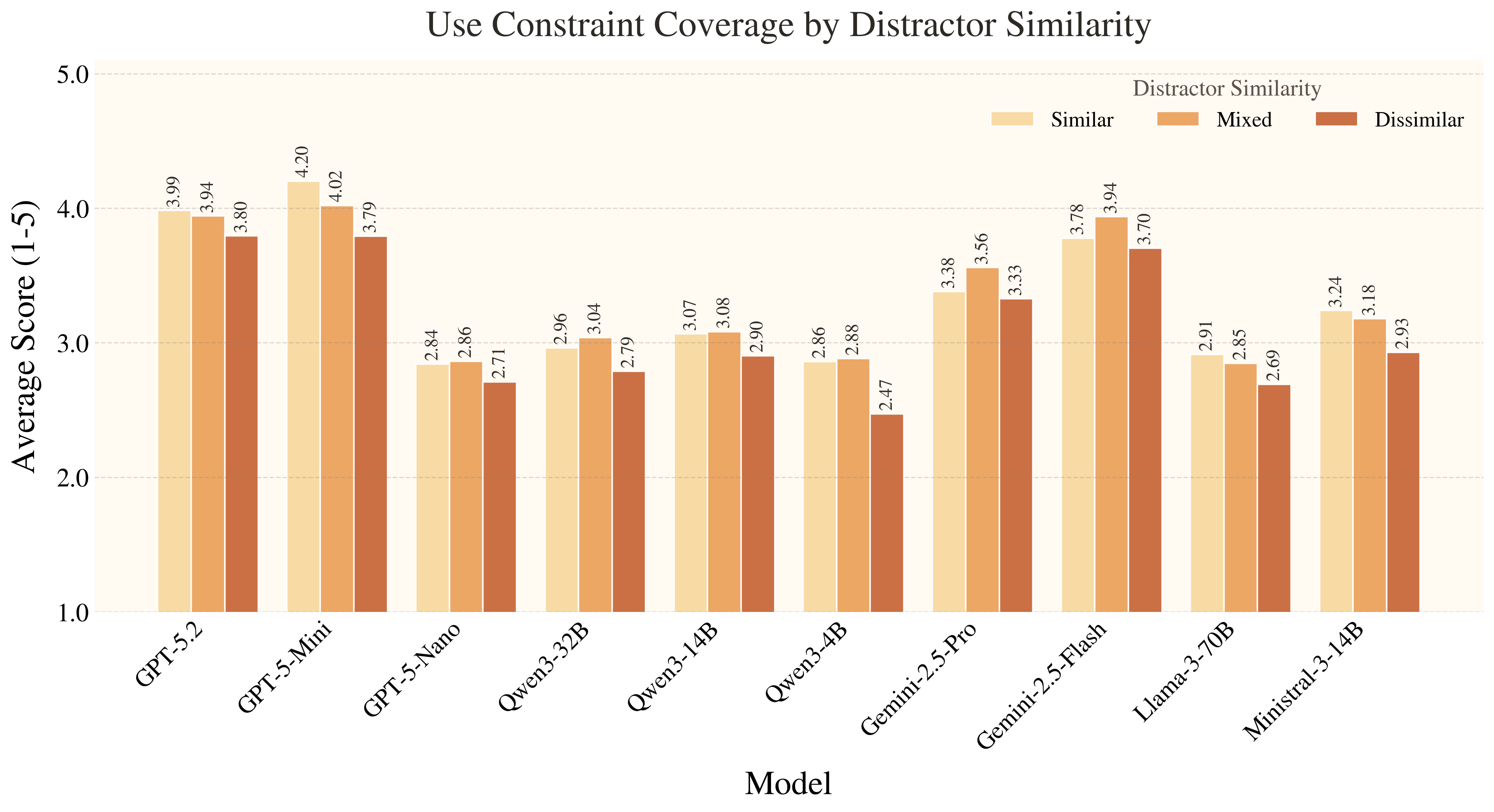}
    \end{subfigure}
    \hfill
    \begin{subfigure}[t]{0.49\linewidth}
        \centering
        \includegraphics[width=\linewidth]{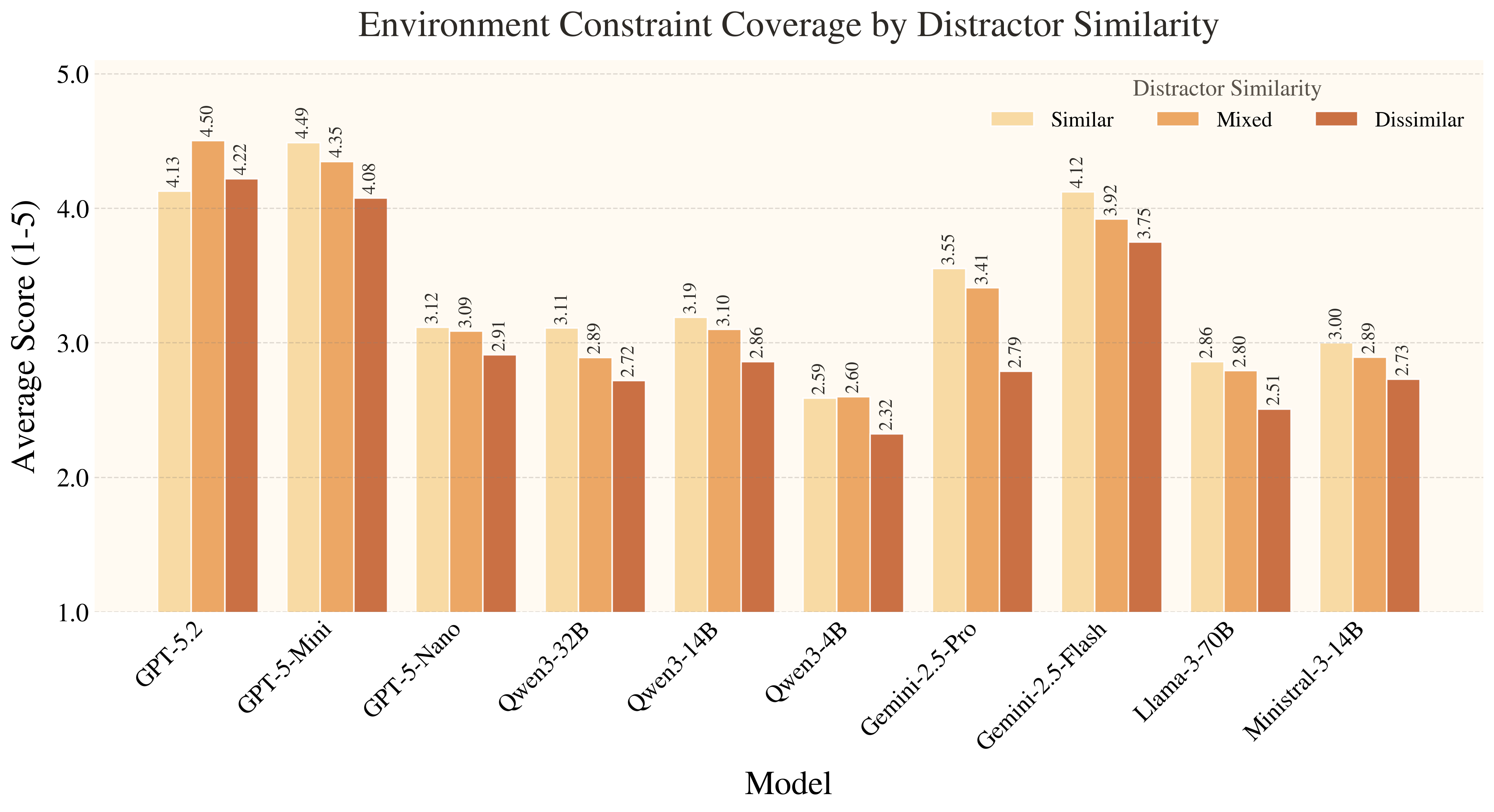}
    \end{subfigure}
    \hfill
    \begin{subfigure}[t]{0.49\linewidth}
        \centering
        \includegraphics[width=\linewidth]{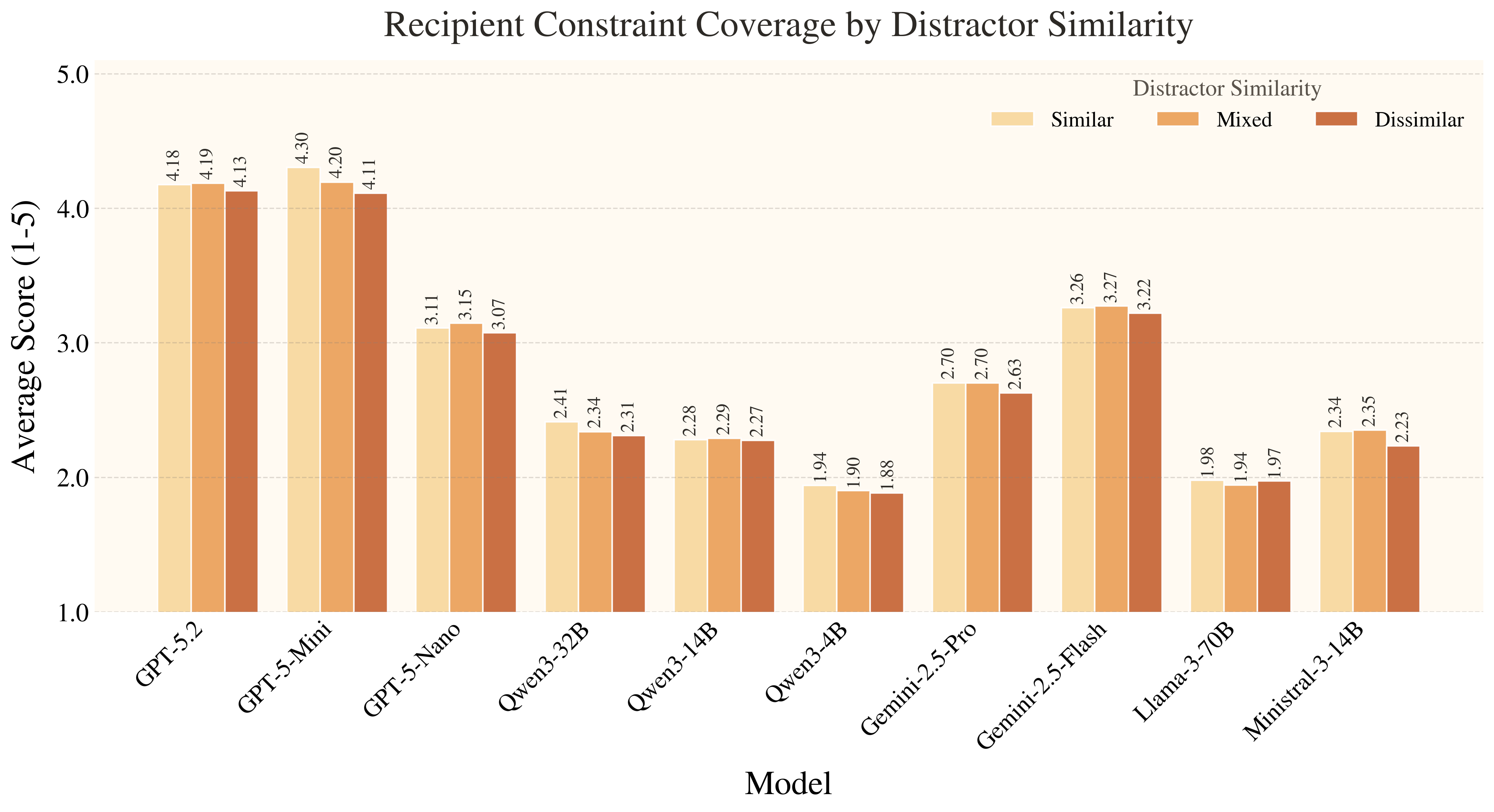}
    \end{subfigure}
    \caption{Fine-grained auxiliary metrics analysis upon different distractor similarity tiers.}
    \label{apdx:aux_metrics_similarity}
\end{figure}

\begin{figure}[t]
    \centering
    \begin{subfigure}[t]{0.49\linewidth}
        \centering
        \includegraphics[width=\linewidth]{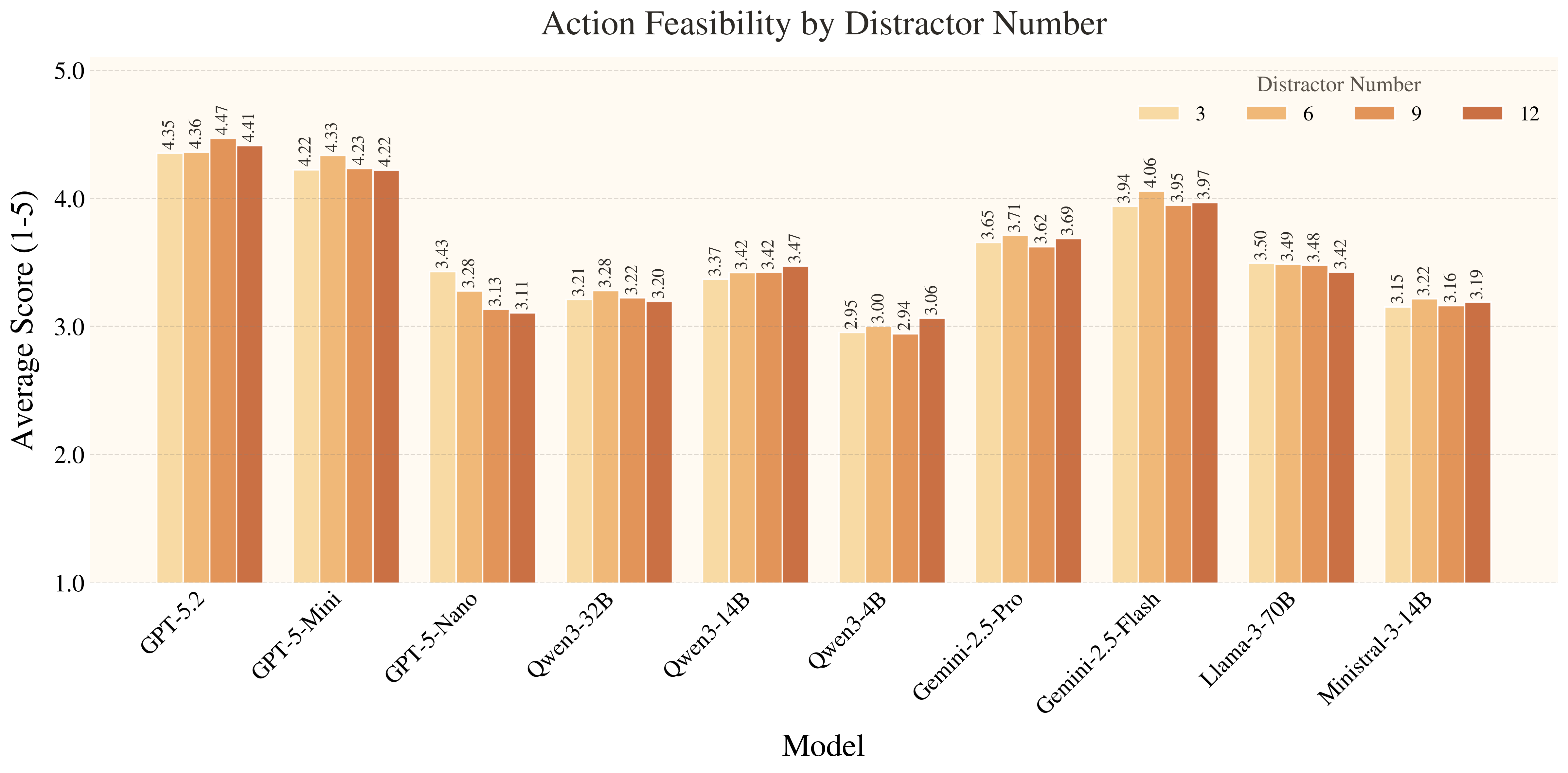}
    \end{subfigure}
    \hfill
    \begin{subfigure}[t]{0.49\linewidth}
        \centering
        \includegraphics[width=\linewidth]{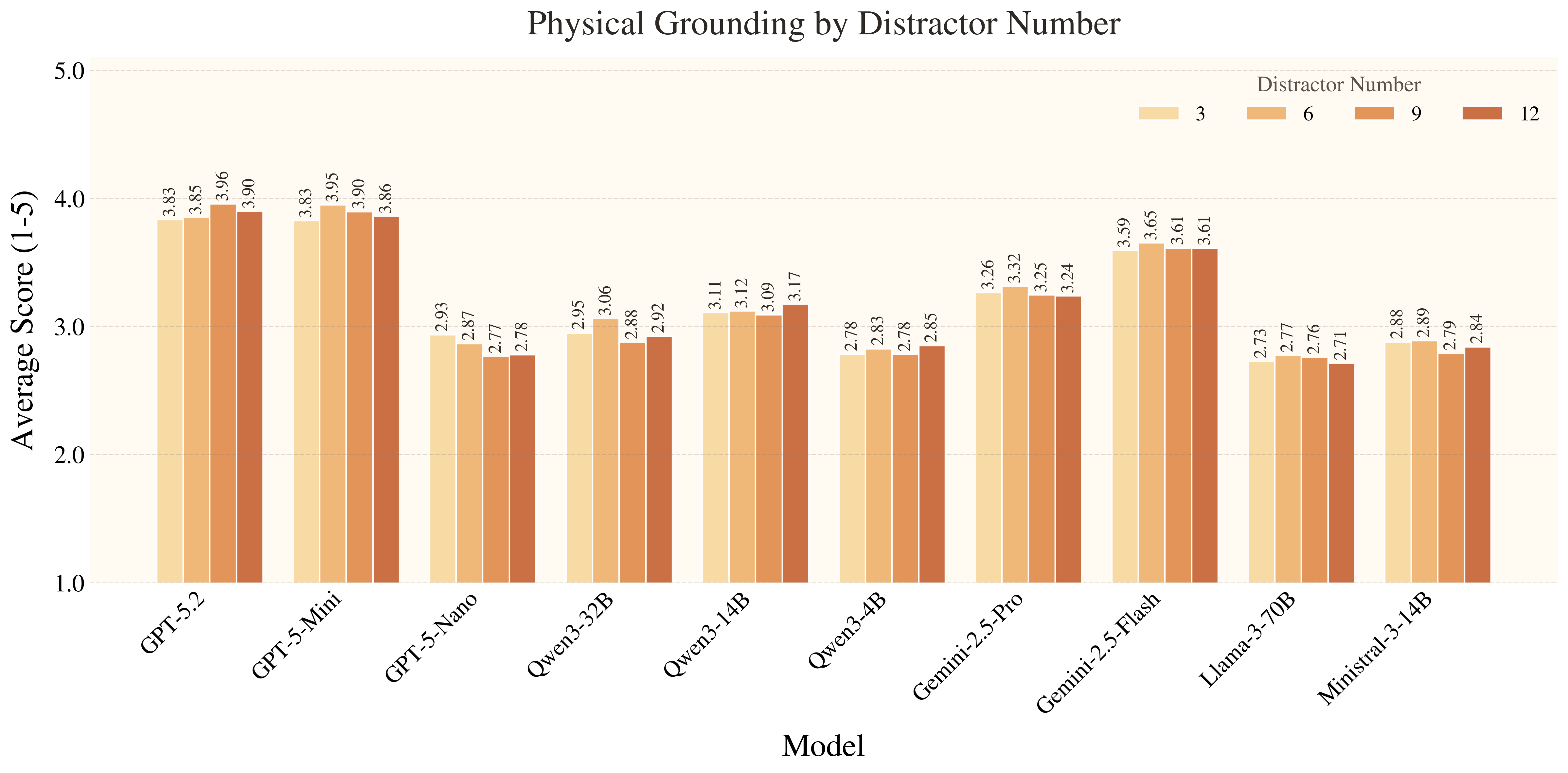}
    \end{subfigure}
    \hfill
    \begin{subfigure}[t]{0.49\linewidth}
        \centering
        \includegraphics[width=\linewidth]{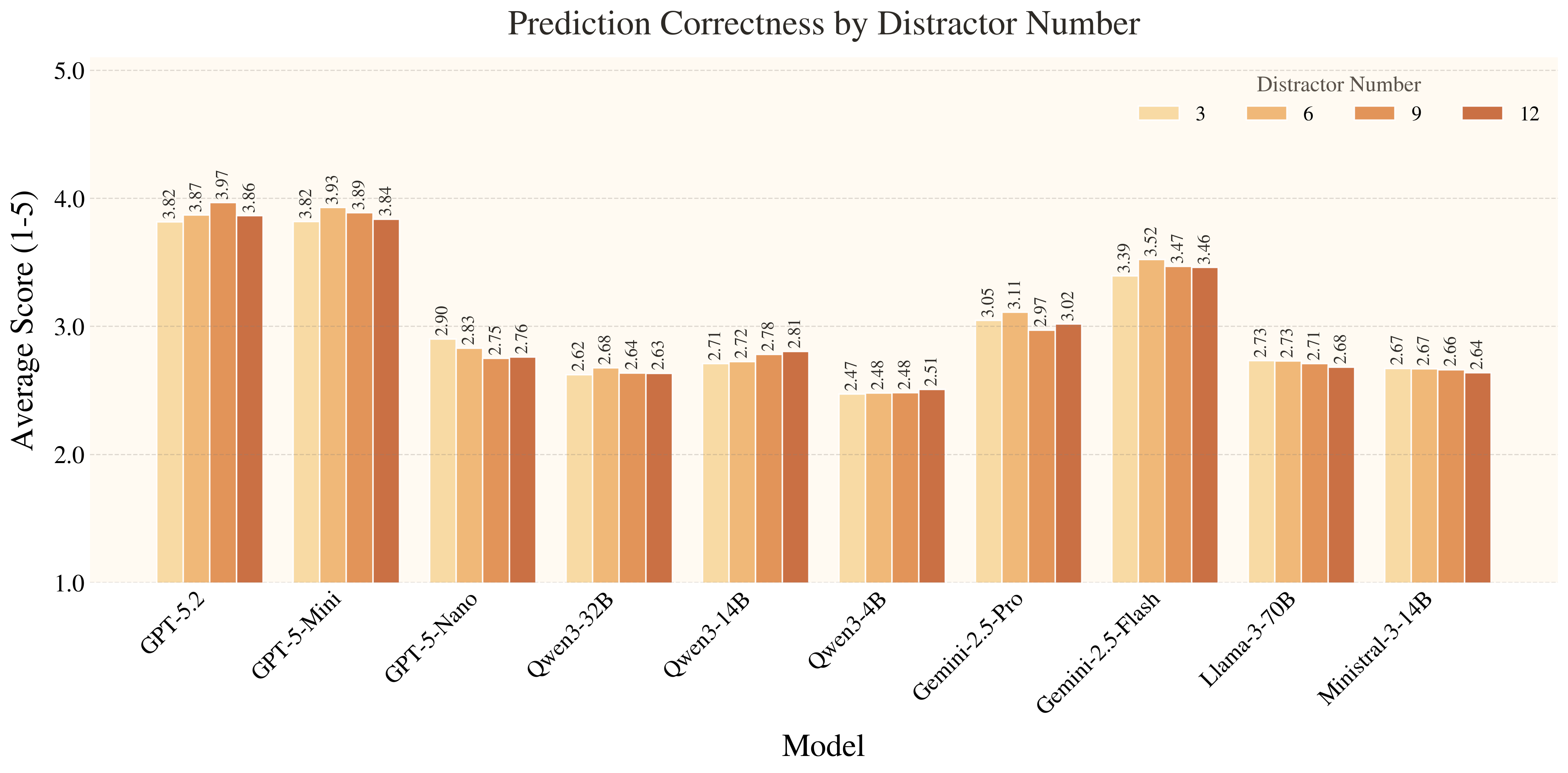}
    \end{subfigure}
    \hfill
    \begin{subfigure}[t]{0.49\linewidth}
        \centering
        \includegraphics[width=\linewidth]{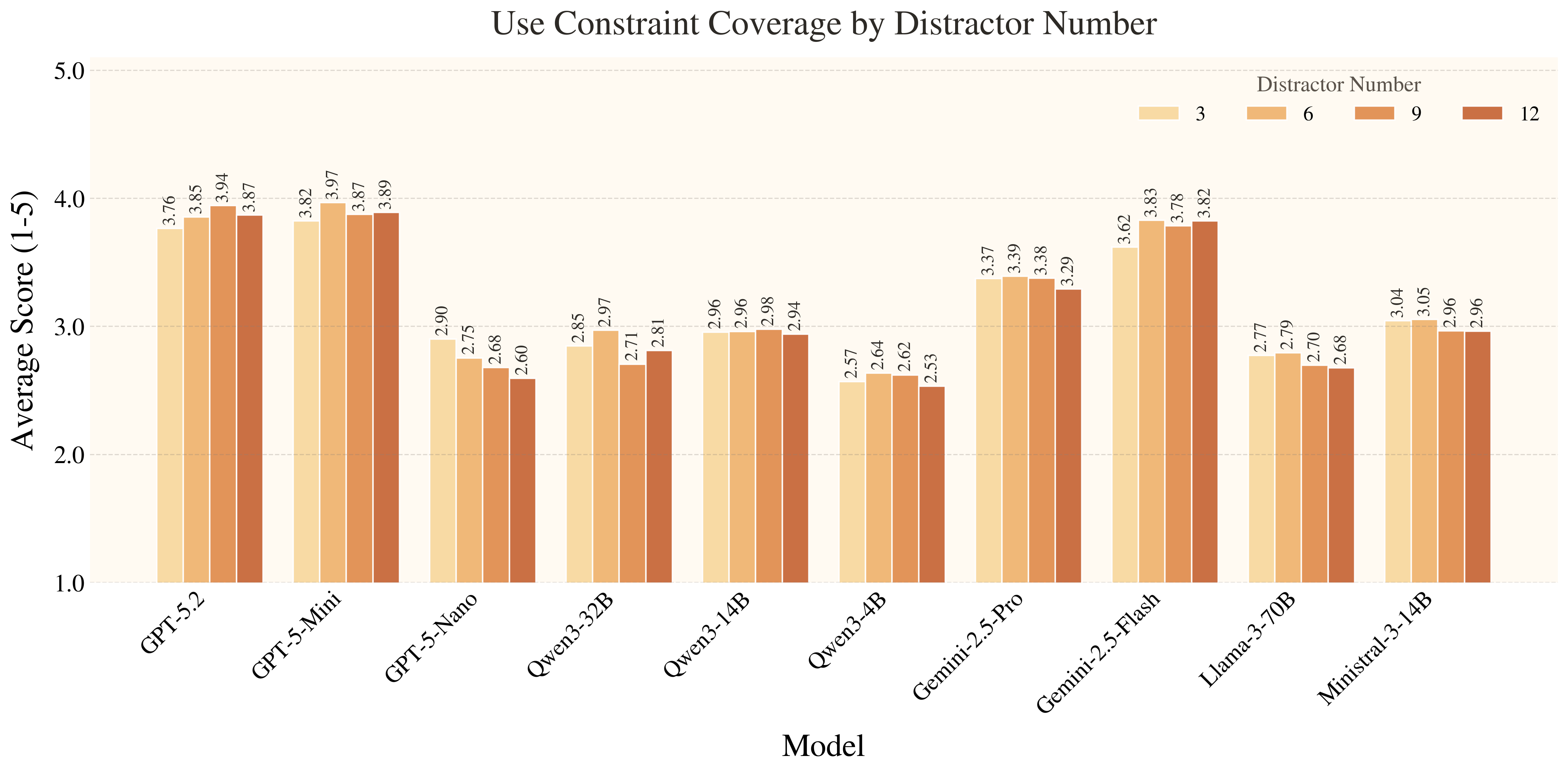}
    \end{subfigure}
    \hfill
    \begin{subfigure}[t]{0.49\linewidth}
        \centering
        \includegraphics[width=\linewidth]{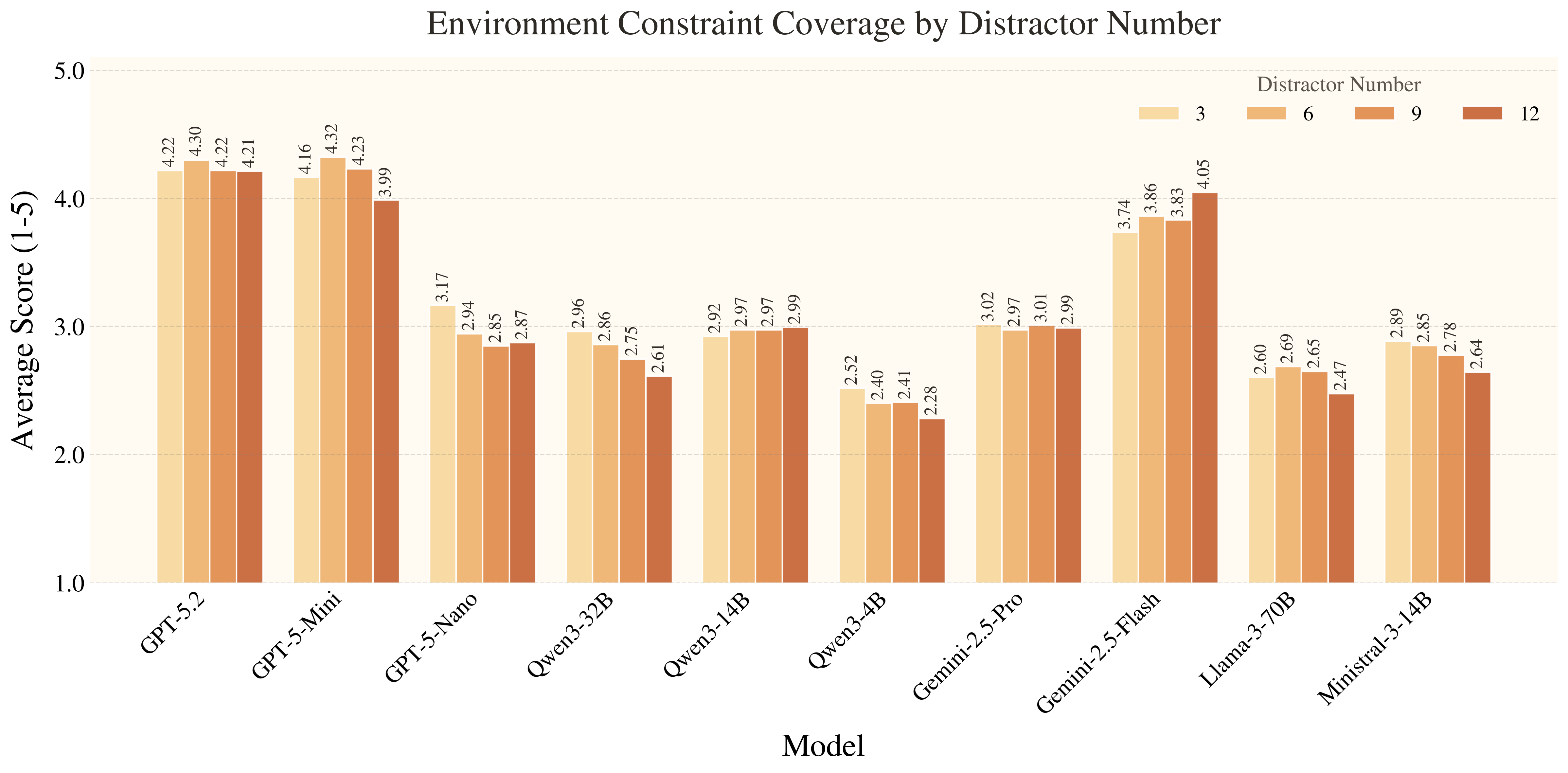}
    \end{subfigure}
    \hfill
    \begin{subfigure}[t]{0.49\linewidth}
        \centering
        \includegraphics[width=\linewidth]{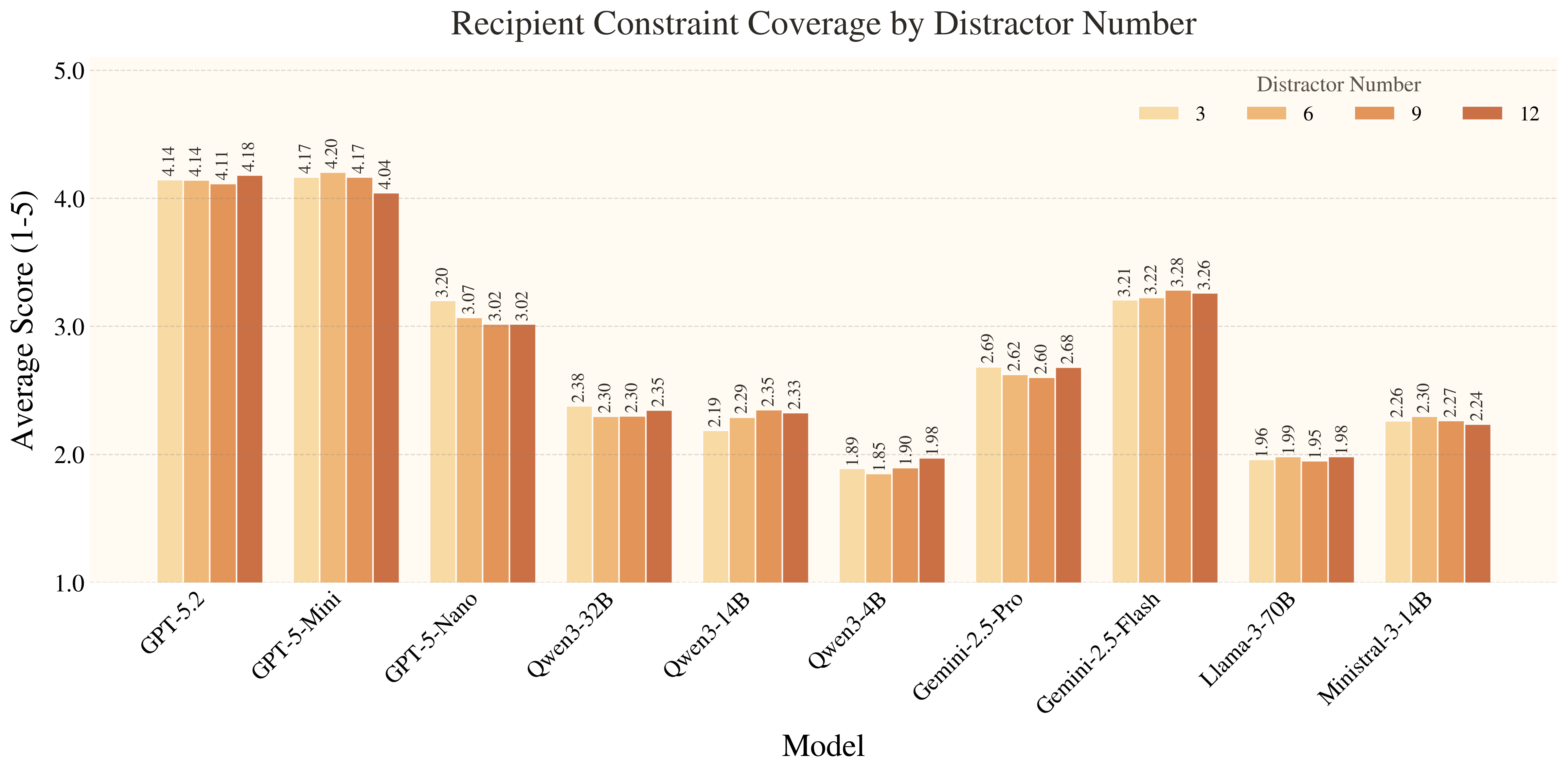}
    \end{subfigure}
    \caption{Fine-grained auxiliary metrics analysis upon different distractor numbers.}
    \label{apdx:aux_metrics_entity_count}
\end{figure}

\begin{figure}[t]
    \centering
    \begin{subfigure}[t]{0.49\linewidth}
        \centering
        \includegraphics[width=\linewidth]{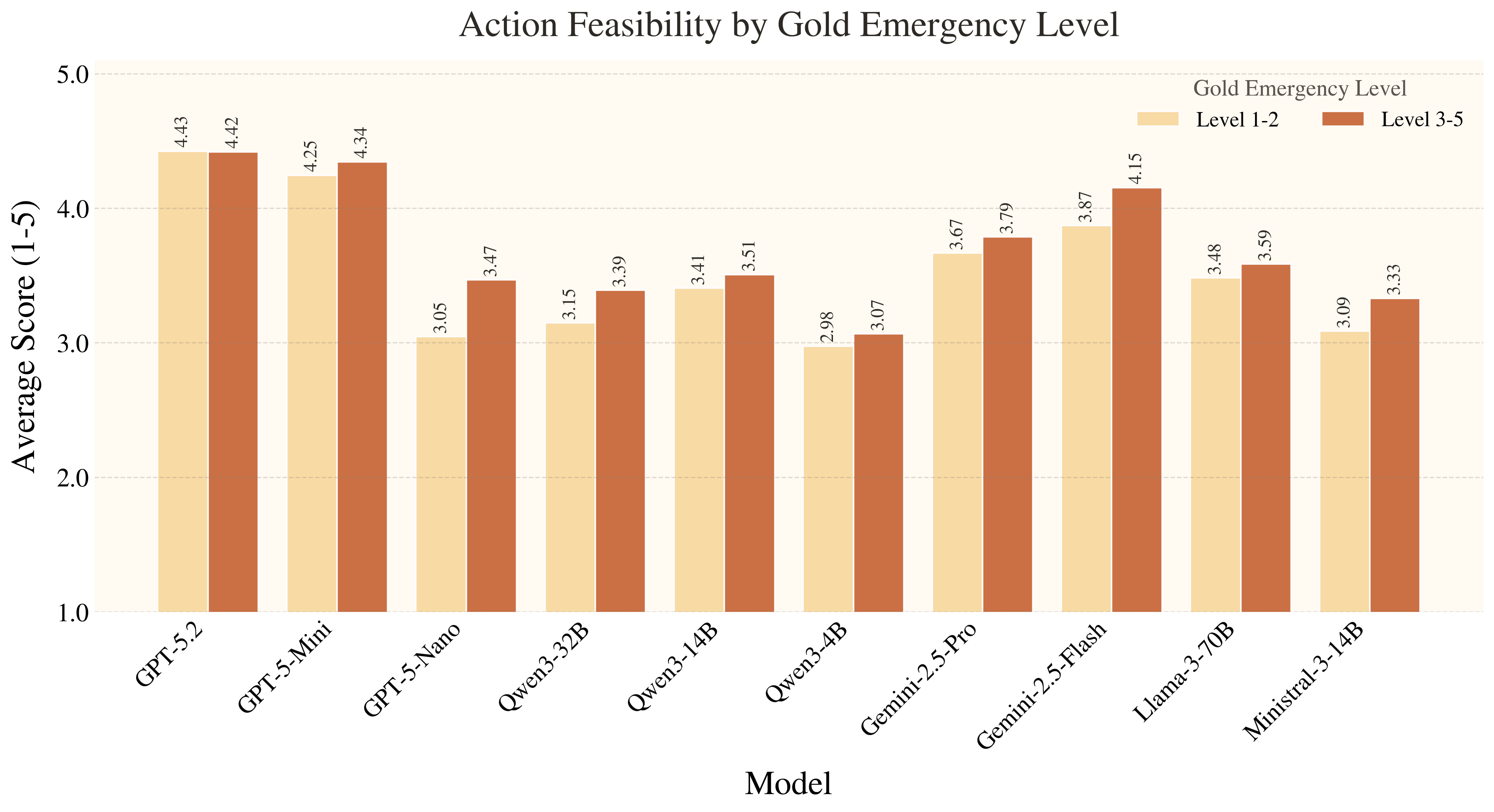}
    \end{subfigure}
    \hfill
    \begin{subfigure}[t]{0.49\linewidth}
        \centering
        \includegraphics[width=\linewidth]{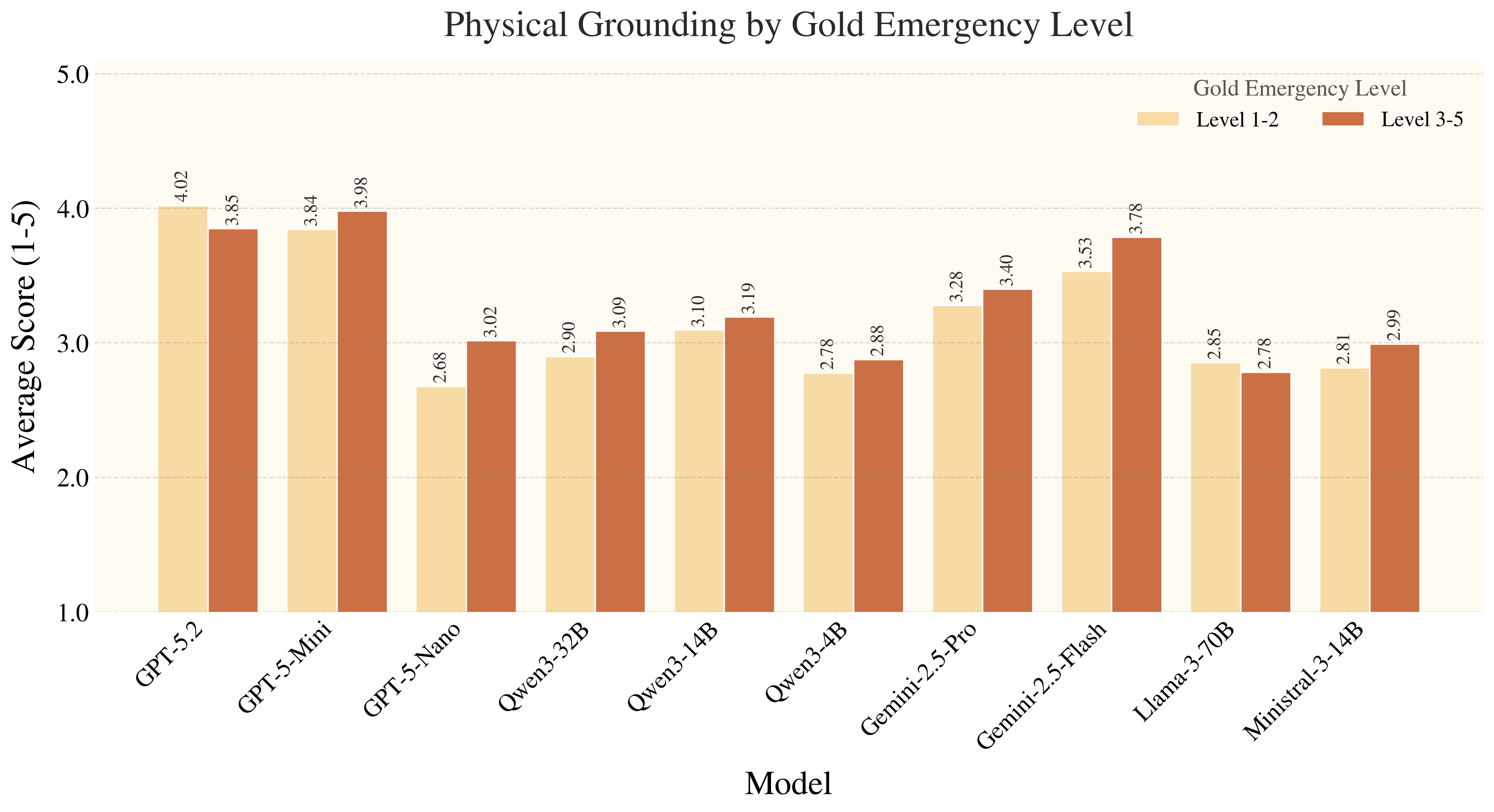}
    \end{subfigure}
    \hfill
    \begin{subfigure}[t]{0.49\linewidth}
        \centering
        \includegraphics[width=\linewidth]{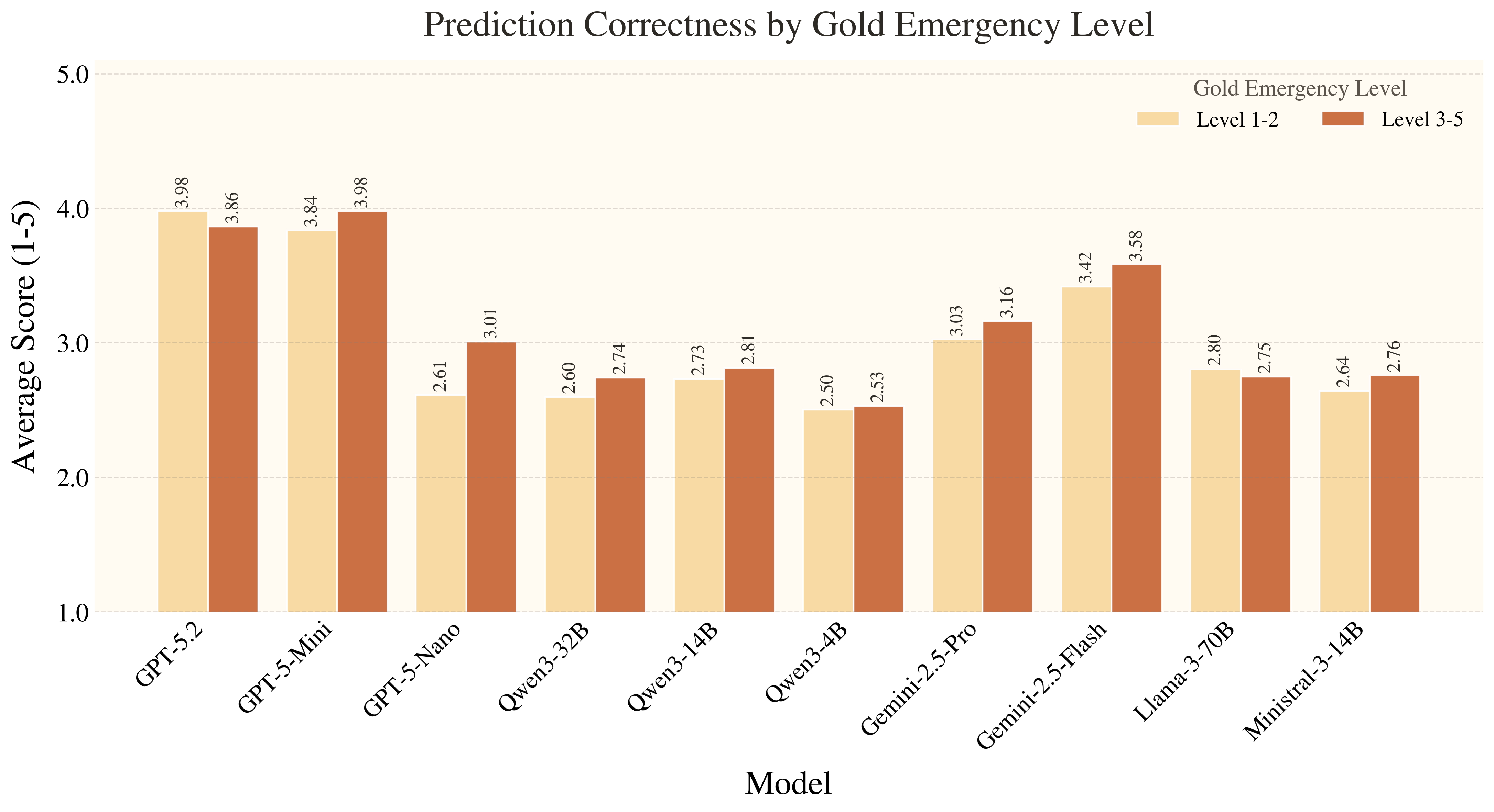}
    \end{subfigure}
    \hfill
    \begin{subfigure}[t]{0.49\linewidth}
        \centering
        \includegraphics[width=\linewidth]{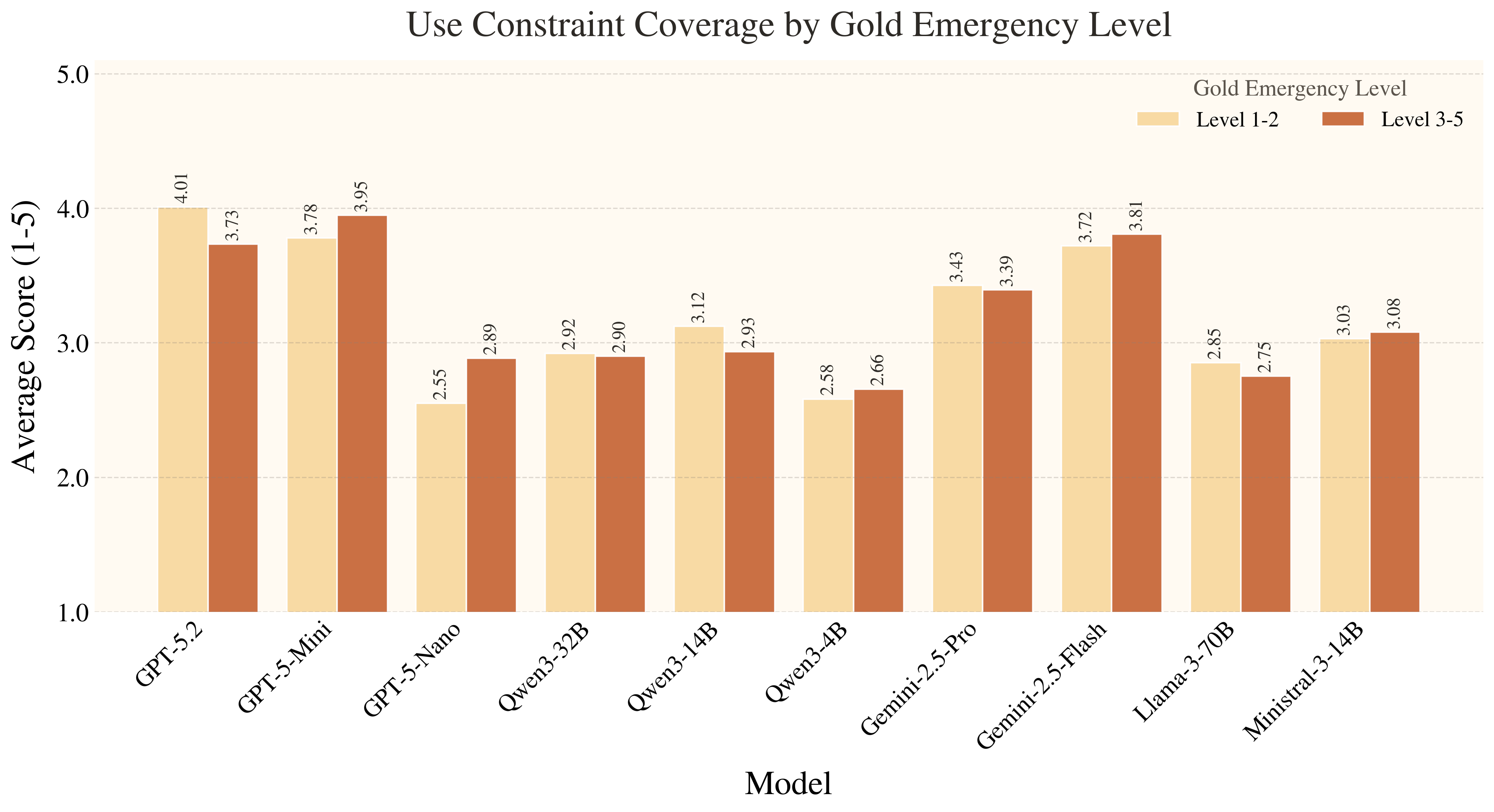}
    \end{subfigure}
    \hfill
    \begin{subfigure}[t]{0.49\linewidth}
        \centering
        \includegraphics[width=\linewidth]{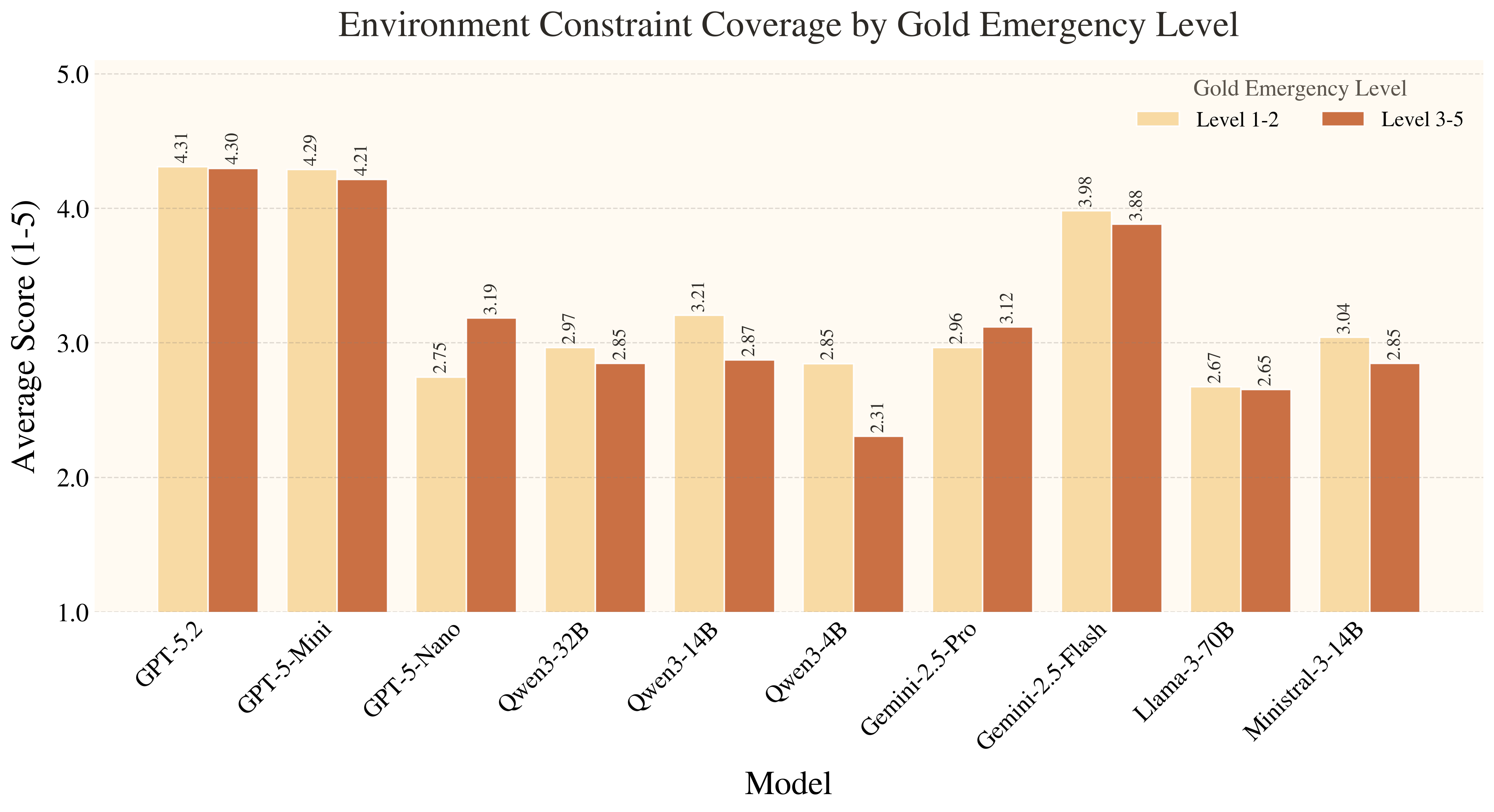}
    \end{subfigure}
    \hfill
    \begin{subfigure}[t]{0.49\linewidth}
        \centering
        \includegraphics[width=\linewidth]{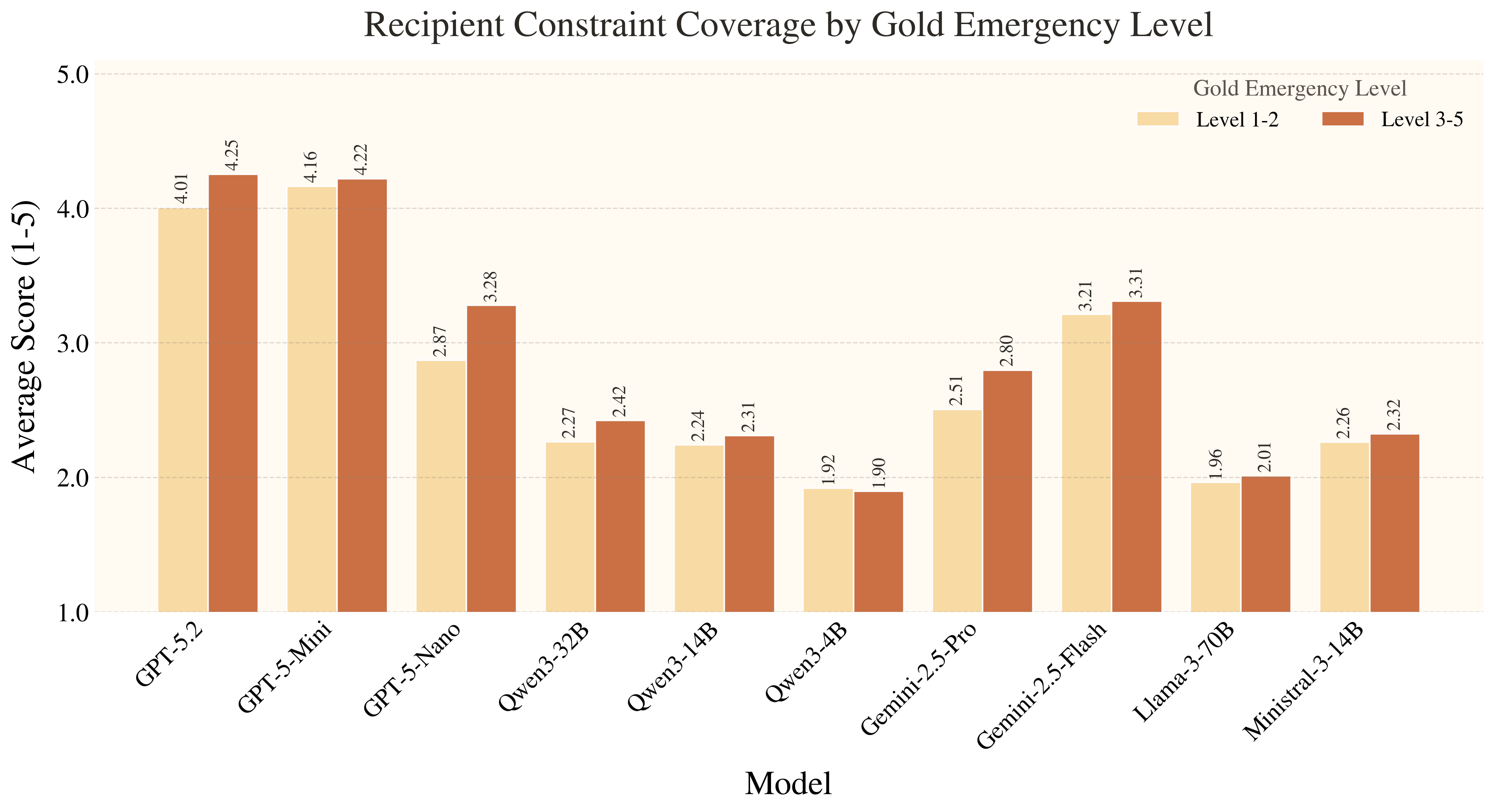}
    \end{subfigure}
    \caption{Fine-grained auxiliary metrics analysis upon two grouped gold emergency level bands.}
    \label{apdx:aux_metrics_level_2bars}
\end{figure}

From \Cref{apdx:aux_metrics_cluster_size} to \Cref{apdx:aux_metrics_level_2bars}, we present the complete results of the fine-grained analysis of the auxiliary metrics on different analysis aspects including gold affordance cluster size, gold affordance emergency level, distractor similarity, and distractor number.

\subsection{Error analysis details}
For error analysis, we use Gemini-3.1-Flash-Lite as the LLM judge on a 10\% sample of failure cases for each model. The temperature is set to zero to ensure deterministic judgments. The judging protocol largely follows the same criteria used in the main-results analysis.

More specifically, the judge is instructed to evaluate each case from two perspectives. First, it assesses the prediction on its own, focusing on constraint satisfaction~\citep{liu2026dynamic}, physical grounding, and action feasibility. Second, it compares the prediction against the gold solution, using the gold supporting rationale introduced during task creation in \Cref{sec:method}. In this comparative step, the judge determines whether the gold solution is indeed preferable, or whether the prediction still has meaningful merits, and summarizes the respective pros and cons of both.

Note that, consistent with the analysis in the main experimental setting and the preliminary analysis in \Cref{sec:preliminaries}, we rescale the judge's original score from the range 0--2 to 1--5 after scoring. The detailed prompts for these two judgment aspects are provided below.

\begin{tcolorbox}[
  enhanced,
  breakable,
  width=0.98\linewidth,
  colback=tzBlueFill,
  colframe=tzBlueBorder,
  boxrule=1.2pt,
  arc=6pt,
  left=5pt,right=5pt,top=4pt,bottom=2pt,
  title={\small System Prompt for Fine-grained Metric-wise Judgment},
  coltitle=white,
  colbacktitle=tzBlueHeader2,
  fonttitle=\bfseries,
]
\small
\begin{lstlisting}[style=jsonTiny]
You are a strict evaluator for practical goal completion.
The predicted entity/part is NOT exactly gold.
Judge whether the predicted "how_to_use" can still accomplish the task reasonably and safely.

Task:
{task.get("task", "")}

Gold solution JSON (only for reference):
{json.dumps(task.get("solution", {}), ensure_ascii=False, indent=2)}

Physical attributes of the predicted part:
{json.dumps(pred_physical_attrs, ensure_ascii=False, indent=2)}

State attributes of the predicted part:
{json.dumps(pred_state_attrs, ensure_ascii=False, indent=2)}

Predicted entity to use: {pred_entity}
Predicted part to use: {pred_part}
Predicted how to use the part:
{how_to_use}

Field-by-field rubric when judging the predicted "how_to_use":
1) environment_condition_covered (0/1/2/"NA"): [Do NOT directly compare with the gold, it's only for reference]
- "NA": in your reasoning, deeply consider for the predicted use of the part to accomplish the task, if the external environment setup is needed; if not, then please give "NA".
- 0: the environment setup is needed but in the predicted use, it is not mentioned or covered at all.
- 1: the environment setup is needed but in the predicted use, it is mentioned but not all are covered.
- 2: the environment setup is needed but in the predicted use, it is covered well and reasonable, aligning with common sense.

2) use_condition_covered (0/1/2/"NA"): [Do NOT directly compare with the gold, it's only for reference]
- "NA": in your reasoning,deeply consider for the predicted use of the part to accomplish the task, if the preparation/access of the part is needed; if not, then please give "NA".
- 0: the preparation/access of the part is needed but in the predicted use, it is not mentioned or covered at all.
- 1: the preparation/access of the part is needed but in the predicted use, it is mentioned but not all are covered.
- 2: the preparation/access of the part is needed but in the predicted use, it is covered well and reasonable, aligning with common sense.

3) recipient_condition_covered (0/1/2/"NA"): [Do NOT directly compare with the gold, it's only for reference]
- "NA": in your reasoning, deeply consider for the predicted use of the part to accomplish the task, if the recipient-side prerequisites are needed; if not, then please give "NA".
- 0: the recipient-side prerequisites are needed but in the predicted use, it is not mentioned or covered at all.
- 1: the recipient-side prerequisites are needed but in the predicted use, it is mentioned but not all are covered.
- 2: the recipient-side prerequisites are needed but in the predicted use, it is covered well and reasonable, aligning with common sense.

4) attributes_grounding (0/1/2): [Compare with the physical and state attributes of the predicted part]
- 0: the predicted action is not grounded in the key enabling attributes of the part, or it violates some attributes of the part.
- 1: the predicted action is mostly grounded in the key enabling attributes of the gold affordance, but not all are covered or implied.
- 2: the predicted action is fully grounded in the key enabling attributes of the gold affordance, and most of them are explicitly mentioned, covered or implied.

5) action_feasibility (0/1/2): [Please only focus on the predicted action itself]
- 0: the action itself is physically impossible, not working at all, very unlikely to be used in practice, or unsafe.
- 1: the action itself is partially workable but still there are some steps that are not plausible, aligned with common sense or not feasible.
- 2: the action itself is operationally correct and complete, almost completely aligned with common sense and feasible in practice.

Evidence policy:
- Every *_reason must quote concrete evidence from predicted how to use the part and other given relevant information.
- If evidence is unclear/missing, default to a stricter outcome (lower score).
- Make your reasoning clear, concise, and to the point for each field, and your score should be based on the evidence.

Please make sure to only return a valid JSON with exactly these 12 fields (no extras, no markdown).
{{
    "environment_condition_covered_reason": "...",
    "environment_condition_covered": 0/1/2/"NA",
    "use_condition_covered_reason": "...",
    "use_condition_covered": 0/1/2/"NA",
    "recipient_condition_covered_reason": "...",
    "recipient_condition_covered": 0/1/2/"NA",
    "attributes_grounding_reason": "...",
    "attributes_grounding": 0/1/2,
    "action_feasibility_reason": "...",
    "action_feasibility": 0/1/2,
}}
\end{lstlisting}
\end{tcolorbox}

\begin{tcolorbox}[
  enhanced,
  breakable,
  width=0.98\linewidth,
  colback=tzBlueFill,
  colframe=tzBlueBorder,
  boxrule=1.2pt,
  arc=6pt,
  left=5pt,right=5pt,top=4pt,bottom=2pt,
  title={\small System Prompt for Gold-Prediction Comparison Judgment},
  coltitle=white,
  colbacktitle=tzBlueHeader2,
  fonttitle=\bfseries,
]
\small
\begin{lstlisting}[style=jsonTiny]
You are a strict evaluator comparing predicted substitution vs ground-truth preference.
You are given the model prediction and the a ground-truth comparison judgment for that predicted entity+part.
Judge how convincing the model prediction remains AFTER reading the comparison evidence.

Task:
{task.get("task", "")}

Predicted entity to use: {pred_entity}
Predicted part to use: {pred_part}
Predicted how to use the part:
{how_to_use}

Ground-truth comparison judgment for the predicted entity+part:
{json.dumps(pred_comp if pred_comp else {}, ensure_ascii=False, indent=2)}

Return ONLY JSON:
{{
  "prediction_reasonable_reason": "...",
  "prediction_reasonable_level": 0 or 1 or 2
}}

Scoring rubric:
- 0: prediction is somewhat reasonable, but ground truth is clearly better after reading the comparison judgment.
- 1: prediction has meaningful merit; trade-offs are close and prediction may be acceptable after reading the comparison judgment.
- 2: prediction is strongly convincing and should replace current gold choice after reading the comparison judgment.

Reasoning requirements:
- Reference available comparison aspects (accessibility, side-effects, willingness, commonness, safety, etc.).
- Consider whether predicted "how to use the part" strengthens or weakens the substitution.
- Penalize unsupported claims.

Please make sure to only return a valid JSON with exactly these 2 fields (no extras, no markdown).
\end{lstlisting}
\end{tcolorbox}

\subsection{Attribution analysis details}

We further conduct the attribution analysis on a randomly sampled 10\% subset of cases using Gemini-3.1-Flash-Lite as the categorization model, with the judging temperature fixed at 0.0. To support the reliability of using Gemini as the judge, we additionally repeat the same categorization on the same sampled subset with Qwen3-32B and GPT-5-Mini. We find that Gemini's predicted primary category overlaps with Qwen3-32B in 83.3\% of cases and with GPT-5-Mini in 90.91\% of cases. This high agreement suggests strong cross-model consistency in the attribution judgments, which supports the validity of the taxonomy assignment and justifies our use of Gemini-3.1-Flash-Lite as the main categorization model.

Specifically, we feed the gold comparison reason into the categorization judge and instruct it to assign exactly one primary contributing factor, along with any additional contributing categories when appropriate. The detailed prompt is shown below:

\begin{tcolorbox}[
  enhanced,
  breakable,
  width=0.98\linewidth,
  colback=tzBlueFill,
  colframe=tzBlueBorder,
  boxrule=1.2pt,
  arc=6pt,
  left=5pt,right=5pt,top=4pt,bottom=2pt,
  title={\small System Prompt for Gold-Prediction Comparison Judgment},
  coltitle=white,
  colbacktitle=tzBlueHeader2,
  fonttitle=\bfseries,
]
\small
\begin{lstlisting}[style=jsonTiny]
You are an expert judge for error analysis of failed physical tool-use predictions.

You will receive one sentence that explains why a prediction was judged unreasonable.
Your job is to classify the failure reason into the taxonomy below.

Categories:

A. Physical / functional invalidity
A1. Hallucinated affordance / wrong object model
- The prediction invents a feature, subpart, or capability the object does not actually have.

A2. Affordance mismatch (wrong geometry / material / mechanics)
- The object is real, but its actual shape, material, or operating principle is fundamentally wrong for the task.

A3. Performance mismatch (too weak / small / unstable / imprecise)
- The object has a somewhat relevant affordance, but not enough capacity, stability, size, precision, or strength to work well.

B. Practical invalidity
B1. Destructive or cannibalizing workaround
- The proposal requires damaging, dismantling, cutting, or sacrificing the donor object or another useful object.

B2. Context / accessibility / process impracticality
- The proposal is too fiddly, inaccessible, complex, unrealistic, or unlikely to be available in the stated situation.

C. Risk / requirement mismatch
C1. Safety / contamination / damage risk
- The proposal is unsafe, unsanitary, electrically risky, scratch-prone, or likely to damage something.

C2. Violates explicit user constraints or the object's primary purpose
- The proposal ignores explicit constraints or repurposes an object in a way that directly conflicts with what the user asked to avoid.

D. Comparative inferiority rather than true failure
D1. Functionally works, but is less preferable in normal use
- The solution is workable, but worse on convenience, accessibility, hygiene, robustness, elegance, or normal practicality.

D2. Functionally competitive / arguably as good as gold
- The solution is actually reasonable and not clearly worse; the supposed error is mostly preference-sensitive.

Instructions:
- Base the judgment only on the input sentence.
- Do brief internal reasoning and then return a JSON object only.
- Use only the small category codes like "A1", "B2", "D1".
- "primary_category" must be exactly one code from the taxonomy.
- "all_categories" must be a list of one or more codes from the taxonomy, including the primary category.
- Include multiple codes only when the sentence clearly supports multiple failure reasons.
- Keep "reason" concise, clear, and to the point.
- Do not output markdown, prose outside JSON, or any extra fields.

Return exactly this schema:
{{
  "reason": "explanation",
  "primary_category": "A1",
  "all_categories": ["A1", ...]
}}

Input reason:
{sentence}
\end{lstlisting}
\end{tcolorbox}

\begin{figure}[t]
    \centering
    \begin{subfigure}[t]{0.37\linewidth}
        \centering
        \includegraphics[width=\linewidth]{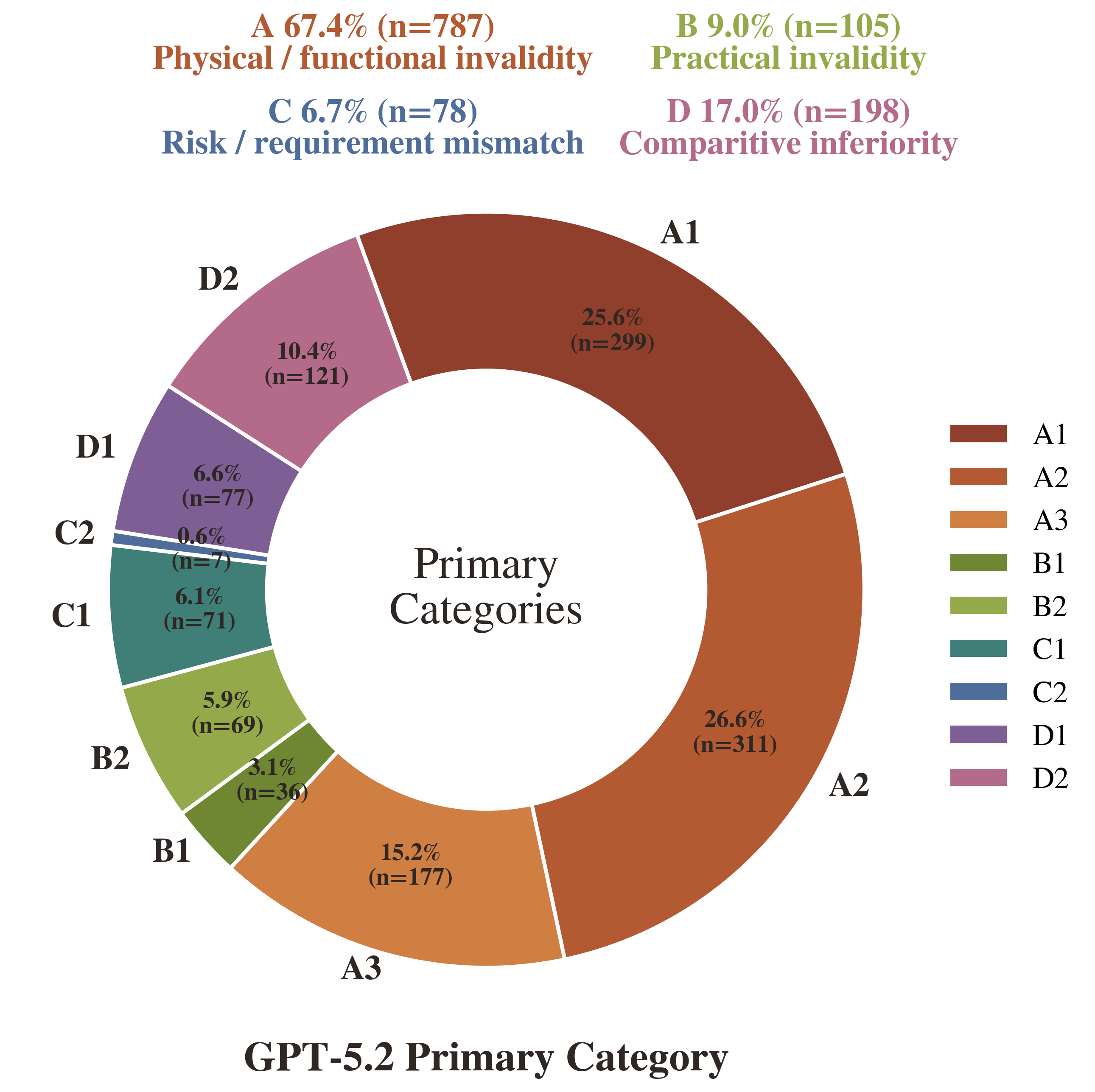}
        \label{fig:attribution_pie_gpt52}
    \end{subfigure}
    \hfill
    \begin{subfigure}[t]{0.6\linewidth}
        \centering
        \includegraphics[width=\linewidth]{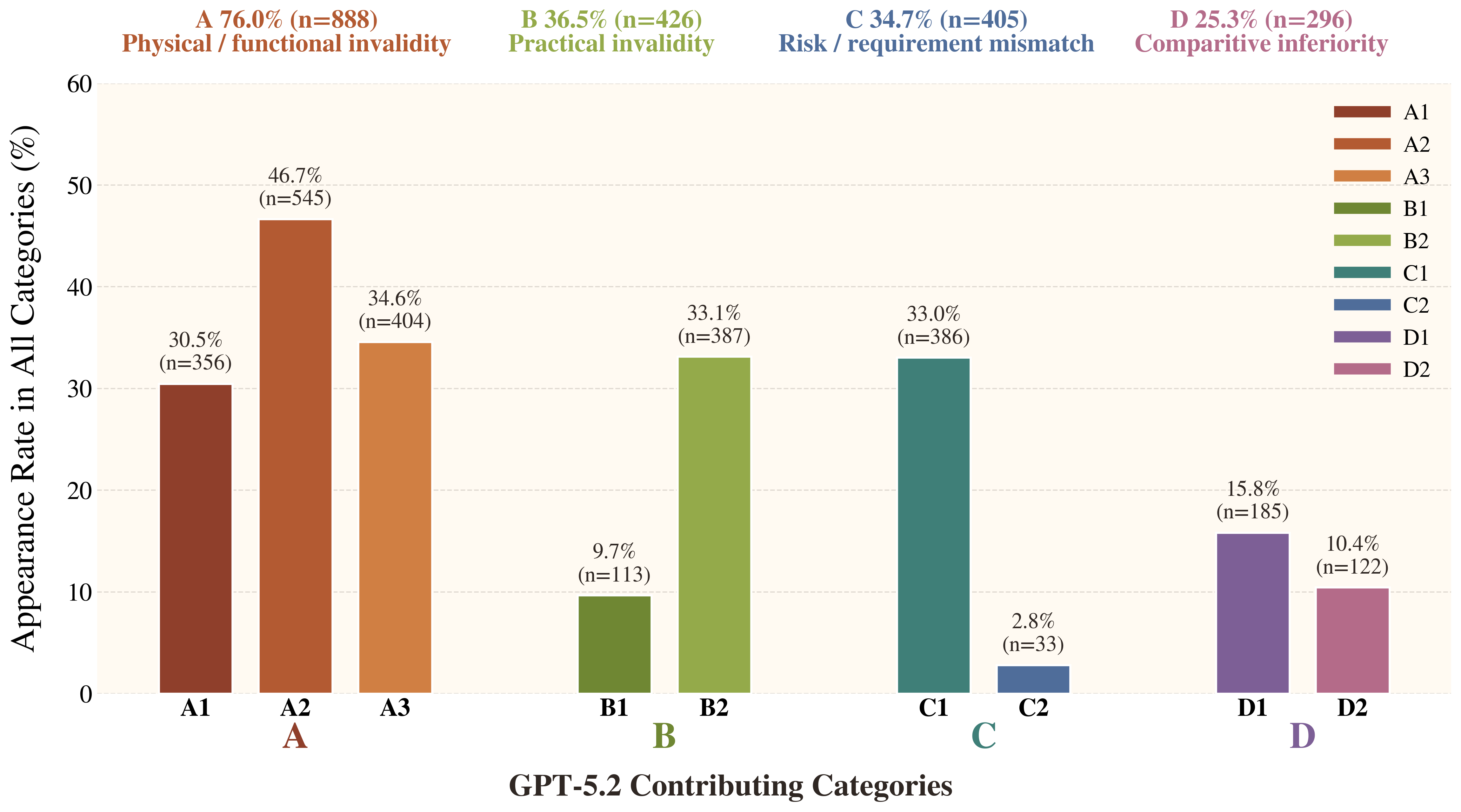}
        \label{fig:attribution_bar_gpt52}
    \end{subfigure}
    \caption{The error attribution analysis of GPT-5.2 respectively on primary category and all contributing reasons.}
    \label{fig:attribution_pie_bar_gpt52}
\end{figure}

\begin{figure}[t]
    \centering
    \begin{subfigure}[t]{0.37\linewidth}
        \centering
        \includegraphics[width=\linewidth]{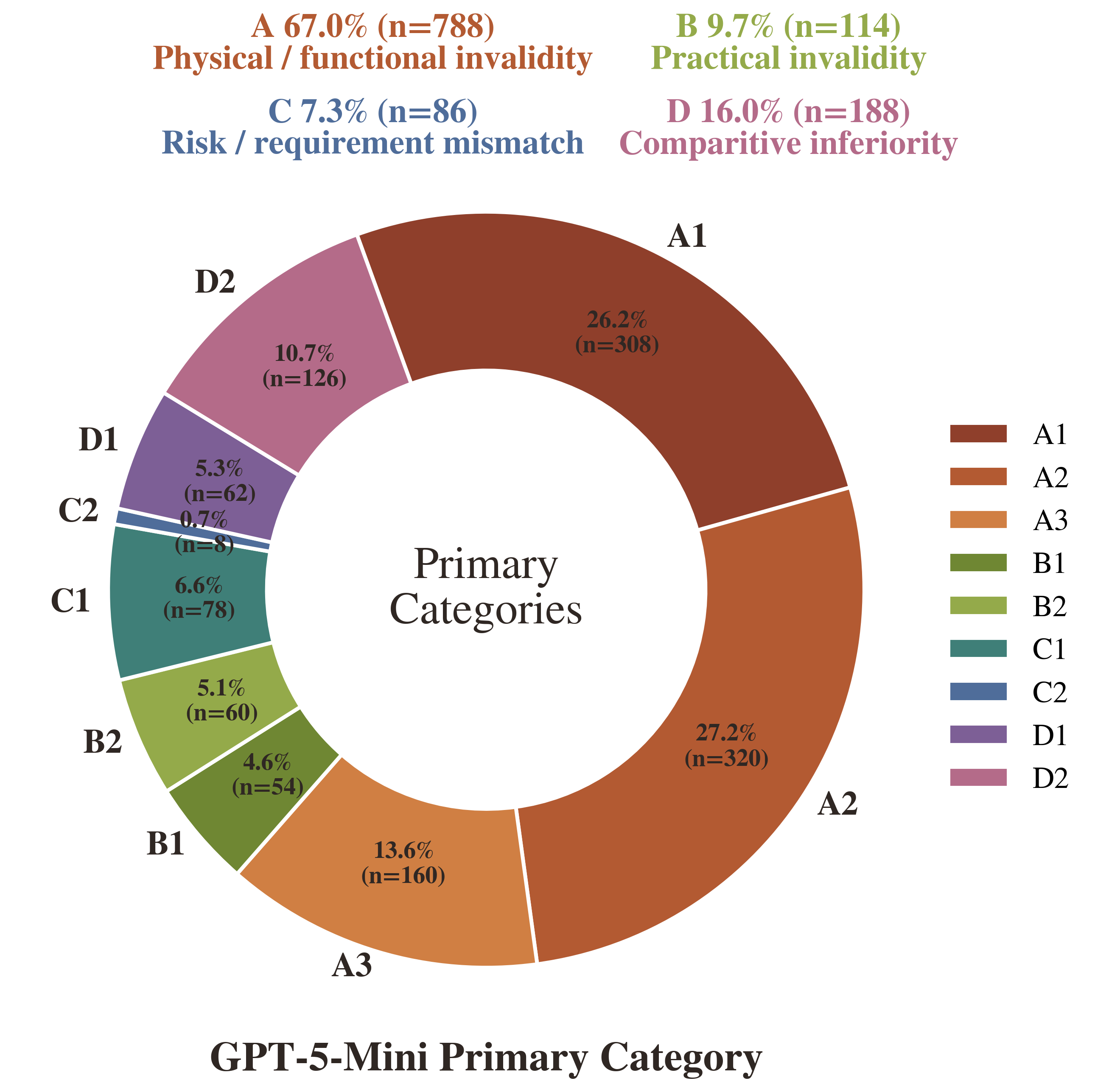}
        \label{fig:attribution_pie_gpt5mini}
    \end{subfigure}
    \hfill
    \begin{subfigure}[t]{0.6\linewidth}
        \centering
        \includegraphics[width=\linewidth]{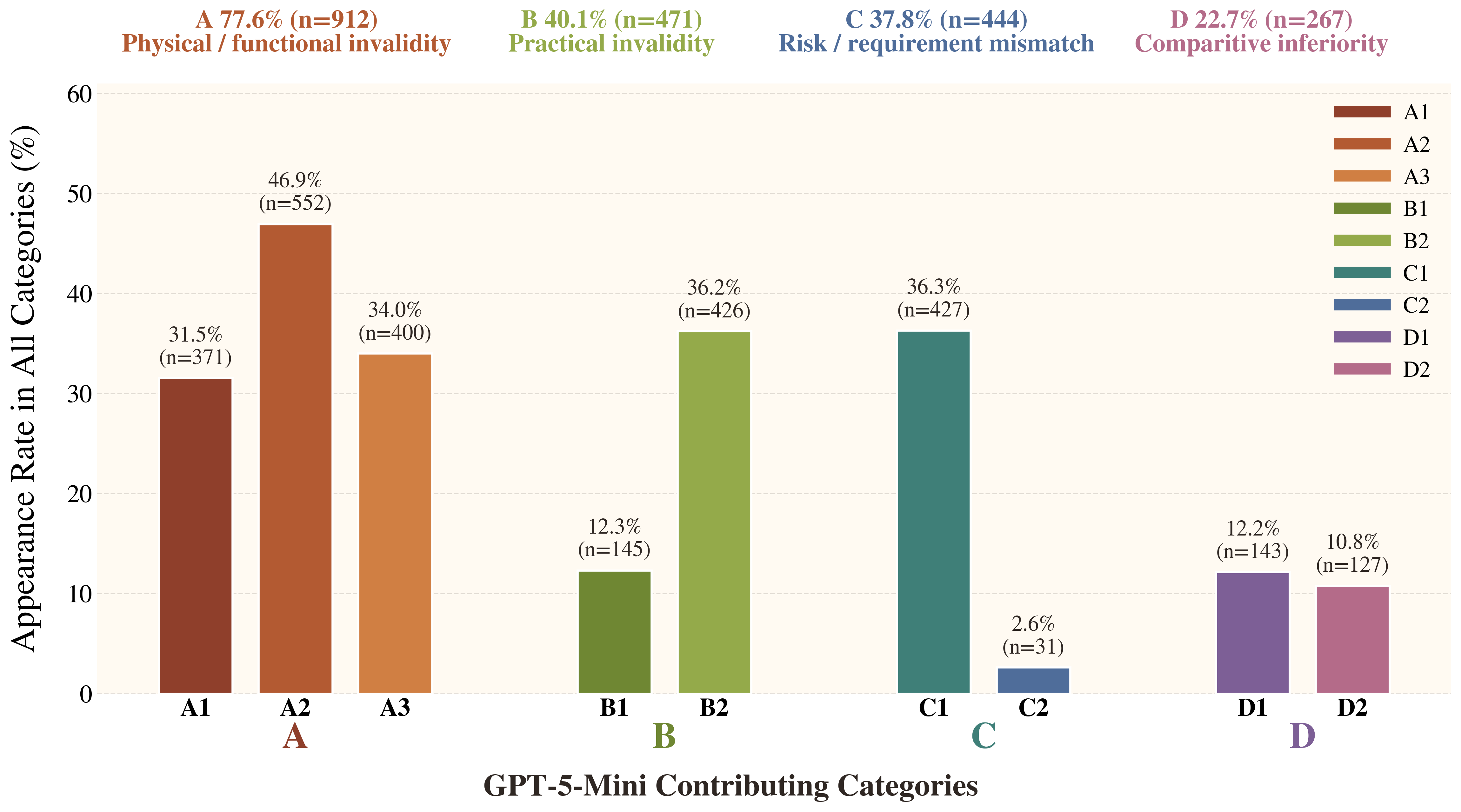}
        \label{fig:attribution_bar_gpt5mini}
    \end{subfigure}
    \caption{The error attribution analysis of GPT-5-Mini respectively on primary category and all contributing reasons.}
    \label{fig:attribution_pie_bar_gpt5mini}
\end{figure}

\begin{figure}[t]
    \centering
    \begin{subfigure}[t]{0.37\linewidth}
        \centering
        \includegraphics[width=\linewidth]{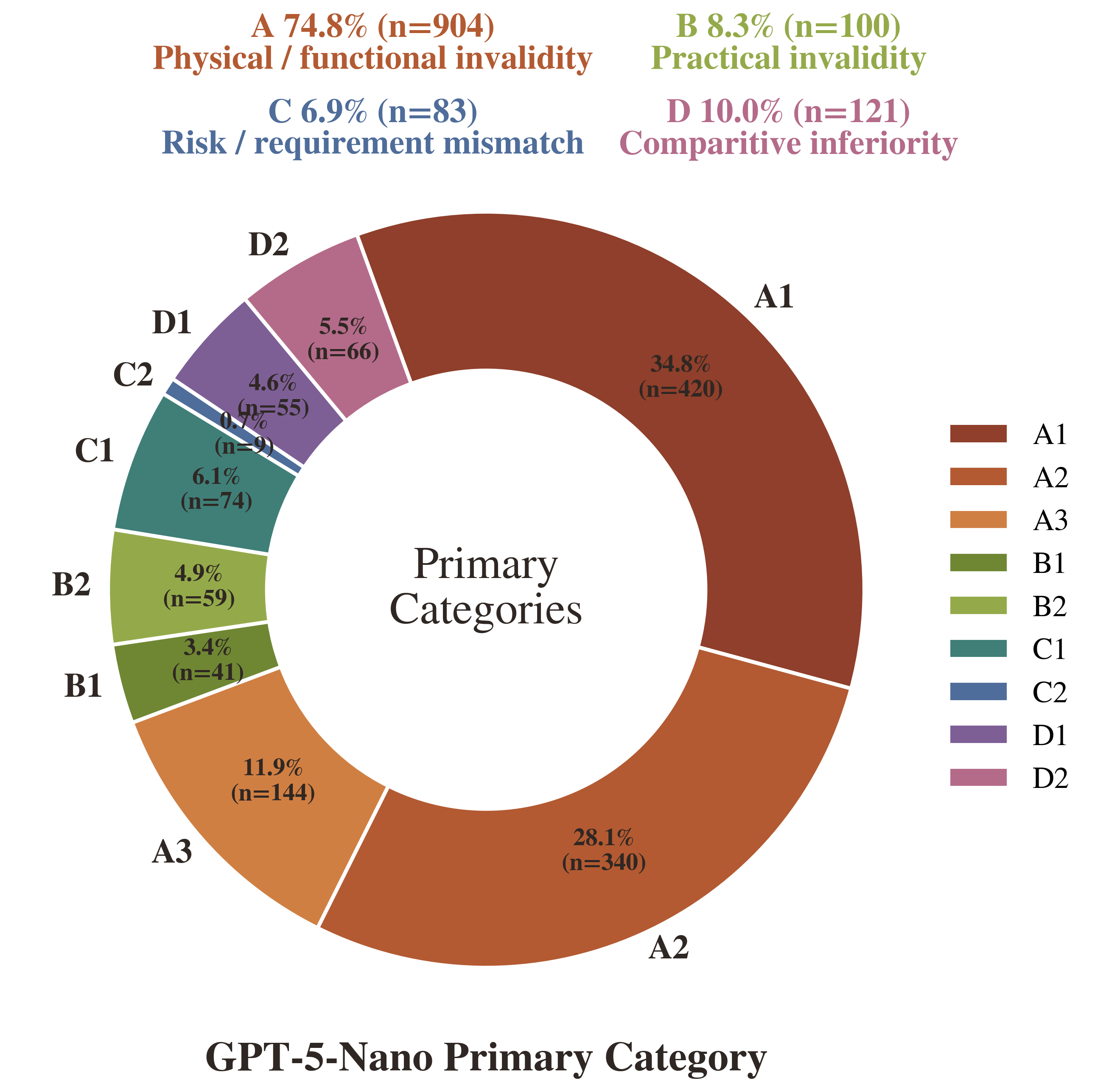}
        \label{fig:attribution_pie_gpt5nano}
    \end{subfigure}
    \hfill
    \begin{subfigure}[t]{0.6\linewidth}
        \centering
        \includegraphics[width=\linewidth]{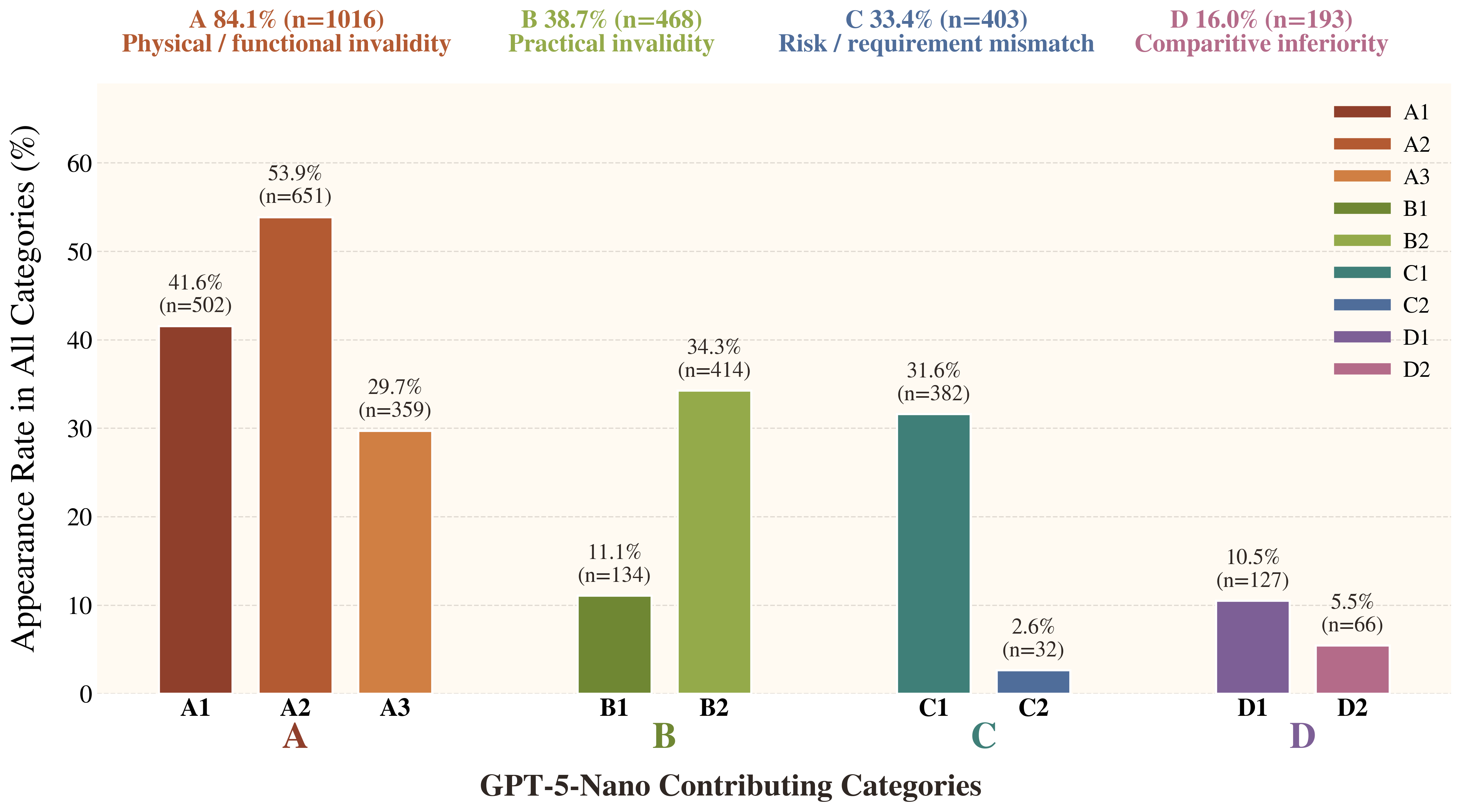}
        \label{fig:attribution_bar_gpt5nano}
    \end{subfigure}
    \caption{The error attribution analysis of GPT-5-Nano respectively on primary category and all contributing reasons.}
    \label{fig:attribution_pie_bar_gpt5nano}
\end{figure}

\begin{figure}[t]
    \centering
    \begin{subfigure}[t]{0.37\linewidth}
        \centering
        \includegraphics[width=\linewidth]{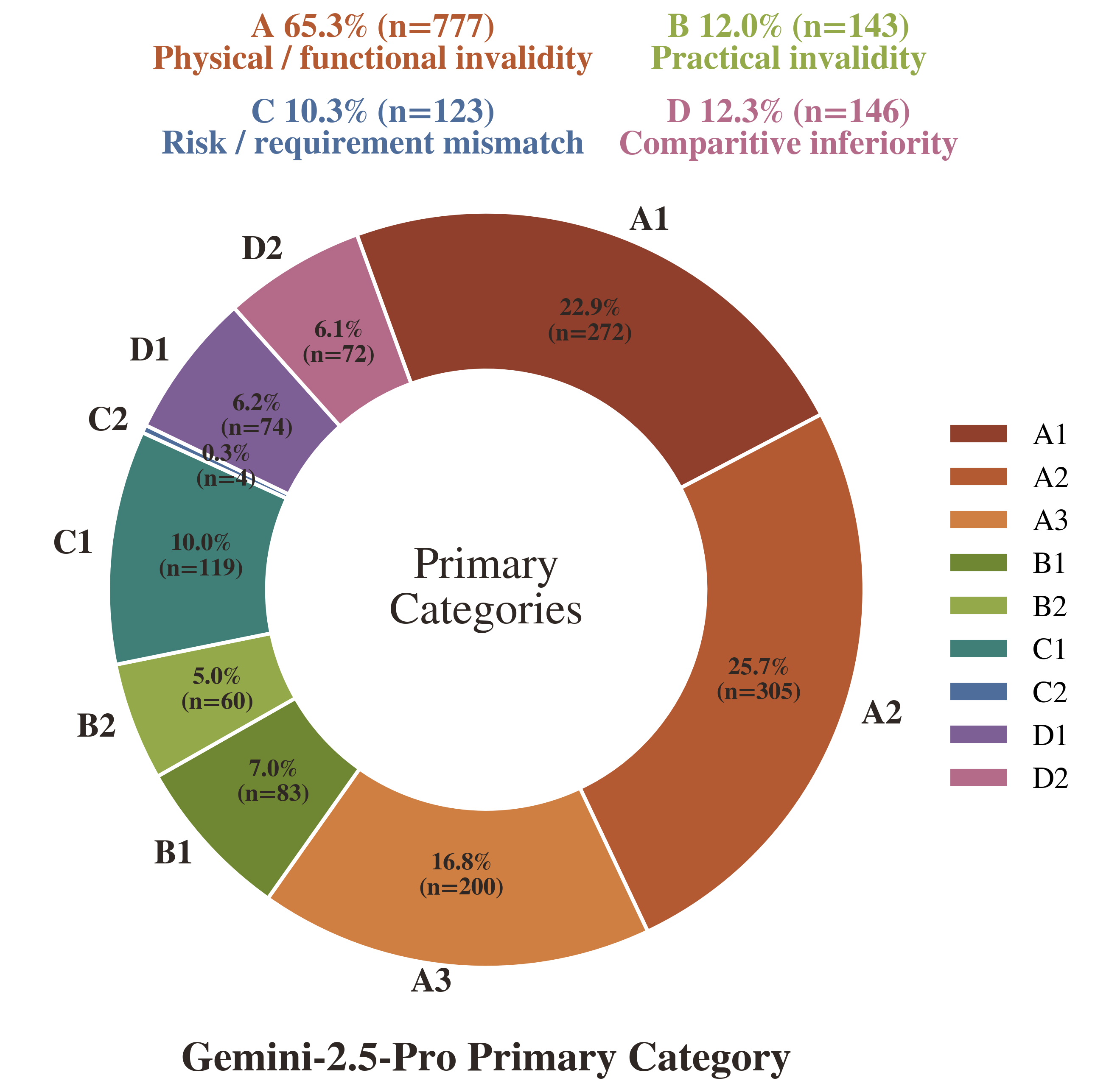}
        \label{fig:attribution_pie_gemini25pro}
    \end{subfigure}
    \hfill
    \begin{subfigure}[t]{0.6\linewidth}
        \centering
        \includegraphics[width=\linewidth]{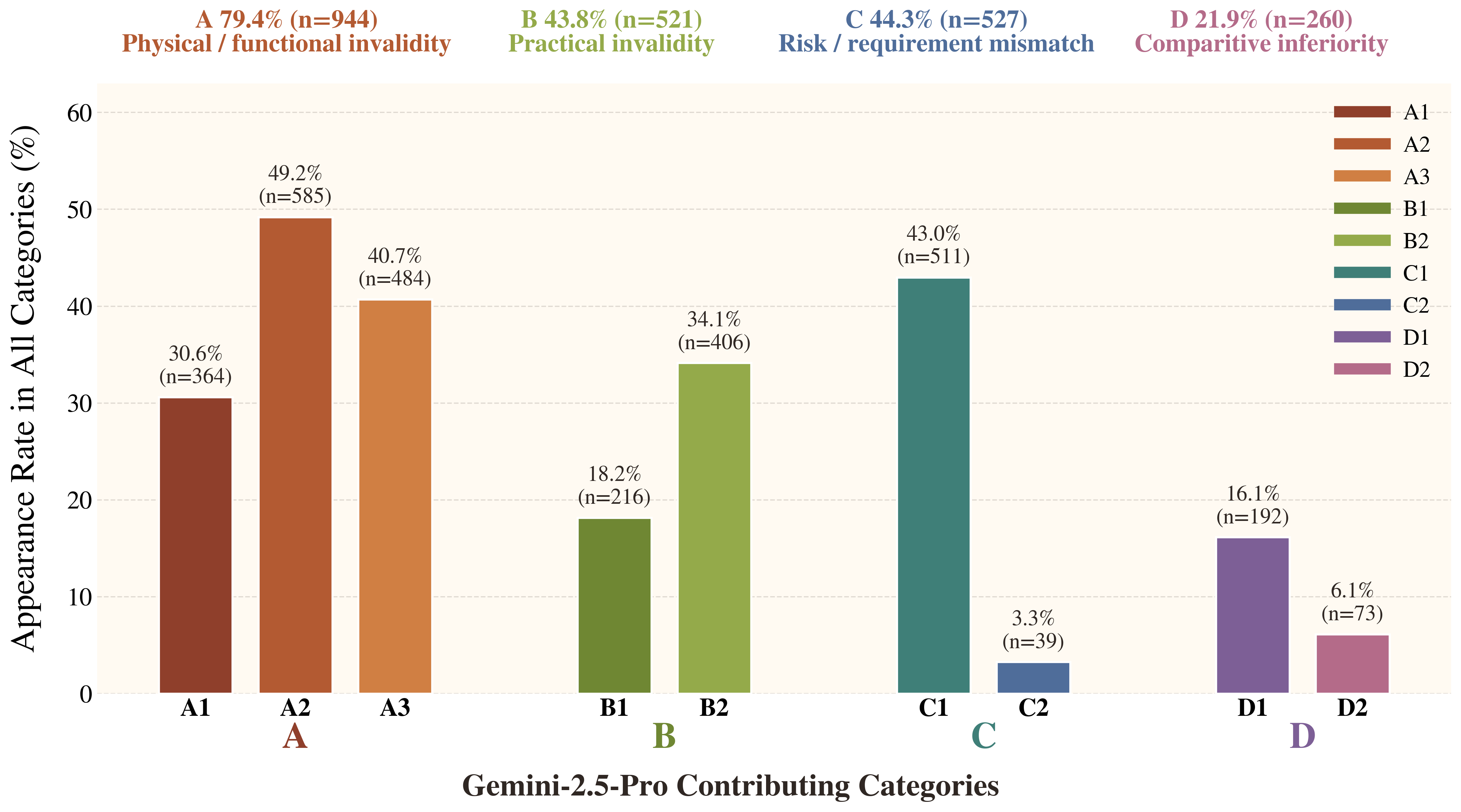}
        \label{fig:attribution_bar_gemini25pro}
    \end{subfigure}
    \caption{The error attribution analysis of Gemini-2.5-Pro respectively on primary category and all contributing reasons.}
    \label{fig:attribution_pie_bar_gemini25pro}
\end{figure}

\begin{figure}[t]
    \centering
    \begin{subfigure}[t]{0.37\linewidth}
        \centering
        \includegraphics[width=\linewidth]{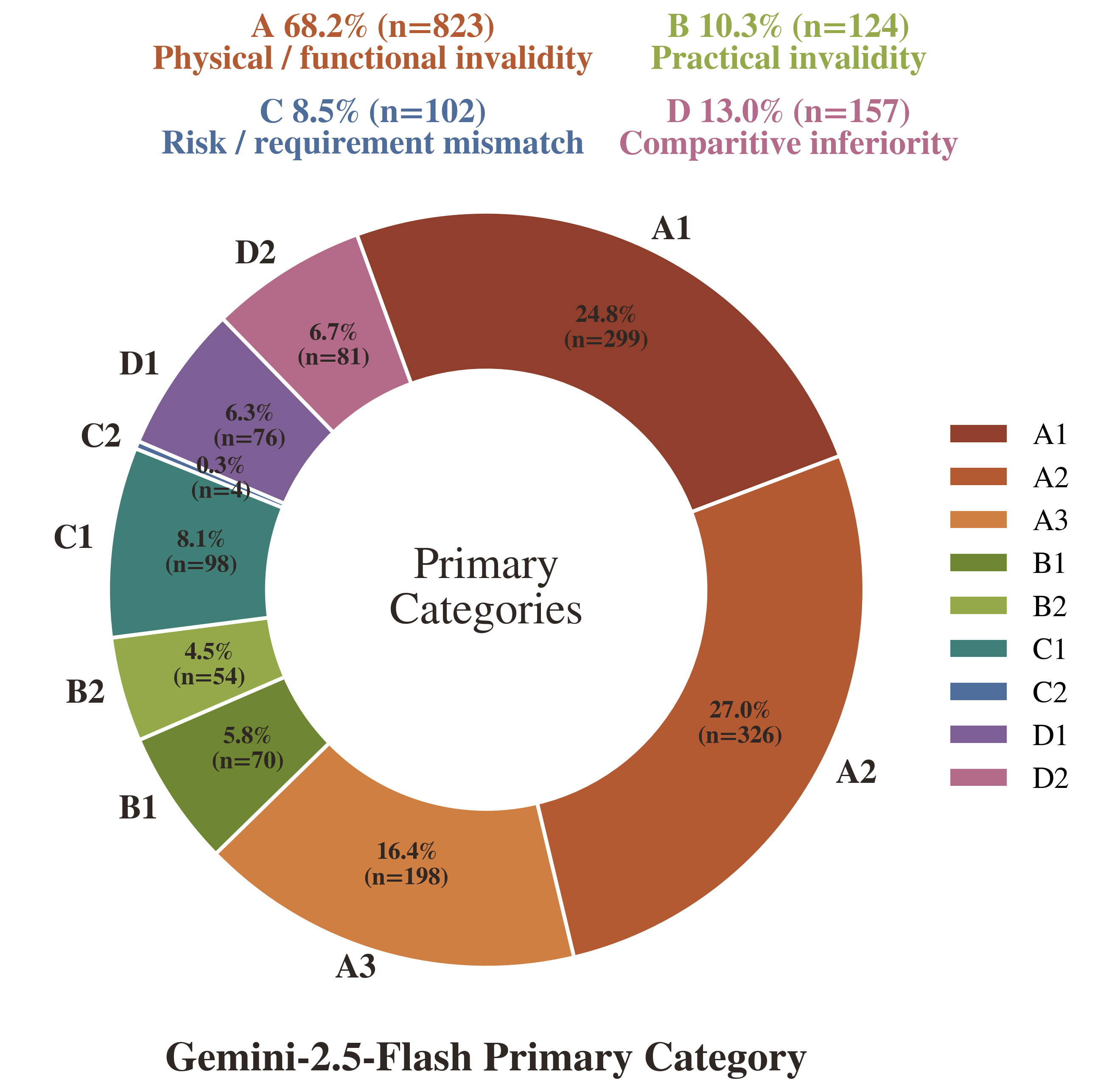}
        \label{fig:attribution_pie_gemini25flash}
    \end{subfigure}
    \hfill
    \begin{subfigure}[t]{0.6\linewidth}
        \centering
        \includegraphics[width=\linewidth]{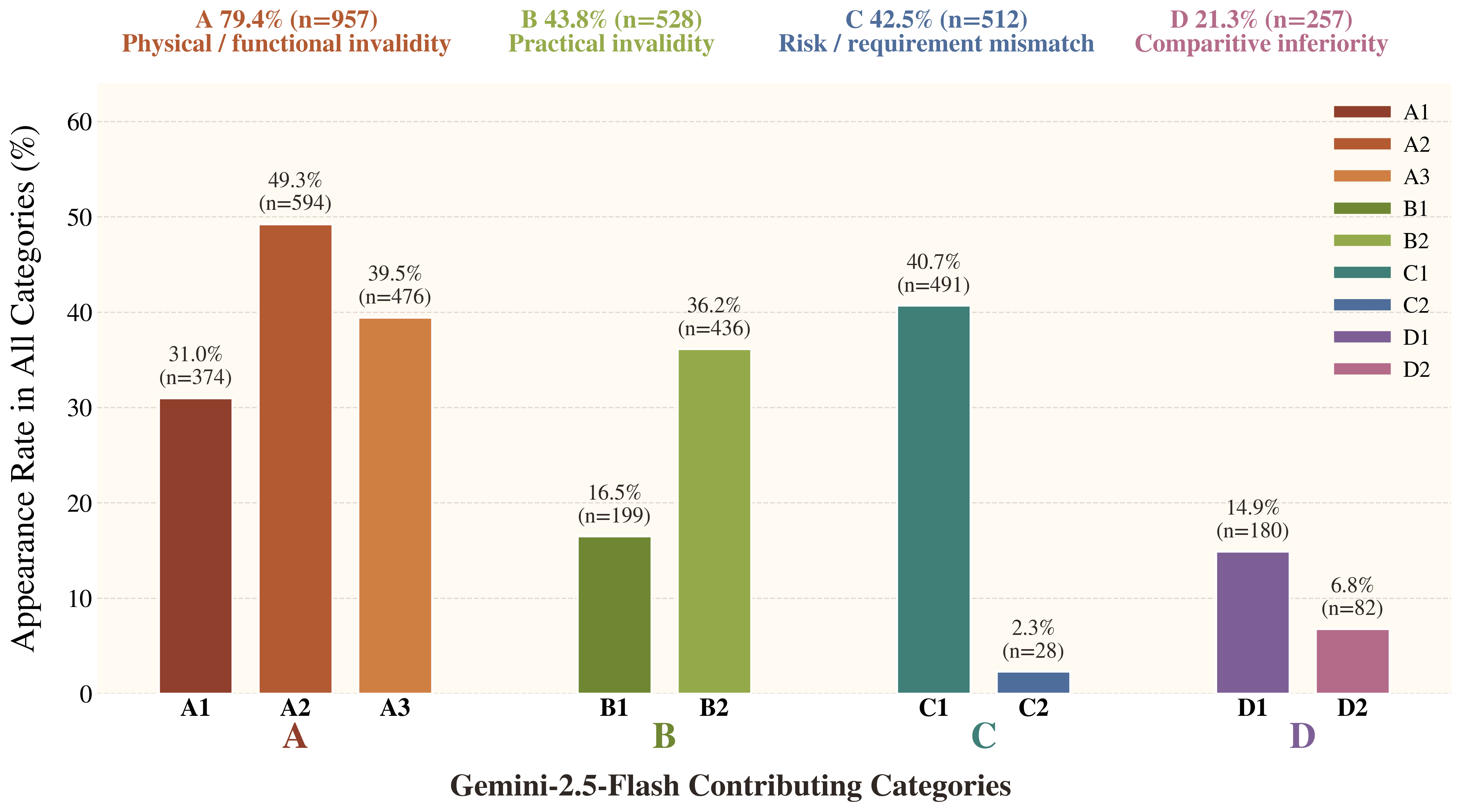}
        \label{fig:attribution_bar_gemini25flash}
    \end{subfigure}
    \caption{The error attribution analysis of Gemini-2.5-Flash respectively on primary category and all contributing reasons.}
    \label{fig:attribution_pie_bar_gemini25flash}
\end{figure}

\begin{figure}[t]
    \centering
    \begin{subfigure}[t]{0.37\linewidth}
        \centering
        \includegraphics[width=\linewidth]{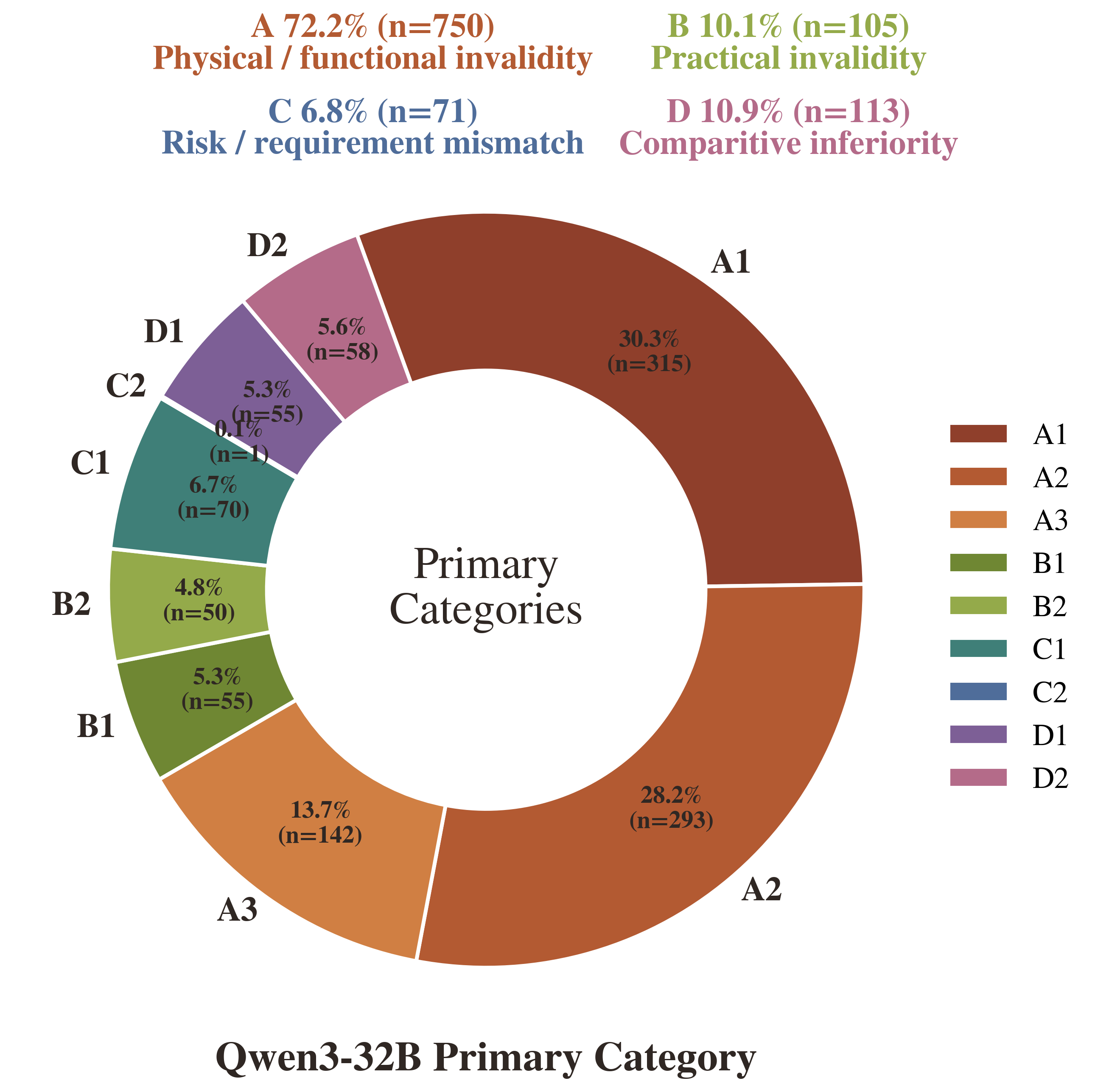}
        \label{fig:attribution_pie_qwen3_32b}
    \end{subfigure}
    \hfill
    \begin{subfigure}[t]{0.6\linewidth}
        \centering
        \includegraphics[width=\linewidth]{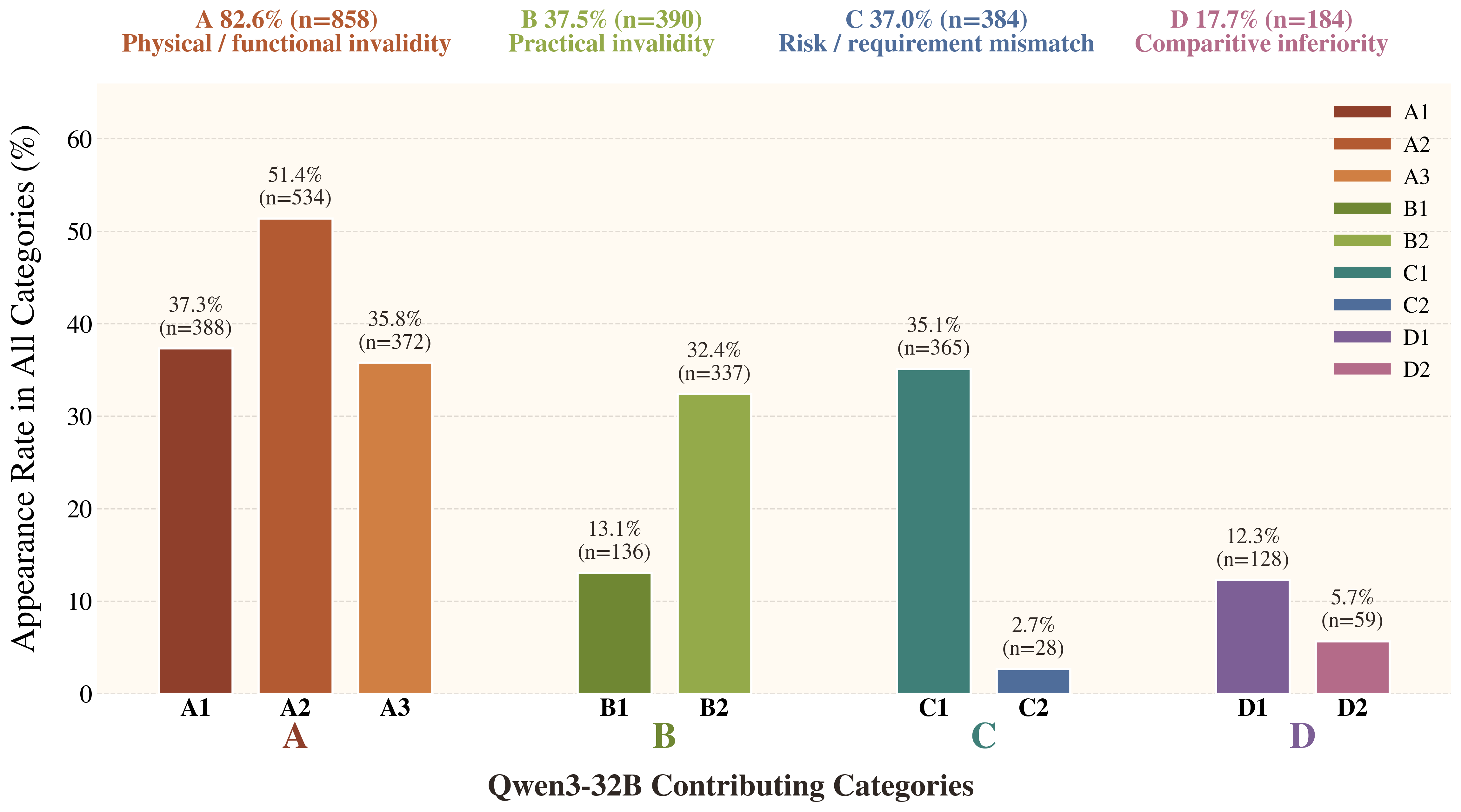}
        \label{fig:attribution_bar_qwen3_32b}
    \end{subfigure}
    \caption{The error attribution analysis of Qwen3-32B respectively on primary category and all contributing reasons.}
    \label{fig:attribution_pie_bar_qwen3_32b}
\end{figure}

\begin{figure}[t]
    \centering
    \begin{subfigure}[t]{0.37\linewidth}
        \centering
        \includegraphics[width=\linewidth]{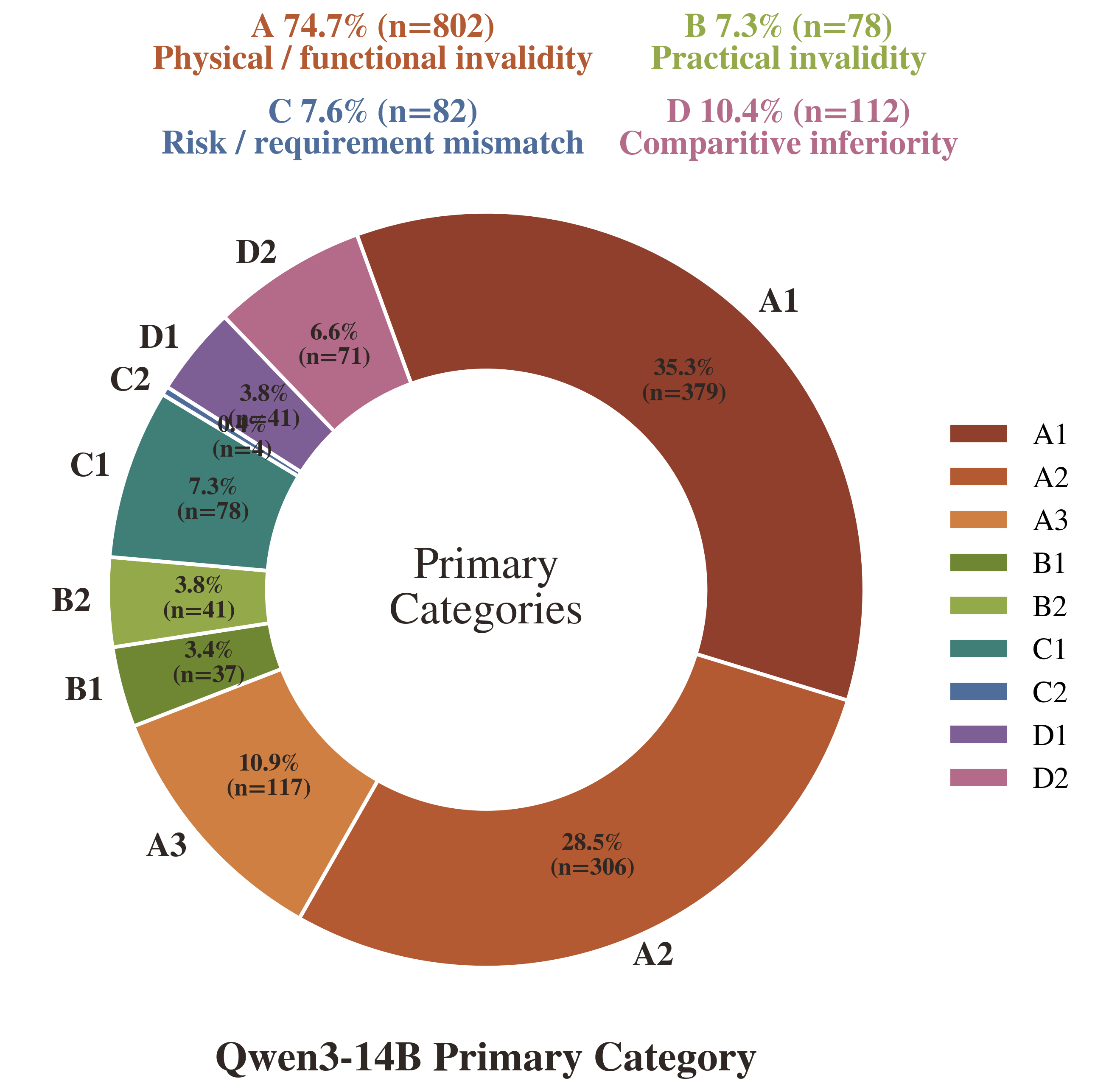}
        \label{fig:attribution_pie_qwen3_14b}
    \end{subfigure}
    \hfill
    \begin{subfigure}[t]{0.6\linewidth}
        \centering
        \includegraphics[width=\linewidth]{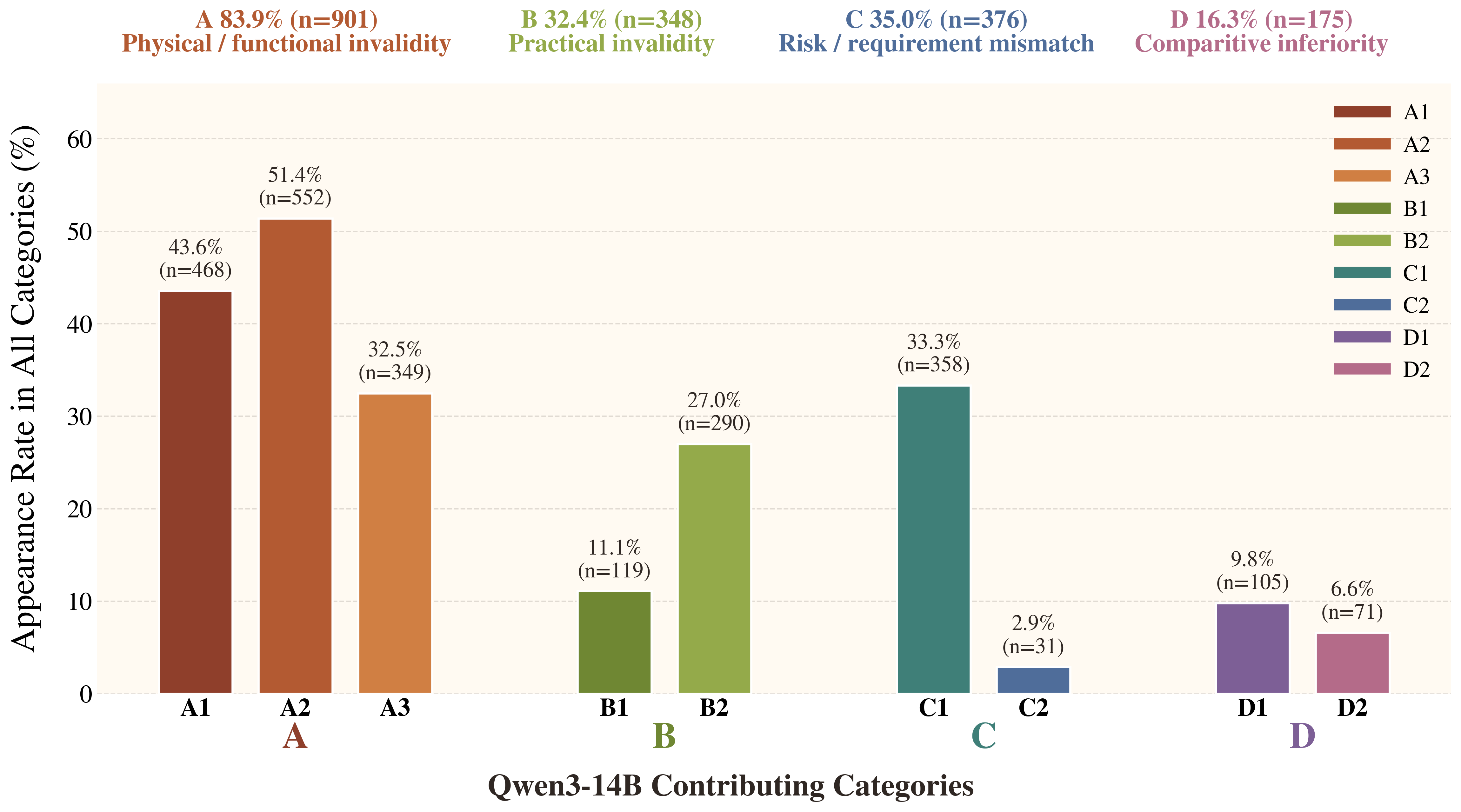}
        \label{fig:attribution_bar_qwen3_14b}
    \end{subfigure}
    \caption{The error attribution analysis of Qwen3-14B respectively on primary category and all contributing reasons.}
    \label{fig:attribution_pie_bar_qwen3_14b}
\end{figure}

\begin{figure}[t]
    \centering
    \begin{subfigure}[t]{0.37\linewidth}
        \centering
        \includegraphics[width=\linewidth]{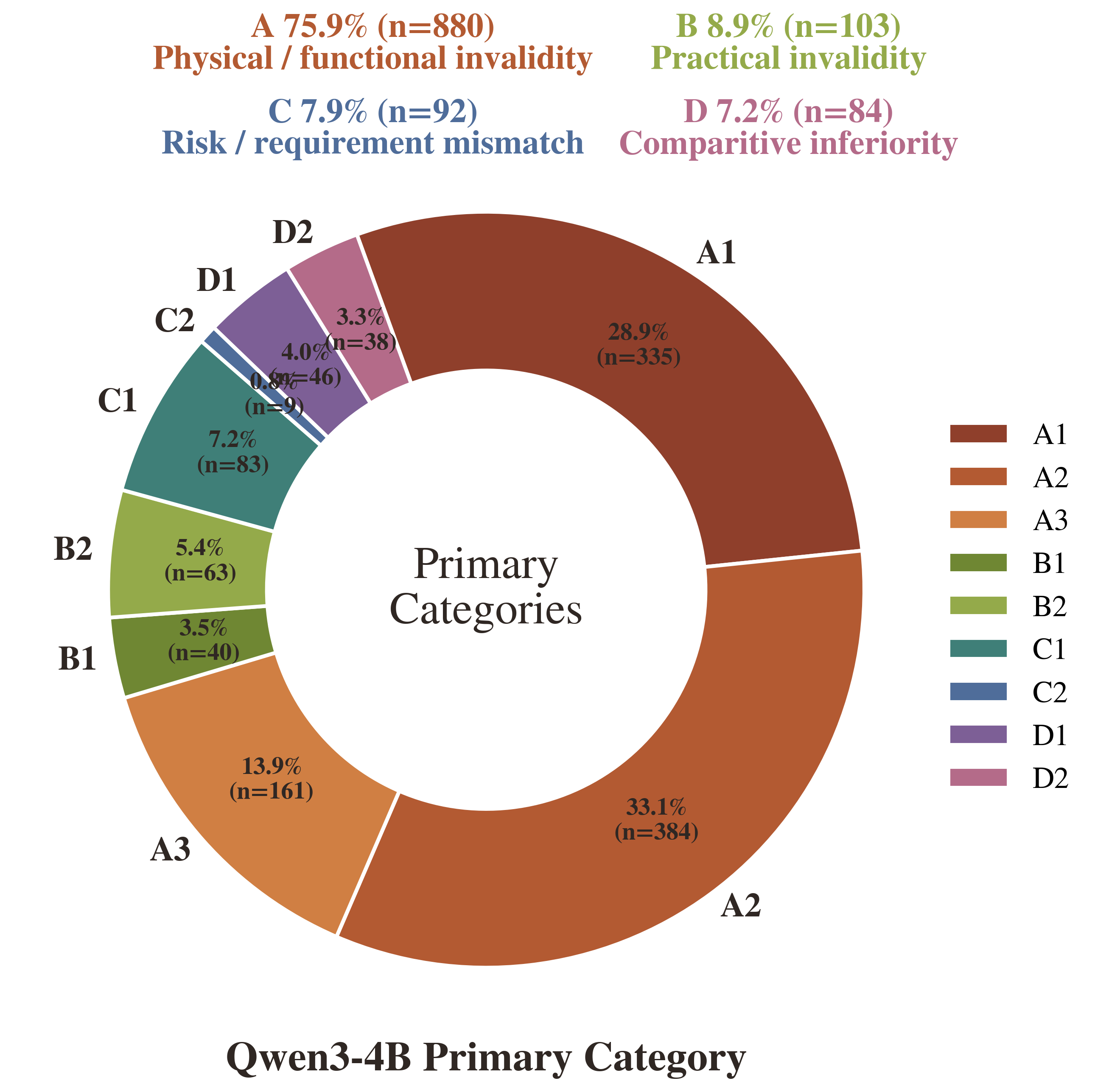}
        \label{fig:attribution_pie_qwen3_4b}
    \end{subfigure}
    \hfill
    \begin{subfigure}[t]{0.6\linewidth}
        \centering
        \includegraphics[width=\linewidth]{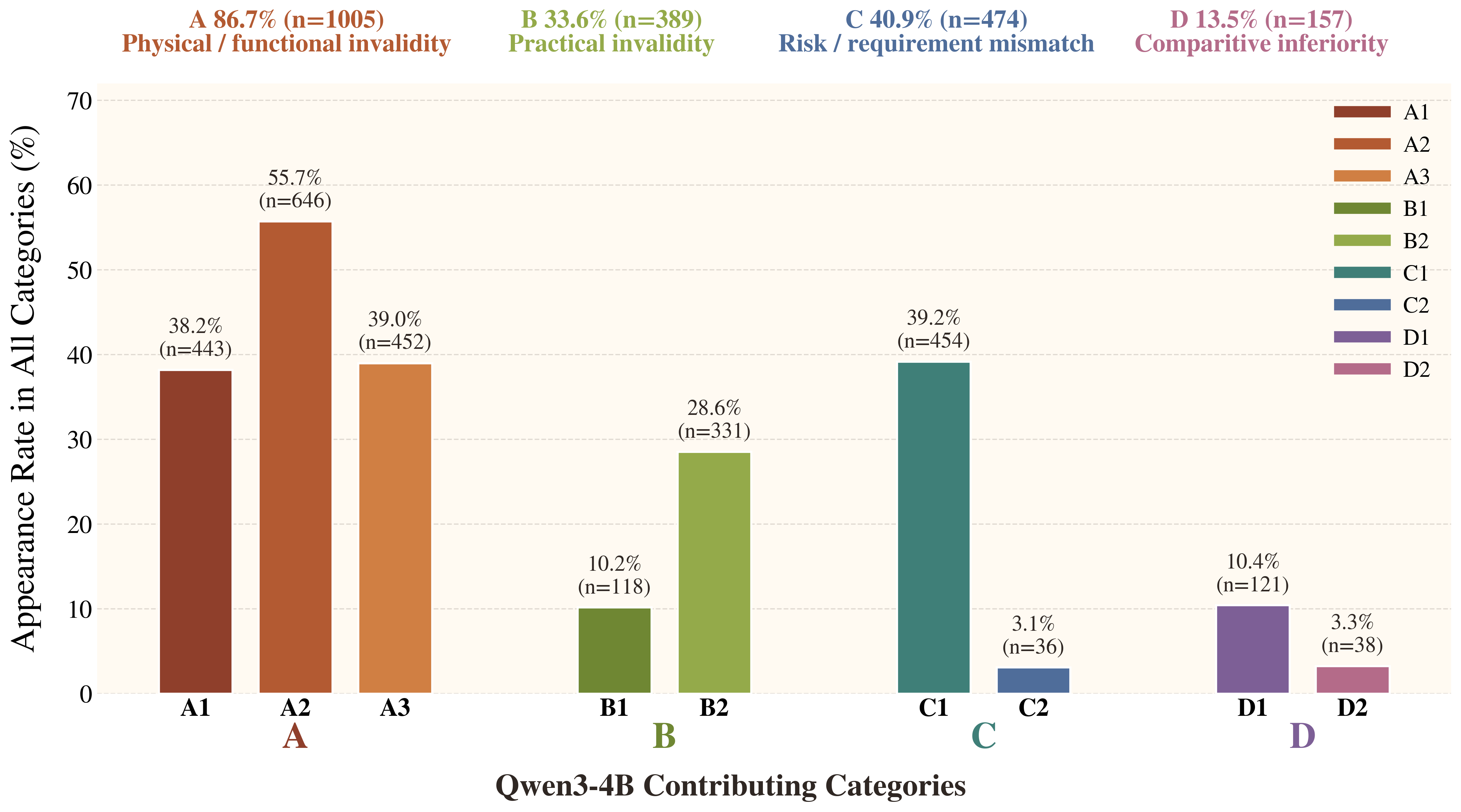}
        \label{fig:attribution_bar_qwen3_4b}
    \end{subfigure}
    \caption{The error attribution analysis of Qwen3-4B respectively on primary category and all contributing reasons.}
    \label{fig:attribution_pie_bar_qwen3_4b}
\end{figure}

\begin{figure}[t]
    \centering
    \begin{subfigure}[t]{0.37\linewidth}
        \centering
        \includegraphics[width=\linewidth]{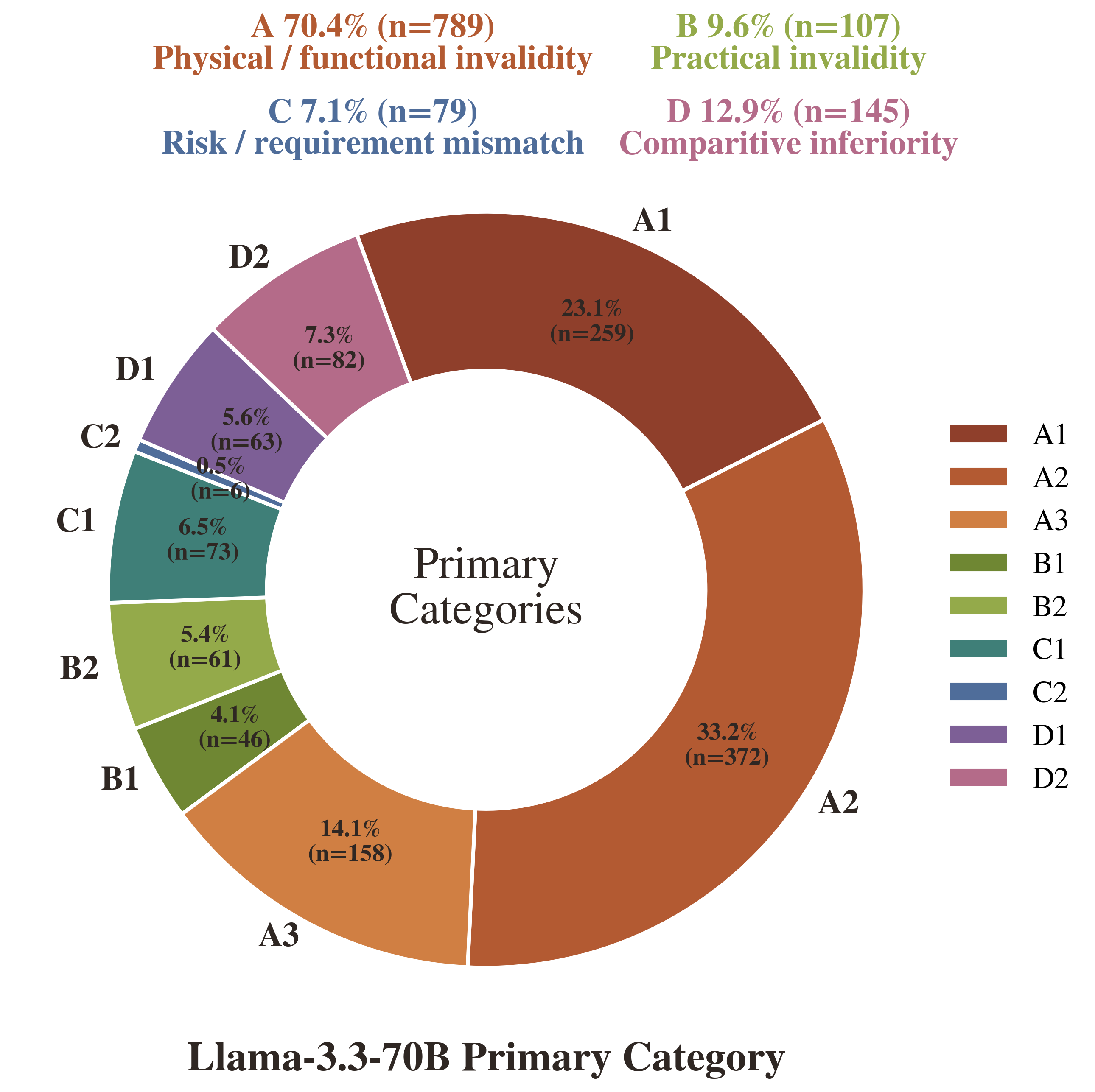}
        \label{fig:attribution_pie_llama33_70b}
    \end{subfigure}
    \hfill
    \begin{subfigure}[t]{0.6\linewidth}
        \centering
        \includegraphics[width=\linewidth]{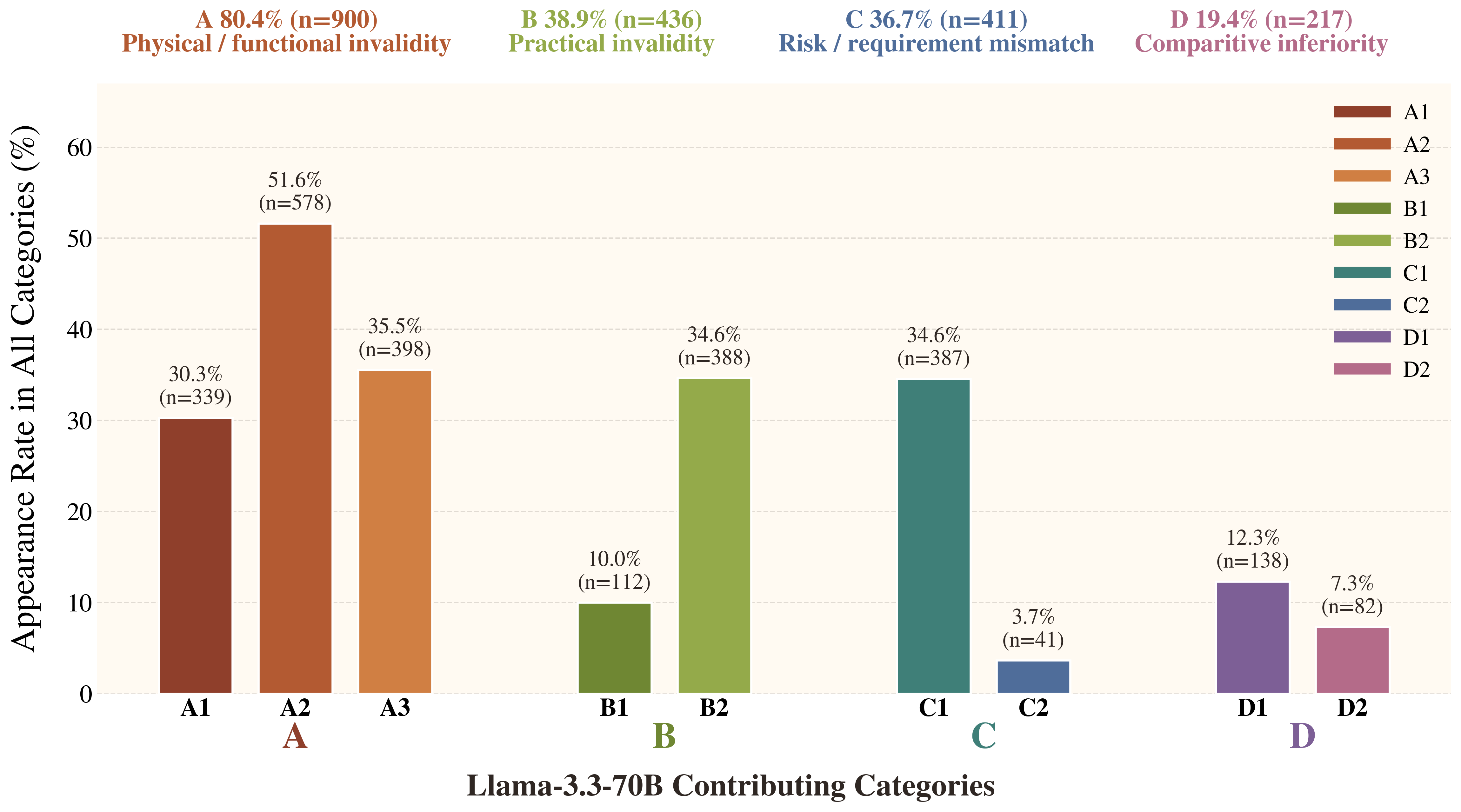}
        \label{fig:attribution_bar_llama33_70b}
    \end{subfigure}
    \caption{The error attribution analysis of Llama-3.3-70B-Instruct respectively on primary category and all contributing reasons.}
    \label{fig:attribution_pie_bar_llama33_70b}
\end{figure}

\begin{figure}[t]
    \centering
    \begin{subfigure}[t]{0.37\linewidth}
        \centering
        \includegraphics[width=\linewidth]{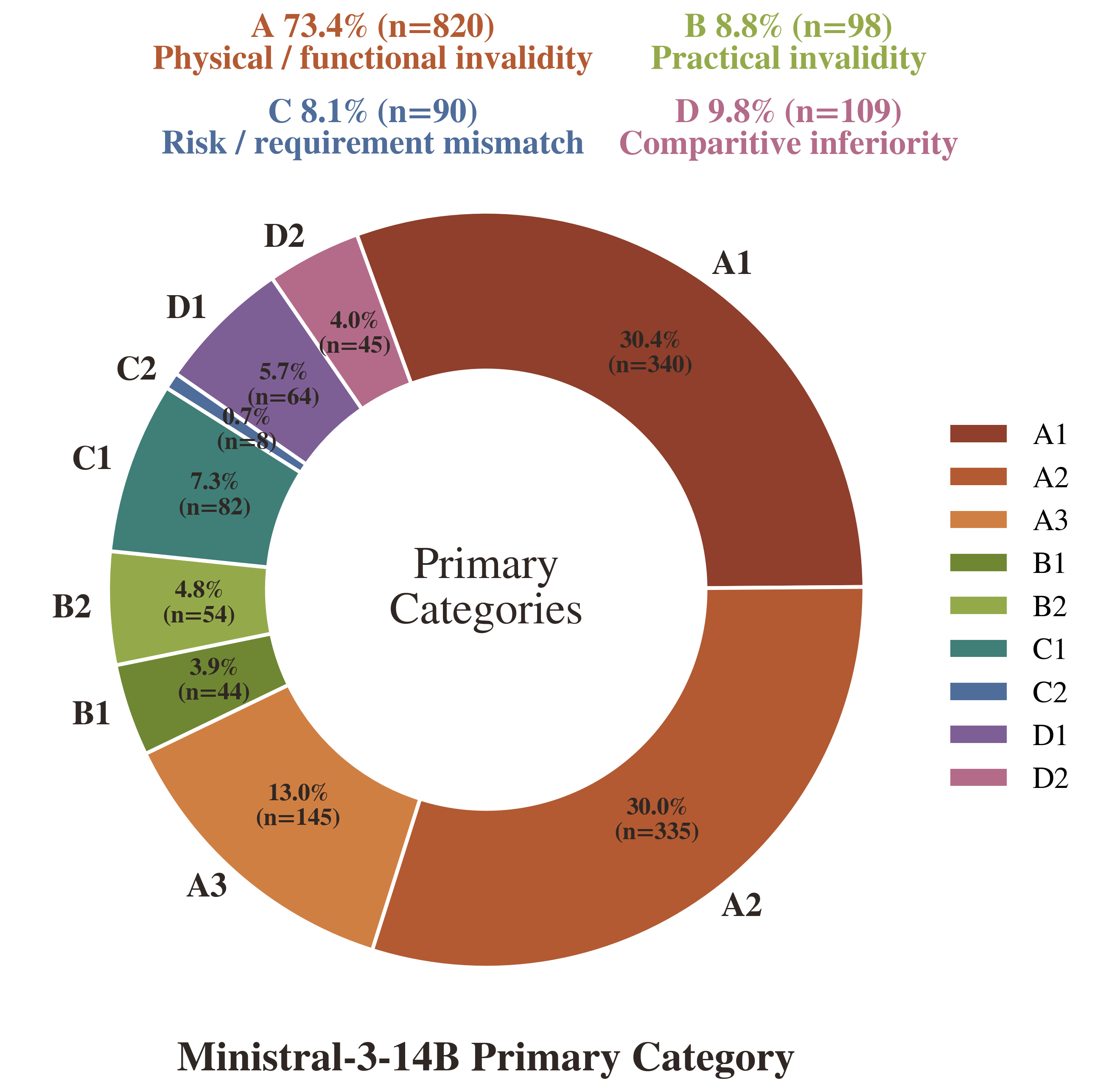}
        \label{fig:attribution_pie_ministral314b}
    \end{subfigure}
    \hfill
    \begin{subfigure}[t]{0.6\linewidth}
        \centering
        \includegraphics[width=\linewidth]{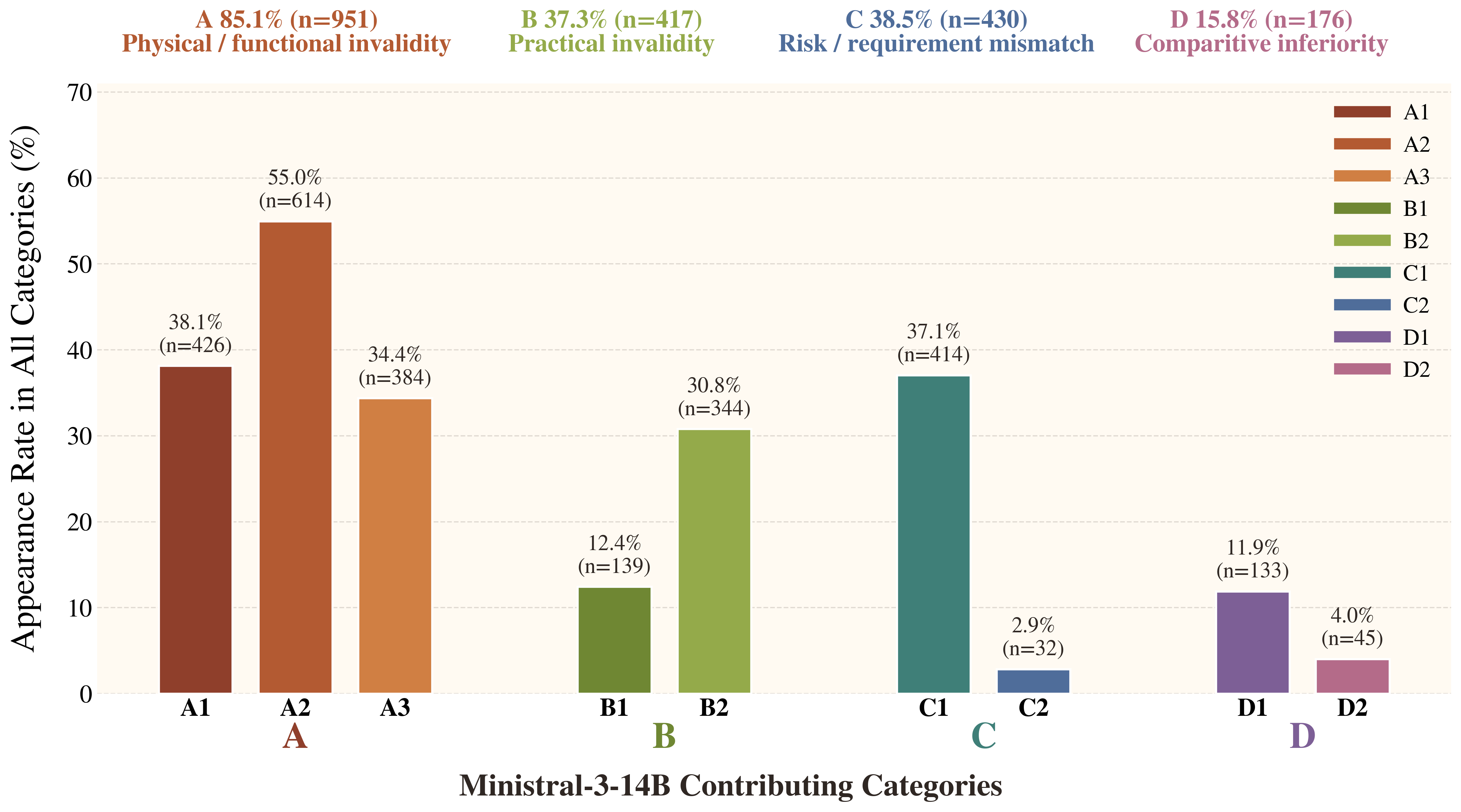}
        \label{fig:attribution_bar_ministral314b}
    \end{subfigure}
    \caption{The error attribution analysis of Ministral-3-14B-Reasoning-2512 respectively on primary category and all contributing reasons.}
    \label{fig:attribution_pie_bar_ministral314b}
\end{figure}

In the main text, we present the attribution analysis aggregated across all models. In this appendix, we further provide the fine-grained category breakdown for each individual model in \Cref{fig:attribution_pie_bar_gpt52} to \Cref{fig:attribution_pie_bar_ministral314b}. Overall, the patterns remain highly consistent with the main-text results: physical invalidity is the dominant failure reason across models, with similar fine-grained trends appearing in the category distributions.

\end{document}